\documentclass[sigconf]{acmart}
\AtBeginDocument{%
  }

\usepackage{stfloats}
\usepackage{amssymb,bm}
\usepackage{multirow}
\usepackage{soul}

\usepackage{subcaption}

\newcommand{\camready}[1]{\textcolor{black}{#1}}

\copyrightyear{2026}
\acmYear{2026}
\setcopyright{cc}
\setcctype{by}
\acmConference[KDD '26]{Proceedings of the 32nd ACM SIGKDD Conference on Knowledge Discovery and Data Mining V.2}{August 09--13, 2026}{Jeju Island, Republic of Korea}
\acmBooktitle{Proceedings of the 32nd ACM SIGKDD Conference on Knowledge Discovery and Data Mining V.2 (KDD '26), August 09--13, 2026, Jeju Island, Republic of Korea}
\acmDOI{10.1145/3770855.3819024}
\acmISBN{979-8-4007-2259-2/2026/08}

\newcommand{\lakefm}{\textsc{LakeFM}}

\begin{document}


\title[\lakefm{}: Toward a Foundation Model for Aquatic Ecosystems]{\lakefm{}: Toward a Foundation Model for Aquatic Ecosystems Using Irregular Multivariate Multi-depth Time Series Data}
\author{Abhilash Neog}
\affiliation{
  \institution{Virginia Tech}
  \city{Blacksburg}
  \state{VA}
  \country{USA}
}
\email{abhilash22@vt.edu}
\author{Sepideh Fatemi}
\affiliation{
  \institution{Virginia Tech}
  \city{Blacksburg}
  \state{VA}
  \country{USA}
}
\email{sepidehfatemi@vt.edu}
\authornote{Both authors contributed equally to this research.}

\author{Medha Sawhney}
\affiliation{
  \institution{Virginia Tech}
  \city{Blacksburg}
  \state{VA}
  \country{USA}
}
\email{medha@vt.edu}
\authornotemark[1]

\author{Kazi Sajeed Mehrab}
\affiliation{
  \institution{Virginia Tech}
  \city{Blacksburg}
  \state{VA}
  \country{USA}
}
\email{ksmehrab@vt.edu}

\author{Aanish Pradhan}
\affiliation{
  \institution{Virginia Tech}
  \city{Blacksburg}
  \state{VA}
  \country{USA}
}
\email{aanishp01@vt.edu}

\author{Bennett J. McAfee}
\affiliation{
  \institution{Grand Valley State University}
  \city{Muskegon}
  \state{MI}
  \country{USA}
}
\email{bennettjmcafee@gmail.com}

\author{Emma Marchisin}
\affiliation{
  \institution{University of Wisconsin - Madison}
  \city{Madison}
  \state{WI}
  \country{USA}
}
\email{marchisin@wisc.edu}


\author{Arka Daw}
\affiliation{
  \institution{Amazon AGI}
  \city{Seattle}
  \state{WA}
  \country{USA}
}
\email{dawark@amazon.com}

\author{Robert Ladwig}
\affiliation{
  \institution{Aarhus University}
  \city{Aarhus}
  \country{Denmark}
}
\email{rladwig@ecos.au.dk}

\author{Cayelan C. Carey}
\affiliation{
  \institution{Virginia Tech}
  \city{Blacksburg}
  \state{VA}
  \country{USA}
}
\email{cayelan@vt.edu}

\author{Paul C Hanson}
\affiliation{
  \institution{University of Wisconsin - Madison}
  \city{Madison}
  \state{WI}
  \country{USA}
}
\email{pchanson@wisc.edu}

\author{Anuj Karpatne}
\authornote{Corresponding author}
\affiliation{
  \institution{Virginia Tech}
  \city{Blacksburg}
  \state{VA}
  \country{USA}
}
\email{karpatne@vt.edu}

\renewcommand{\shortauthors}{Neog, et al.}
\begin{abstract}
  Understanding and forecasting lake dynamics is critical for monitoring water quality and ecosystem health across lakes and reservoirs. While machine learning methods have been recently applied to ecological time-series data, existing works assume regular sampling in time and depth, and struggle to generalize across lakes with heterogeneous variables, depths, and observation patterns. To address these limitations, we introduce \textsc{LakeFM}, a foundation model for aquatic systems, pre-trained on large-scale ecological datasets comprising both simulated and observed lakes. Through extensive empirical evaluation, we show that \textsc{LakeFM} learns meaningful representations spanning broader lake-level characteristics, and achieves competitive or often superior-forecasting performance compared to existing time-series foundation and non-foundation models, while producing physically plausible predictions consistent with real-world lake dynamics.
  \\
\noindent\camready{
\textbf{Project page}: 
\href{https://abhilash-neog.github.io/lakefm.github.io/}
{\textbf{abhilash-neog.github.io/lakefm.github.io/}}
}

\end{abstract}

\ccsdesc[500]{Computing methodologies~Machine learning}


\keywords{Foundation Model, Time Series, AI4Science}


\maketitle

\section{Introduction}
Monitoring the health of inland water bodies such  as lakes and reservoirs is essential for ensuring sustainable and equitable use of Earth's freshwater reserves. 
Lakes are governed by rich physical and biogeochemical processes that vary across geographies and time, creating unique opportunities for machine learning (ML) methods to model their temporal evolution across depths using ecological time-series data. For example, there is a growing body of work on modeling the temperature of water in lakes
\cite{daw2022physics,physicsrnn,ladwig2024modular}.
However, modeling a single variate only provides a partial view to the complex interactions of processes governing the quality of water in lakes, observed at varying depths, frequencies, subsets of variables, and levels of reliability from one site (lake) to another. 
While recent benchmarking efforts such as LakeBeD-US \cite{mcafee2025lakebed} have harmonized water quality observations across multiple monitoring programs resulting in over 500M observations across 17 variables from 21 lakes, it is still plagued with high degrees of missing values, uneven sampling frequencies, and highly variable depth and variate coverage across sites. \textit{This sparsity and heterogeneity in lake measurements, which is intrinsic to real-world environmental monitoring, severely limits the 
ability of ML methods to scale to broader collections of lakes using irregular multi-variate multi-depth time series data.}

At the same time, the broader ML community has made significant progress in developing time series (TS) foundation models such as Chronos 2 \cite{ansari2025chronos} and Moment \cite{goswami2024moment} that learn task-agnostic representations from large, heterogeneous corpora for generic time-series forecasting.
However, aquatic sciences still lacks a foundation model capable of unifying information across multiple lakes and variates with irregular frequencies and depths.
Moreover, most TS foundation models either focus solely on univariate signals or assume clean and densely sampled data that are difficult to find in ecology, where data is multivariate and inherently sparse across space and time. While recent efforts \cite{physics_fm_aqua, willard2022daily, willard2021predicting} have explored building large-scale foundation models for multiple lake systems, they are restricted to predicting a small number of  variates
with fixed sets of inputs at regular time scales without missing values.

Motivated by this gap, we ask the following questions: {\textbf{(1)} \textit{Can we build a foundation model for aquatic sciences that learns generic lake processes across a broad collection of lakes and variables, while retaining site-specific nuances?}} {\textbf{(2)} \textit{Can we use such a foundation model to forecast lake dynamics using any subset of variables available at a lake with irregular observations across time and depth?} \textbf{(3)} \textit{Can we extract feature representations of lakes that capture their static and time-varying characteristics, revealing novel information about their similarity  and temporal evolution at macro-system scales?}}

To answer these questions, we introduce \textsc{LakeFM}, a foundation model pre-trained on a \textit{large-scale ecological dataset containing over 1.5 million samples} comprising a mixture of synthetic data (over 1000 diverse lake simulations) from physics-based simulations and real-world observations coming from 21 lakes in the LakeBeD-US dataset \cite{mcafee2025lakebed} with significant sparsity (\textit{60-70\% on average}). To robustly handle irregular spatio-temporal data (which is common in scientific systems), \lakefm{} is designed to operate on an irregular grid unlike most temporal prediction (or time-series) models. Specifically, we model the data as a one-dimensional sequence of events or tokens, where each variate observation at each depth and time is treated as an event (we refer to it as a token throughout the paper). Every event/token is distinguished by its individual embedding that takes into account contextual metadata in the form of temporal, variate and depth information. Furthermore, to effectively capture both time-invariant (static lake characteristics) and time-variant (dynamic lake behavior) factors, we decouple the representation space into separate static and dynamic embeddings, and jointly optimize contrastive learning objectives with prediction losses over these spaces. Overall, \lakefm{} attempts to establish a practical step towards scalable and generalizable modeling of lake ecosystems. Our main contributions are as follows.
\noindent\begin{enumerate}
    \item We propose \lakefm{}, a foundation model that can ingest irregular, multi-variate multi-depth data, with competitive forecasting performance on both seen and unseen lakes while also demonstrating an emergent ability to adhere to aquatic physical laws.
    \item We present novel insights about the static characteristics and temporal evolution of lakes using learned lake-specific embeddings, and highlight how \lakefm{} representations effectively align with different ecological axes.
    \item We present case studies on forecasting performance under variate and depth masking scenarios, showing how \lakefm{}'s ability to handle partial inputs reveals novel insights about variable interactions in lakes.
\end{enumerate}
\vspace{-1ex}
\section{Related Works}
\label{sec:related_works}

Time-series forecasting models, including statistical approaches and deep learning architectures such as PatchTST \cite{nie2022time} and iTransformer \cite{liu2023itransformer} have shown strong performance on benchmark TS datasets. However, these models are domain- or dataset-specific, and hence struggle to generalize across ecosystems or variate configurations.
Scientific datasets, particularly in ecology and environmental modeling, involve unique challenges: missing values, irregular sampling, and multi-resolution measurements across time and depth. Models like mTAN \cite{shukla2021multi} and ContiFormer \cite{chen2023contiformer} attempt to address these issues through neural ODEs, temporal embeddings, or attention over irregular grids. However, these methods are often task-specific, rely on carefully engineered architectures, and do not scale well to large multi-lake or multi-variable ecosystems. While techniques like MissTSM \cite{neog2025masking} provide a model agnostic approach to handle missing values, it is not very computationally scalable. 

Recent Time Series Foundation Models (TSFM) aim to generalize across diverse time-series tasks by learning from large corpora of univariate (examples include MOMENT \cite{goswami2024moment}, Chronos \cite{chronos}, LPTM \cite{prabhakar2024large}, etc.) or multivariate signals (examples include Chronos 2 \cite{ansari2025chronos}, Toto \cite{cohen2024toto}. However, there are certain limitations.
Crucially, most current TSFMs operate under the assumption that data is fully observed or regularly sampled. While Chronos 2 can handle some sparsity, it remains ill-equipped for the highly irregular sampling intervals common in scientific datasets. This limitation creates a heavy dependency on external imputation methods. In scientific domains where data is significantly sparse, specialized imputation models like SAITS \cite{du2023saits} or CSDI \cite{tashiro2021csdi} often suffer from poor performance due to a lack of sufficient training signals, which subsequently degrades the accuracy of downstream forecasting models. Our approach overcomes this by considering each time, variate and depth observation as a token, thus converting the multi-variate multi-depth data as a list of tuples, thus, facilitating model training under partial observations and irregular sampling.
\vspace{2ex}

\section{Methodology}
\textbf{Background and Notations:}
{Let $\mathcal{D} = \{\mathcal{D}_1, \dots, \mathcal{D}_N\}$ denote a collection of $N$ lakes, where each lake $\mathcal{D}_i$ contains a multivariate, multi-depth time series: $\mathcal{D}_i = \left\{(\mathbf{x}_t^{(i)}, \mathbf{m}_t^{(i)}, \ell_i) \right\}_{t=1}^{T_i}$}. {Here, $\mathbf{x}_t^{(i)} \in \mathbb{R}^{V \times D}$ represents observations of $V$ variables across $D$ \textit{depth levels} for lake $i$ at time $t$. The time intervals between consecutive steps $t$ and $t+1$ are \textit{irregular}, varying dynamically from one lake to another (e.g., from daily to bi-weekly and even monthly observations). Further, the binary mask $\mathbf{m}_t^{(i)} \in \{0,1\}^{V \times D}$ indicates missing values across \textit{variables} and \textit{depths} at a time $t$, and $\ell_i$ denotes a categorical site-specific identifier of every lake, used for contrastive training.}

We formulate the task of \textit{probabilistic forecasting} for modeling lake systems as follows. Given a history of observations across a set of variables over $L$ irregular timesteps of a lake, the goal is to model the conditional distribution of all its lake variables over a time horizon $H$. To solve this problem, we employ an encoder-decoder framework where an encoder $f_\theta$ first maps the historical context $\{\mathbf{x}_t^{(i)}\}_{t=1}^{L}$ into a latent feature representation $\mathbf{z}_i \in \mathbb{R}^d$. This feature representation is subsequently processed by a decoder $g_\phi$ to generate the parameters of the future distribution of lake variables.

\begin{figure*}[t]
  \centering
    \includegraphics[width=0.8\linewidth]{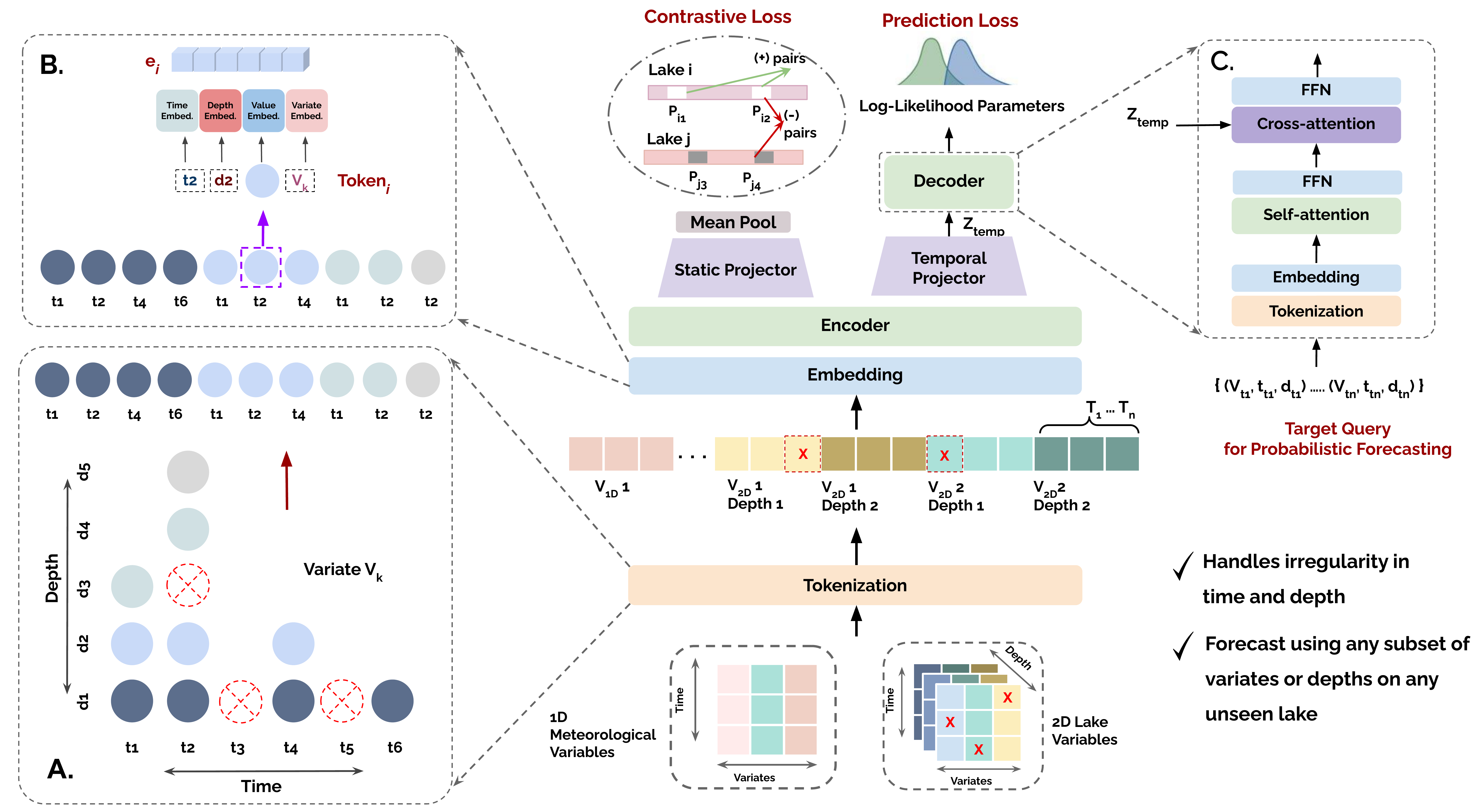}
  \caption{Overview of \textsc{LakeFM}. Tokenization and embedding of irregular multi-variable, multi-depth time-series data shown on the left. Overall Model architecture showing decoupled static and dynamic representation learning with joint forecasting and contrastive objectives in the middle, with the decoder shown on the right.}
  \label{fig:lakefm_overview}
\end{figure*}

\noindent\textbf{\textsc{LakeFM} Architecture:} Figure \ref{fig:lakefm_overview} shows the architecture of \textsc{LakeFM} that operates as an encoder-decoder framework. The overall framework comprises of four major components: (i) {tokenization and embedding}, (ii) {encoder layers}, (iii) static and temporal feature disentanglement strategy, and (iv) decoding and query-based forecasting strategy. We describe 
each of the components them in the following.


\subsection{Input Tokenization and Embedding}

To handle the heterogeneous and irregular nature of ecological data available in lake ecosystems, we adopt a token-centric representation as described in Figure \ref{fig:lakefm_overview}(A). Unlike regular grid-based approaches that require fixed depth levels, we treat every individual measurement---whether from a specific depth in the water column (2D variables) or a surface meteorological driver (1D variables)---as a distinct observation tuple containing time-variate-depth information. This allows our model to naturally ingest data with varying time intervals, subsets of variables, and depth resolutions without imputation or explicit handling of missing data.

\noindent\textbf{{Tokenization}:}
We represent the raw time series data for a specific lake $i$ as a set of observations $\mathcal{O}_i$, where each observation $o_k \in \mathcal{O}_i$ is defined as a tuple:
$o_k = (t_k, v_k, d_k, x_k)$
where $t_k$ is the absolute timestamp, $v_k \in \mathcal{V}$ is the variable identifier (e.g., temperature, DO, or air temperature), $d_k \in \mathbb{R}$ is the continuous depth measurement (where $d_k=0$ denotes surface/meteorological variables), and $x_k \in \mathbb{R}$ is the measured scalar value.
{Each observation $o_k \in \mathcal{C}_L$ is treated as a distinct token, where $\mathcal{C}_L$ is the context set comprising of all observations over the last $L$ timesteps. To form the input sequence $S$, we flatten the set $\mathcal{C}_L$ and sort the tokens first by variable ID, then by depth, and within each $(\text{variable}, \text{depth})$ series, we sort by absolute time as follows:
$S = [o_1, o_2, \dots, o_M],$
where $M$ is the total number of observed triplets $(t, v, d)$ across the $L$ timesteps.}

\noindent\textbf{{Embedding Layer}:}
To map the discrete observations into a latent space suitable for the Transformer backbone, we construct a composite embedding $e_k \in \mathbb{R}^{d_{\text{model}}}$ for each token $k$ (see Figure \ref{fig:lakefm_overview}(B)). This is formed by concatenating embeddings for time, depth, variable identity, and the scalar value as follows:
\begingroup
\small
\setlength{\abovedisplayskip}{3pt}
\setlength{\belowdisplayskip}{3pt}
\setlength{\abovedisplayshortskip}{2pt}
\setlength{\belowdisplayshortskip}{2pt}
\begin{equation}
    e_k = [E_{\text{time}}(t_k) \, ; \, E_{\text{depth}}(d_k) \, ; \, E_{\text{var}}(v_k) \, ; \, E_{\text{val}}(x_k)].
\end{equation}
Rather than summing these embeddings with the input token representation, we concatenate them to form the final token representation as we empirically found that concatenation leads to better performance. Specifically, concatenation preserves the semantic distinction between different embedding types, allowing the model to attend over heterogeneous subspaces independently, while summation tends to blur these roles in a shared latent space. We describe each of the four embeddings in the following.

\noindent\textbf{Time Embedding ($E_{\text{time}}$):} We utilize sinusoidal positional encodings to represent continuous time. For a timestamp $t_k$, the $j$-th dimension is given by:

\begingroup
\small
\setlength{\abovedisplayskip}{3pt}
\setlength{\belowdisplayskip}{3pt}
\setlength{\abovedisplayshortskip}{2pt}
\setlength{\belowdisplayshortskip}{2pt}
\begin{align}
    E_{\text{time}}(t_k)_j = \begin{cases} 
    \sin(2\pi t_k / 10000^{j/d_{\text{time}}}), & j \text{ is even} \\
    \cos(2\pi t_k / 10000^{(j-1)/d_{\text{time}}}), & j \text{ is odd}
    \end{cases}
\end{align}
where $t_k$ $\in$ [0,1] is the normalized time (or day of the year).


\noindent\textbf{Depth Embedding ($E_{\text{depth}}$):} Depth embeddings are generated using Fourier feature encoding, where every scalar depth $d_k \in \mathbb{R}$ is projected to a vector of sinusoidal components. Specifically, we apply $K$ frequency bands to produce: 
\begin{equation}
\begin{aligned}
E_{\text{depth}}(d_k)
    = \big[\, d_k \,;\,
    &\sin(\omega_0 d_k), \cos(\omega_0 d_k), \dots, \\
    &\sin(\omega_{K-1} d_k), \cos(\omega_{K-1} d_k)
    \big],
\end{aligned}
\end{equation}
where $\omega_k = \frac{2^k \pi}{\mathrm{max\_res}}$ for $k = 0, \dots, K{-}1$ frequency bands and $\mathrm{max\_res}$ is the maximum value of input used to scale frequencies, where the raw input $d_k$ is optionally concatenated (when enabled in the configuration). This Fourier feature vector is then linearly projected to the model depth embedding space via a learned matrix $\mathbf{W}_{\text{depth}} \in \mathbb{R}^{d_{\text{depth}} \times (1+2K)}$, yielding $\tilde{E}_{\text{depth}}(d_k) = \mathbf{W}_{\text{depth}}\, E_{\text{depth}}(d_k)$.

\noindent\textbf{Variate Embedding ($E_{\text{var}}$):} We employ a learnable lookup table (or embedding layer) to project the categorical variable identifier $v_k$ into a vector $E_{\text{var}}(v_k) \in \mathbb{R}^{d_{\text{var}}}$.

\noindent\textbf{Value Embedding ($E_{\text{val}}$):} 
To embed the continuous scalar measurement $x_k$, we utilize a SwiGLU-style gated projection, similar to the approach used in ~\citet{shi2024time}. This mechanism allows the model to non-linearly modulate the importance of the scalar input. Formally, let $\mathbf{W}_a, \mathbf{W}_b \in \mathbb{R}^{1 \times d_{\text{val}}}$ be learnable weight matrices and $\mathbf{b}_a, \mathbf{b}_b \in \mathbb{R}^{d_{\text{val}}}$ be bias vectors. The value embedding is then computed as:
\begin{equation}
    E_{\text{val}}(x_k) = \text{Swish}(x_k \mathbf{W}_a + \mathbf{b}_a) \odot (x_k \mathbf{W}_b + \mathbf{b}_b),
\end{equation}
where $\odot$ denotes the element-wise Hadamard product, and $\text{Swish}(\mathbf{z}) = \mathbf{z} \odot \sigma(\mathbf{z})$ is the Swish activation function.

 \subsection{Encoder Layers} 
 The encoder layers of \textsc{LakeFM} are a stack of Transformer blocks that transform a sequence of input tokens of length $S$ into feature representation outputs $\mathbf{H} \in \mathbb{R}^{S \times d}$. 
 To handle the irregularity in data, we adopt Rotary Position Embeddings (RoPE)~\cite{su2024roformer} for modeling relative temporal dependencies among tokens and a binary \textit{attention bias} similar to the approach used in  \citet{moirai} to explicitly differentiate intra- and inter-variate interactions.
 {Specifically, since standard self-attention treats all token interactions uniformly, we introduce a learnable additive bias to the attention logits to inject structural knowledge regarding variable identity. Let $v_i$ denote the variable identifier for token $i$. We define a binary indicator, $\delta_{ij} = \mathbb{I}[v_i = v_j]$, which equals 1 if tokens $i$ and $j$ belong to the same variable, and 0 otherwise. For every attention head $h$, we learn two scalar bias terms: $b_{\text{intra}}^{(h)}$, for same-variable interactions, and $b_{\text{inter}}^{(h)}$, for cross-variable interactions. The specific bias $b_{ij}^{(h)}$ for a query-key pair is determined by:
$
b_{ij}^{(h)} = \delta_{ij} \cdot b_{\text{intra}}^{(h)} + (1 - \delta_{ij}) \cdot b_{\text{inter}}^{(h)}.$
The attention score $s_{ij}^{(h)}$ is then computed as:
$
s_{ij}^{(h)} = \frac{\tilde{\mathbf{q}}_i^{(h)\top} \tilde{\mathbf{k}}_j^{(h)}}{\sqrt{d_h}} + b_{ij}^{(h)} + m_{ij},
$
where $\tilde{\mathbf{q}}$ and $\tilde{\mathbf{k}}$ denotes the RoPE-rotated queries/keys, $d_h$ is the head dimension, and $m_{ij}$ represents the padding mask. This formulation allows \textsc{LakeFM} to learn distinct attention patterns for temporal dynamics (intra-variate) versus variable correlations (inter-variate) within distinct heads, while preserving the relative temporal information via RoPE.}

\subsection{ {Static \& Temporal Feature Disentanglement}}

 {A key objective of \textsc{LakeFM} is to simultaneously model the temporal dynamics of lake state variables while capturing the intrinsic, time-invariant signatures unique to each lake site. 
 To achieve this, we introduce two parallel linear projectors that map the shared encoder output $\mathbf{H} \in \mathbb{R}^{S \times d}$ into specialized subspaces as follows.}

 {We consider a static feature projector $\mathbf{W}_{\text{stat}} \in \mathbb{R}^{d_{\text{stat}} \times d}$ and a temporal feature projector $\mathbf{W}_{\text{temp}} \in \mathbb{R}^{d_{\text{temp}} \times d}$ defined as:
$\mathbf{Z_{\text{stat}}} = \mathbf{H} \mathbf{W}_{\text{stat}}^\top, \quad \mathbf{Z_{\text{temp}}} = \mathbf{H} \mathbf{W}_{\text{temp}}^\top.
$
This formulation creates a learnable \textit{soft partition} of the encoded feature outputs. The static representations $\mathbf{Z_{\text{stat}}}$ are subsequently aggregated via mean pooling to generate a point-wise representation. Specifically, given token embeddings $\mathbf{Z^{\text{stat}}_{(i)}} = [\mathbf{z^{(i)}_1}, \ldots, \mathbf{z^{(i)}_S}] \in \mathbb{R}^{S \times d_{\text{stat}}}$ and mask $\mathbf{m^{(i)}} \in \{0,1\}^S$, we compute point-wise representations $\mathbf{z_i}$ as:
\begin{equation}
\begin{split}
\mathbf{\bar{z}^{(i)}} &= \frac{1}{\sum_{t=1}^{S} m^{(i)}_{t}} \sum_{t=1}^{S} m^{(i)}_{t}\, \mathbf{z^{(i)}_{t}} \in \mathbb{R}^{d_{\text{stat}}}, \\
\mathbf{z_i} &= g_{\text{proj}}(\mathbf{\bar{z}^{(i)}}),
\end{split}
\end{equation}
where $g_{proj}$ is a small projection head, and $\mathbf{z_i}$ is the final representation used during contrastive training. Conversely, the temporal representations $\mathbf{Z_{\text{temp}}}$ retain temporal information and serve as the input to the decoding and forecasting heads. By not enforcing explicit orthogonality, the network retains the flexibility to share information between static and temporal subspaces where beneficial, while allowing task-specific losses to drive the specialization of features.}

\subsection{Query-Based Forecasting Strategy}
Traditional time-series decoders typically operate on a regular grid utilizing a fixed feed-forward network, which is ill-suited for the irregular sampling inherent in aquatic ecosystems. To address this, \textsc{LakeFM} adopts a \textit{query-based forecasting} strategy (see Figure \ref{fig:lakefm_overview}(C)) described in the following.

 We define a set of target queries $\mathbf{Q_{\text{target}}} = \{(t_k, v_k, d_k)\}_{k=1}^{K}$ representing the specific spatiotemporal points where predictions are required. Here, $t_k$ of the target queries continue along the same axis as the $t_k$ of input tokens. However, unlike encoder inputs, these queries do not contain observed scalar values. We generate target embeddings $E_{\text{target}}$ using the same embedding layers used in the encoder, effectively acting as query prompts: ``What is the value of variable $v$ at depth $d$ and time $t$?''
 
 The decoder is a stack of Transformer blocks operating on target query embeddings, built from the target time, variate, and depth metadata defined as, $\mathbf{q_k} = [E_{time}(t_k); E_{var}(v_k); E_{depth}(d_k)] \in \mathbb{R}^{d_{query}}$, and $\mathbf{\tilde{q_k}} = \mathbf{W_q}\mathbf{q_k} \in \mathbb{R}^{d_temp}, \mathbf{W_q} \in \mathbb{R}^{d_{temp}\times d_{query}}.$ Each layer first applies self-attention over the target tokens (with padding masks) followed by a feed‑forward network, and then applies cross attention from the target queries to the encoder’s temporal representations (historical context), again followed by a feed‑forward network. The output $\mathbf{U}$ is defined as:
$
    \mathbf{U} = \texttt{Dec}(\mathbf{Q}, \mathbf{Z_{temp}}) \in \mathbb{R}^{S_t\times d_{temp}},$
where $S_t$ is the number of target tokens and $\mathbf{Q}$ is a stack of all target queries. Both self-attention and cross-attention layers use the same structured attentions as the encoder (variable-wise bias and temporal projections), enabling the decoder to integrate dependencies among prediction points while selectively retrieving relevant information from the encoded history. This formulation enables flexible inference at arbitrary unobserved points. \textit{Consequently, the output grid can be user-defined (either regular or irregular) with variable context and horizon lengths}.

\noindent\textbf{Probabilistic Prediction:}
To effectively capture predictive uncertainty and model the heavy-tailed noise distributions often observed in environmental data, we employ a probabilistic output head as follows. We map the final decoder representations via a feed-forward network to the parameters of a Student-$t$ distribution: $\Theta = (\mu, \sigma, \nu)$, where $\mu$ is the location, $\sigma$ is the scale, and $\nu$ represents the degrees of freedom. Specifically, we compute, $\mathbf{\mu} = \mathbf{W}_\mu^\top \mathbf{U} + \mathbf{b_\mu} \in \mathbb{R}^{S_t}$, and
$\mathbf{\sigma} = \texttt{softplus}(\mathbf{W_\sigma^\top} \mathbf{U} + \mathbf{b_\sigma}) + \epsilon \in \mathbb{R}^{S_t}$, where $\epsilon > 0.$
Crucially, rather than fixing the distribution or assuming normality, we explicitly learn the degrees of freedom of the distribution, $\nu$, as a dynamic parameter. This grants \textsc{LakeFM} the flexibility to adaptively transition between heavy-tailed regimes (low $\nu$) and Gaussian-like regimes (as $\nu \to \infty$), which is essential for modeling the heterogeneous behaviors of scientific variables. Specifically, for a given variate $k$, we compute,

\begingroup
\small
\setlength{\abovedisplayskip}{3pt}
\setlength{\belowdisplayskip}{3pt}
\setlength{\abovedisplayshortskip}{2pt}
\setlength{\belowdisplayshortskip}{2pt}
\begin{equation}
    \nu_k = \texttt{softplus}(\underbrace{f_{\text{base}}(E_{\text{var}}(v_k))}_{b_{v_k}} + \underbrace{f_{\text{temp}}(\mathbf{u}_k)}_{a_k}) + c,
    \label{eq:nu}
\end{equation}
\noindent where $E_{\text{var}}(v_k)$ is the variate embedding, $f_{\text{base}}$ and $f_{\text{temp}}$ are small linear heads, $\mathbf{u}_k$ is the decoder output and $c > 2$ is an empirically chosen constant for numerical stability. $a_{k}$ represents a per-token non-variate-specific $\nu$ that can be noisy and vary by time and depth for the same variable, whereas $b_{v_{k}}$ represents a variate-specific $\nu$, which learns per-variable uncertainty characteristics but doesn't adapt to different conditions (e.g., temp. at surface vs. at the bottom). Hence, we consider a sum of the base $\nu$ ($b_{v_{k}}$) and the context-aware refinement/adjustment ($a_{k}$) in our formulation (Eq. \ref{eq:nu}).


\noindent\textbf{Loss Functions:} {\textsc{LakeFM} is pre-trained to jointly optimize a probabilistic forecasting objective and a lake-wise contrastive objective.
For the forecasting component, we minimize the negative log-likelihood of the ground truth values under the predicted Student-$t$ distributions.
Let $\mathcal{I}$ denote the set of indices $(k, t)$ for all valid (unmasked) target tokens.
Given the predicted parameters $(\mu_{t}^{(k)}, \sigma_{t}^{(k)}, \nu_{t}^{(k)})$ from the projection head, the forecasting loss is:
\begin{equation}
\mathcal{L}_{\text{forecast}}
  = - \frac{1}{|\mathcal{I}|} \sum_{(k,t) \in \mathcal{I}}
      \log \mathcal{T}(y_{t}^{(k)} \mid \mu_{t}^{(k)}, \sigma_{t}^{(k)}, \nu_{t}^{(k)}),
\end{equation}}

 To encourage lake-specific representations, we adopt a contrastive learning objective. For each batch of $B$ samples we obtain corresponding representations $\{\mathbf{z}_1, \dots, \mathbf{z}_B\}$ and lake identifiers $\{\ell_1, \dots, \ell_B\}$. We treat samples from the same lake as positives and those from different lakes as negatives. Each representation is $\ell_2$-normalized, and we construct a weight matrix $w_{ij} = \mathbf{1}[\ell_i = \ell_j]$ that encodes lake-wise positives. The contrastive loss uses a weighted InfoNCE formulation with temperature $\tau$:
\begingroup
\small
\setlength{\abovedisplayskip}{4pt}
 \setlength{\belowdisplayskip}{4pt}
\begin{align}
\begin{gathered}
\mathcal{L}^{(i)} = 
    - \sum_{j} w_{ij} \left( \frac{z_i^\top z_j}{\tau} 
    - \log \sum_{k} \exp(z_i^\top z_k/\tau) \right)
    \big/ \sum_{j} w_{ij}, \\
    \quad \mathcal{L}_{\text{contrast}} = \frac{1}{B} \sum_{i=1}^{B} \mathcal{L}^{(i)}, \quad i = 1, \dots, B
\end{gathered}
\label{eq:hardcl_indiv}
\end{align}
\endgroup
 
 {The final pre-training objective combines forecasting and contrastive learning $\mathcal{L}_{\text{total}} =
  \mathcal{L}_{\text{forecast}} + \lambda_t \, \mathcal{L}_{\text{contrast}},$
where $\lambda_t$ is a time-varying weight. In our implementation, $\lambda_t$ is obtained by
combining (i) a short warmup schedule over the first few epochs, and (ii) an adaptive scaling rule to keep the magnitude of the contrastive term comparable to the forecasting loss.

\section{Experimental Setup}
\label{sec:exp_setup}

\textbf{Dataset.} The LakeFM model is pretrained over a mixture of realworld and simulation datasets. The real-world data is obtained from the LakeBeD-US dataset \cite{mcafee2025lakebed}, which consists of 500 million unique observations spanning 21 lakes across the United States and exhibit significant sparsity (60-70\% on average). The collection of simulation datasets comprises of two types of simulations - (a) WQHanson Simulations, consisting of 4 simulation lakes, generated using the process-based water quality model \cite{hanson2023}, and (b) FCR Simulations, consisting of 1000 simulations, generated using the GLM-AED process-based model \cite{hipseyGLM}. Please refer to Appendix \ref{sec:dataset} for more details on the datasets.

\noindent\textbf{Pretraining and Evaluation Setup.} We partition the LakeBeD-US data into an \textit{(a) In-Distribution (ID) set}, consisting of 15 lakes and an \textit{(b) Out-of-Distribution set}, comprising 6 lakes. \lakefm{} is pretrained on the ID lakes (using the first 70\% data of each lake), together with a subset of simulation lakes. We evaluate the model under two settings - \textit{(a) In-Distribution (ID)} evaluation, where we test it on the final 20\% (in time) held out data of each ID lake, and \textit{(b) Zero-shot generalization}, where we evaluate the model on the six entirely unseen OOD lakes. Please refer to the Appendix \ref{sec:lakebedus} for more details on the ID and OOD set partitioning.

\noindent\textbf{Baselines.} We evaluate \lakefm{} against three primary classes of baselines to assess its performance: (a) Time-Series Foundation Models (TSFMs), including two multivariate forecasting model: Chronos 2 \cite{ansari2025chronos}, \camready{MOIRAI \cite{moirai}}, and univariate models: LPTM \cite{prabhakar2024large}, and MOMENT \cite{goswami2024moment}, (b) a non-foundation or local model, iTransformer \cite{liu2023itransformer}, \camready{and (c) Irregularly-sampled time-series (IMTS) models: HyperIMTS \cite{hyperimts}, ReIMTS \cite{reimts}. The goal behind this selection is to ensure a comprehensive comparison against both general-purpose pretrained models, local forecasting models and forecasting models that can inherently handle irregularly sampled data. Detailed descriptions of the baseline and \lakefm{} implementation are provided in Appendix \ref{appendix:lakefm_implementaion_details}.}

\section{Results and Discussions}
\label{sec:results}

\subsection{Comparing Forecasting Performance}
\label{sec:forecasting}
Figure \ref{fig:iid_ood_lake_barplot} compares the overall lake-wise MSE (across all variates) of \lakefm{} \camready{and non-IMTS baselines} for five In-Distribution (ID) lakes and five Out-of-distribution (OOD) lakes (\camready{see Table \ref{tab:lakes_metadata_split} in Appendix \ref{sec:lakebedus} for details of abbreviated lake names used throughout the paper}). We can see that in the ID setting, \lakefm{} consistently shows lowest MSE across all lakes, while baselines like \camready{iTransformer show high variability on BM and GL4}. On the OOD lakes, \lakefm{} shows best zero-shot performance on all lakes except \camready{TR}. Note that the performance of iTransformer varies widely across the OOD lakes, since it only relies on local data from a specific lake for training and does not utilize transfer of knowledge across lakes in contrast to \lakefm{} and other foundation models. Tables \ref{tab:lakebed_full_mse_mae_OOD} and \ref{tab:lakebed_full_mse_mae_ID} in the Appendix \ref{sec:full_results} provide a detailed comparison of \lakefm{} and baselines for every variate-lake combination. While there are some variate-lake combinations where baselines are performing better, \lakefm{} shows the \textit{best overall rank} of \camready{$2.0$} across all OOD lakes and \camready{$2.03$} across all ID lakes in terms  of lake-wise MSE. Figure \ref{fig:prediction_plot} shows an example time-series of the Chlorophyll-a variable over lake BM comparing the test predictions of \lakefm{} with baselines. 

A key practical difference between \lakefm{} and many forecasting baselines is that \lakefm{} \textit{does not require any imputation} of lake data as it can directly work with irregular multi-variate time-series data, a feature common in many ecological applications. In contrast, standard TSFM baselines rely on the accuracy of imputation-based pre-processing techniques to transform data onto regularly gridded formats, which can be unstable in the presence of sparse data. \lakefm{} thus provides a novel paradigm shift for sharing information across disparate lakes with varying forms of irregularities in time, space, and variates, going beyond typical single-lake and single-variate analyses presented in previous works.

\camready{To further evaluate this imputation-free setting, we compare \lakefm{} with recent irregular/missing time-series (IMTS) baselines across all ID and OOD lakes. For readability, Figure~\ref{fig:imts_bar_plot} visualizes a subset of lakes, while the complete lake-wise and variate-wise results are provided in Tables \ref{tab:imts_ood} and \ref{tab:imts_id} in Appendix \ref{sec:full_results}. Across the full evaluation, \lakefm{} obtains the best average rank among the evaluated IMTS baselines, with ranks of $1.0$ and $1.27$ under the OOD and ID settings, respectively, indicating that the gains of \lakefm{} are not only due to avoiding standard imputation pipelines, but also reflect its ability to share information across heterogeneous lakes while directly modeling irregular observations.}

\begin{figure}[tb]
    \centering
    \includegraphics[width=0.38\textwidth]{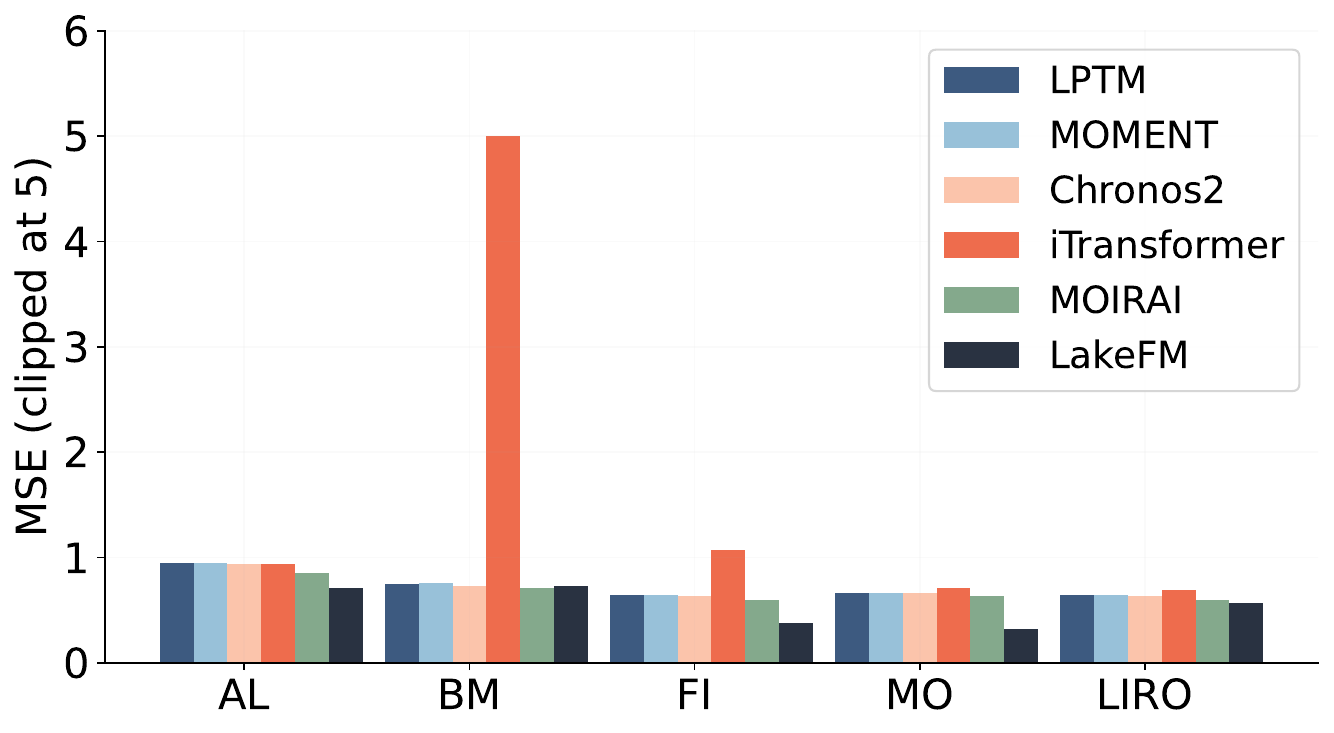}
    \includegraphics[width=0.38\textwidth]{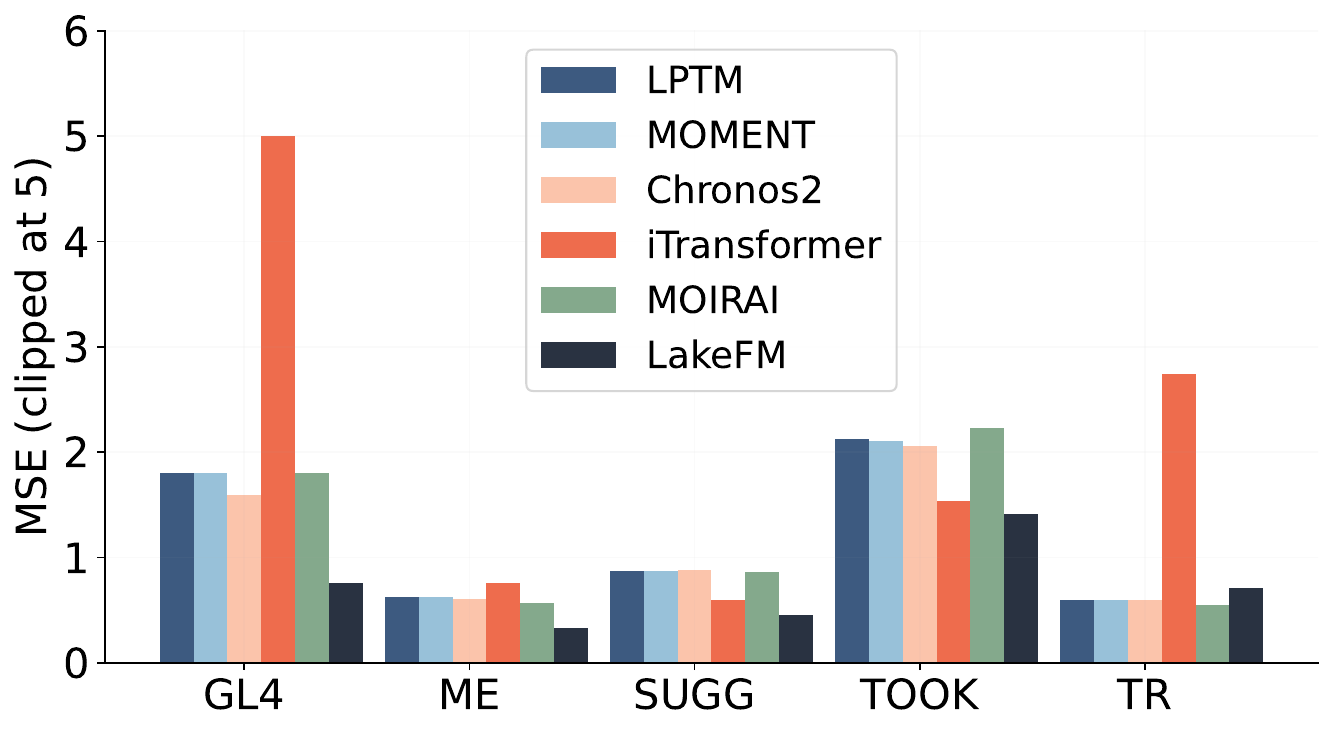}
    \caption{Overall lake-wise prediction performance (MSE) comparison between \lakefm{} and baselines:
    (top) ID lakes; (bottom) OOD lakes.}
    \label{fig:iid_ood_lake_barplot}
\end{figure}

\begin{figure}[tb]
    \centering
    \includegraphics[width=0.45\textwidth]{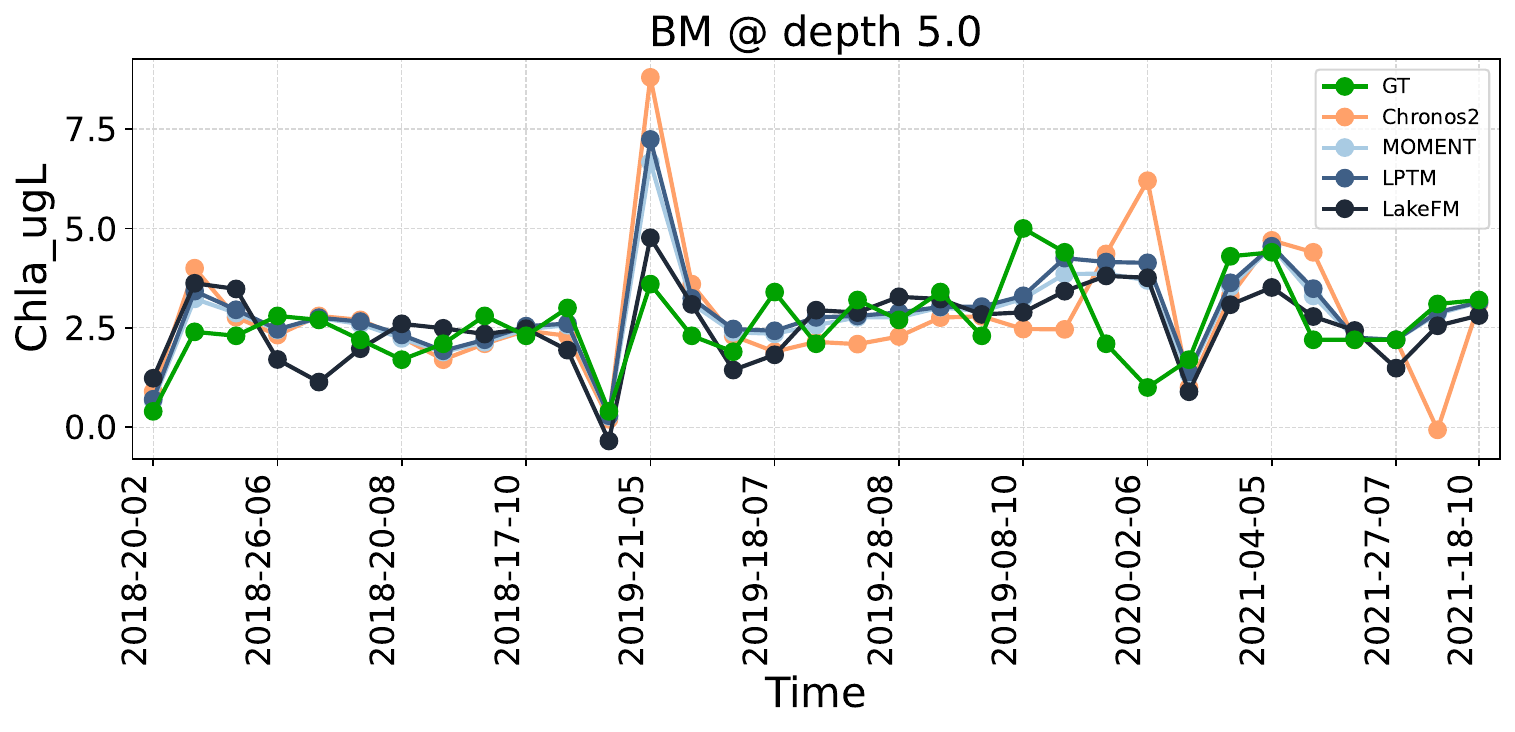}
    \caption{An example of time-series forecasts of chlorophyll-a for lake BM at 5m depth. The corresponding Mean Squared Error (MSE) values are: Chronos 2 (1.11), LPTM (1.23), MOMENT (1.24), and LakeFM (1.07).}
    \label{fig:prediction_plot}
\end{figure}

\begin{figure}[tb]
    \centering
    \includegraphics[width=0.38\textwidth]{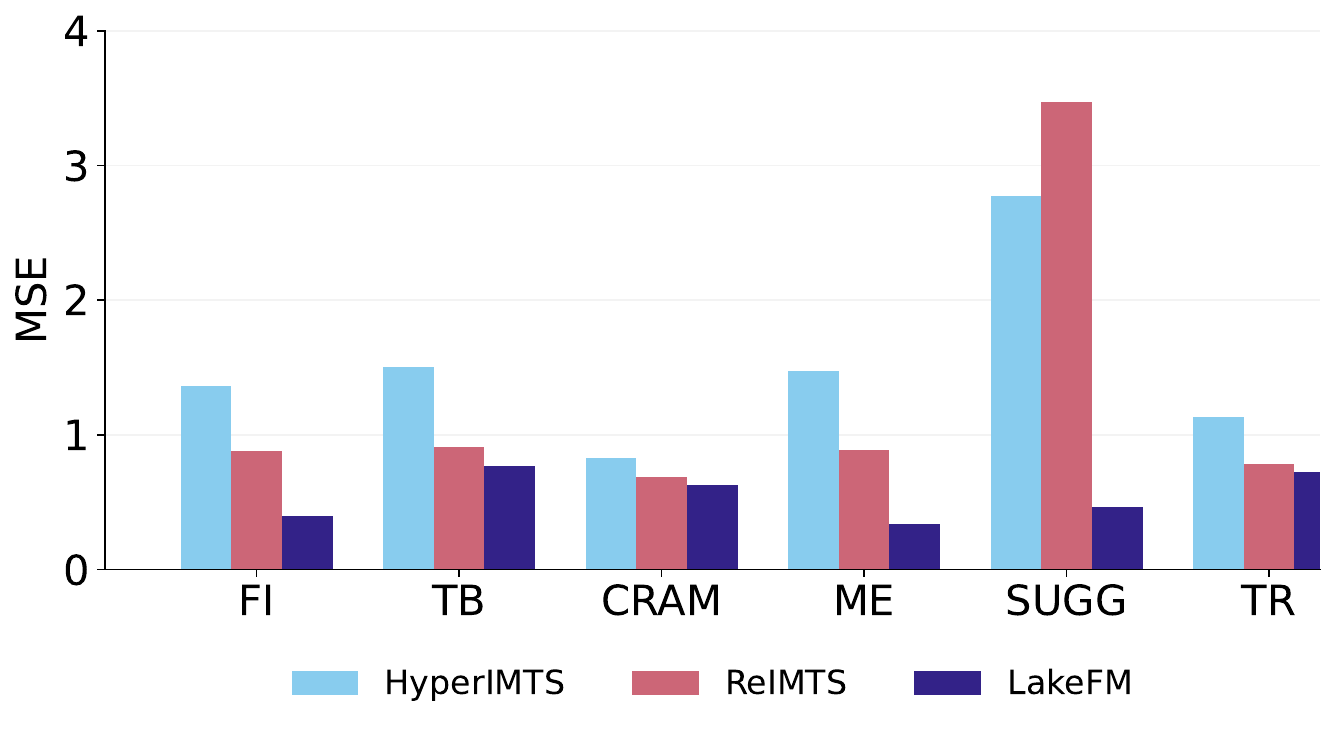}
    \caption{Overall lake-wise prediction performance (MSE) comparison between \lakefm{} and IMTS baselines:
    ID lakes: FI, TB, CRAM; OOD lakes: ME, TR, SUGG}
    \label{fig:imts_bar_plot}
\end{figure}
\subsection{Discovering Novel Insights of Lake Variate Interactions}

A unique feature of \lakefm{} is that it can be applied on a new lake with data available on any subset of variates and depths. This enables \lakefm{} to be used not only as a forecasting tool but as a novel \textit{discovery engine} for analyzing the interactions among lake variates at varying depths in relation to prediction performance. We specifically study the following two questions.

\subsubsection{{How Does Masking a Lake Variate Affect Forecasts of Other Variates?}}
\label{sec:vd_masking}
To study this question, we conduct experiments where we mask one or more variates in the context window during inference on a target lake and observe the forecasting performance of \lakefm{}, effectively relying on the cross-variate interactions learned during training. 
Figure \ref{fig:prla_masking_main}
shows time-series plots of one such experiment for lake PRLA, where we either mask out Dissolved Oxygen (DO) or Water Temperature (Temp) and observe their impacts on DO forecasts. 
\camready{We can see that masking DO leads to a larger increase in DO MSE than masking Temp, which can be intuitively explained based on the auto-correlation structure in DO. Quantitatively, DO masking results in a higher DO MSE of 12.57, compared to 11.00 under Temp masking. However, the uncertainty behavior reveals a more nuanced trend. Although Temp masking yields a lower MSE, it produces a higher CRPS of 2.52, and the corresponding forecast plots show narrower prediction intervals that often fail to capture the true values. This indicates that the model becomes more confident despite making inaccurate predictions. In contrast, DO masking increases the MSE but yields a lower CRPS of 1.93, with the forecast plots showing wider prediction intervals around the true trajectory. This suggests that the model responds appropriately to missing information by assigning greater uncertainty when a critical covariate is missing.} \camready{Appendix \ref{sec:appendix-variate-masking}} provides more visualizations of time-series forecasts with masked variables across multiple lakes revealing similar trends. This is generating \textit{novel hypotheses} about the effects of DO and Temp on the accuracy and uncertainty of forecasting lake variables that can be scientifically verified by ecologists in subsequent studies.

To further quantify the interactions between lake variates $(i, j)$ at any lake, we conduct two experiments. First, we mask variate $i$ and study the increase in MSE of variate $j$ compared to the no masking baseline. Second, we consider masking all variates except $i$ and measure the increase in MSE of variate $j$. \camready{Appendix \ref{sec:appendix-single-variate-masking} and \ref{sec:appendix-single-variate-input}} provide results for both these experiments with ecological observations of some of the common trends across lakes. 

\begin{figure}[!htbp]
    \centering
    \begin{subfigure}{\columnwidth}
        \centering
        \includegraphics[width=0.80\linewidth, height=3cm, keepaspectratio=false]{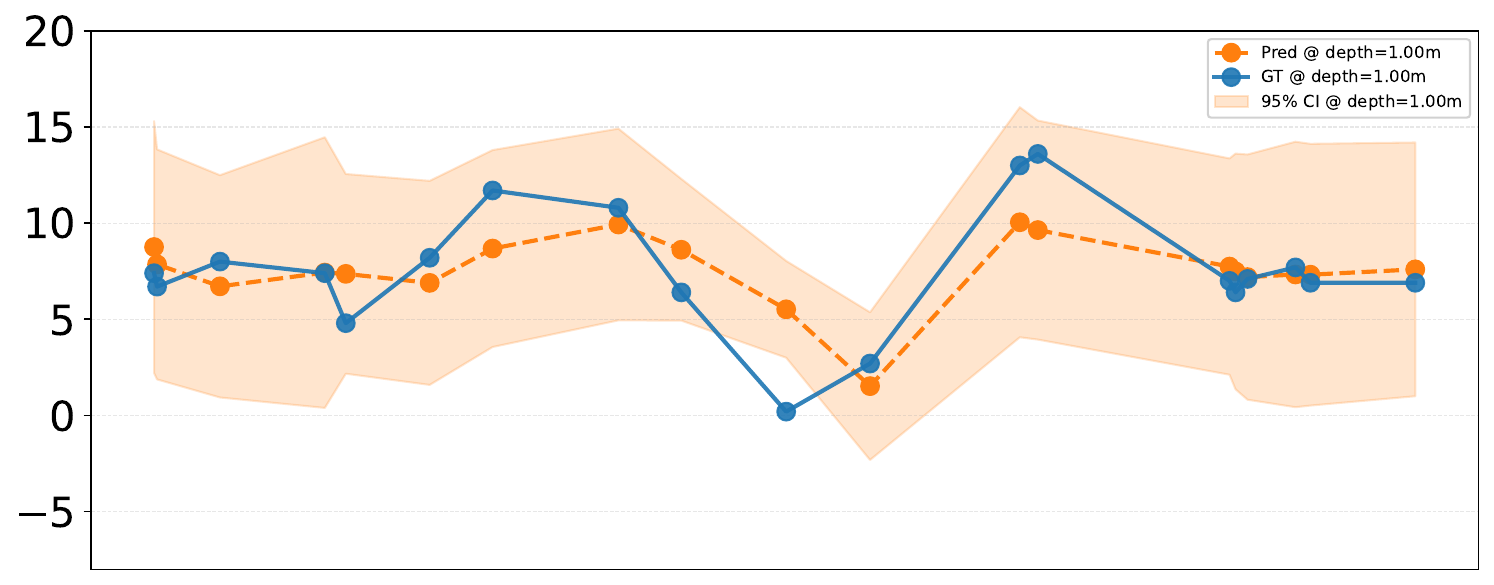}
        \caption{No masking}
        \label{fig:top_plot}
    \end{subfigure}

    \begin{subfigure}{\columnwidth}
        \centering
        \includegraphics[width=0.80\linewidth, height=3cm, keepaspectratio=false]{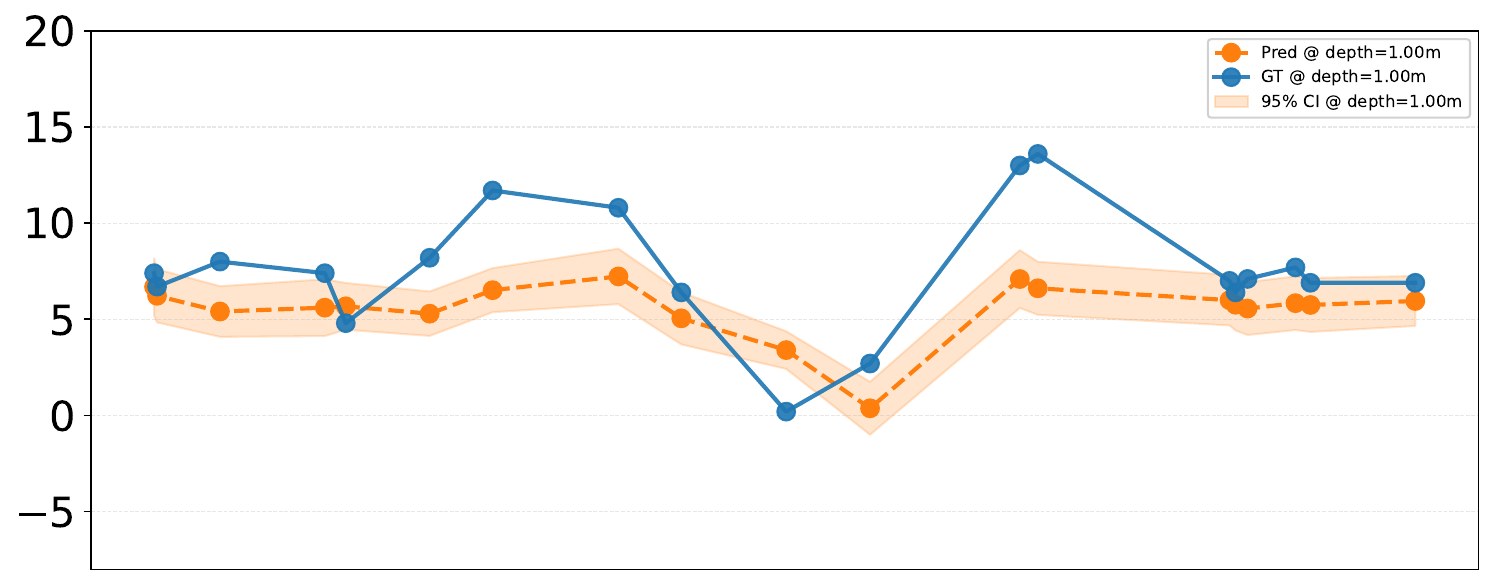}
        \caption{Temp Masked}
        \label{fig:middle_plot}
    \end{subfigure}

    \begin{subfigure}{\columnwidth}
        \centering
        \includegraphics[width=0.80\linewidth, height=3cm, keepaspectratio=false]{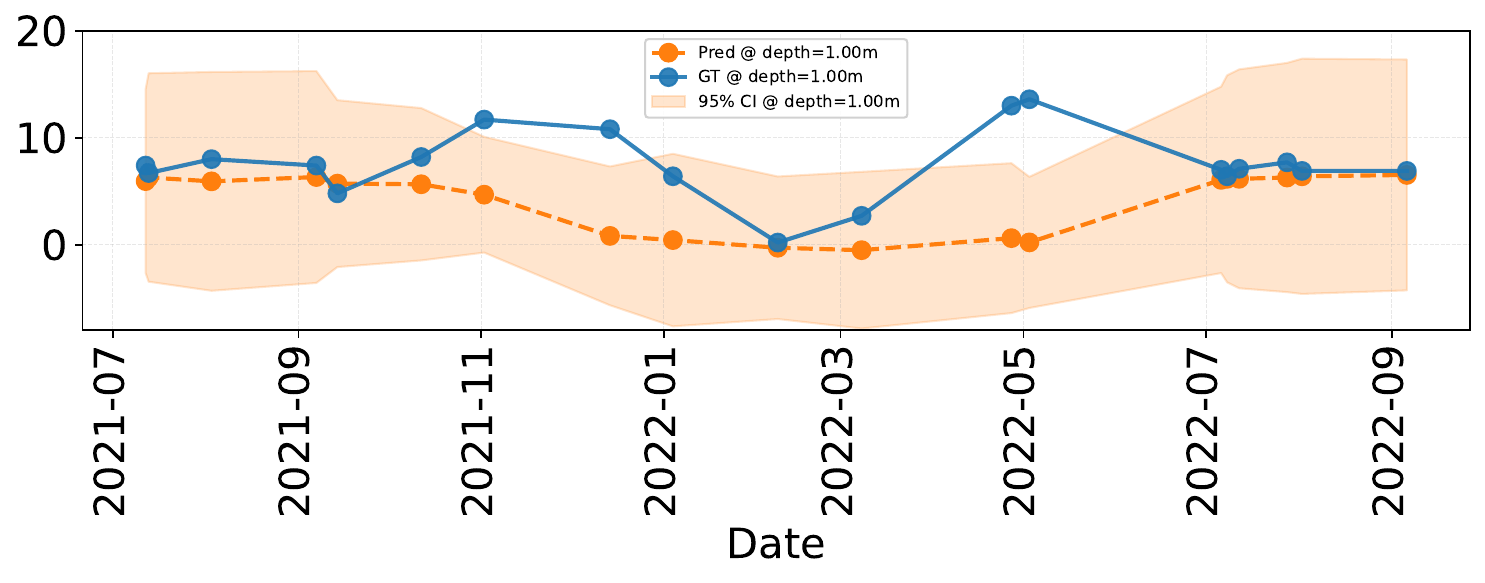}
        \caption{DO Masked}
        \label{fig:bottom_plot}
    \end{subfigure}
    \caption{Visualizing DO forecasts under masked and no masking scenarios for Lake PRLA at depth 1.0m}
    \label{fig:prla_masking_main}
\end{figure}

\begin{table}[t]
\centering
\small
\begin{minipage}[t]{0.48\textwidth}
\centering
\captionof{table}{Impact of masking shallow ($Z_s$) and deep ($Z_d$) depth information on forecasting performance (MSE) across lakes.}
\setlength{\tabcolsep}{4pt}
\renewcommand{\arraystretch}{1.2}
\resizebox{1.0\linewidth}{!}{
\fontsize{12pt}{12pt}\selectfont
\begin{tabular}{llcccc}
\toprule
& & \multicolumn{2}{c}{\textbf{$Z_{\text{shallow}}$ (MSE)}}
& \multicolumn{2}{c}{\textbf{$Z_{\text{deep}}$ (MSE)}} \\
\cmidrule(lr){3-4} \cmidrule(lr){5-6}
\textbf{Lake} & \textbf{Configuration}
& \textbf{Chronos 2}
& \textbf{LakeFM}
& \textbf{Chronos 2}
& \textbf{LakeFM} \\
\midrule
\multirow{3}{*}{CRAM}
& $Z_d$ \checkmark, $Z_s$ \checkmark
& 1.02 & \textbf{0.12}
& 1.24 & \textbf{0.18} \\
& $Z_d$ $\times$, $Z_s$ \checkmark
& 1.02 & \textbf{0.11}
& --   & {0.20} \\
& $Z_d$ \checkmark, $Z_s$ $\times$
& --   & {0.20}
& 1.24 & \textbf{0.14} \\
\midrule
\multirow{3}{*}{BARC}
& $Z_d$ \checkmark, $Z_s$ \checkmark
& 0.41 & \textbf{0.05}
& \textbf{0.20} & 0.38 \\
& $Z_d$ $\times$, $Z_s$ \checkmark
& 0.40 & \textbf{0.05}
& -- & {0.40} \\
& $Z_d$ \checkmark, $Z_s$ $\times$
& -- & {0.04}
& \textbf{0.19} & {0.43} \\
\bottomrule
\end{tabular}}
\label{tab:zdepth_ablation}
\end{minipage}
\end{table}


\subsubsection{{Shallow Layers vs. Deep Layers: What Matters More For Forecasting?}}
Similar to variate-masking, we study the effect of masking out all variates in the context window at shallow layers ($Z_{shallow}$) compared to deep layers $Z_{deep}$ on forecasting performance (\camready{see Appendix \ref{sec:appendix-depth-masking}} for details on how shallow and deep layers are defined for different lakes).  
Table \ref{tab:zdepth_ablation} shows the results of this masking experiment on two lakes (CRAM and BARC) comparing \lakefm{} with Chronos 2. We can see from the results of \lakefm{} that the shallow layers in the context window contain more predictive information about shallow layer forecasts across both lakes, and the same is true for deep layers. On the other hand, Chronos 2 does not register any significant difference in MSE values by masking shallow or deep layers. Also note that Chronos 2 requires a variate to be present in the context window to compute its forecast. Hence, it is unable to analyze the impact of masking shallow (or deep) layers upon itself.

\subsection{Analyzing Consistency with Physical Laws}
\label{sec:physical_consistency}

\begin{figure}[htbp]
    \centering
    \includegraphics[width=0.227\textwidth]{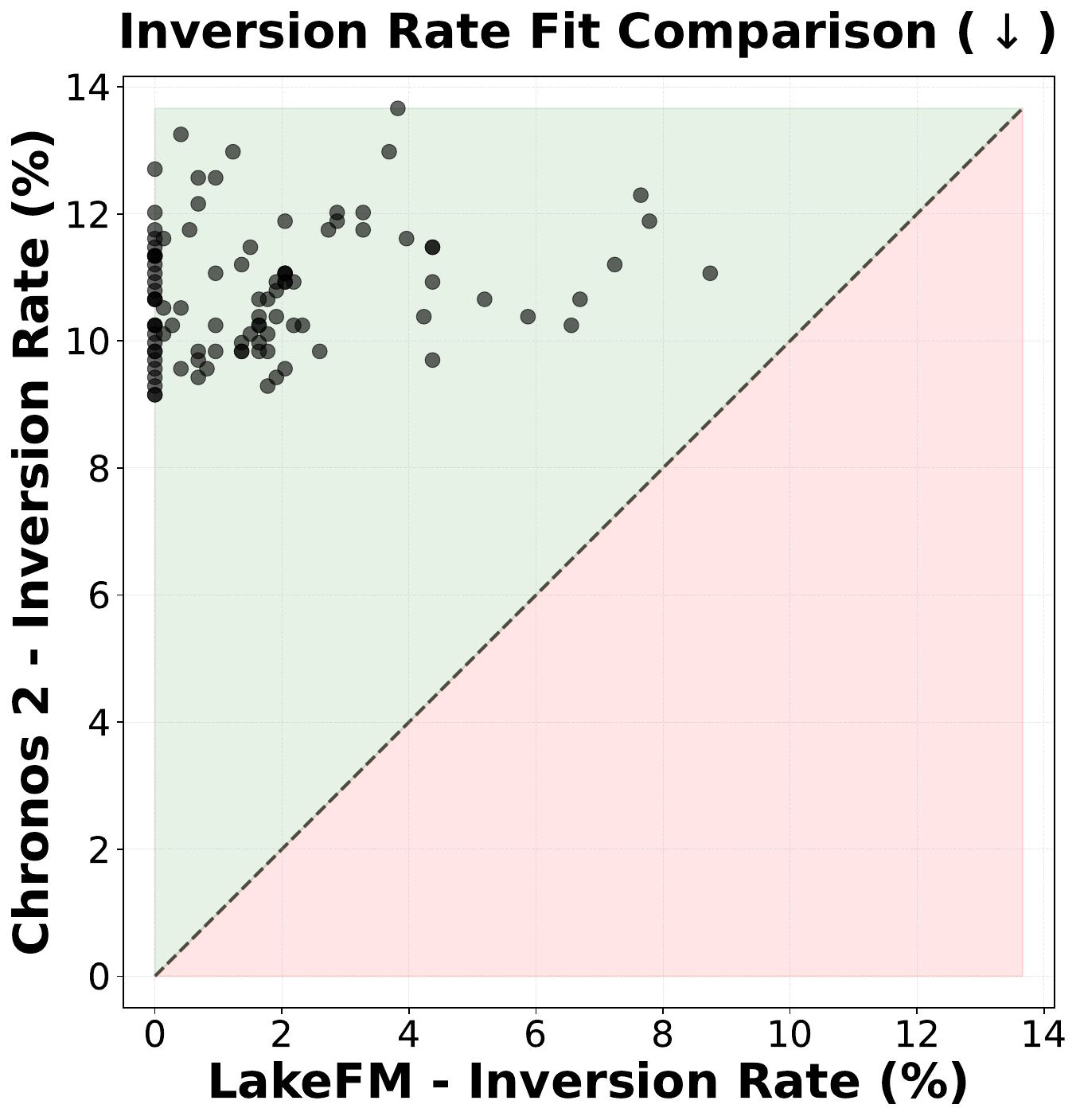}
    \hfill
    \centering
    \includegraphics[width=0.233\textwidth]{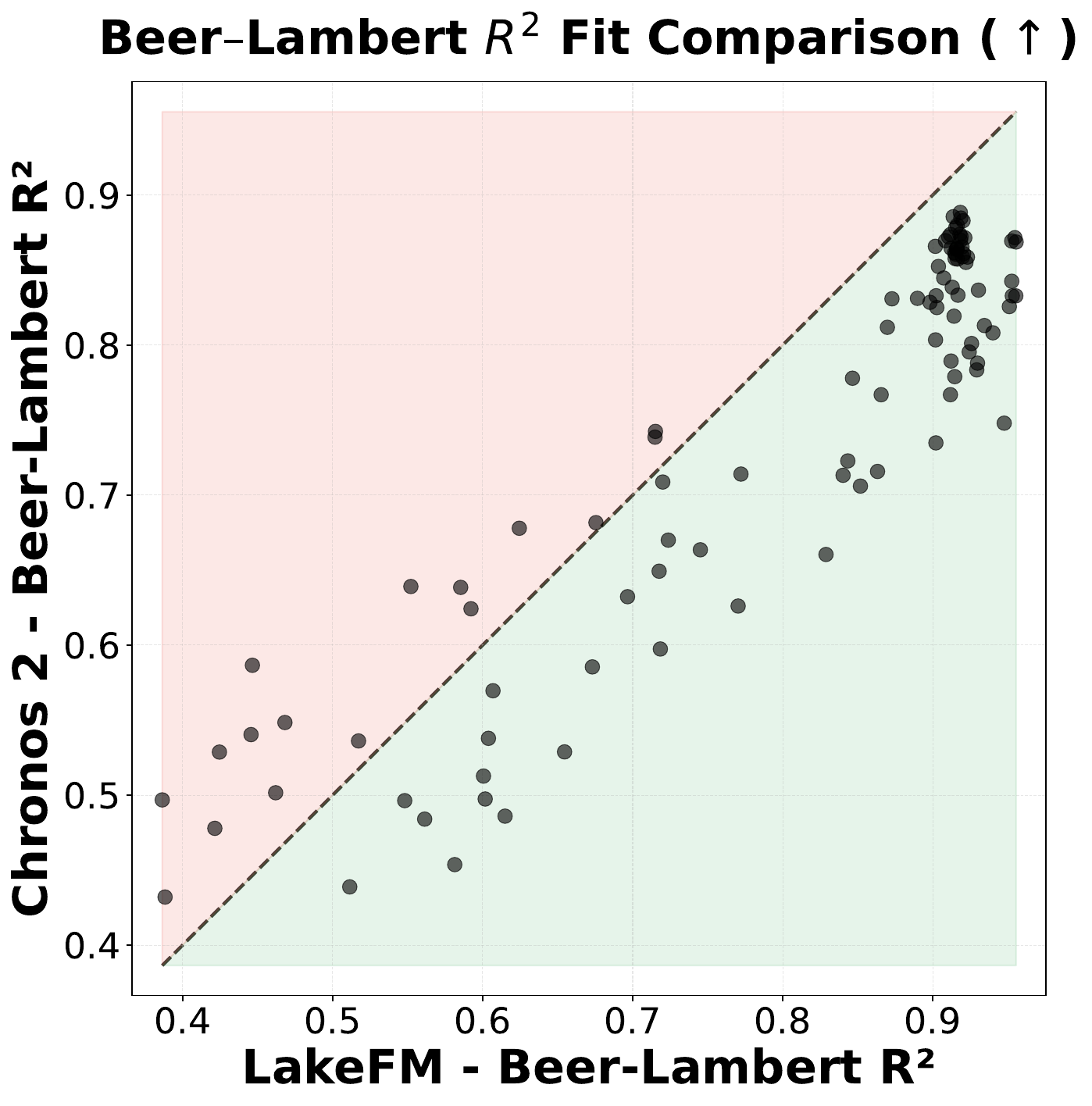}
    \caption{Comparing physical consistency of LakeFM \& Chronos~2  across 100 simulated lakes: (\textit{left}) inversion rate with thermal stratification law ($\downarrow$); (\textit{right}) Pearson $R^2$ with Beer-Lambert law ($\downarrow$).}
    \label{fig:lakefm_chronos_comparison}
\end{figure}
We evaluate the \textit{emergent ability} of \lakefm{} to comply with two physical laws of aquatic systems that it has not been trained for, as described in the following.  


\noindent \textbf{(1) Thermal Stratification Law.} A fundamental property of lakes is that during summer, lake temperature varies monotonically with depth, thus maintaining a vertical gradient with depth ($T_z >= T_{z+1}$). Deviations from this monotonic rule indicate inversion (which is physically inconsistent). We quantify this using the \textit{Inversion rate}, defined as the average number of depth-wise inversions($T_z < T_{z+1}$) per day. Lower inversion rate means higher physical consistency. 

\noindent \textbf{(2) Beer-Lambert Law} \cite{beer1852determination} states that light intensity decreases exponentially with depth due to biomass in the water column. We evaluate this relationship by computing the Pearson $R^2$ between predicted Chlorophyll-a and Light Attenuation (higher is better). 

Figure \ref{fig:lakefm_chronos_comparison} compares the inversion rate and $R^2$ values of \lakefm{} and Chronos2 over 100 unseen simulated lakes. We can see that \lakefm{} shows higher consistency with both physical laws than Chronos2 over a large majority of lakes (shaded green).
For additional comparisons of physical consistency, \camready{see the Appendix \ref{appendix_physical}}.


\subsection{Interpreting \lakefm{} Embeddings}
\label{sec:lake_emb_insight}

\begin{figure}[htbp]
    \centering
    \begin{subfigure}{0.42\textwidth}
        \includegraphics[width=\linewidth]{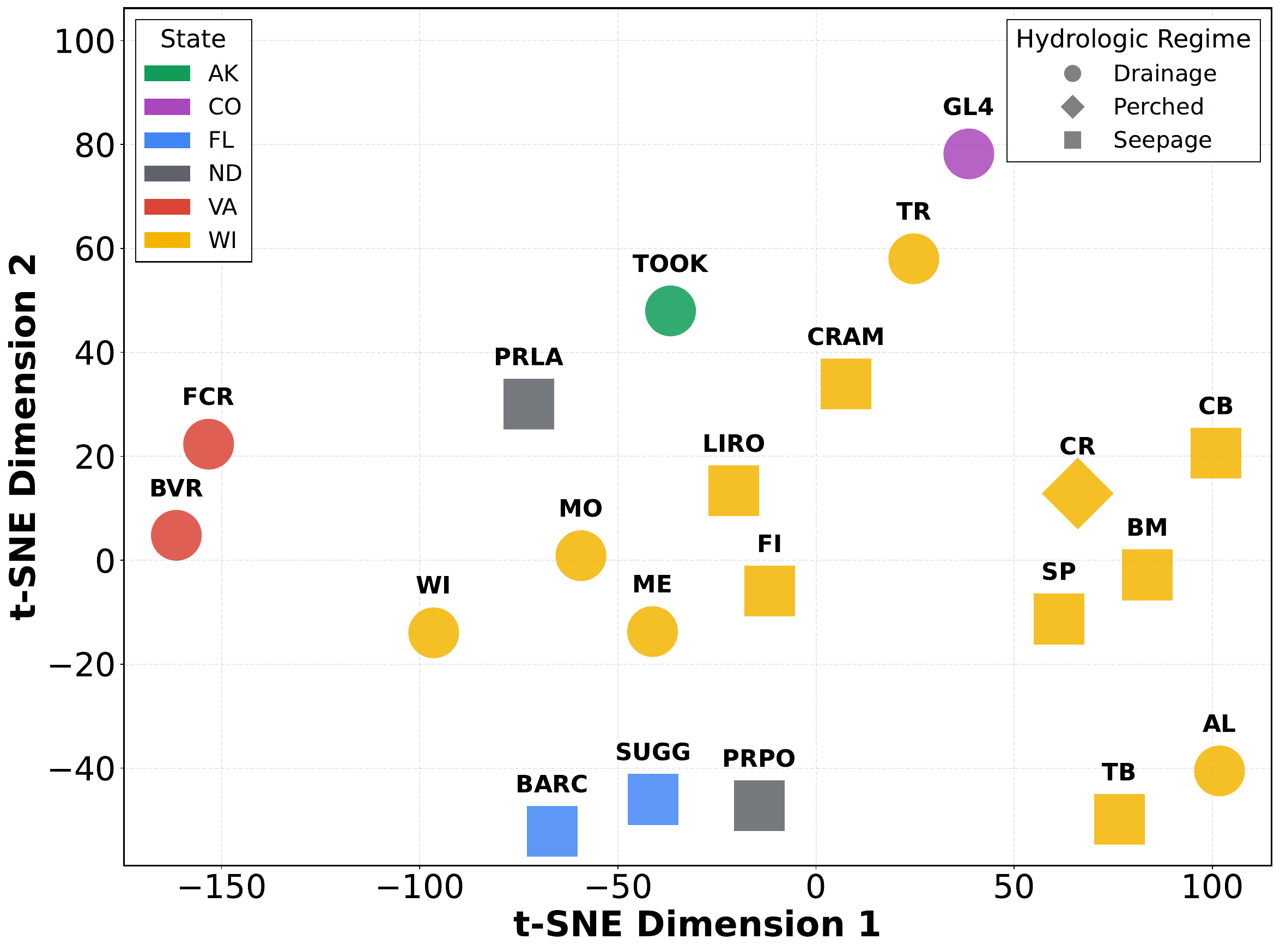}
        \end{subfigure}
    \caption{Static \lakefm{} embeddings of observed lakes categorized by location (State) and hydrologic regime.}
    \label{fig:lake_tsne_state_hydro}
\end{figure}

\begin{figure}[htbp]
    \centering
    \begin{subfigure}{0.43\textwidth}
        \includegraphics[width=\linewidth]{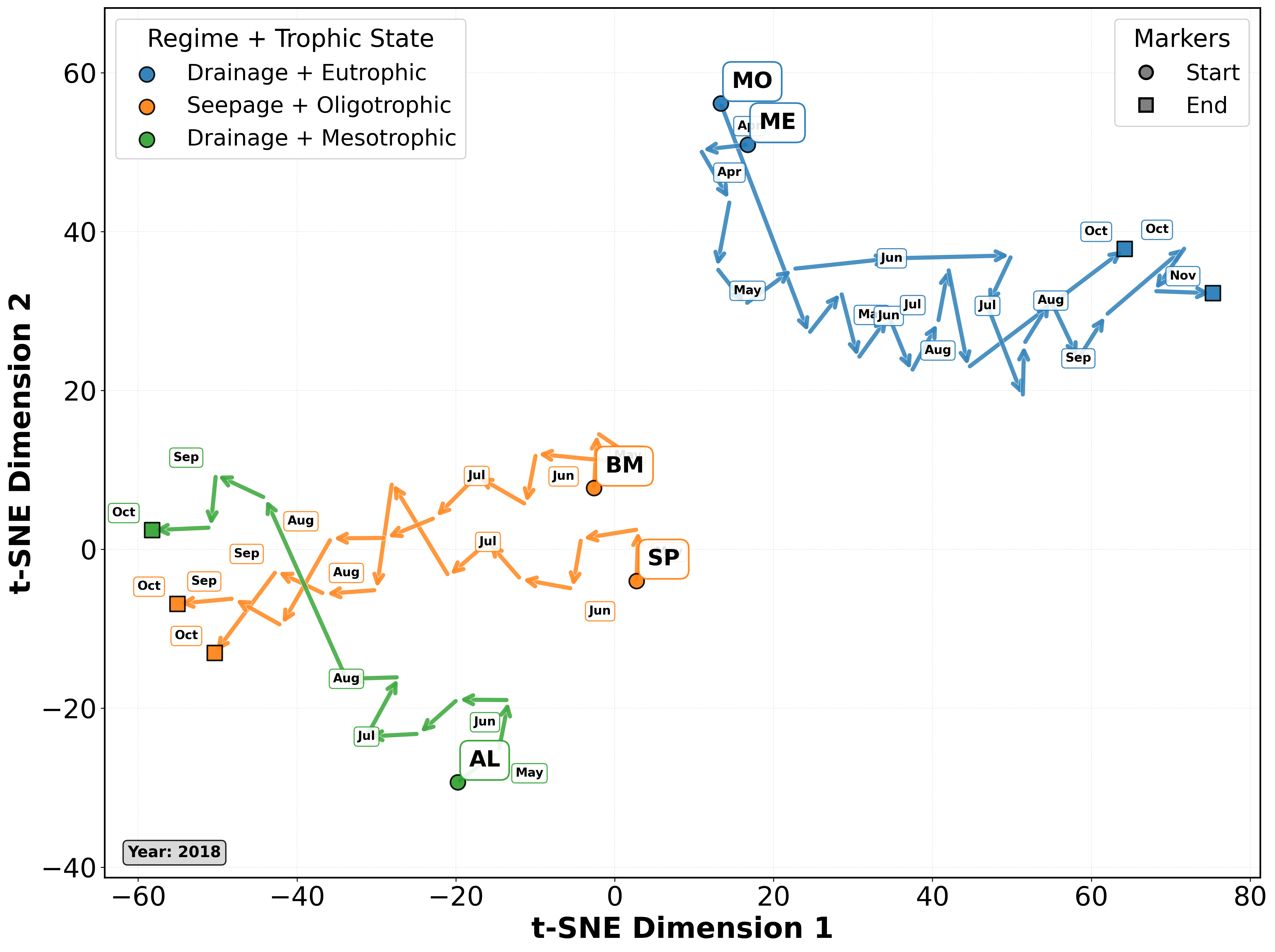}
    \end{subfigure}
    \caption{Trajectories of \lakefm{} Dynamic Embeddings for lakes AL, BM, SP, MO, and ME in 2018.}
    \label{fig:lakebed_tsne_multilake_2018}
\end{figure}

\noindent \textbf{Static Embeddings.} Figure \ref{fig:lake_tsne_state_hydro} shows the static lake-level embeddings learned by \lakefm{} (from its static projector) using 2D t-SNE.
We can see that lakes from similar geographic locations (US State) are closer to each other, with finer variations within lakes in each state determined by the hydrologic regimes of lakes. For example, for Wisconsin (WI) lakes, \lakefm{} is able to separate lakes with drainage (MO, ME, and WI) from those with seepage (SP, BM, CB). 
\camready{Figure \ref{fig:lake_tsne_static_trophic_hydro_app} in Appendix \ref{appendix:lake_embeddings_viz}} presents additional visualizations of \lakefm{} embeddings based on lake trophic state, that further helps to differentiate lakes such as AL (mesotrophic) from MO and ME (eutrophic).
Note that neither of these lake metadata were used in the training of \lakefm{}, demonstrating \lakefm{}'s ability to produce meaningful static embeddings aligned with known lake properties. 
\noindent \textbf{Time-varying Embeddings.} We examine the ability of \lakefm{} to produce dynamic embeddings of lakes (from its temporal projector) that help differentiate their temporal trajectories.
Figure~\ref{fig:lakebed_tsne_multilake_2018} shows the embedding trajectories of 5 lakes in Wisconsin for 2018 using 2D t-SNE, colored on the basis of hydrologic regime \& trophic state. 
We can see that both MO and ME (eutrophic lakes with drainage regime) exhibit closely aligned seasonal trajectories in the embedding space, while SP and BM, oligotrophic lakes with seepage-dominated regime, form a distinct group. While AL is geographically close to SP and BM, it is ecologically different in terms of hydrologic regime and trophic state, and thus follows a different trajectory than the other two. 
\camready{Figure \ref{fig:lakebed_tsne_multilake_2021} in Appendix \ref{appendix:lake_embeddings_viz}} includes an additional visualization for 2021, with similar observations,
suggesting that \lakefm{} jointly encodes time-invariant lake characteristics and time-varying ecological behavior.

\label{sec:inverse_modeling}
\begin{figure}[htbp]
\centering\includegraphics[width=0.45\textwidth]{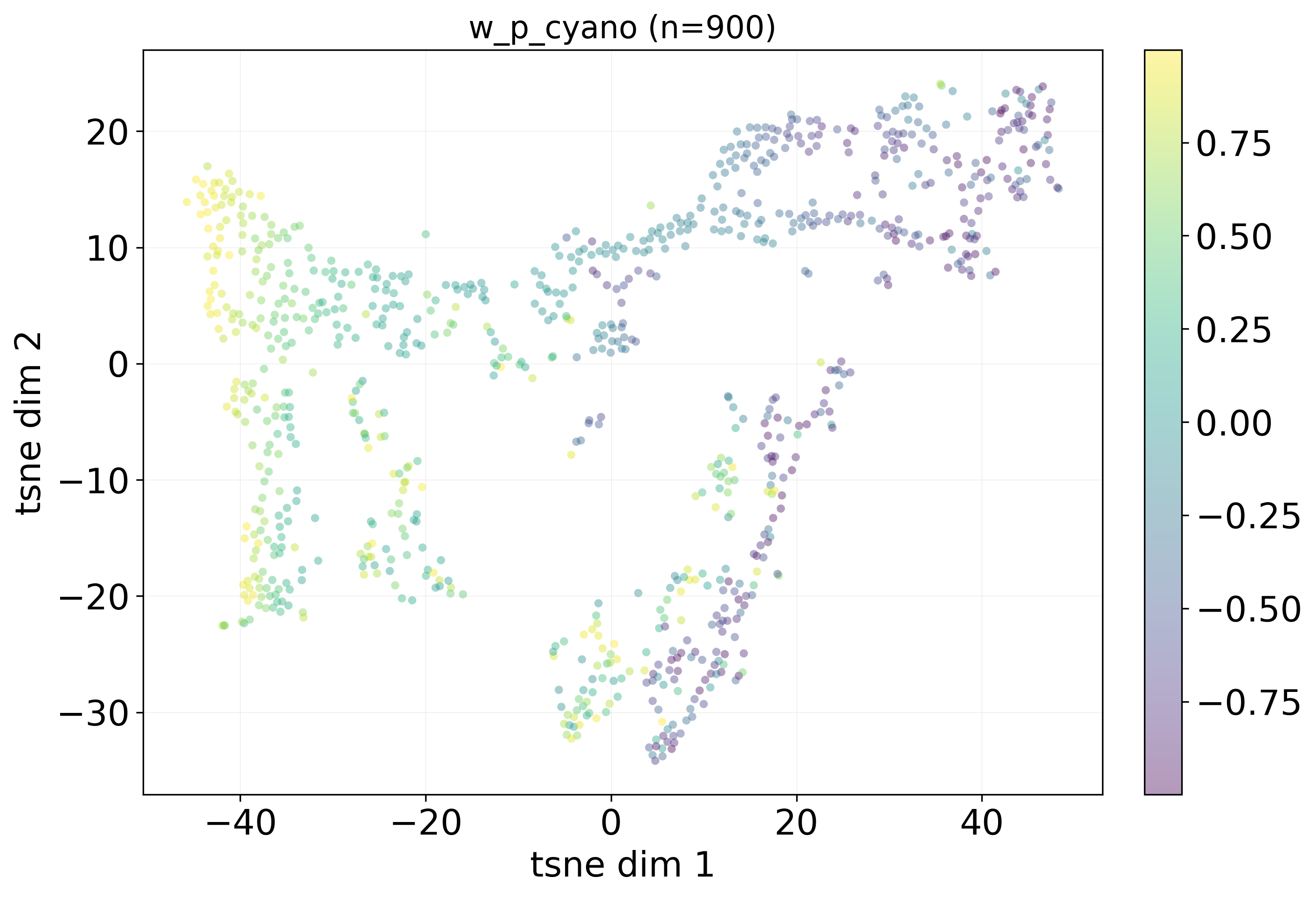}
    \caption{\lakefm{} representation for 900 unseen simulation lakes, each corresponding to a different cyanobacteria value.}
    \label{fig:lakefmgradients}
\end{figure}

\noindent \textbf{Embeddings of Simulated Lakes.} We investigate whether \lakefm{}’s embeddings of simulated lakes encode information of process-based parameters used to generate the simulations.
Figure~\ref{fig:lakefmgradients} shows the static embeddings of 900 unseen simulated lakes generated with varying input parameter configurations (\camready{see Appendix \ref{sec:appendix-fcr-simulations}} for simulation details), where we focus on the cyanobacteria-related parameter $w\_p\_cyano$ to color every point in the embedding space, which modulates phytoplankton dynamics. 
We can see a clear gradient across the embedding space with respect to this parameter, with clear separation of low, intermediate and high $w\_p\_cyano$ values. 
We further analyze the trajectory embeddings of simulated lakes in \camready{Appendix \ref{appendix_inverse_learning}}, showing consistent trends of temporal dynamics for lakes within the same range of $w\_p\_cyano$ values.



\subsection{Ablations}
\label{appendix_ablations}
\subsubsection{\camready{Model Ablations}}

\camready{Figure~\ref{fig:model_ablation} compares \textsc{LakeFM} against model ablations, evaluating the contribution of contrastive learning, variate-specific likelihoods, and probabilistic training in \textsc{LakeFM}.}

\noindent\textbf{\camready{Without Contrastive Loss.}} \camready{We remove the contrastive objective 
and train the model solely with the probabilistic forecasting loss. This ablation leads to a consistent degradation in performance across all held-out lakes, highlighting that enforcing the model to learn time-invariant lake representations help in \lakefm{}'s forecasting performance.}
\\
\camready{\textbf{Variate-specific Degrees of Freedom (DoF)}. We evaluate the impact of learning variate-wise DoF within the Student-$t$ distribution versus a shared or fixed DoF. Results show a strict decrease in performance without variate-specific parameterization. This confirms that different limnological variables (e.g., highly volatile Chlorophyll-a vs. stable Deep-water Temperature) possess distinct heavy-tail characteristics that require individualized distributional modeling.\\
\textbf{Student-$t$ vs. Gaussian Likelihood}. Replacing the Student-$t$ distribution with a standard Normal distribution resulted in significantly higher MSE. This degradation highlights that environmental time-series frequently violate normality assumptions; the Student-$t$ provides the necessary flexibility to handle the outliers and heteroskedasticity inherent in lake ecosystems.\\
\textbf{Probabilistic vs. Point-Estimation (MSE Loss)}. Training with a deterministic MSE loss (non-probabilistic) yielded poor performance across held-out lakes. This highlights the high degree of epistemic uncertainty in lake modeling. A deterministic approach fails to capture the uncertain nature of future states, whereas our probabilistic framework provides a more resilient objective for zero-shot transfer.\\
\textbf{Continuous Depth Embedding Ablation}. To isolate the impact of our continuous depth embedding, we conducted an ablation study where depth is flattened into discrete, independent variates (i.e., treating each unique depth-variable pair as a new variate). From Table \ref{tab:depth_embedding_ablation}, we observe that treating depth as discrete independent variates degrades performance. Continuous depth modeling provides a spatial coordinate system that allows the model to learn vertical gradients and generalize to unseen depths. Treating depth as independent categories makes this challenging and also leads to extremely large, sparse input matrices.}
\\

\begin{table}[h]
\centering
\caption{\camready{Comparing LakeFM with and without continuous-depth modeling (\textsc{W/o depth embed}) that treats each unique (depth, variable) pair as an independent discrete variate}}
\label{tab:depth_embedding_ablation}
\renewcommand{\arraystretch}{0.85}
\resizebox{0.9\columnwidth}{!}{%
\begin{tabular}{cc*{4}{c}}
\toprule
\textbf{Lake} & \textbf{Variant} & \textbf{WaterTemp\_C} & \textbf{Water\_DO\_mg\_per\_L} & \textbf{Chla\_ugL} & \textbf{par} \\
\midrule
\multirow{2}{*}{\textbf{ME}}
& \textsc{W/o depth embed} & 9.76 & 4.27 & -- & -- \\
& \textsc{LakeFM}     & \textbf{0.21} & \textbf{0.45} & -- & -- \\
\midrule
\multirow{2}{*}{\textbf{TR}}
& \textsc{W/o depth embed} & 9.34 & 10.49 & -- & 13.00 \\
& \textsc{LakeFM}     & \textbf{0.85} & \textbf{0.45} & -- & \textbf{0.97} \\
\midrule
\multirow{2}{*}{\textbf{BM}}
& \textsc{W/o depth embed} & 7.69 & 8.12 & \textbf{0.88} & 12.80 \\
& \textsc{LakeFM}     & \textbf{0.48} & \textbf{0.69} & 1.07 & \textbf{0.95} \\
\midrule
\multirow{2}{*}{\textbf{CRAM}}
& \textsc{W/o depth embed} & 6.895 & 1.035 & -- & -- \\
& \textsc{LakeFM}     & \textbf{0.62} & \textbf{0.52} & -- & -- \\
\bottomrule
\end{tabular}%
}
\end{table}




\begin{figure}[htbp]
    \centering
    \begin{subfigure}{0.4\textwidth}
        \centering
        \includegraphics[width=\textwidth]{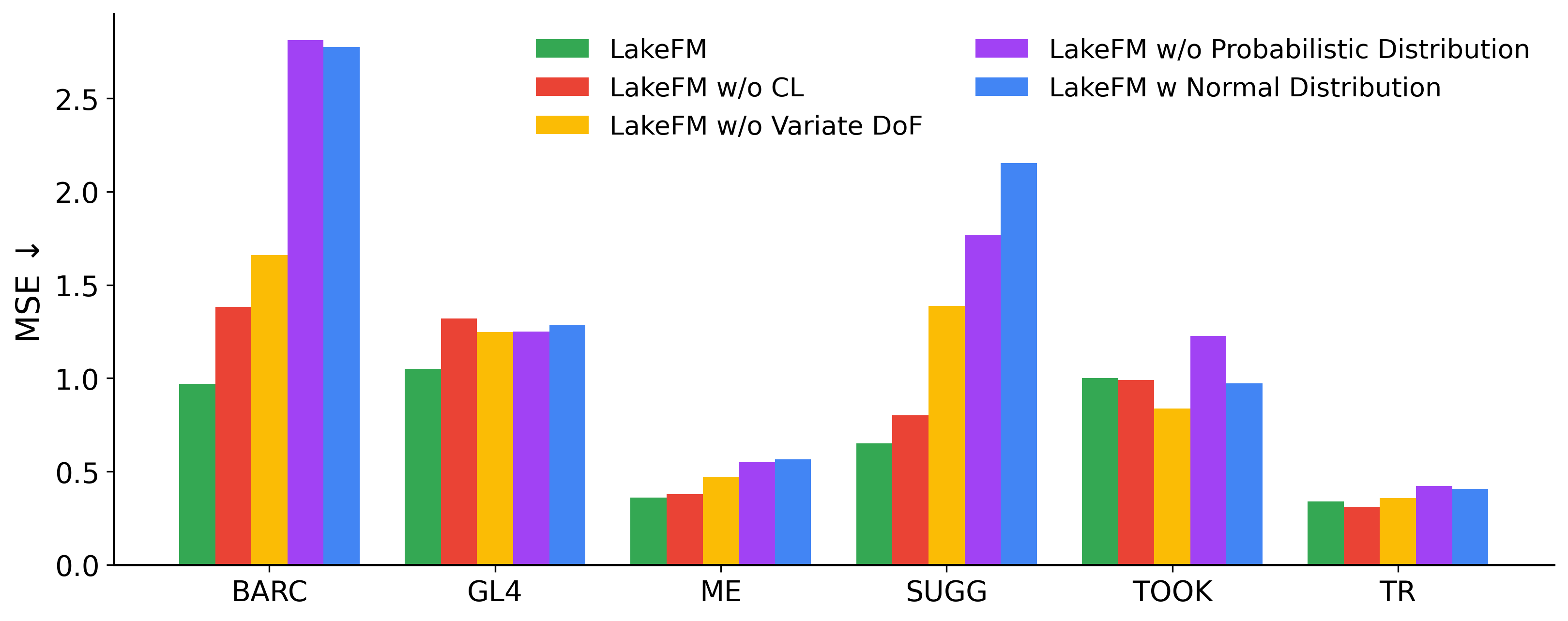}
        \caption{Model ablations}
        \label{fig:model_ablation}
    \end{subfigure}
    \hfill
    \begin{subfigure}{0.4\textwidth}
        \centering
        \includegraphics[width=\textwidth]{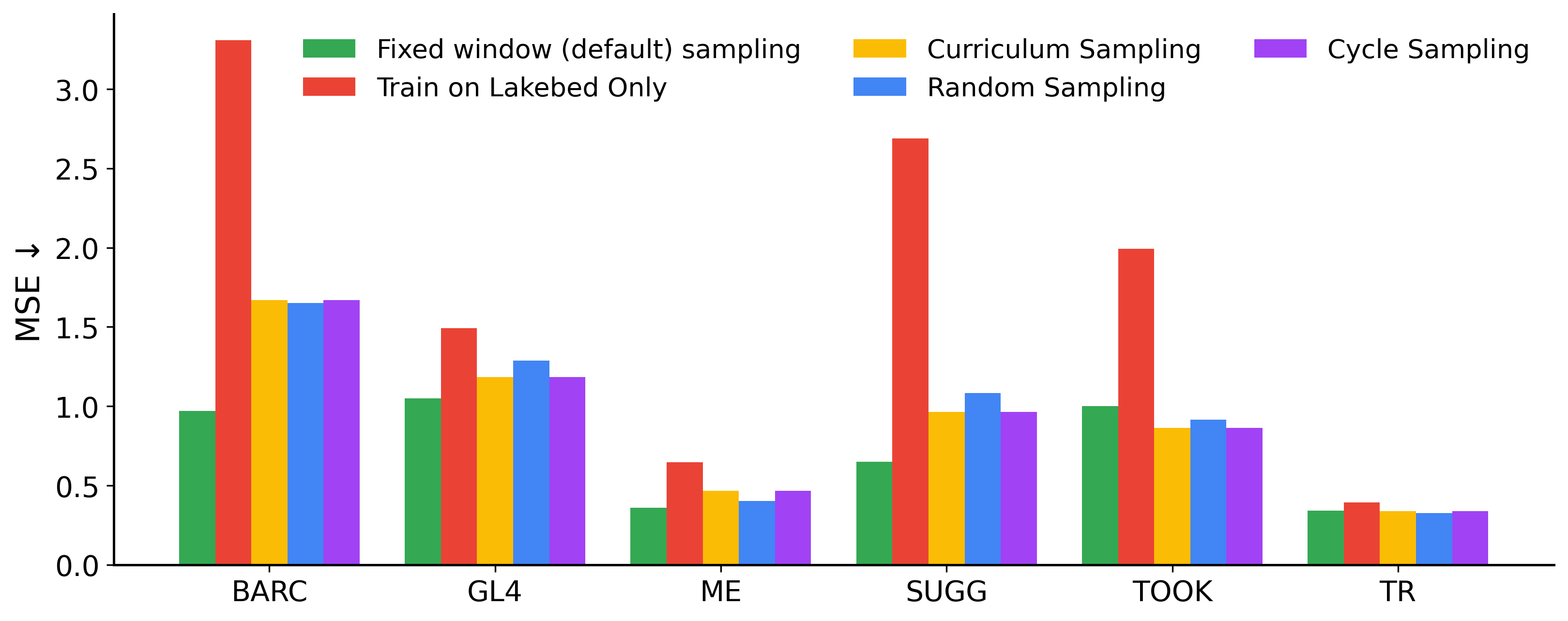}
        \caption{Training ablations}
        \label{fig:training_ablation}
    \end{subfigure}

    \caption{Model and training ablations evaluated using MSE across six held-out lakes}
    \label{fig:ood_ablations}
\end{figure}

\subsubsection{\camready{Training Ablations}}
\camready{Figure~\ref{fig:training_ablation} compares training strategy ablations, analyzing the impact of simulation-based pretraining and temporal sampling strategies on OOD generalization.}

\noindent\textbf{\camready{Impact of Simulation Data in Pre-training.}} \camready{We compare LakeFM trained exclusively on the LakeBeD (real-world) dataset against our standard pipeline which includes synthetic simulation data. The significantly lower generalizability of the real-only model indicates that pre-training on physically-grounded simulations is critical for learning a robust representation of lake dynamics that transfers to unseen basins.\\
\textbf{Sampling Strategies (Curriculum vs. Cycle vs. Random)}. We experimented with various window-sampling strategies during training: Curriculum (increasing window sizes), Cycle (alternating context/prediction lengths), and Random. We found that complex scheduling provided negligible benefits over standard random sampling, suggesting the model's robustness is driven more by data diversity than the order of temporal exposure.\\}

\vspace{-2ex}
\section{\camready{Computational Cost Analysis}}
\label{sec:compute_cost}

\camready{We report the computational cost of \lakefm{} and the baselines in Table~\ref{tab:compute_cost}, with measurements obtained on the ME dataset using a single H200 GPU and batch size 32.}


\begin{table}[htbp]
\centering
\caption{Computational cost comparison of \lakefm{} and baseline models, measured on the ME dataset.}
\label{tab:compute_cost}
\setlength{\tabcolsep}{5pt}
\renewcommand{\arraystretch}{1.15}
\small
\resizebox{\columnwidth}{!}{%
\begin{tabular}{lccccc}
\toprule
\textbf{Metric} & \textbf{Chronos-2} & \textbf{MOMENT} & \textbf{LPTM} & \textbf{MOIRAI} & \textbf{\lakefm{}} \\
\midrule
\textbf{Peak VRAM (GB)} & 0.5 & 2.0 & 0.9 & 5.0 & 4.2 \\
\textbf{Inf.\ throughput (samples/sec)} & 44.65 & 171.54 & 490.04 & 121.48 & 176.64 \\
\textbf{\# Params} & 120M & 341.28M & 109.73M & 310.97M & 7M \\
\bottomrule
\end{tabular}%
}
\end{table}

\noindent\camready{Treating each observation as a distinct token increases the input sequence length; however, the computational cost analysis shows that this is a highly parameter-efficient trade-off that enables the model to handle extreme irregularities that baselines cannot. \lakefm{} achieves competitive results using significantly fewer (7M) parameters than the general-purpose TSFMs. While peak VRAM is slightly higher than some baselines, it remains well within the limits of standard modern hardware, including low-end GPUs (e.g., NVIDIA V100). Moreover, \lakefm{}'s inference throughput is comparable to or better than several production-grade state-of-the-art FMs such as MOMENT, Chronos~2, and MOIRAI.}

\section{How does \lakefm{} Advance the Science of Aquatic Systems?}


The conventional approach of modeling lake systems is to use process-based models, which require expert calibration of lake-specific parameters using custom data from every lake, which is hard to obtain at operational scales. ML offers a completely different solution for this problem by training a model over a large collection of lakes that can be transferred to any lake without expert calibration of parameters. However, a major challenge in harnessing the generalization power of ML models across lakes is the irregularity in data,  a challenge that no other TSFM is able to address despite its wide prevalence across many ecological applications.

For the first time, LakeFM is enabling scalable knowledge transfer across a large collection of lakes (real and simulated), overcoming the challenges of sampling differences within variables across depth and time. \textit{\lakefm{} is a first-step toward macro-system understanding of lake ecology using diverse and heterogeneous lake data.}
Another advantage of LakeFM for aquatic sciences is its ability to work with masked variables, which neither process-based nor existing ML models in this domain are able to handle. The variable masking experiments of \lakefm{} are revealing novel insights about the interactions of variables in lakes that require further scientific investigation. They have the ability to inform which variables to collect and at what depths when working on new lakes, to maximize forecasting performance. Finally, the embedding visualizations are revealing novel interpretations of the static and dynamic characteristics of lakes, jointly accounting for changes in geography, hydrological regime, and trophic states.

\section{Conclusion}

We present \lakefm{}, a domain-specific foundation model for lake ecosystems developed through an inter-disciplinary collaboration between ML researchers and ecologists that is able to handle irregular multivariate, multi-depth time-series across diverse lake systems. By jointly modeling variable interactions and site-level dynamics, \lakefm{} enables reliable zero-shot cross-lake generalization and recovers physically and ecologically meaningful information. 
We hope our work inspires future research in building scientific foundation models that are tailored to the needs of application domains and are not only trained with supervision contained in data but also with the physical principles underlying scientific phenomena.

\section*{Limitations and Ethical Considerations}


To the best of our knowledge, this work poses no major ethical concerns, as it focuses on forecasting and representation learning from environmental time-series data. Potential limitations stem primarily from data quality and coverage. \camready{LakeFM’s reliability in environmental conditions that deviate significantly from the training distribution, e.g., regions with different thermal and ice-cover dynamics than US, remains a key boundary condition. Specifically, different climatic contexts can introduce variate scales that fall outside the observed training statistics. In such scenarios, the model's reliance on learned statistical dependencies and multi-variate correlations may limit its extrapolative accuracy}. 

\section*{GenAI Disclosure}

Generative AI tools were used to assist with editing and improving the readability of the manuscript, including grammar, phrasing, and presentation of the manuscript. All scientific content, experimental design, results, analysis, and conclusions were conceived, verified, and finalized by the authors.

\begin{acks}
\camready{We sincerely thank Mary E. Lofton from the Department of Biology, Virginia Tech for preparing and curating the FCR simulations (comprising 1000 simulation lake datasets) used in this study. This work was supported in part by NSF awards \#2213549, \#2213550, and \#2239328. We are also grateful to computing resources from Bridges-2 at Pittsburgh Supercomputing Center available through NAIRR pilot award \#240161. We are also grateful to the Advanced Research Computing (ARC) Center at Virginia Tech for providing access to GPU compute resources for this project. 
}
\end{acks}


\bibliographystyle{ACM-Reference-Format}
\balance
\bibliography{ref}


\begin{thebibliography}{32}


\ifx \showCODEN    \undefined \def \showCODEN     #1{\unskip}     \fi
\ifx \showISBNx    \undefined \def \showISBNx     #1{\unskip}     \fi
\ifx \showISBNxiii \undefined \def \showISBNxiii  #1{\unskip}     \fi
\ifx \showISSN     \undefined \def \showISSN      #1{\unskip}     \fi
\ifx \showLCCN     \undefined \def \showLCCN      #1{\unskip}     \fi
\ifx \shownote     \undefined \def \shownote      #1{#1}          \fi
\ifx \showarticletitle \undefined \def \showarticletitle #1{#1}   \fi
\ifx \showURL      \undefined \def \showURL       {\relax}        \fi
\providecommand\bibfield[2]{#2}
\providecommand\bibinfo[2]{#2}
\providecommand\natexlab[1]{#1}
\providecommand\showeprint[2][]{arXiv:#2}

\bibitem[Ansari et~al\mbox{.}(2025)]%
        {ansari2025chronos}
\bibfield{author}{\bibinfo{person}{Abdul~Fatir Ansari}, \bibinfo{person}{Oleksandr Shchur}, \bibinfo{person}{Jaris K{\"u}ken}, \bibinfo{person}{Andreas Auer}, \bibinfo{person}{Boran Han}, \bibinfo{person}{Pedro Mercado}, \bibinfo{person}{Syama~Sundar Rangapuram}, \bibinfo{person}{Huibin Shen}, \bibinfo{person}{Lorenzo Stella}, \bibinfo{person}{Xiyuan Zhang}, {et~al\mbox{.}}} \bibinfo{year}{2025}\natexlab{}.
\newblock \showarticletitle{Chronos-2: From univariate to universal forecasting}.
\newblock \bibinfo{journal}{\emph{arXiv preprint arXiv:2510.15821}} (\bibinfo{year}{2025}).
\newblock


\bibitem[Ansari et~al\mbox{.}(2024)]%
        {chronos}
\bibfield{author}{\bibinfo{person}{Abdul~Fatir Ansari}, \bibinfo{person}{Lorenzo Stella}, \bibinfo{person}{Caner Turkmen}, \bibinfo{person}{Xiyuan Zhang}, \bibinfo{person}{Pedro Mercado}, \bibinfo{person}{Huibin Shen}, \bibinfo{person}{Oleksandr Shchur}, \bibinfo{person}{Syama~Sundar Rangapuram}, \bibinfo{person}{Sebastian~Pineda Arango}, \bibinfo{person}{Shubham Kapoor}, {et~al\mbox{.}}} \bibinfo{year}{2024}\natexlab{}.
\newblock \showarticletitle{Chronos: Learning the language of time series}.
\newblock \bibinfo{journal}{\emph{arXiv preprint arXiv:2403.07815}} (\bibinfo{year}{2024}).
\newblock


\bibitem[Beer and Beer(1852)]%
        {beer1852determination}
\bibfield{author}{\bibinfo{person}{August Beer} {and} \bibinfo{person}{P Beer}.} \bibinfo{year}{1852}\natexlab{}.
\newblock \showarticletitle{Determination of the absorption of red light in colored liquids}.
\newblock \bibinfo{journal}{\emph{Annalen der Physik und Chemie}} \bibinfo{volume}{86}, \bibinfo{number}{5} (\bibinfo{year}{1852}), \bibinfo{pages}{78--88}.
\newblock


\bibitem[Chen et~al\mbox{.}(2023)]%
        {chen2023contiformer}
\bibfield{author}{\bibinfo{person}{Yuqi Chen}, \bibinfo{person}{Kan Ren}, \bibinfo{person}{Yansen Wang}, \bibinfo{person}{Yuchen Fang}, \bibinfo{person}{Weiwei Sun}, {and} \bibinfo{person}{Dongsheng Li}.} \bibinfo{year}{2023}\natexlab{}.
\newblock \showarticletitle{Contiformer: Continuous-time transformer for irregular time series modeling}.
\newblock \bibinfo{journal}{\emph{Advances in Neural Information Processing Systems}}  \bibinfo{volume}{36} (\bibinfo{year}{2023}), \bibinfo{pages}{47143--47175}.
\newblock


\bibitem[Cohen et~al\mbox{.}(2024)]%
        {cohen2024toto}
\bibfield{author}{\bibinfo{person}{Ben Cohen}, \bibinfo{person}{Emaad Khwaja}, \bibinfo{person}{Kan Wang}, \bibinfo{person}{Charles Masson}, \bibinfo{person}{Elise Ram{\'e}}, \bibinfo{person}{Youssef Doubli}, {and} \bibinfo{person}{Othmane Abou-Amal}.} \bibinfo{year}{2024}\natexlab{}.
\newblock \showarticletitle{Toto: Time series optimized transformer for observability}.
\newblock \bibinfo{journal}{\emph{arXiv preprint arXiv:2407.07874}} (\bibinfo{year}{2024}).
\newblock


\bibitem[Corman et~al\mbox{.}(2023)]%
        {corman2023high}
\bibfield{author}{\bibinfo{person}{Jessica Corman}, \bibinfo{person}{Jacob Zwart}, \bibinfo{person}{Jennifer Klug}, \bibinfo{person}{Denise Bruesewitz}, \bibinfo{person}{Elvira de Eyto}, \bibinfo{person}{Marcus Klaus}, \bibinfo{person}{Lesley Knoll}, \bibinfo{person}{James Rusak}, \bibinfo{person}{Michael Vanni}, \bibinfo{person}{Maria~Belen Alfonso}, {et~al\mbox{.}}} \bibinfo{year}{2023}\natexlab{}.
\newblock \showarticletitle{High-frequency dissolved oxygen, water temperature, wind speed, and radiation data; stream and in-lake nutrient concentration data; and daily metabolism and nutrient loading estimates for 16 lakes in North America and Northern Europe.}
\newblock  (\bibinfo{year}{2023}).
\newblock


\bibitem[Daw et~al\mbox{.}(2022)]%
        {daw2022physics}
\bibfield{author}{\bibinfo{person}{Arka Daw}, \bibinfo{person}{Anuj Karpatne}, \bibinfo{person}{William~D Watkins}, \bibinfo{person}{Jordan~S Read}, {and} \bibinfo{person}{Vipin Kumar}.} \bibinfo{year}{2022}\natexlab{}.
\newblock \showarticletitle{Physics-guided neural networks (pgnn): An application in lake temperature modeling}.
\newblock In \bibinfo{booktitle}{\emph{Knowledge guided machine learning}}. \bibinfo{publisher}{Chapman and Hall/CRC}, \bibinfo{pages}{353--372}.
\newblock


\bibitem[Du et~al\mbox{.}(2023)]%
        {du2023saits}
\bibfield{author}{\bibinfo{person}{Wenjie Du}, \bibinfo{person}{David C{\^o}t{\'e}}, {and} \bibinfo{person}{Yan Liu}.} \bibinfo{year}{2023}\natexlab{}.
\newblock \showarticletitle{Saits: Self-attention-based imputation for time series}.
\newblock \bibinfo{journal}{\emph{Expert Systems with Applications}}  \bibinfo{volume}{219} (\bibinfo{year}{2023}), \bibinfo{pages}{119619}.
\newblock


\bibitem[Goswami et~al\mbox{.}(2024)]%
        {goswami2024moment}
\bibfield{author}{\bibinfo{person}{Mononito Goswami}, \bibinfo{person}{Konrad Szafer}, \bibinfo{person}{Arjun Choudhry}, \bibinfo{person}{Yifu Cai}, \bibinfo{person}{Shuo Li}, {and} \bibinfo{person}{Artur Dubrawski}.} \bibinfo{year}{2024}\natexlab{}.
\newblock \showarticletitle{Moment: A family of open time-series foundation models}.
\newblock \bibinfo{journal}{\emph{arXiv preprint arXiv:2402.03885}} (\bibinfo{year}{2024}).
\newblock


\bibitem[Hanson et~al\mbox{.}(2023)]%
        {hanson2023}
\bibfield{author}{\bibinfo{person}{P.~C. Hanson}, \bibinfo{person}{R. Ladwig}, \bibinfo{person}{C. Buelo}, \bibinfo{person}{E.~A. Albright}, \bibinfo{person}{A.~D. Delany}, {and} \bibinfo{person}{C.~C. Carey}.} \bibinfo{year}{2023}\natexlab{}.
\newblock \showarticletitle{Legacy Phosphorus and Ecosystem Memory Control Future Water Quality in a Eutrophic Lake}.
\newblock \bibinfo{journal}{\emph{Journal of Geophysical Research: Biogeosciences}} \bibinfo{volume}{128}, \bibinfo{number}{12} (\bibinfo{year}{2023}), \bibinfo{pages}{e2023JG007620}.
\newblock
\href{https://doi.org/10.1029/2023JG007620}{doi:\nolinkurl{10.1029/2023JG007620}}


\bibitem[Hipsey et~al\mbox{.}(2019)]%
        {hipseyGLM}
\bibfield{author}{\bibinfo{person}{M.~R. Hipsey}, \bibinfo{person}{L.~C. Bruce}, \bibinfo{person}{C. Boon}, \bibinfo{person}{B. Busch}, \bibinfo{person}{C.~C. Carey}, \bibinfo{person}{D.~P. Hamilton}, \bibinfo{person}{P.~C. Hanson}, \bibinfo{person}{J.~S. Read}, \bibinfo{person}{E. de Sousa}, \bibinfo{person}{M. Weber}, {and} \bibinfo{person}{L.~A. Winslow}.} \bibinfo{year}{2019}\natexlab{}.
\newblock \showarticletitle{A General Lake Model (GLM 3.0) for linking with high-frequency sensor data from the Global Lake Ecological Observatory Network (GLEON)}.
\newblock \bibinfo{journal}{\emph{Geoscientific Model Development}} \bibinfo{volume}{12}, \bibinfo{number}{1} (\bibinfo{year}{2019}), \bibinfo{pages}{473--523}.
\newblock


\bibitem[Jia et~al\mbox{.}(2019)]%
        {physicsrnn}
\bibfield{author}{\bibinfo{person}{Xiaowei Jia}, \bibinfo{person}{Jared Willard}, \bibinfo{person}{Anuj Karpatne}, \bibinfo{person}{Jordan Read}, \bibinfo{person}{Jacob Zwart}, \bibinfo{person}{Michael Steinbach}, {and} \bibinfo{person}{Vipin Kumar}.} \bibinfo{year}{2019}\natexlab{}.
\newblock \showarticletitle{Physics guided RNNs for modeling dynamical systems: A case study in simulating lake temperature profiles}.
\newblock  (\bibinfo{year}{2019}), \bibinfo{pages}{558--566}.
\newblock


\bibitem[Ladwig et~al\mbox{.}(2024)]%
        {ladwig2024modular}
\bibfield{author}{\bibinfo{person}{Robert Ladwig}, \bibinfo{person}{Arka Daw}, \bibinfo{person}{Elen~A Albright}, \bibinfo{person}{Cal Buelo}, \bibinfo{person}{Anuj Karpatne}, \bibinfo{person}{Michael~Frederick Meyer}, \bibinfo{person}{Abhilash Neog}, \bibinfo{person}{Paul~C Hanson}, {and} \bibinfo{person}{Hilary~A Dugan}.} \bibinfo{year}{2024}\natexlab{}.
\newblock \showarticletitle{Modular Compositional Learning Improves 1D Hydrodynamic Lake Model Performance by Merging Process-Based Modeling With Deep Learning}.
\newblock \bibinfo{journal}{\emph{Journal of Advances in Modeling Earth Systems}} \bibinfo{volume}{16}, \bibinfo{number}{1} (\bibinfo{year}{2024}), \bibinfo{pages}{e2023MS003953}.
\newblock


\bibitem[Langman et~al\mbox{.}(2010)]%
        {langman2010control}
\bibfield{author}{\bibinfo{person}{OC Langman}, \bibinfo{person}{PC Hanson}, \bibinfo{person}{SR Carpenter}, {and} \bibinfo{person}{YH Hu}.} \bibinfo{year}{2010}\natexlab{}.
\newblock \showarticletitle{Control of dissolved oxygen in northern temperate lakes over scales ranging from minutes to days}.
\newblock \bibinfo{journal}{\emph{Aquatic Biology}} \bibinfo{volume}{9}, \bibinfo{number}{2} (\bibinfo{year}{2010}), \bibinfo{pages}{193--202}.
\newblock


\bibitem[Li et~al\mbox{.}(2026)]%
        {reimts}
\bibfield{author}{\bibinfo{person}{Boyuan Li}, \bibinfo{person}{Zhen Liu}, \bibinfo{person}{Yicheng Luo}, {and} \bibinfo{person}{Qianli Ma}.} \bibinfo{year}{2026}\natexlab{}.
\newblock \showarticletitle{Learning Recursive Multi-Scale Representations for Irregular Multivariate Time Series Forecasting}.
\newblock \bibinfo{journal}{\emph{arXiv preprint arXiv:2602.21498}} (\bibinfo{year}{2026}).
\newblock


\bibitem[Li et~al\mbox{.}(2025)]%
        {hyperimts}
\bibfield{author}{\bibinfo{person}{Boyuan Li}, \bibinfo{person}{Yicheng Luo}, \bibinfo{person}{Zhen Liu}, \bibinfo{person}{Junhao Zheng}, \bibinfo{person}{Jianming Lv}, {and} \bibinfo{person}{Qianli Ma}.} \bibinfo{year}{2025}\natexlab{}.
\newblock \showarticletitle{Hyperimts: Hypergraph neural network for irregular multivariate time series forecasting}.
\newblock \bibinfo{journal}{\emph{arXiv preprint arXiv:2505.17431}} (\bibinfo{year}{2025}).
\newblock


\bibitem[Liu et~al\mbox{.}(2023)]%
        {liu2023itransformer}
\bibfield{author}{\bibinfo{person}{Yong Liu}, \bibinfo{person}{Tengge Hu}, \bibinfo{person}{Haoran Zhang}, \bibinfo{person}{Haixu Wu}, \bibinfo{person}{Shiyu Wang}, \bibinfo{person}{Lintao Ma}, {and} \bibinfo{person}{Mingsheng Long}.} \bibinfo{year}{2023}\natexlab{}.
\newblock \showarticletitle{itransformer: Inverted transformers are effective for time series forecasting}.
\newblock \bibinfo{journal}{\emph{arXiv preprint arXiv:2310.06625}} (\bibinfo{year}{2023}).
\newblock


\bibitem[McAfee et~al\mbox{.}(2025)]%
        {mcafee2025lakebed}
\bibfield{author}{\bibinfo{person}{Bennett~J McAfee}, \bibinfo{person}{Aanish Pradhan}, \bibinfo{person}{Abhilash Neog}, \bibinfo{person}{Sepideh Fatemi}, \bibinfo{person}{Robert~T Hensley}, \bibinfo{person}{Mary~E Lofton}, \bibinfo{person}{Anuj Karpatne}, \bibinfo{person}{Cayelan~C Carey}, {and} \bibinfo{person}{Paul~C Hanson}.} \bibinfo{year}{2025}\natexlab{}.
\newblock \showarticletitle{LakeBeD-US: a benchmark dataset for lake water quality time series and vertical profiles}.
\newblock \bibinfo{journal}{\emph{Earth System Science Data}} \bibinfo{volume}{17}, \bibinfo{number}{7} (\bibinfo{year}{2025}), \bibinfo{pages}{3141--3165}.
\newblock


\bibitem[Neog et~al\mbox{.}(2026)]%
        {neog2025masking}
\bibfield{author}{\bibinfo{person}{Abhilash Neog}, \bibinfo{person}{Arka Daw}, \bibinfo{person}{Sepideh Fatemi}, \bibinfo{person}{Medha Sawhney}, \bibinfo{person}{Aanish Pradhan}, \bibinfo{person}{Mary~E Lofton}, \bibinfo{person}{Bennett~J McAfee}, \bibinfo{person}{Adrienne Breef-Pilz}, \bibinfo{person}{Heather~L Wander}, \bibinfo{person}{Dexter~W Howard}, {et~al\mbox{.}}} \bibinfo{year}{2026}\natexlab{}.
\newblock \showarticletitle{Investigating a Model-Agnostic and Imputation-Free Approach for Irregularly-Sampled Multivariate Time-Series Modeling}.
\newblock \bibinfo{journal}{\emph{Transactions on Machine Learning Research}} (\bibinfo{year}{2026}).
\newblock


\bibitem[Nie et~al\mbox{.}(2022)]%
        {nie2022time}
\bibfield{author}{\bibinfo{person}{Yuqi Nie}, \bibinfo{person}{Nam~H Nguyen}, \bibinfo{person}{Phanwadee Sinthong}, {and} \bibinfo{person}{Jayant Kalagnanam}.} \bibinfo{year}{2022}\natexlab{}.
\newblock \showarticletitle{A time series is worth 64 words: Long-term forecasting with transformers}.
\newblock \bibinfo{journal}{\emph{arXiv preprint arXiv:2211.14730}} (\bibinfo{year}{2022}).
\newblock


\bibitem[Prabhakar~Kamarthi and Prakash(2024)]%
        {prabhakar2024large}
\bibfield{author}{\bibinfo{person}{Harshavardhan Prabhakar~Kamarthi} {and} \bibinfo{person}{B~Aditya Prakash}.} \bibinfo{year}{2024}\natexlab{}.
\newblock \showarticletitle{Large Pre-trained time series models for cross-domain Time series analysis tasks}.
\newblock \bibinfo{journal}{\emph{Advances in Neural Information Processing Systems}}  \bibinfo{volume}{37} (\bibinfo{year}{2024}), \bibinfo{pages}{56190--56214}.
\newblock


\bibitem[Pradhan et~al\mbox{.}(2024)]%
        {pradhan_lakebed}
\bibfield{author}{\bibinfo{person}{Aanish Pradhan}, \bibinfo{person}{Bennett~J. McAfee}, \bibinfo{person}{Abhilash Neog}, \bibinfo{person}{Sepideh Fatemi}, \bibinfo{person}{Mary~E. Lofton}, \bibinfo{person}{Cayelan~C. Carey}, \bibinfo{person}{Anuj Karpatne}, {and} \bibinfo{person}{Paul~C. Hanson}.} \bibinfo{year}{2024}\natexlab{}.
\newblock \bibinfo{title}{{LakeBeD}-{US}: {Computer} {Science} {Edition} - a benchmark dataset for lake water quality time series and vertical profiles}.
\newblock
\href{https://doi.org/10.57967/hf/3771}{doi:\nolinkurl{10.57967/hf/3771}}


\bibitem[Shi et~al\mbox{.}(2024)]%
        {shi2024time}
\bibfield{author}{\bibinfo{person}{Xiaoming Shi}, \bibinfo{person}{Shiyu Wang}, \bibinfo{person}{Yuqi Nie}, \bibinfo{person}{Dianqi Li}, \bibinfo{person}{Zhou Ye}, \bibinfo{person}{Qingsong Wen}, {and} \bibinfo{person}{Ming Jin}.} \bibinfo{year}{2024}\natexlab{}.
\newblock \showarticletitle{Time-moe: Billion-scale time series foundation models with mixture of experts}.
\newblock \bibinfo{journal}{\emph{arXiv preprint arXiv:2409.16040}} (\bibinfo{year}{2024}).
\newblock


\bibitem[Shukla and Marlin(2021)]%
        {shukla2021multi}
\bibfield{author}{\bibinfo{person}{Satya~Narayan Shukla} {and} \bibinfo{person}{Benjamin~M Marlin}.} \bibinfo{year}{2021}\natexlab{}.
\newblock \showarticletitle{Multi-time attention networks for irregularly sampled time series}.
\newblock \bibinfo{journal}{\emph{arXiv preprint arXiv:2101.10318}} (\bibinfo{year}{2021}).
\newblock


\bibitem[Staehr et~al\mbox{.}(2010)]%
        {staehr2010lake}
\bibfield{author}{\bibinfo{person}{Peter~A Staehr}, \bibinfo{person}{Darren Bade}, \bibinfo{person}{Matthew~C Van~de Bogert}, \bibinfo{person}{Gregory~R Koch}, \bibinfo{person}{Craig Williamson}, \bibinfo{person}{Paul Hanson}, \bibinfo{person}{Jonathan~J Cole}, {and} \bibinfo{person}{Tim Kratz}.} \bibinfo{year}{2010}\natexlab{}.
\newblock \showarticletitle{Lake metabolism and the diel oxygen technique: state of the science}.
\newblock \bibinfo{journal}{\emph{Limnology and Oceanography: Methods}} \bibinfo{volume}{8}, \bibinfo{number}{11} (\bibinfo{year}{2010}), \bibinfo{pages}{628--644}.
\newblock


\bibitem[Su et~al\mbox{.}(2024)]%
        {su2024roformer}
\bibfield{author}{\bibinfo{person}{Jianlin Su}, \bibinfo{person}{Murtadha Ahmed}, \bibinfo{person}{Yu Lu}, \bibinfo{person}{Shengfeng Pan}, \bibinfo{person}{Wen Bo}, {and} \bibinfo{person}{Yunfeng Liu}.} \bibinfo{year}{2024}\natexlab{}.
\newblock \showarticletitle{Roformer: Enhanced transformer with rotary position embedding}.
\newblock \bibinfo{journal}{\emph{Neurocomputing}}  \bibinfo{volume}{568} (\bibinfo{year}{2024}), \bibinfo{pages}{127063}.
\newblock


\bibitem[Tashiro et~al\mbox{.}(2021)]%
        {tashiro2021csdi}
\bibfield{author}{\bibinfo{person}{Yusuke Tashiro}, \bibinfo{person}{Jiaming Song}, \bibinfo{person}{Yang Song}, {and} \bibinfo{person}{Stefano Ermon}.} \bibinfo{year}{2021}\natexlab{}.
\newblock \showarticletitle{Csdi: Conditional score-based diffusion models for probabilistic time series imputation}.
\newblock \bibinfo{journal}{\emph{Advances in neural information processing systems}}  \bibinfo{volume}{34} (\bibinfo{year}{2021}), \bibinfo{pages}{24804--24816}.
\newblock


\bibitem[Willard et~al\mbox{.}(2021)]%
        {willard2021predicting}
\bibfield{author}{\bibinfo{person}{Jared~D Willard}, \bibinfo{person}{Jordan~S Read}, \bibinfo{person}{Alison~P Appling}, \bibinfo{person}{Samantha~K Oliver}, \bibinfo{person}{Xiaowei Jia}, {and} \bibinfo{person}{Vipin Kumar}.} \bibinfo{year}{2021}\natexlab{}.
\newblock \showarticletitle{Predicting water temperature dynamics of unmonitored lakes with meta-transfer learning}.
\newblock \bibinfo{journal}{\emph{Water Resources Research}} \bibinfo{volume}{57}, \bibinfo{number}{7} (\bibinfo{year}{2021}), \bibinfo{pages}{e2021WR029579}.
\newblock


\bibitem[Willard et~al\mbox{.}(2022)]%
        {willard2022daily}
\bibfield{author}{\bibinfo{person}{Jared~D Willard}, \bibinfo{person}{Jordan~S Read}, \bibinfo{person}{Simon Topp}, \bibinfo{person}{Gretchen~JA Hansen}, {and} \bibinfo{person}{Vipin Kumar}.} \bibinfo{year}{2022}\natexlab{}.
\newblock \showarticletitle{Daily surface temperatures for 185,549 lakes in the conterminous United States estimated using deep learning (1980--2020)}.
\newblock \bibinfo{journal}{\emph{Limnology and Oceanography Letters}} \bibinfo{volume}{7}, \bibinfo{number}{4} (\bibinfo{year}{2022}), \bibinfo{pages}{287--301}.
\newblock


\bibitem[Woo et~al\mbox{.}(2024)]%
        {moirai}
\bibfield{author}{\bibinfo{person}{G Woo}, \bibinfo{person}{C Liu}, \bibinfo{person}{A Kumar}, \bibinfo{person}{C Xiong}, \bibinfo{person}{S Savarese}, {and} \bibinfo{person}{D Sahoo}.} \bibinfo{year}{2024}\natexlab{}.
\newblock \showarticletitle{Unified training of universal time series forecasting transformers. arXiv 2024}.
\newblock \bibinfo{journal}{\emph{arXiv preprint arXiv:2402.02592}} (\bibinfo{year}{2024}).
\newblock


\bibitem[Xia et~al\mbox{.}(2012)]%
        {xia_nldas_2012}
\bibfield{author}{\bibinfo{person}{Youlong Xia}, \bibinfo{person}{Kenneth Mitchell}, \bibinfo{person}{Michael Ek}, \bibinfo{person}{Justin Sheffield}, \bibinfo{person}{Brian Cosgrove}, \bibinfo{person}{Eric Wood}, \bibinfo{person}{Lifeng Luo}, \bibinfo{person}{Charles Alonge}, \bibinfo{person}{Helin Wei}, \bibinfo{person}{Jesse Meng}, \bibinfo{person}{Ben Livneh}, \bibinfo{person}{Dennis Lettenmaier}, \bibinfo{person}{Victor Koren}, \bibinfo{person}{Qingyun Duan}, \bibinfo{person}{Kingtse Mo}, \bibinfo{person}{Yun Fan}, {and} \bibinfo{person}{David Mocko}.} \bibinfo{year}{2012}\natexlab{}.
\newblock \showarticletitle{Continental-scale water and energy flux analysis and validation for the {North} {American} {Land} {Data} {Assimilation} {System} project phase 2 ({NLDAS}-2): 1. {Intercomparison} and application of model products}.
\newblock \bibinfo{journal}{\emph{Journal of Geophysical Research: Atmospheres}} \bibinfo{volume}{117}, \bibinfo{number}{D3} (\bibinfo{year}{2012}).
\newblock
\showISSN{2156-2202}
\href{https://doi.org/10.1029/2011JD016048}{doi:\nolinkurl{10.1029/2011JD016048}}


\bibitem[Yu et~al\mbox{.}(2025)]%
        {physics_fm_aqua}
\bibfield{author}{\bibinfo{person}{Runlong Yu}, \bibinfo{person}{Chonghao Qiu}, \bibinfo{person}{Robert Ladwig}, \bibinfo{person}{Paul Hanson}, \bibinfo{person}{Yiqun Xie}, {and} \bibinfo{person}{Xiaowei Jia}.} \bibinfo{year}{2025}\natexlab{}.
\newblock \showarticletitle{Physics-Guided Foundation Model for Scientific Discovery: An Application to Aquatic Science}.
\newblock \bibinfo{journal}{\emph{arXiv preprint arXiv:2502.06084}} (\bibinfo{year}{2025}).
\newblock


\end{thebibliography}
\appendix
\onecolumn
\section{Dataset Details}
\label{sec:dataset}


We pretrain and evaluate \textsc{LakeFM} on three datasets (spanning over 530+ million observations) that together span both real-world (21 observed lakes) and process-based simulation datasets (1000+ diverse lake simulations). Each dataset contributes unique strengths to the modeling framework, as described below.
\subsection{LakeBeD-US}
\label{sec:lakebedus}

Our primary observational dataset is LakeBeD-US \cite{mcafee2025lakebed, pradhan_lakebed}, consisting of over 500 million unique lake water quality observations collected between 1981 and 2024. The data span 21 U.S. lakes and include both high- and low-frequency measurements. In this work, we utilize only the low-frequency measurements. The dataset features 17 variables organized into three categories: (1) static attributes, such as lake morphology and geographic location; (2) one-dimensional (1D) variables that vary over time (e.g., Secchi depth, inflow); and (3) two-dimensional (2D) variables that vary over both time and depth. This rich observational dataset captures diverse temporal and spatial lake dynamics.

\textbf{In-Distribution vs. Out-of-Distribution}
\label{app:lakes_id_ood_split}To evaluate generalization beyond the training distribution, we split ther LakeBeD-US lakes into in-distribution (ID) and Out of distribution (OOD) groups using a feature-based notion of lake similarity. Each lake is represented by a vector of geographic and physical attributes (latitude, longitude, surface area, mean depth, maximum depth, and elevation). Since these attributes have different units and scales, we standardize all features to zero mean and unit variance before computing similarities.

We then apply Principal Component Analysis (PCA) to obtain a compact representation capturing the dominant modes of variability across lakes, and perform $k$-means clustering in this normalized feature space. The number of clusters is selected using the silhouette score, favoring clusterings that are both tight and well-separated. 
To define an OOD subset, we select a small set of lakes that are maximally dissimilar from the bulk of the dataset in the learned PCA space (i.e., located in sparse or well-separated regions relative to cluster structure). The remaining lakes, which lie in denser regions of the feature space, are treated as ID. Figure \ref{fig:lake-similarity-pca} displays the results of the PCA with the proportion of variance explained by each principal component. Table \ref{tab:lakes_metadata_split} shows the splitting of the lakes as well as information about ecological state of the lake.

\begin{figure}[h]
    \centering
    \includegraphics[keepaspectratio, width = 0.5\textwidth]{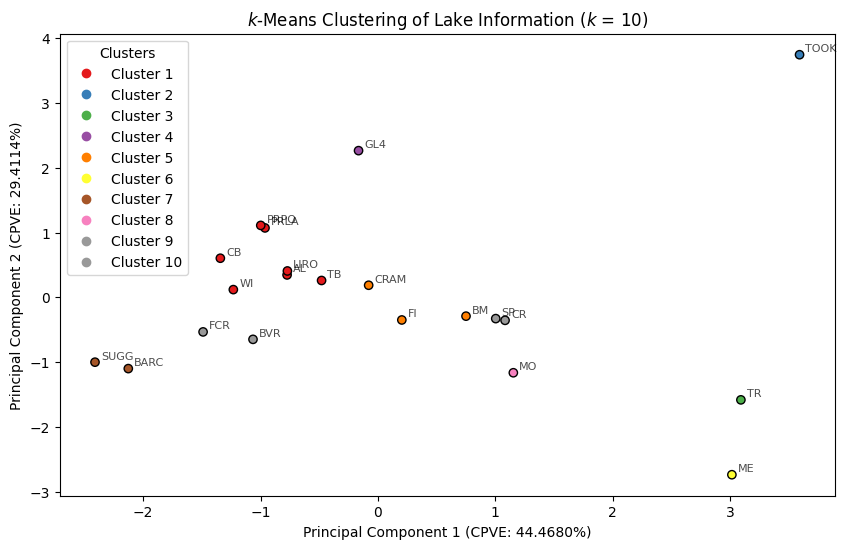}
    \caption{PCA of lake similarity with clusters obtained by $k$-means clustering.}
    \label{fig:lake-similarity-pca}
\end{figure}

\begin{table}[h]
\centering
\caption{List of lakes used in the study, split into OOD and ID groups, with locations, hydrology types, and trophic states.}
\label{tab:lakes_metadata_split}
\setlength{\tabcolsep}{4pt}
\renewcommand{\arraystretch}{1.1}
\small
\begin{tabular}{l l l l l l}
\toprule
\textbf{Group} & \textbf{Lake Name} & \textbf{Abbr.} & \textbf{Location} & \textbf{Hydrology} & \textbf{Trophic State} \\
\midrule
\multirow{6}{*}{\textbf{OOD Lakes}} 
& Lake Barco & BARC & Putnam County, FL, USA & Seepage & Oligotrophic \\
& Green Lake 4 & GL4 & Boulder County, CO, USA & Drainage & Oligotrophic \\
& Lake Mendota & ME & Dane County, WI, USA & Drainage & Eutrophic \\
& Lake Suggs & SUGG & Putnam County, FL, USA & Seepage & Mesotrophic \\
& Toolik Lake & TOOK & North Slope Borough, AK, USA & Drainage & Oligotrophic \\
& Trout Lake & TR & Vilas County, WI, USA & Drainage & Oligotrophic \\
\midrule
\multirow{15}{*}{\textbf{ID Lakes}} 
& Allequash Lake & AL & Vilas County, WI, USA & Drainage & Mesotrophic \\
& Big Muskellunge Lake & BM & Vilas County, WI, USA & Seepage & Oligotrophic \\
& Beaverdam Reservoir & BVR & Roanoke County, VA, USA & Drainage & Meso-Eutrophic \\
& Crystal Bog & CB & Vilas County, WI, USA & Seepage & Dystrophic \\
& Crystal Lake & CR & Vilas County, WI, USA & Perched & Oligotrophic \\
& Crampton Lake & CRAM & Vilas County, WI, USA & Seepage & Oligotrophic \\
& Falling Creek Reservoir & FCR & Roanoke County, VA, USA & Drainage & Eutrophic \\
& Fish Lake & FI & Dane County, WI, USA & Seepage & Mesotrophic \\
& Little Rock Lake & LIRO & Vilas County, WI, USA & Seepage & Mesotrophic \\
& Lake Monona & MO & Dane County, WI, USA & Drainage & Eutrophic \\
& Prairie Lake & PRLA & Stutsman County, ND, USA & Seepage & Dystrophic \\
& Prairie Pothole & PRPO & Stutsman County, ND, USA & Seepage & Dystrophic \\
& Sparkling Lake & SP & Vilas County, WI, USA & Seepage & Oligotrophic \\
& Lake Suggs & TB & Vilas County, WI, USA & Seepage & Dystrophic \\
& Lake Wingra & WI & Dane County, WI, USA & Drainage & Eutrophic \\
\bottomrule
\end{tabular}
\end{table}

\begin{table}[!htbp]
\caption{Variable names and corresponding water quality description}
\centering
\begin{tabular}{ll}
\toprule
Variable name & Water Quality Variable \\
\midrule
WaterTemp\_C & Water temperature \\
SRP\_ugL & Soluble reactive phosphorus \\
DIN\_ugL & Dissolved inorganic nitrogen \\
LightAttenuation\_Kd & Light attenuation coefficient \\
Chla\_ugL & Chlorophyll-a \\
AirTemp\_C & Air temperature \\
Shortwave\_Wm2 & Shortwave (solar) radiation \\
Inflow\_cms & Inflow \\
sum\_Longwave\_Radiation\_Downwelling\_wattPerMeterSquared & Downwelling longwave radiation \\
median\_Ten\_Meter\_Elevation\_Wind\_Speed\_meterPerSecond & Wind speed at 10-meter elevation \\
sum\_Precipitation\_millimeterPerDay & Precipitation \\
TOC\_load\_g\_per\_d & Total organic carbon load \\
TP\_load\_g\_per\_d & Total phosphorus load \\
Water\_DO\_mg\_per\_L & Dissolved oxygen \\
Water\_DOC\_mg\_per\_L & Dissolved organic carbon \\
Water\_POC\_mg\_per\_L & Particulate organic carbon \\
Water\_TP\_mg\_per\_L & Total phosphorus \\
Water\_Secchi\_m & Secchi depth \\
Discharge\_m3\_per\_d & Discharge \\
par & Photosynthetically active radiation \\
nh4 & Ammonium \\
tn & Total nitrogen \\
no3no2 & Nitrate and nitrite \\
\bottomrule
\end{tabular}
\label{tab:variable_names}
\end{table}
\subsection{FCR Simulations}
\label{sec:appendix-fcr-simulations}
The FCR Simulation datasets were generated using the General Lake Model coupled with the AED water quality module \citep[GLM-AED;][]{hipseyGLM}, and comprises 1,000 process-based model runs at Falling Creek Reservoir (FCR), VA, spanning daily resolution from December 1, 2016, to December 31, 2020. Each run represents a distinct ecological scenario defined by a unique set of phytoplankton trait parameters, sampled using Latin hypercube sampling. Six parameters were varied across three phytoplankton groups-cyanobacteria, green algae, and diatoms—including group-specific growth rates and sinking rates.
Model outputs include five key water quality variables: water temperature, soluble reactive phosphorus (SRP), dissolved inorganic nitrogen (DIN), chlorophyll-a (Chla), and the light attenuation coefficient (Kd). These are reported at seven depths (0.1, 1.6, 3.8, 5, 6.2, 8, and 9 m), corresponding to observational depths in FCR. Additionally, meteorological driver variables (e.g., AirTemp, Shortwave, Inflow) are included. Each row represents a specific date and depth, enabling detailed analysis of how phytoplankton trait variation influences ecosystem dynamics, particularly nutrient-light-temperature interactions and emergent biogeochemical patterns.

\subsection{WQHanson Simulations}
The WQHansonSim dataset is a set of lake water quality simulation datasets covering four lakes: Green Lake, Lake Mendota, Prairie Lake, and Trout Lake. The synthetic data were created using a process-based water quality model \cite{hanson2023} driven by meteorological forcing data from the second phase of the North American Land Data Assimilation System \citep[NLDAS-2;][]{xia_nldas_2012}. Each simulation underwent a 60-year burn-in period to allow slow-changing ecosystem states to reach dynamic equilibrium, followed by a 20-year simulation period. The outputs are structured as daily time series, with each row representing a unique date-depth combination. 

Each record includes six core water quality variables: water temperature, dissolved oxygen, dissolved organic carbon, particulate organic carbon, total phosphorus, and depth, alongside the corresponding date. Depths are lake-specific and selected to reflect stratification layers, representing both the epilimnion and hypolimnion (e.g., 5 m and 23 m for Trout Lake)—allowing for realistic modeling of thermal and chemical compositions among layers of the lake.

\begin{table}[t]
\centering
\caption{Overview of available lake variables (2D) for each lake across all datasets that forms the vocabulary of \textsc{LakeFM}. 
}
\small
\setlength{\tabcolsep}{6pt}
\renewcommand{\arraystretch}{1.15}
\begin{tabular}{lll}
\toprule
\textbf{Dataset} & \textbf{Lake ID} & \textbf{Available Variables} \\
\midrule
LakeBedUS & AL   & Water\_DO, WaterTemp, Chla, par, Secchi \\
LakeBedUS & BVR  & Water\_DO, Water\_DOC, SRP, Water\_TP, WaterTemp, Secchi \\
LakeBedUS & CRAM & Water\_DO, WaterTemp, Secchi \\
LakeBedUS & FI   & Water\_DO, WaterTemp, Secchi \\
LakeBedUS & MO   & Water\_DO, WaterTemp, Secchi \\
LakeBedUS & BARC & Water\_DO, WaterTemp, Secchi \\
LakeBedUS & BM   & Water\_DO, WaterTemp, Chla, par, Secchi \\
LakeBedUS & CB   & Water\_DO, WaterTemp, Chla, par, Secchi \\
LakeBedUS & CR   & Water\_DO, WaterTemp, Chla, par, Secchi \\
LakeBedUS & FCR  & Water\_DO, Water\_DOC, SRP, Water\_TP, WaterTemp, Secchi \\
LakeBedUS & GL4  & Water\_DO, WaterTemp, no3no2, par, Secchi \\
LakeBedUS & LIRO & Water\_DO, WaterTemp, Secchi \\
LakeBedUS & ME   & Water\_DO, WaterTemp, Secchi \\
LakeBedUS & PRLA & Water\_DO, WaterTemp, Secchi \\
LakeBedUS & PRPO & Water\_DO, WaterTemp, Secchi \\
LakeBedUS & SP   & Water\_DO, WaterTemp, Chla, par, Secchi \\
LakeBedUS & SUGG & Water\_DO, WaterTemp, Secchi \\
LakeBedUS & TB   & Water\_DO, WaterTemp, Chla, par, Secchi \\
LakeBedUS & TOOK & Water\_DO, WaterTemp, Inflow, Secchi \\
LakeBedUS & TR   & Water\_DO, WaterTemp, par, Secchi \\
LakeBedUS & WI   & Water\_DO, WaterTemp, Secchi \\
\midrule
WQHansonSim & All &
Water\_DO, Water\_DOC, Water\_POC, Water\_TP, Secchi, AirTemp, Shortwave \\
\midrule
FcrSimPhy & All &
Chla, WaterTemp, DIN\_ugL, LightAttenuation\_Kd, Inflow, AirTemp, Shortwave \\
\bottomrule
\end{tabular}
\label{tab:variables_per_lake}
\end{table}

\section{Implementation Details}
\label{appendix:lakefm_implementaion_details}

\subsection{\camready{LakeFM}} 
\camready{LakeFM uses a 12-layer Transformer encoder with 4 attention heads and hidden size $d_{\text{model}}=128$. Each scalar token combines learned variable (48-d), depth (32-d), and value (96-d) embeddings, along with a sinusoidal time embedding (16-d). The encoder uses RoPE for relative positional information, pre-norm Transformer blocks, SwiGLU activations, and dropout rates of 0.1 globally, 0.05 for attention, and 0.05 for heads. The contrastive projection head uses attention pooling with projection dimension 128. Forecasting is performed with a probabilistic decoder head that outputs Student-$t$ parameters, with scale and degrees of freedom constrained using softplus and clamping. We train with an Adam-style optimizer using learning rate $5\times10^{-5}$, weight decay $4\times10^{-4}$, gradient clipping at 1.0, and cosine learning-rate scheduling without warmup.}

\noindent\textbf{Hyper\-parameter tuning} We perform hyper-parameter sweeps involving the following parameters: \textit{enc\_layers}, \textit{num\_heads}, \textit{weight\_decay}, \textit{embed\_dim}, \textit{attention\_dropout}, \textit{head\_dropout}, \textit{variate\_embed\_dim}, \textit{depth\_embed\_dim}, \textit{contrastive\_loss\_weight}. 

 \noindent\textbf{Contrastive Sampling Strategy.} We adopt a custom balanced sampling strategy, built on top of PyTorch’s \texttt{DistributedSampler}, to construct batches for contrastive pretraining. Each batch consists of multiple anchor-positive groups, where each anchor is paired with \texttt{P\_pos = 4} positive samples from the same lake. For e.g., for a total \texttt{batch\_size = 64}, this allows up to 12 such anchor-positive sets per batch, with the remaining slots filled by negative samples drawn from different lakes. Positive and negative pools are precomputed per lake for efficiency, and sampling is performed with deterministic seeding to support reproducibility across distributed processes. This sampling strategy ensures within-lake similarity and across-lake contrast, enabling the model to learn lake-discriminative representations.

\noindent\textbf{Hardware}. We use a combination of NVIDIA H100 and A100 GPUs for pretraining and carrying out the experiments

\subsection{Baselines}\label{sec:baseline_details}


\camready{We use the Samay Time-series Foundational Models Library for foundation model baselines~\cite{prabhakar2024large} and the official iTransformer implementation. Foundation models are evaluated zero-shot without lake-specific fine-tuning. For non-foundation baselines, including iTransformer and IMTS models, ID evaluation uses the same 7:1:2 split as \lakefm{}. For OOD evaluation, these non-foundation baselines are trained on 60\% of each OOD lake, and all models, including zero-shot foundation models, are evaluated on the same remaining 40\%. Separate iTransformer, ReIMTS, and HyperIMTS models are trained for each lake. All evaluations use a 30-day context and 14-day prediction horizon. For baselines that cannot directly handle missing values, irregular observations are linearly interpolated along time. For cross-model comparisons, we report normalized metrics by re-standardizing the denormalized model predictions using the same ground-truth statistics across all models.}




\section{Additional Results}
\subsection{Physical Consistency Experiments} \label{appendix_physical}
We evaluate iTransformer and compare against \lakefm{} for the Physical consistency experiments. We hold out the data from years 2017 and 2018 as training data (for iTransformer) and evaluate the models on 2019 and 2020 data (same evaluation split as used for Chronos 2 comparison). Figure \ref{fig:lakefm_itrans_comparison} shows that \lakefm{} outperforms iTransformer in 97\% of the cases in the $R^2$ comparison, and in 98\% of the cases in the vertical thermal stratification experiment.
\begin{figure}[htbp]
    \centering
    \begin{minipage}{0.48\textwidth}
        \centering
        \includegraphics[width=0.71\linewidth]{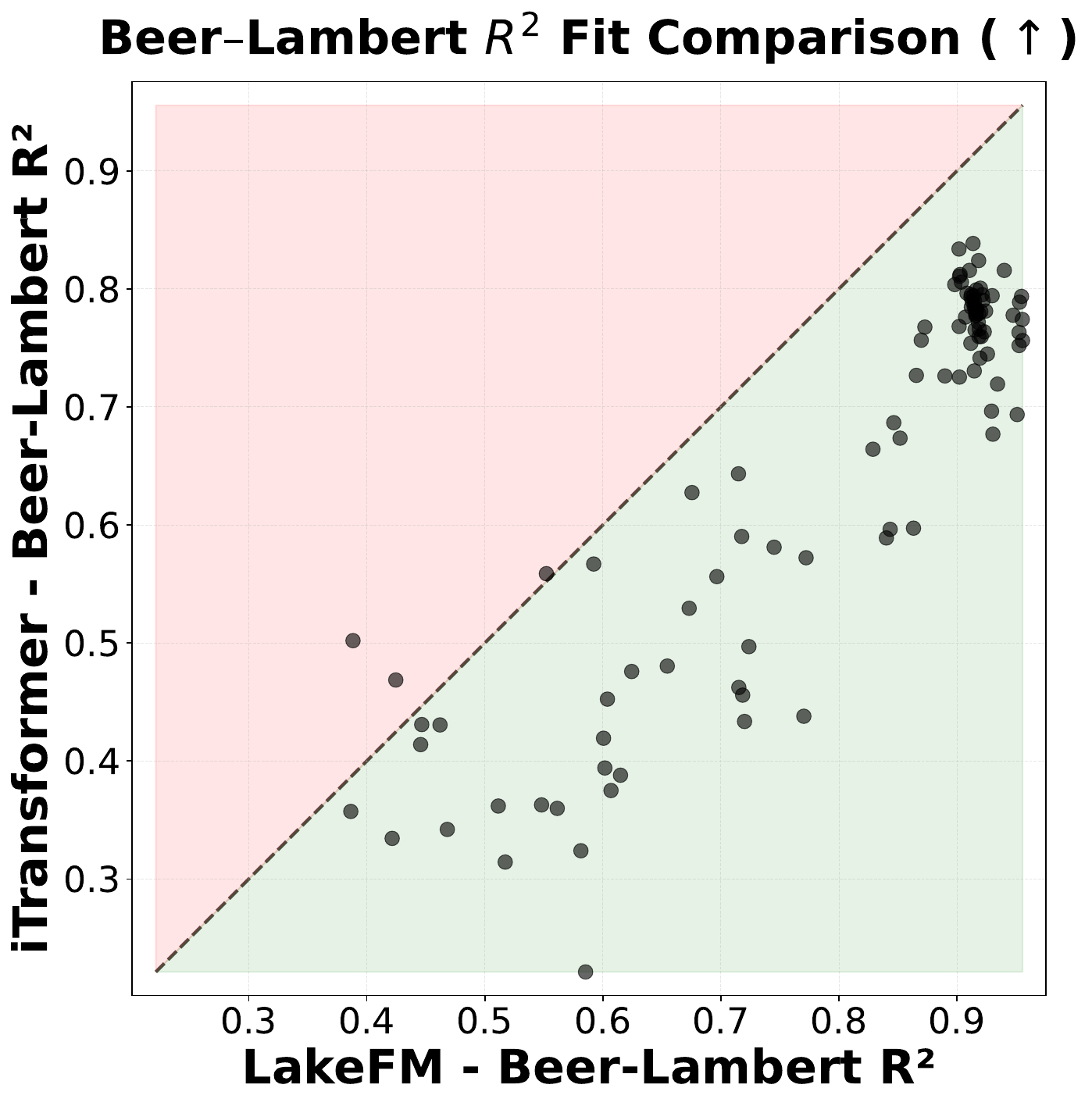}
        \caption*{(a) Pearson $R^2$ comparison for LakeFM and iTransformer}
        \label{fig:bl_lakefm_itrans}
    \end{minipage}
    \hfill
    \begin{minipage}{0.48\textwidth}
        \centering
        \includegraphics[width=0.7\linewidth]{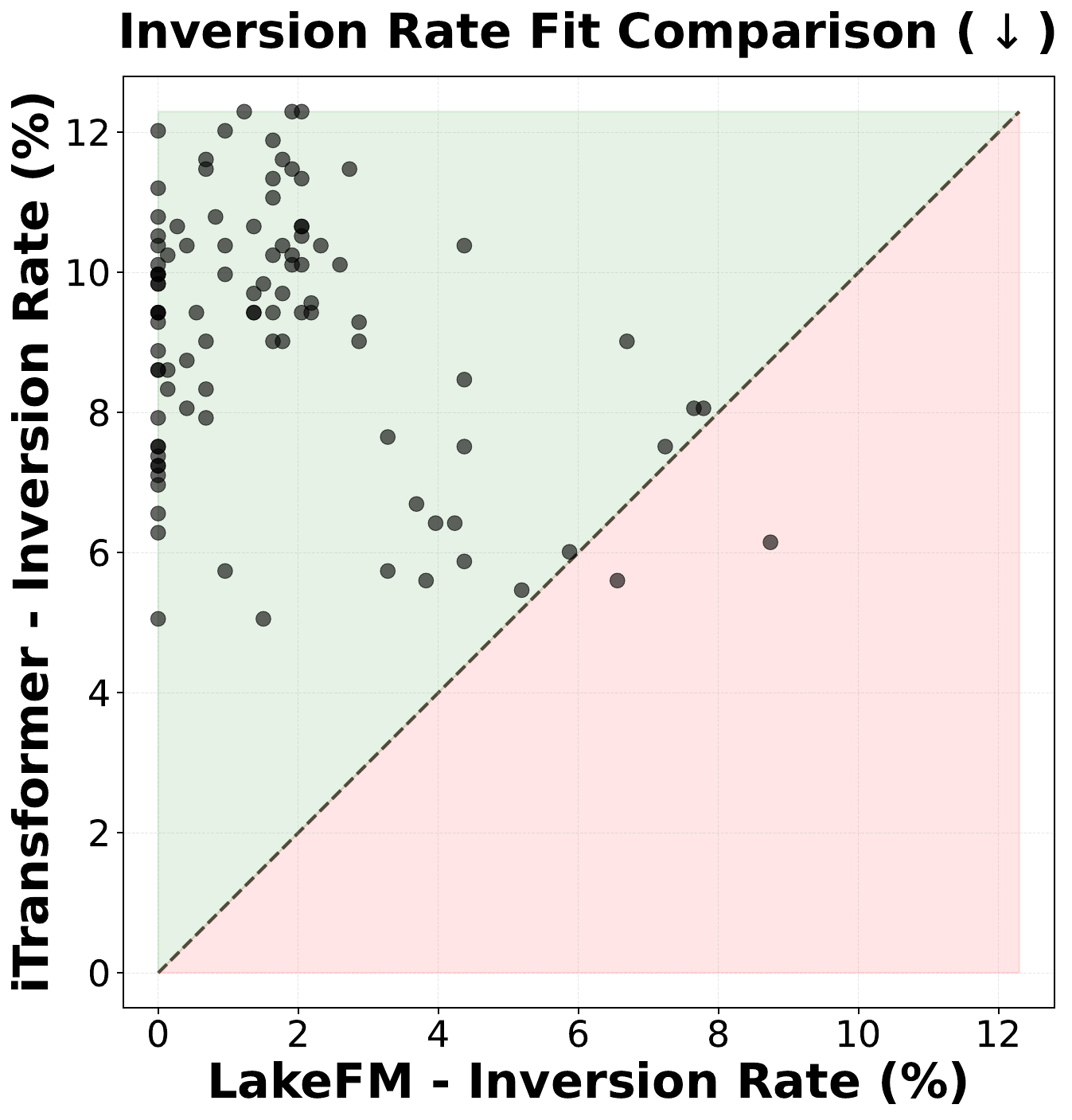}
        \caption*{(b) Inversion rate comparison for LakeFM and iTransformer.}
        \label{fig:inv_lakefm_itrans}
    \end{minipage}
    \caption{Comparison of LakeFM and iTransformer on the Beer-Lambert Law and Vertical stratification tests, evaluated across 100 simulation lake datasets.}
    \label{fig:lakefm_itrans_comparison}
\end{figure}

\subsection{Insights from Time-varying Lake Embeddings}
\label{appendix:lake_embeddings_viz}
Figure \ref{fig:lakebed_tsne_multilake_2021} shows trends in 2021 that mirror those observed in 2018 (Figure \ref{fig:lakebed_tsne_multilake_2018}). In both years, the lake pairs (ME, MO) and (SP, BM) exhibit closely aligned trajectories within each pair, while remaining well separated in the embedding space, reflecting differences in hydrologic regime, trophic state, and regional context. Consistent with 2018, AL's trajectory lies near SP and BM due to geographic proximity, yet remains distinct, consistent with differences in trophic state.
\label{appendix_lake_embed_insight}

\begin{figure}[!htbp]
    \centering
    \begin{minipage}{0.48\textwidth}
        \centering
        \includegraphics[width=0.96\linewidth]{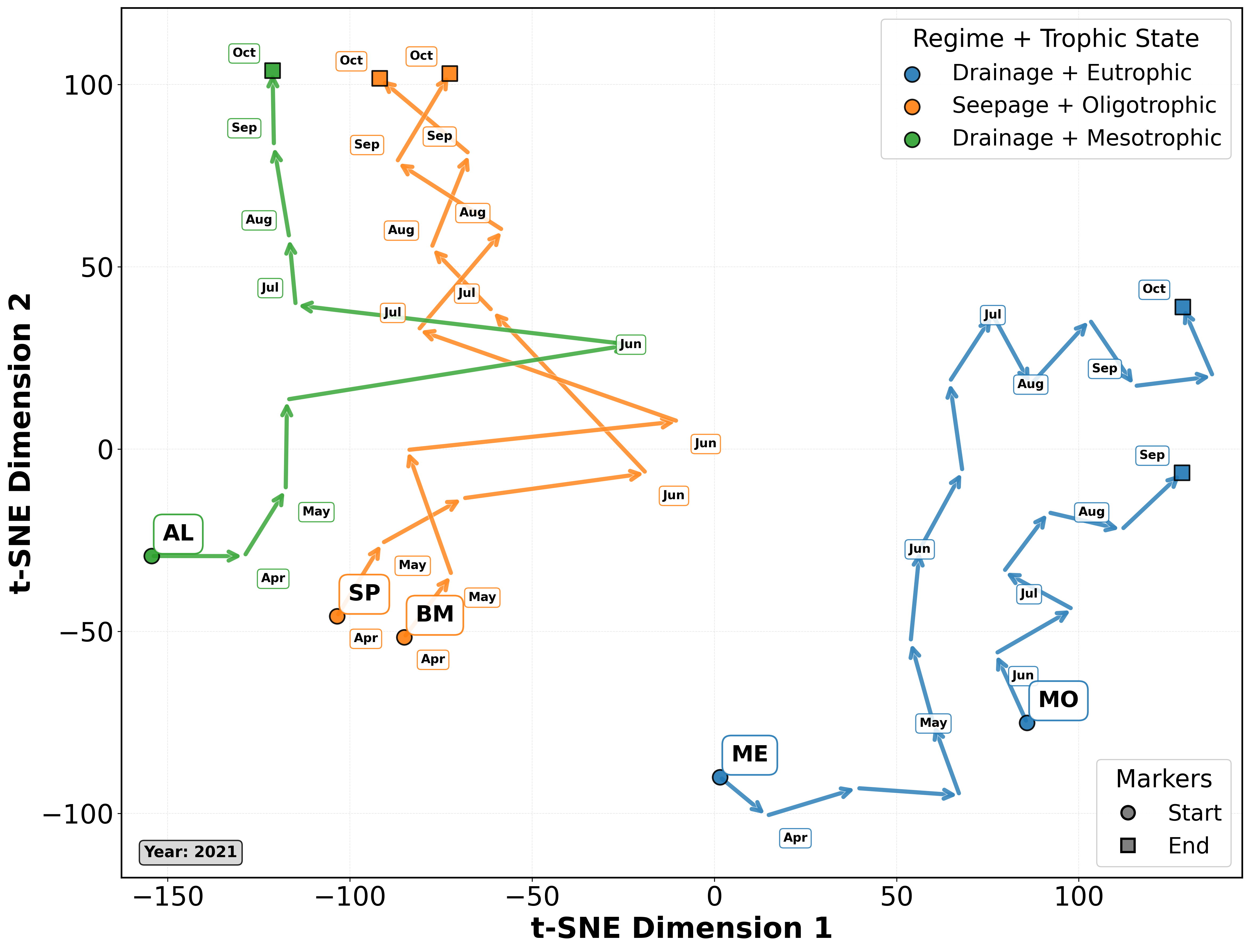}
        \caption{Lake Embedding trajectories comparing the combination of the hydrologic and trophic states in 2021. }
        \label{fig:lakebed_tsne_multilake_2021}
    \end{minipage}
    \hfill 
    \begin{minipage}{0.5\textwidth}
        \centering
        \includegraphics[width=0.96\linewidth]{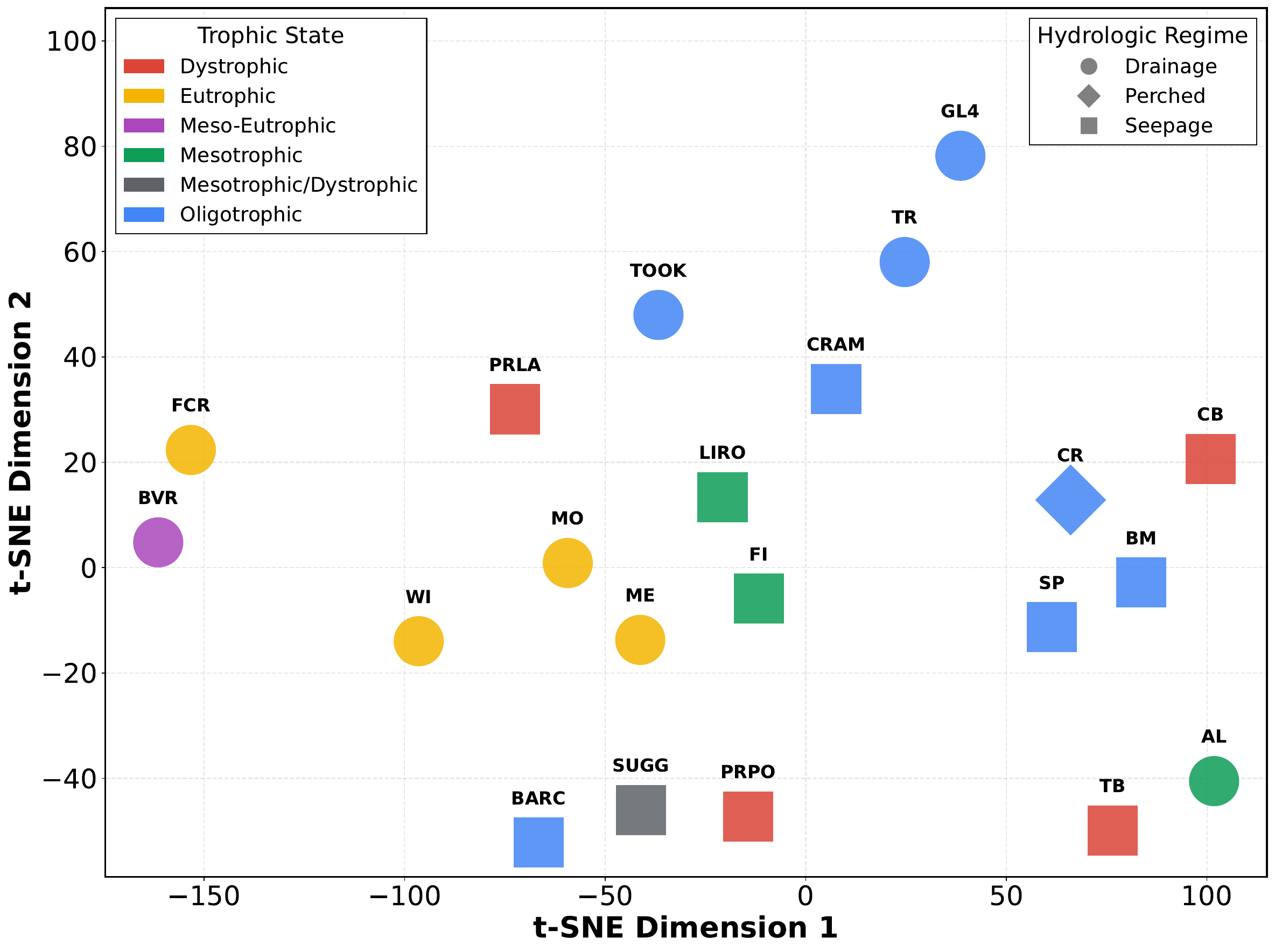}
        \caption{Static Lake Embedding representations categorized according to their respective trophic state and hydrologic regime}
        \label{fig:lake_tsne_static_trophic_hydro_app}
    \end{minipage}
\end{figure}

\begin{figure}[!htbp]
    \centering
    \begin{subfigure}[b]{0.48\textwidth}
        \centering
        \includegraphics[width=\linewidth]{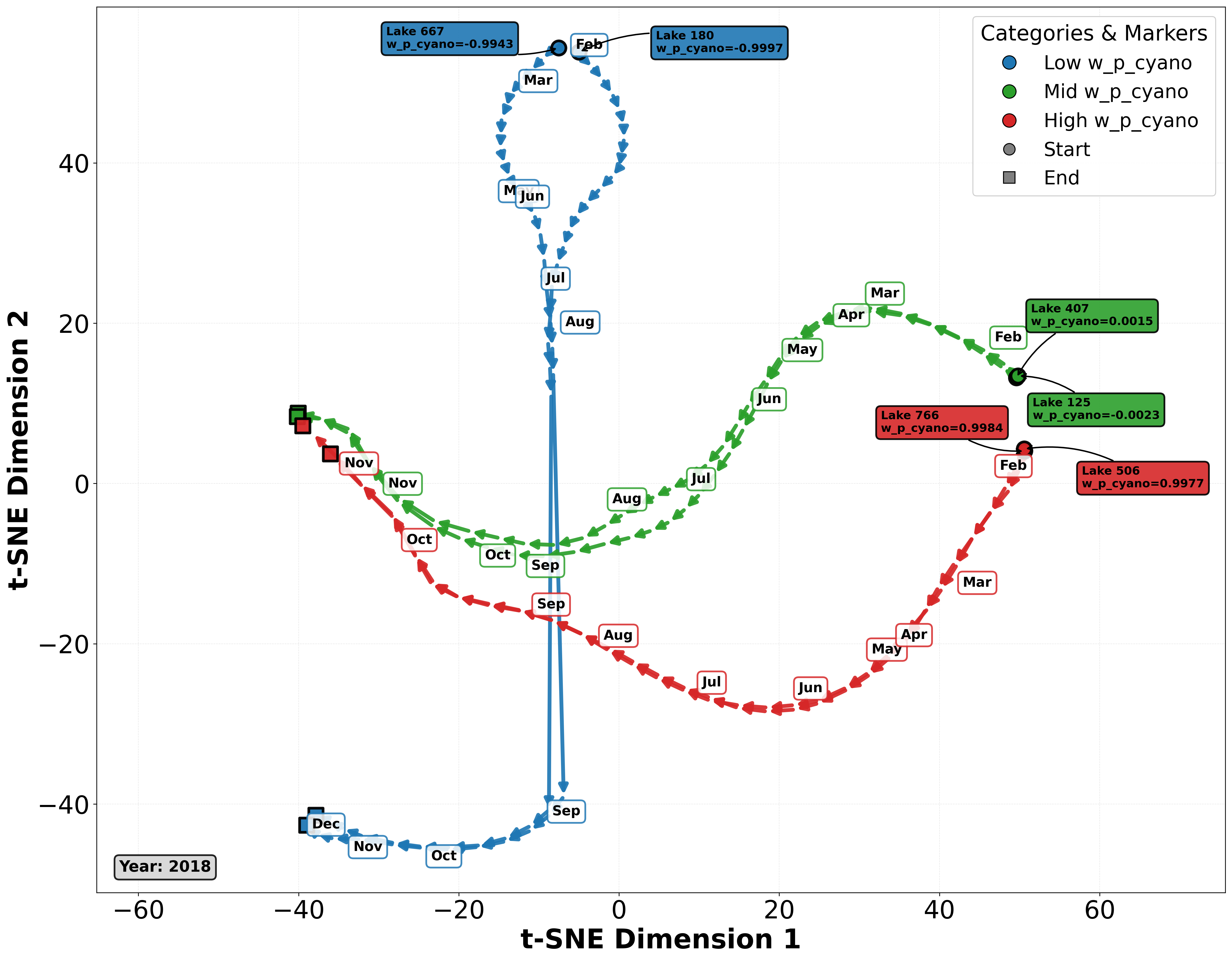}
        \caption{Trajectories for 2018.}
        \label{fig:fcr_cynao_bucket_2018}
    \end{subfigure}
    \hfill
    \begin{subfigure}[b]{0.48\textwidth}
        \centering
        \includegraphics[width=\linewidth]{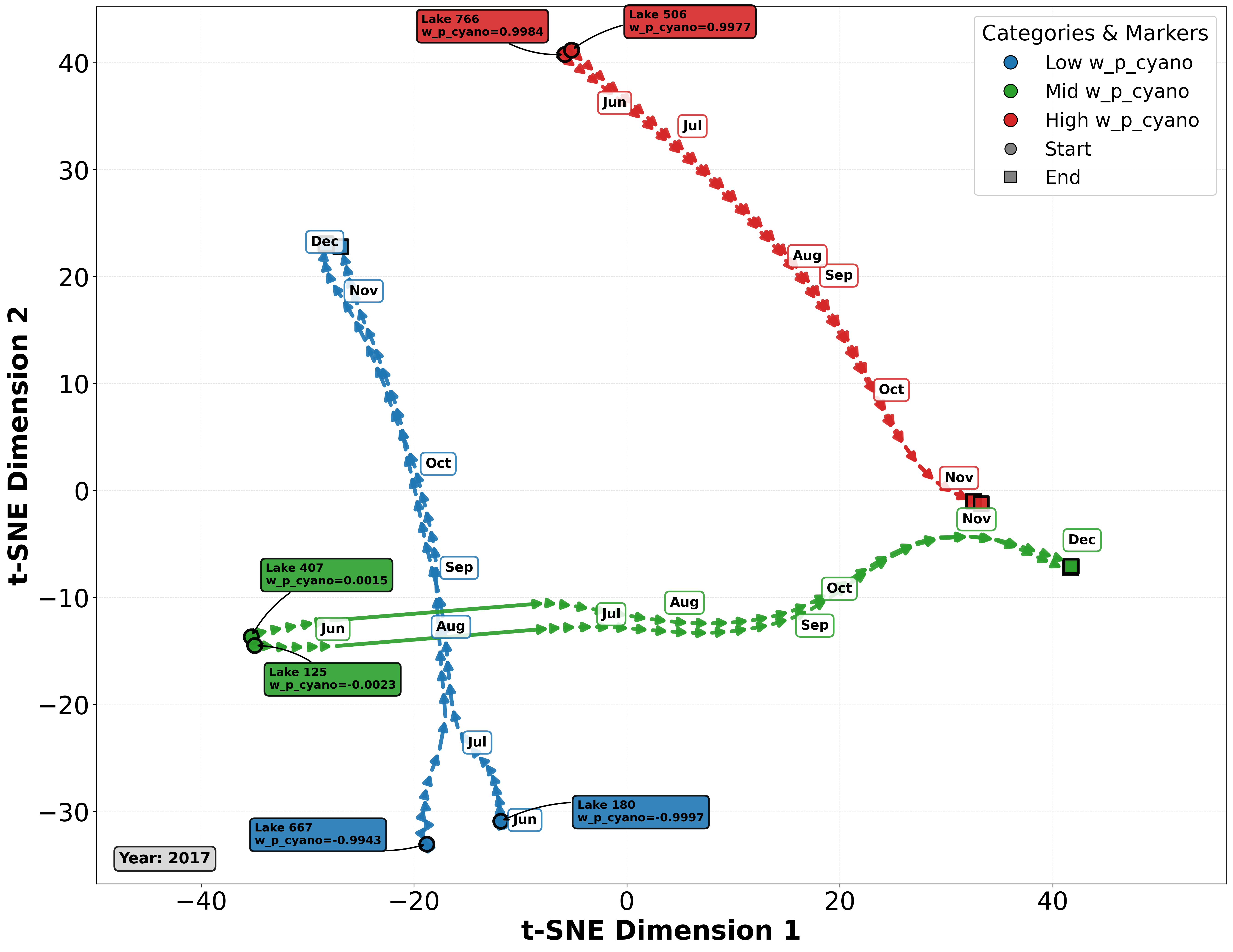}
        \caption{Trajectories for 2017.}
        \label{fig:fcr_cynao_bucket_2017}
    \end{subfigure}

    \caption{Dynamic Embedding-based trajectories for simulation lakes sampled from low, intermediate, and high $w\_p\_cyano$ groups for (a) 2018 and (b) 2017.}
    \label{fig:fcr_cyano_combined}
\end{figure}
\subsection{Embeddings from Simulated Lakes}
\label{appendix_inverse_learning}

To further assess whether the implicit parameter sensitivity extends beyond static representations, we analyze the temporal evolution of dynamic embeddings for lakes sampled from low, intermediate, and high $w\_p\_cyano$ groups. Figure ~\ref{fig:fcr_cyano_combined} show embedding trajectories for two representative lakes from each group over a full annual cycle. Within each $w\_p\_cyano$ group, lakes follow highly similar seasonal trajectories, while trajectories diverge systematically across regimes. This indicates that \lakefm{} captures parameter-conditioned dynamical behavior, organizing lake evolution according to the underlying phytoplankton growth parameter groups.
\subsection{\camready{Additional Results on Non-US Lakes}}
\label{sec:non_us_lakes}

\camready{The current OOD evaluation is limited in real-world breadth, hence, we obtained additional external lake datasets from \cite{corman2023high} and evaluated LakeFM and the baselines on them. Table~\ref{tab:euro_lakes_do_mse} reports DO prediction results (MSE) across six additional lakes, including one from Canada (Harp) and five from Sweden. These results provide additional evidence that LakeFM can generalize beyond the original benchmark and across broader real-world lake settings.}

\begin{table}[htbp]
\centering
\caption{Dissolved Oxygen (DO) prediction MSE on non-US lakes.}
\label{tab:euro_lakes_do_mse}
\setlength{\tabcolsep}{6pt}
\renewcommand{\arraystretch}{1.15}
\small
\begin{tabular}{lccccc}
\toprule
\textbf{Lake} & \textbf{LPTM} & \textbf{MOMENT} & \textbf{Chronos2} & \textbf{MOIRAI} & \textbf{LakeFM} \\
\midrule
Harp           & 1.419 & {1.408} & \underline{0.805}    & 1.62e2         & \textbf{0.558}  \\
Lillsjoliden   & 1.446 & \underline{1.432} & 1.896             & 6.87e1         & \textbf{1.388} \\
Mangstrettjarn & 0.634             & 0.634             & \underline{0.557} & 4.55e1         & \textbf{0.381} \\
Nastjarn       & 5.609             & \underline{5.487} & 1.12e1            & 4.85e2         & \textbf{5.335} \\
Ovre           & {1.556} & 1.566             & \underline{1.163}    & 1.14e2         & \textbf{1.059}          \\
Struptjarn     & 1.828             & 1.795             & \underline{1.790}    & 7.59e1         & \textbf{1.712} \\
\bottomrule
\end{tabular}
\end{table}
\subsection{Qualitative Analysis - Variate Masking}
\label{sec:appendix-variate-masking}

Figures \ref{fig:prla_masking}, \ref{fig:barc_masking} and \ref{fig:cb_masking} visually show the time-series forecasts of variates across lakes under no masking and masking conditions of variates.

\begin{figure*}[!htbp]
    \centering

    \begin{subfigure}[t]{0.32\textwidth}
        \centering
        \includegraphics[width=\linewidth]{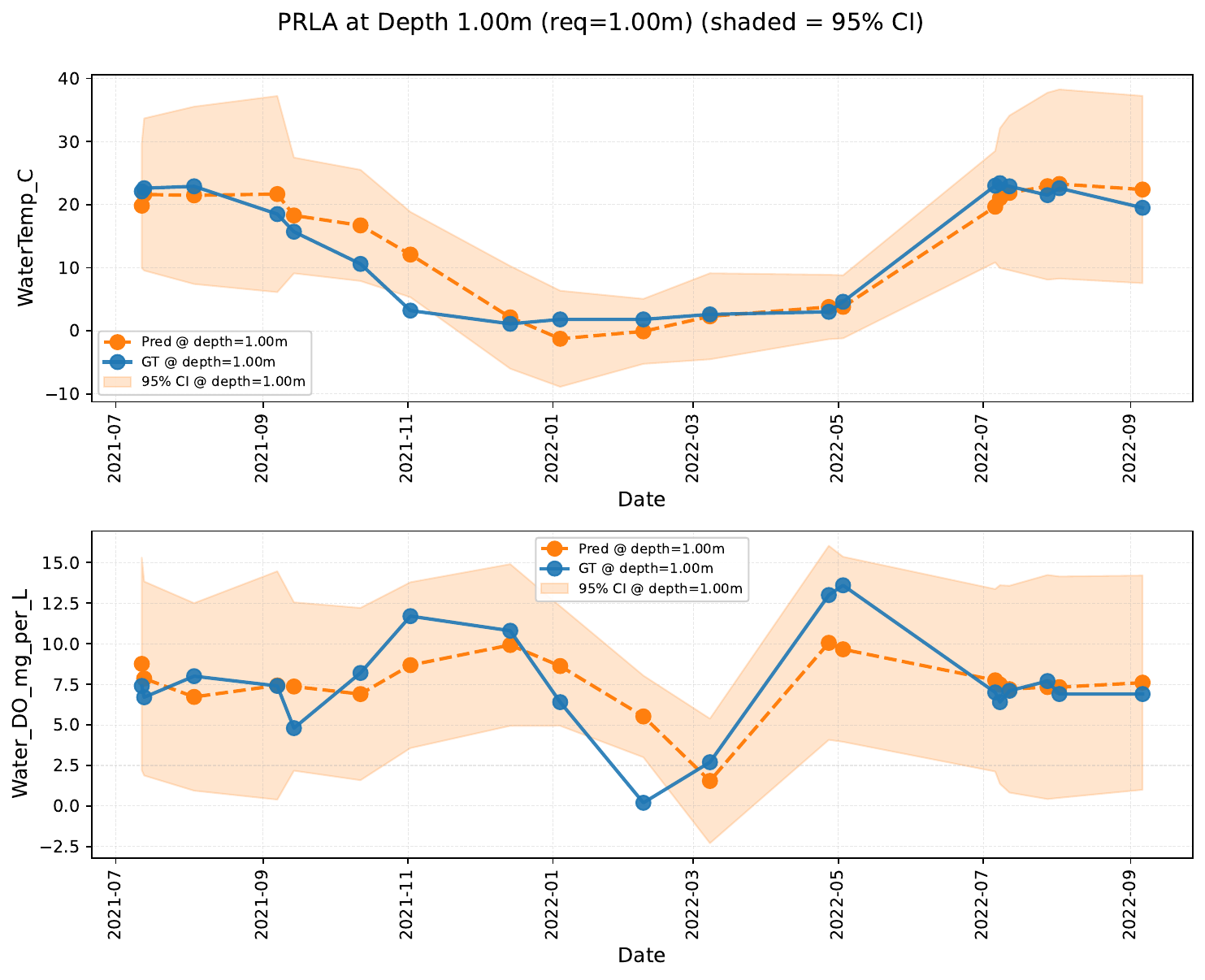}
        \caption{No Masking. PRLA @ 1.0m}
    \end{subfigure}
    \hfill
    \begin{subfigure}[t]{0.32\textwidth}
        \centering
        \includegraphics[width=\linewidth]{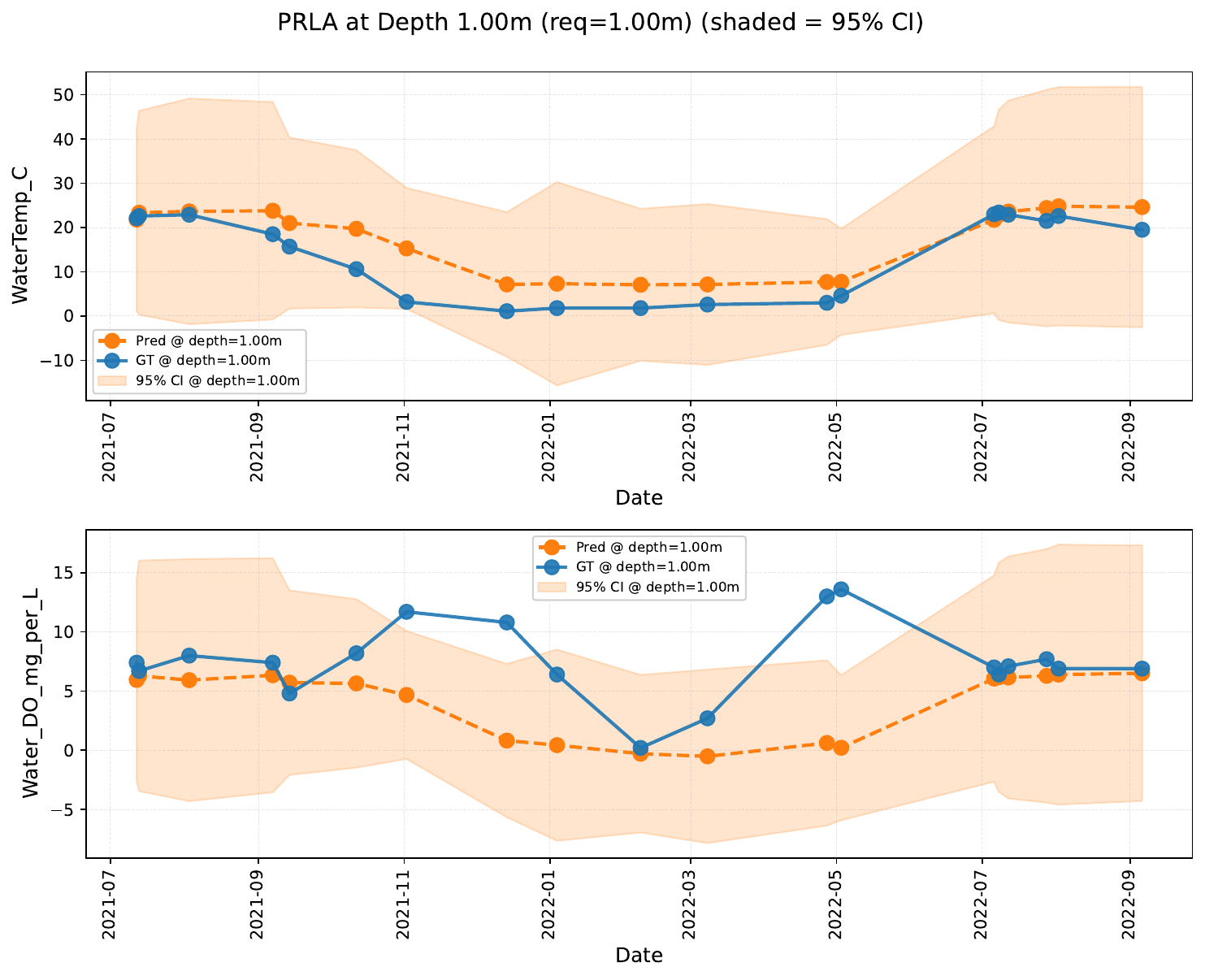}
        \caption{DO Masked. PRLA @ 1.0m}
    \end{subfigure}
    \hfill
    \begin{subfigure}[t]{0.32\textwidth}
        \centering
        \includegraphics[width=\linewidth]{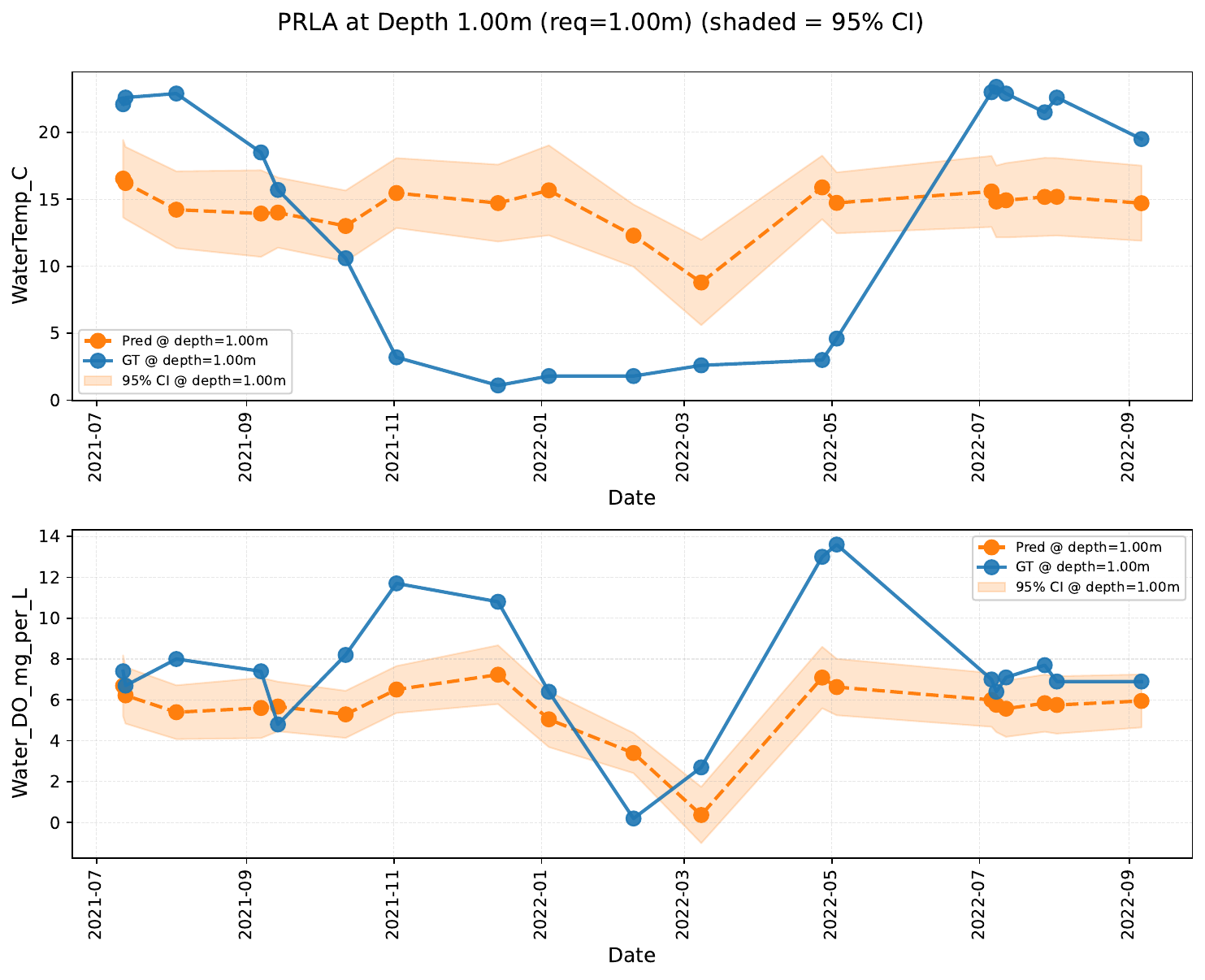}
        \caption{Temp Masked. PRLA @ 1.0m}
    \end{subfigure}

    \begin{subfigure}[t]{0.32\textwidth}
        \centering
        \includegraphics[width=\linewidth]{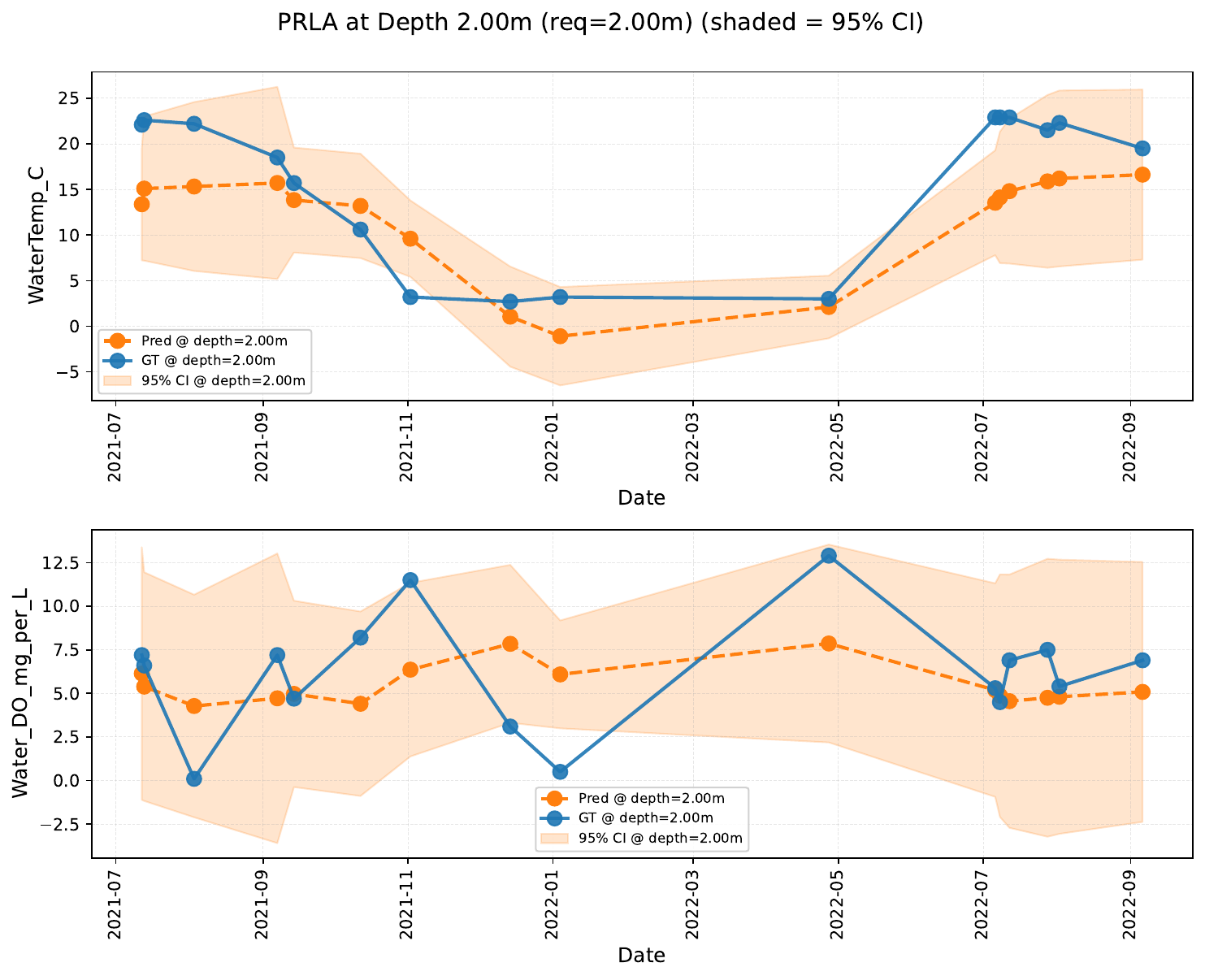}
        \caption{No Masking. PRLA @ 2.0m}
    \end{subfigure}
    \hfill
    \begin{subfigure}[t]{0.32\textwidth}
        \centering
        \includegraphics[width=\linewidth]{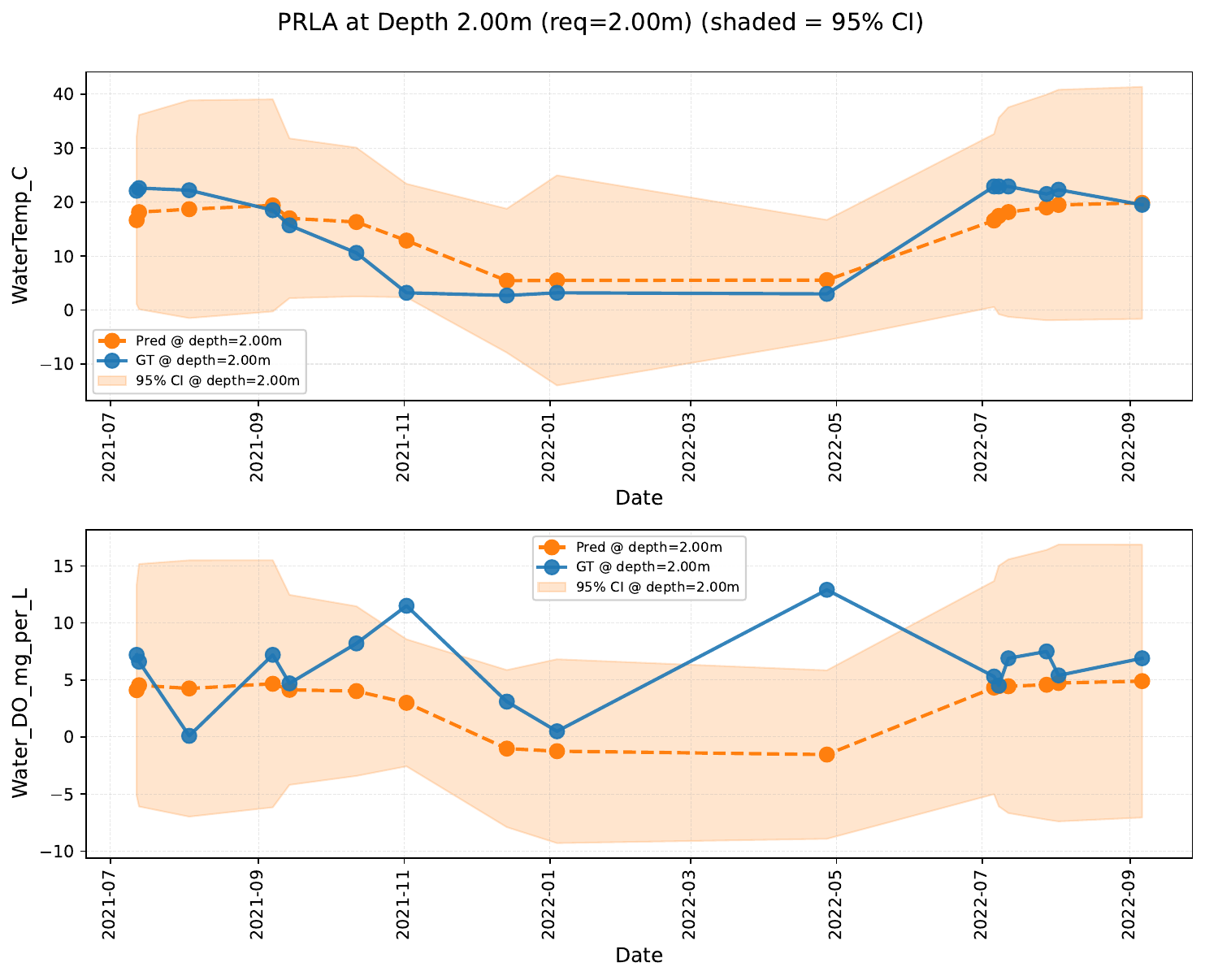}
        \caption{DO Masked. PRLA @ 2.0m}
    \end{subfigure}
    \hfill
    \begin{subfigure}[t]{0.32\textwidth}
        \centering
        \includegraphics[width=\linewidth]{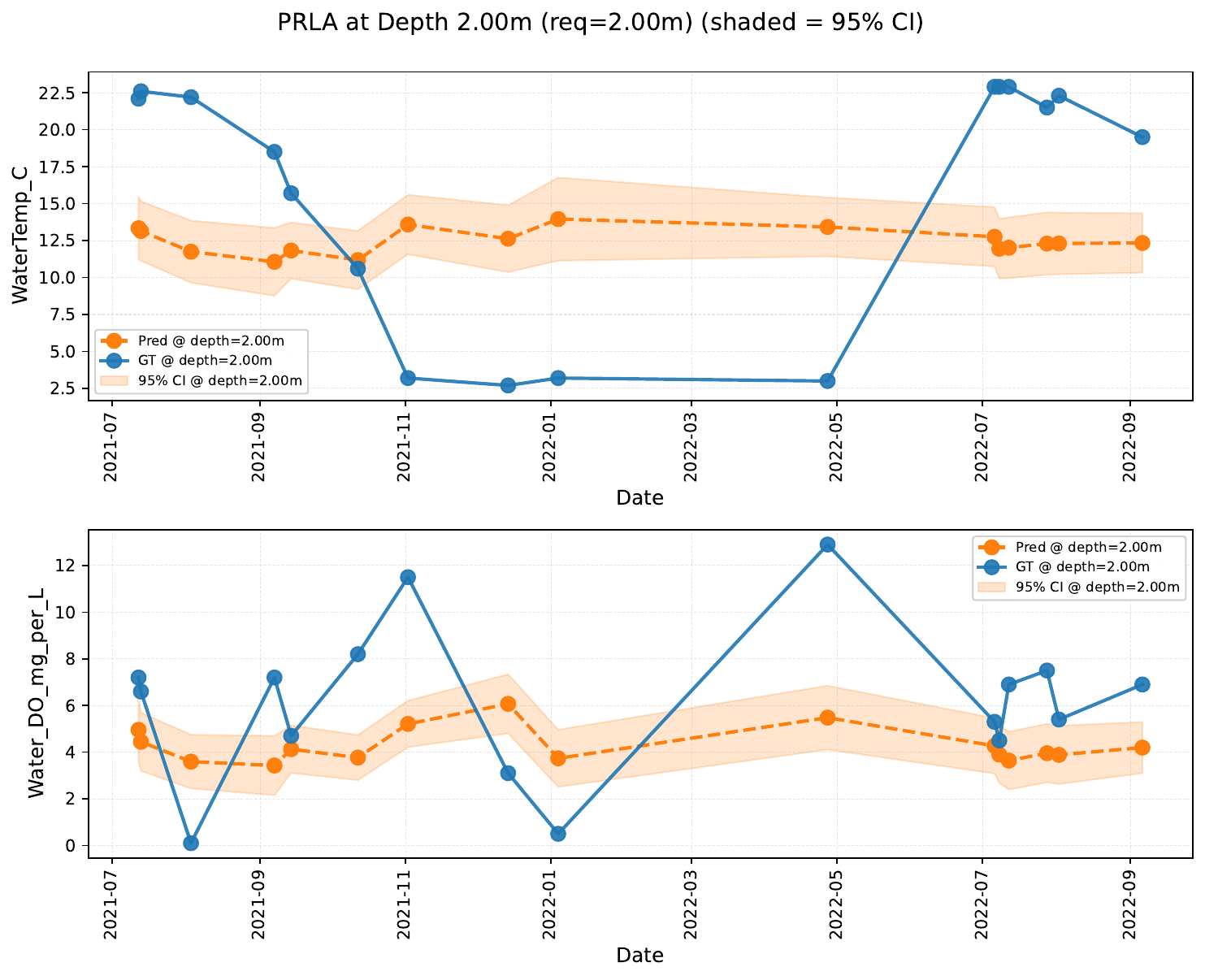}
        \caption{Temp Masked. PRLA @ 2.0m}
    \end{subfigure}


    \caption{Variate Masking - No masking vs Masked prediction Plots for Lake PRLA}
    \label{fig:prla_masking}
\end{figure*}

\begin{figure*}[t]
    \centering

    \begin{subfigure}[t]{0.32\textwidth}
        \centering
        \includegraphics[width=\linewidth]{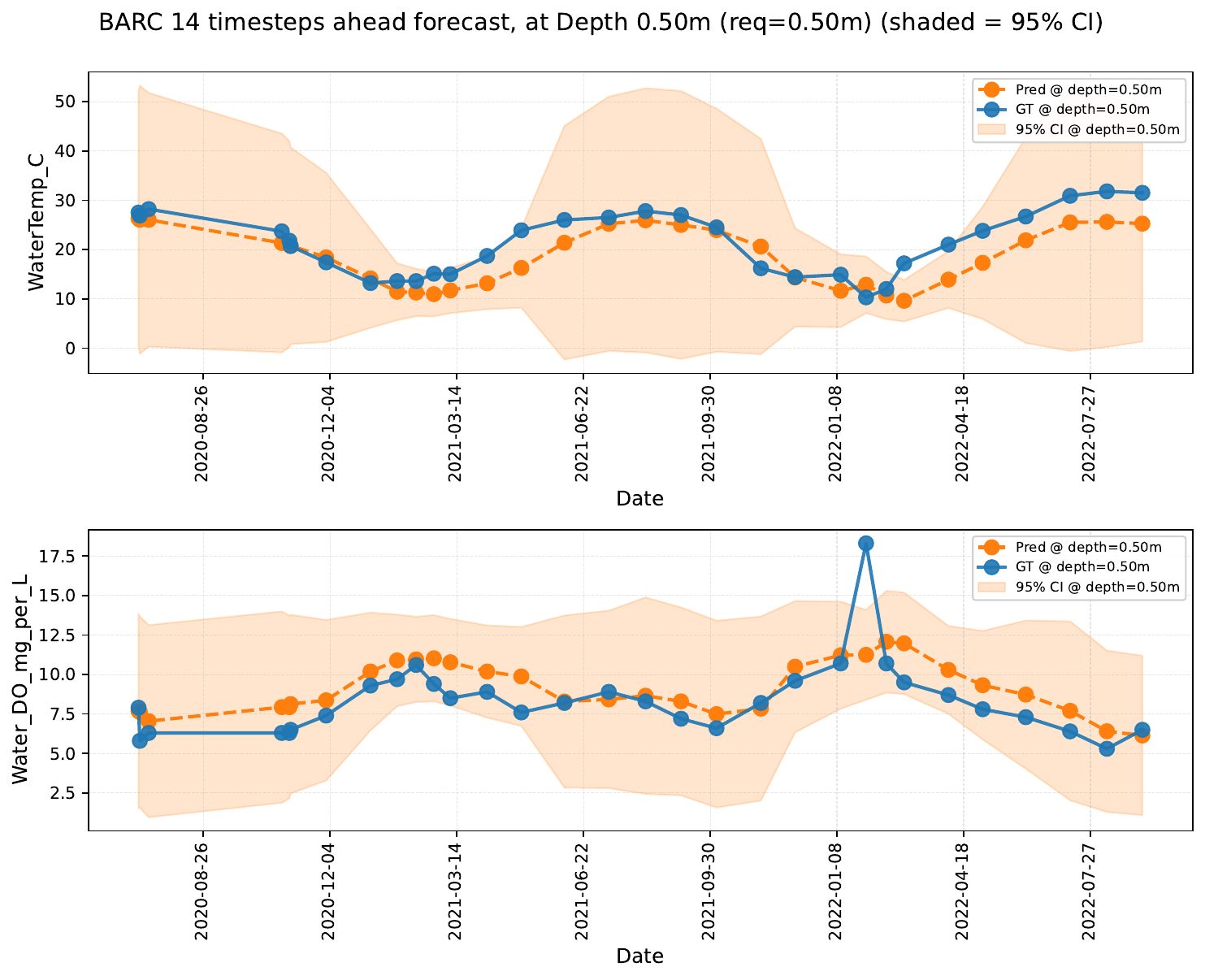}
        \caption{No Masking. BARC @ 0.5m}
    \end{subfigure}
    \hfill
    \begin{subfigure}[t]{0.32\textwidth}
        \centering
        \includegraphics[width=\linewidth]{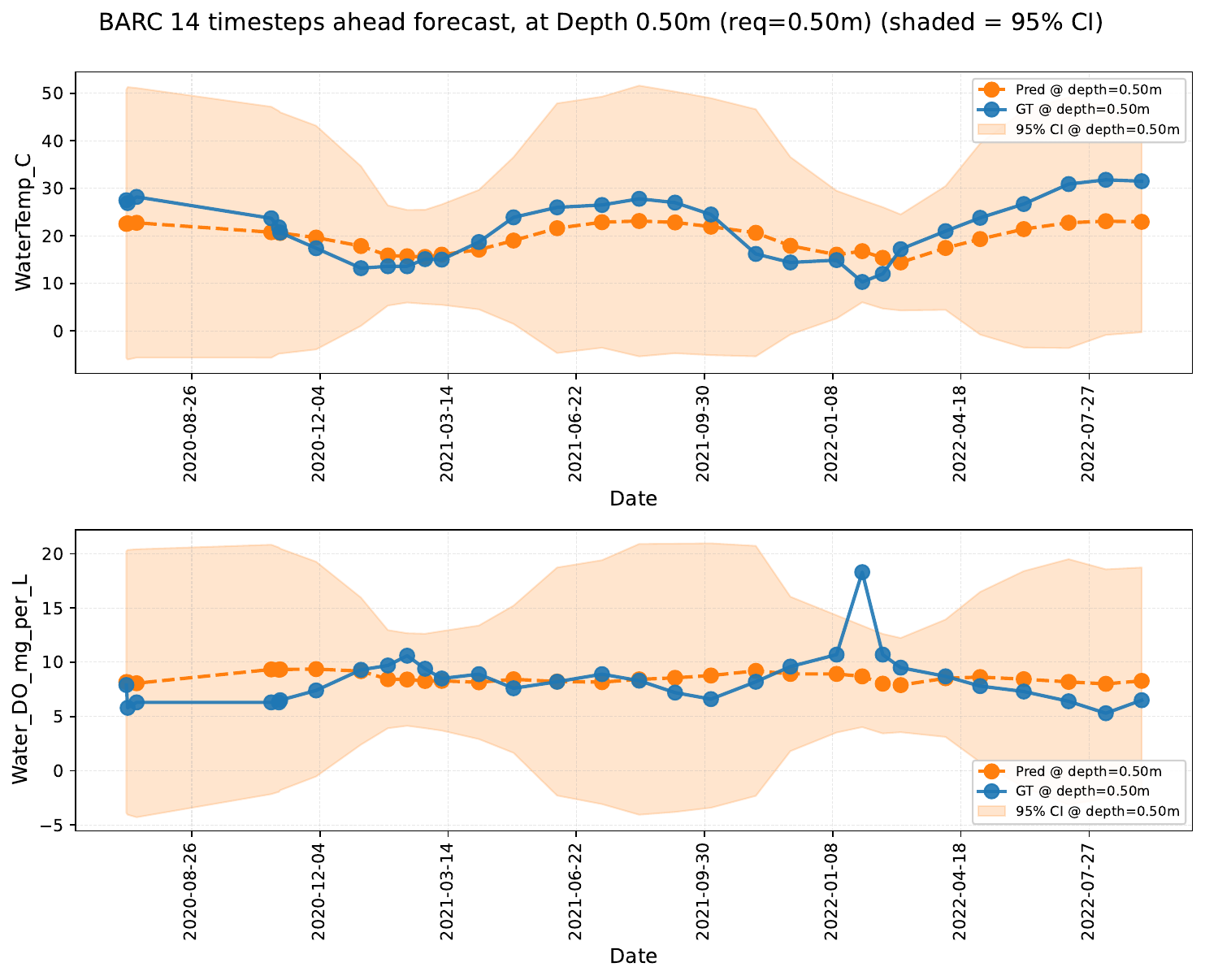}
        \caption{DO Masked. BARC @ 0.5m}
    \end{subfigure}
    \hfill
    \begin{subfigure}[t]{0.32\textwidth}
        \centering
        \includegraphics[width=\linewidth]{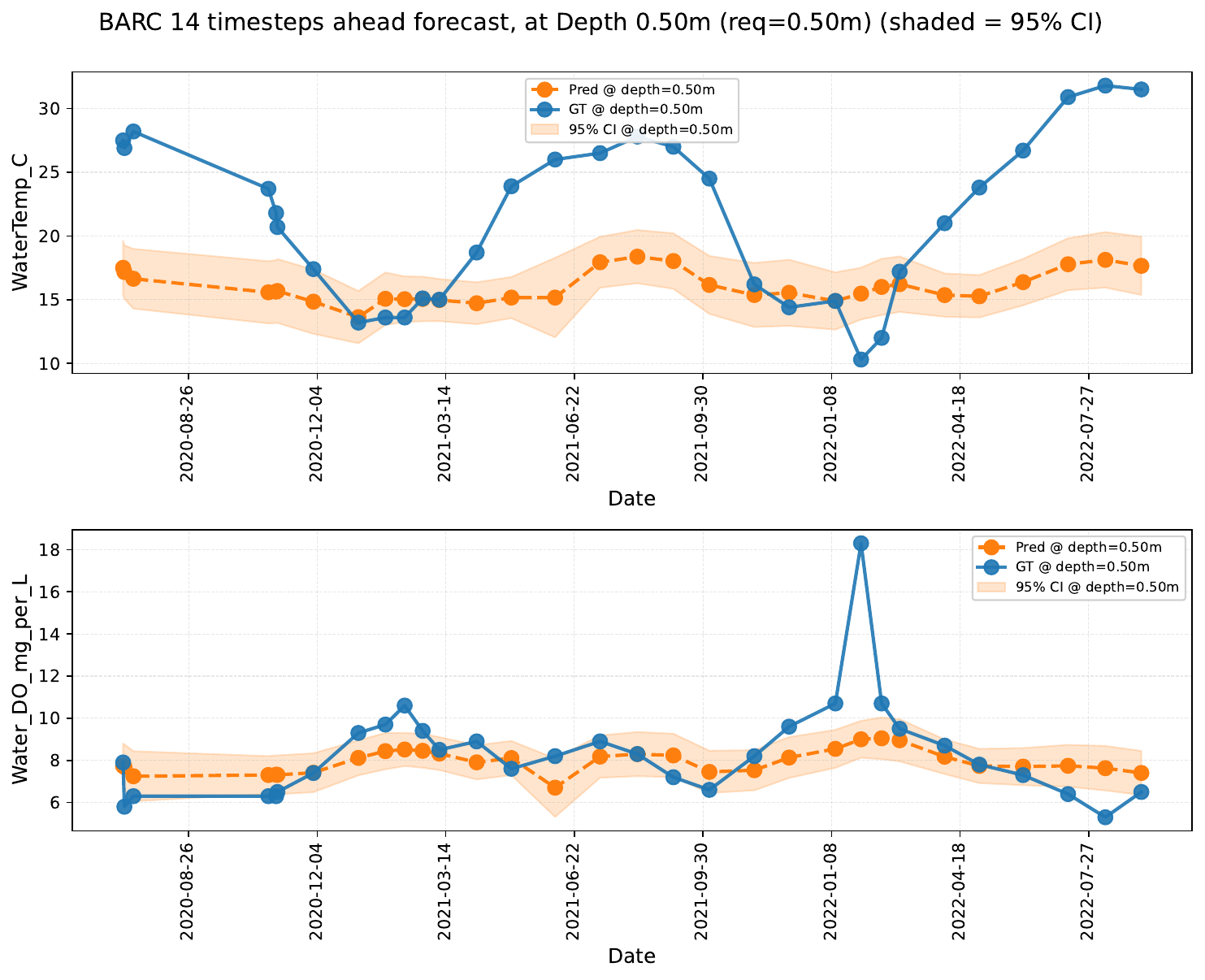}
        \caption{Temp Masked. BARC @ 0.5m}
    \end{subfigure}

    \begin{subfigure}[t]{0.32\textwidth}
        \centering
        \includegraphics[width=\linewidth]{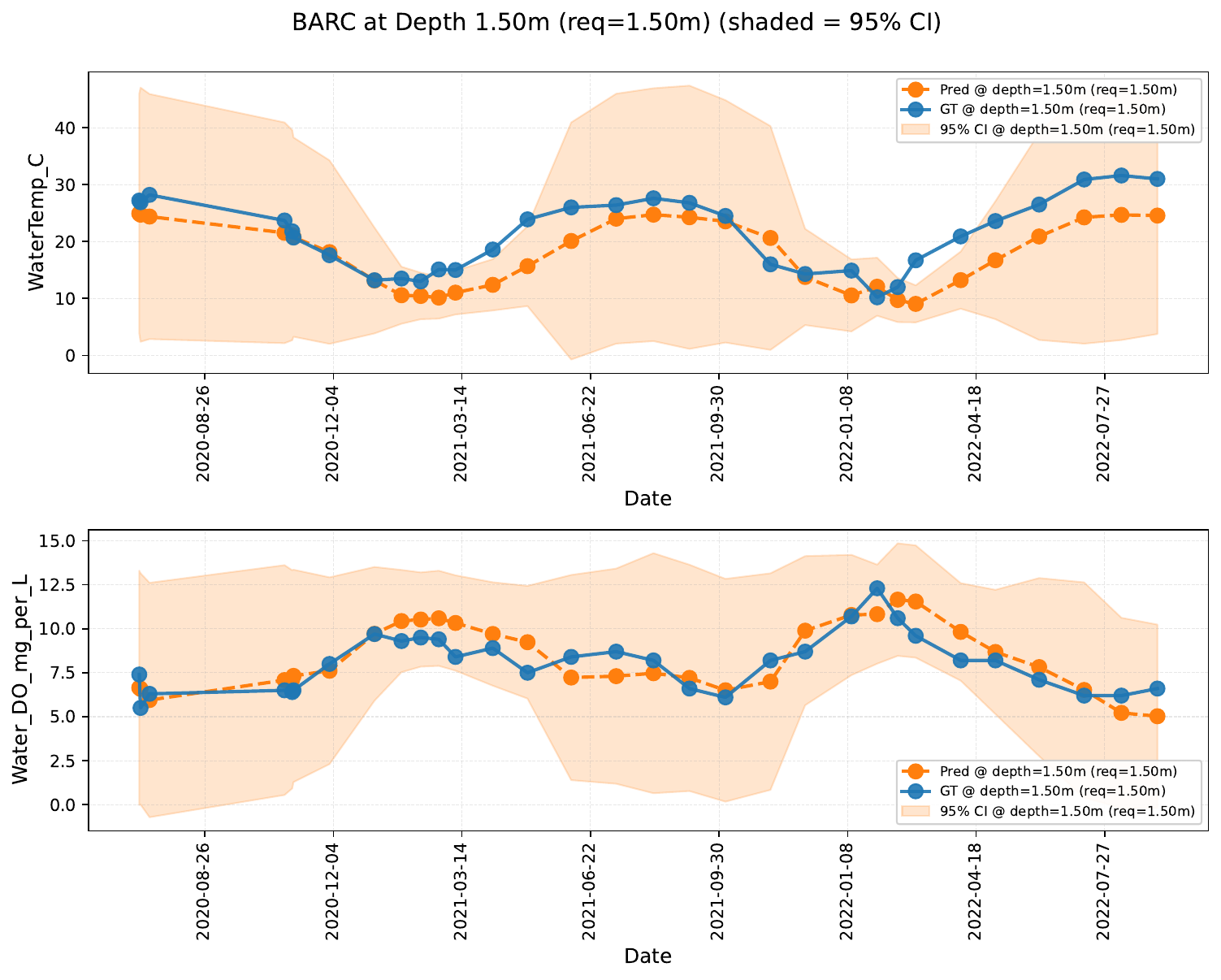}
        \caption{No Masking. BARC @ 1.5m}
    \end{subfigure}
    \hfill
    \begin{subfigure}[t]{0.32\textwidth}
        \centering
        \includegraphics[width=\linewidth]{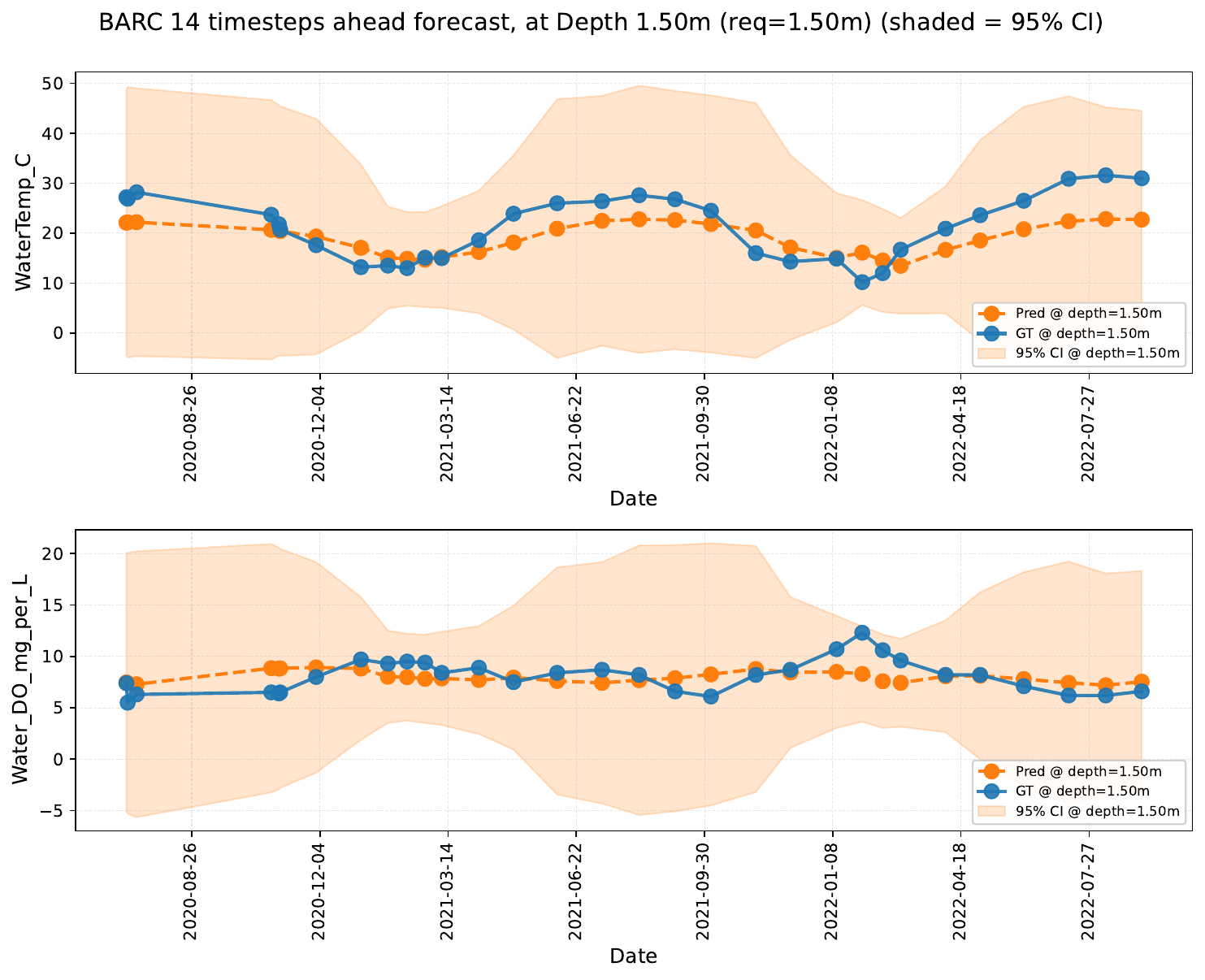}
        \caption{DO Masked. BARC @ 1.5m}
    \end{subfigure}
    \hfill
    \begin{subfigure}[t]{0.32\textwidth}
        \centering
        \includegraphics[width=\linewidth]{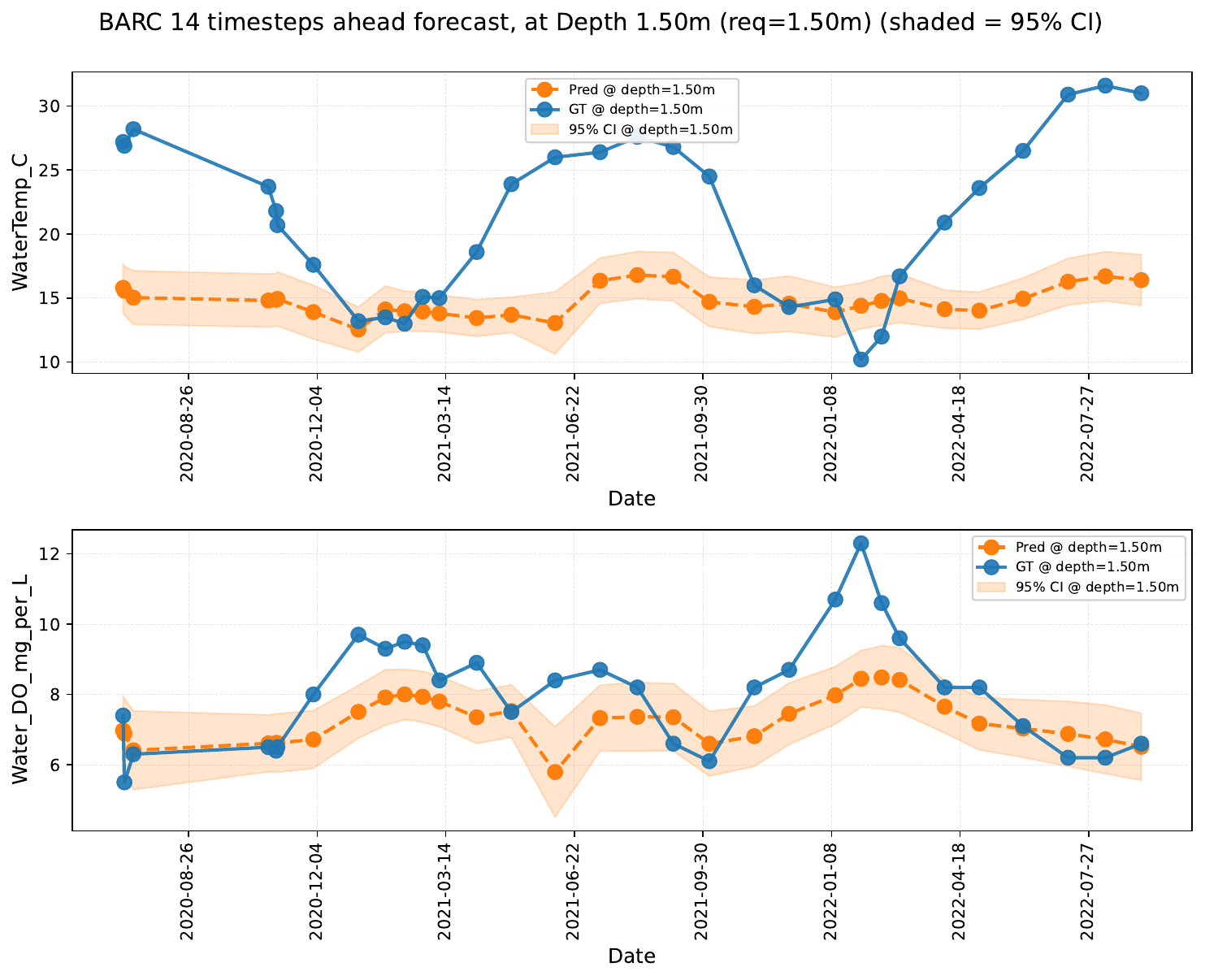}
        \caption{Temp Masked. BARC @ 1.5m}
    \end{subfigure}


    \caption{Variate Masking - No masking vs Masked prediction Plots for Lake BARC}
    \label{fig:barc_masking}
\end{figure*}

\begin{figure*}[t]
    \centering
    \begin{subfigure}[t]{0.32\textwidth}
        \centering
        \includegraphics[width=\linewidth]{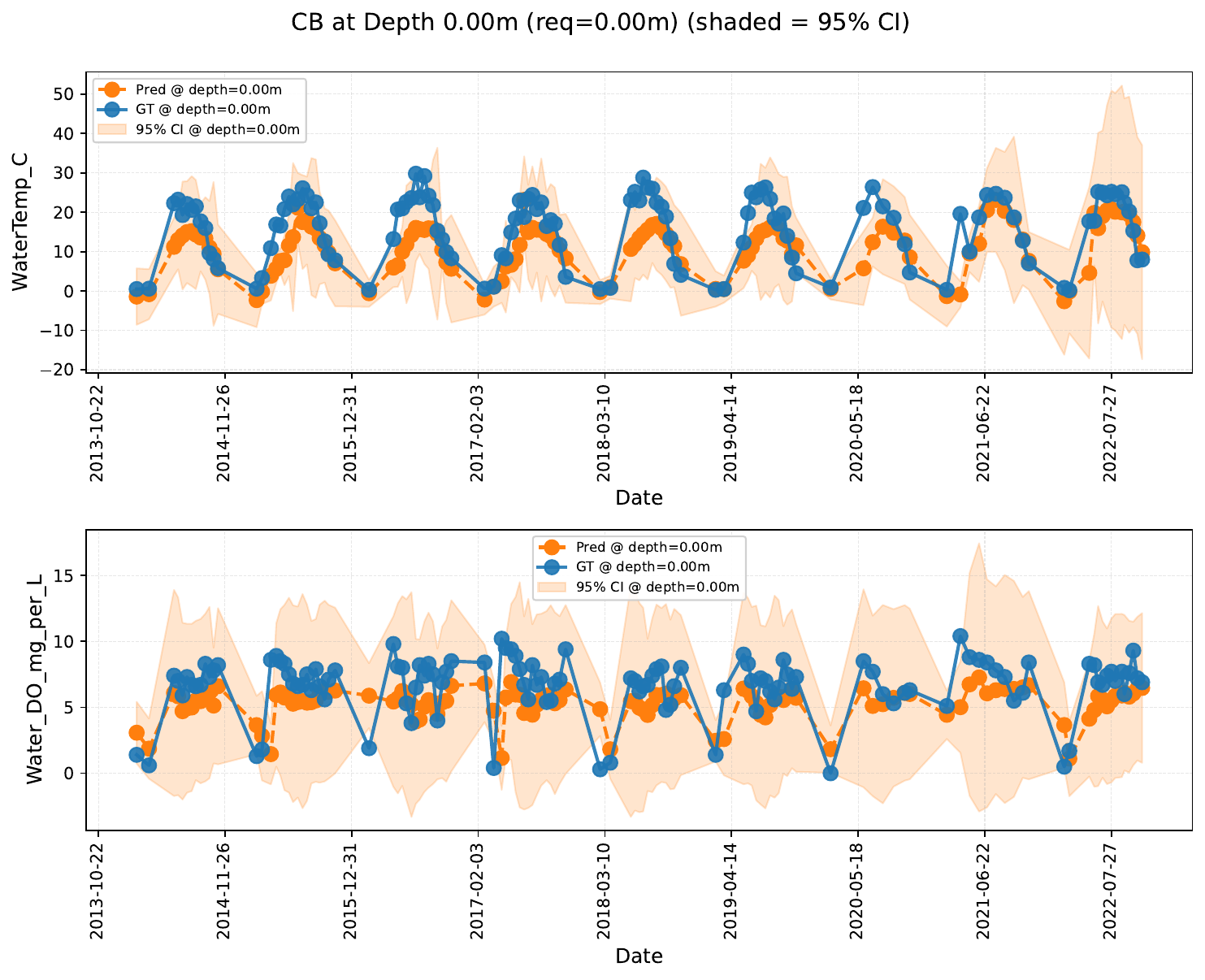}
        \caption{No Masking. CB @ 0.0m}
    \end{subfigure}
    \hfill
    \begin{subfigure}[t]{0.32\textwidth}
        \centering
        \includegraphics[width=\linewidth]{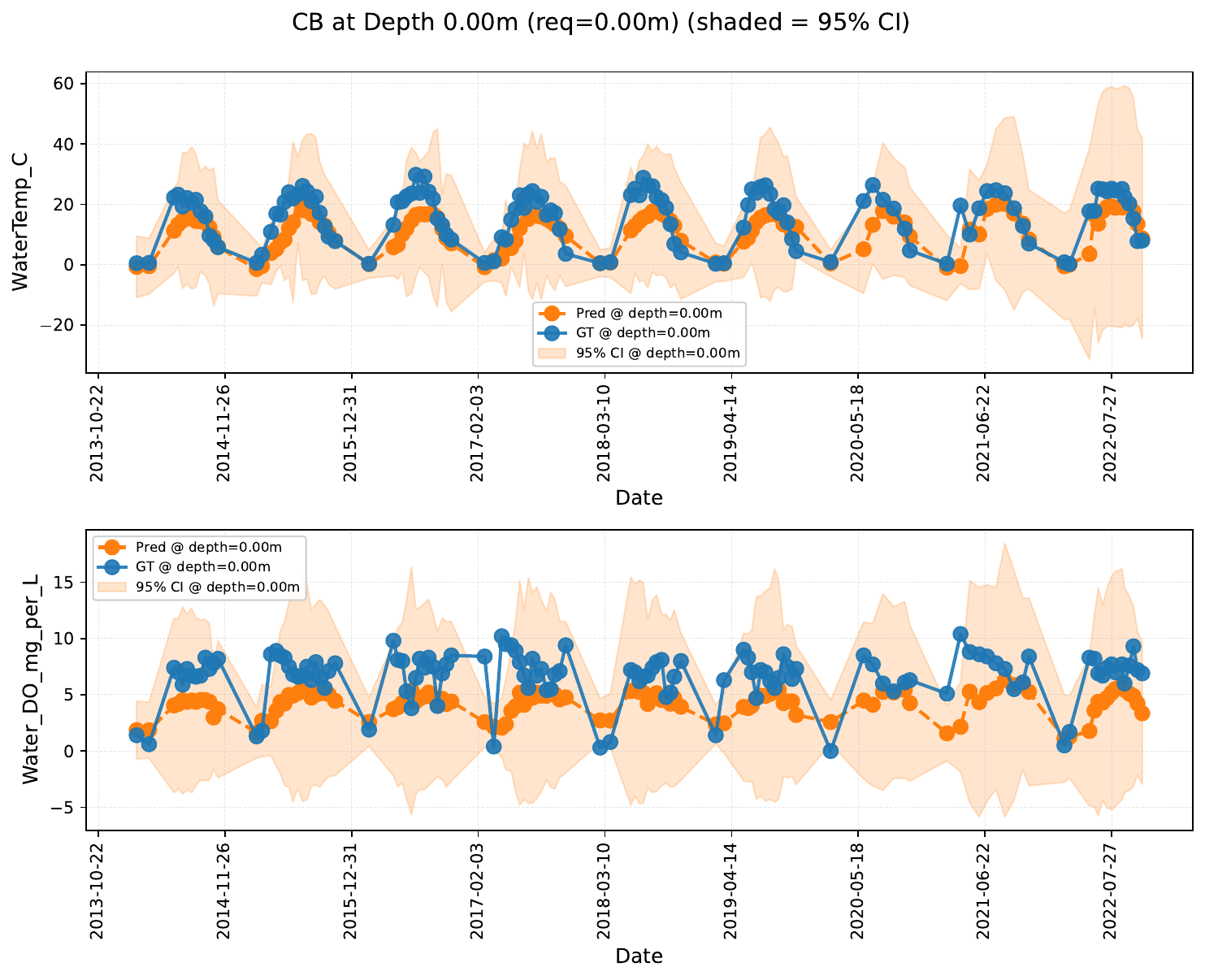}
        \caption{DO Masked. CB @ 0.0m}
    \end{subfigure}
    \hfill
    \begin{subfigure}[t]{0.32\textwidth}
        \centering
        \includegraphics[width=\linewidth]{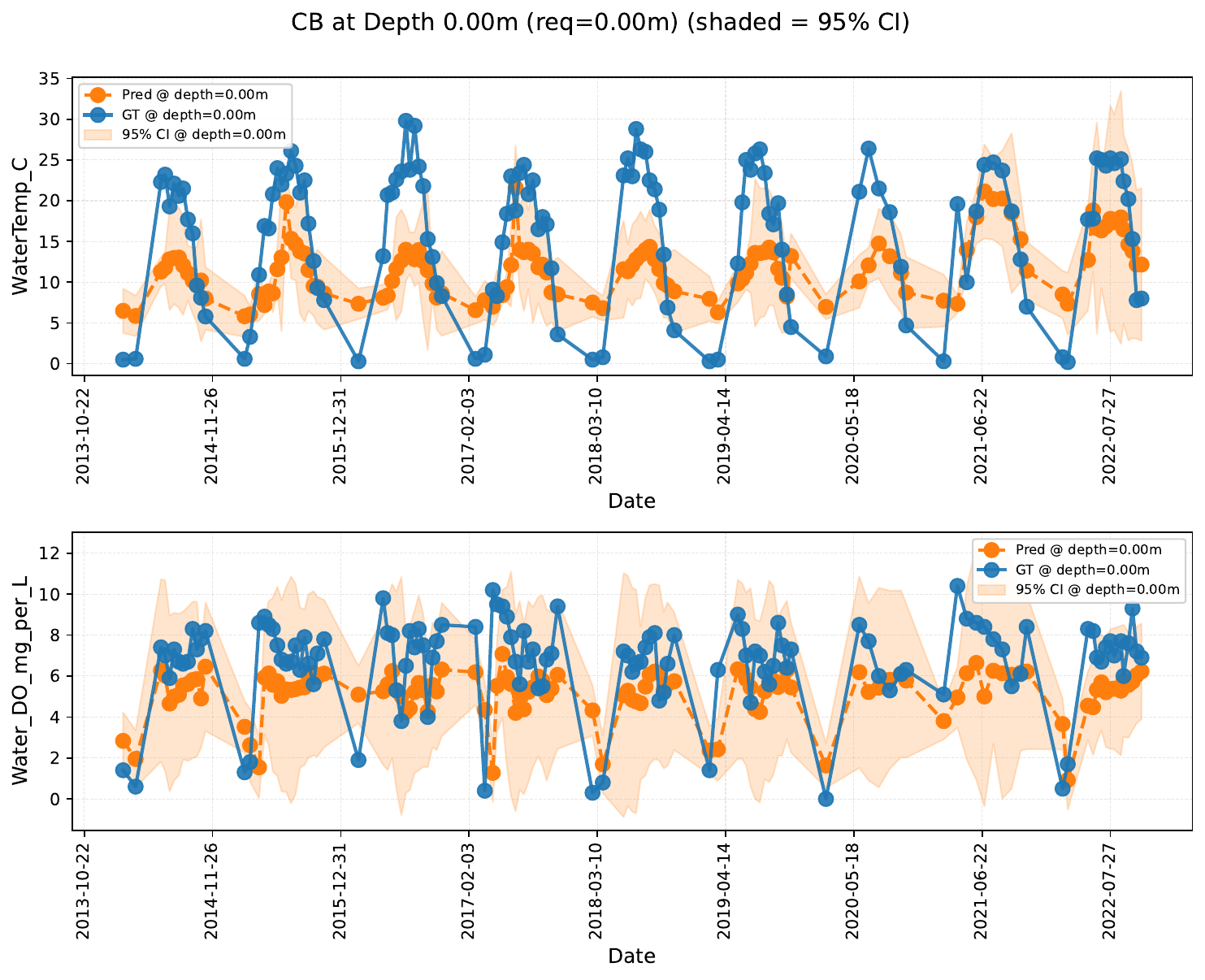}
        \caption{Temp Masked. CB @ 0.0m}
    \end{subfigure}

    \begin{subfigure}[t]{0.32\textwidth}
        \centering
        \includegraphics[width=\linewidth]{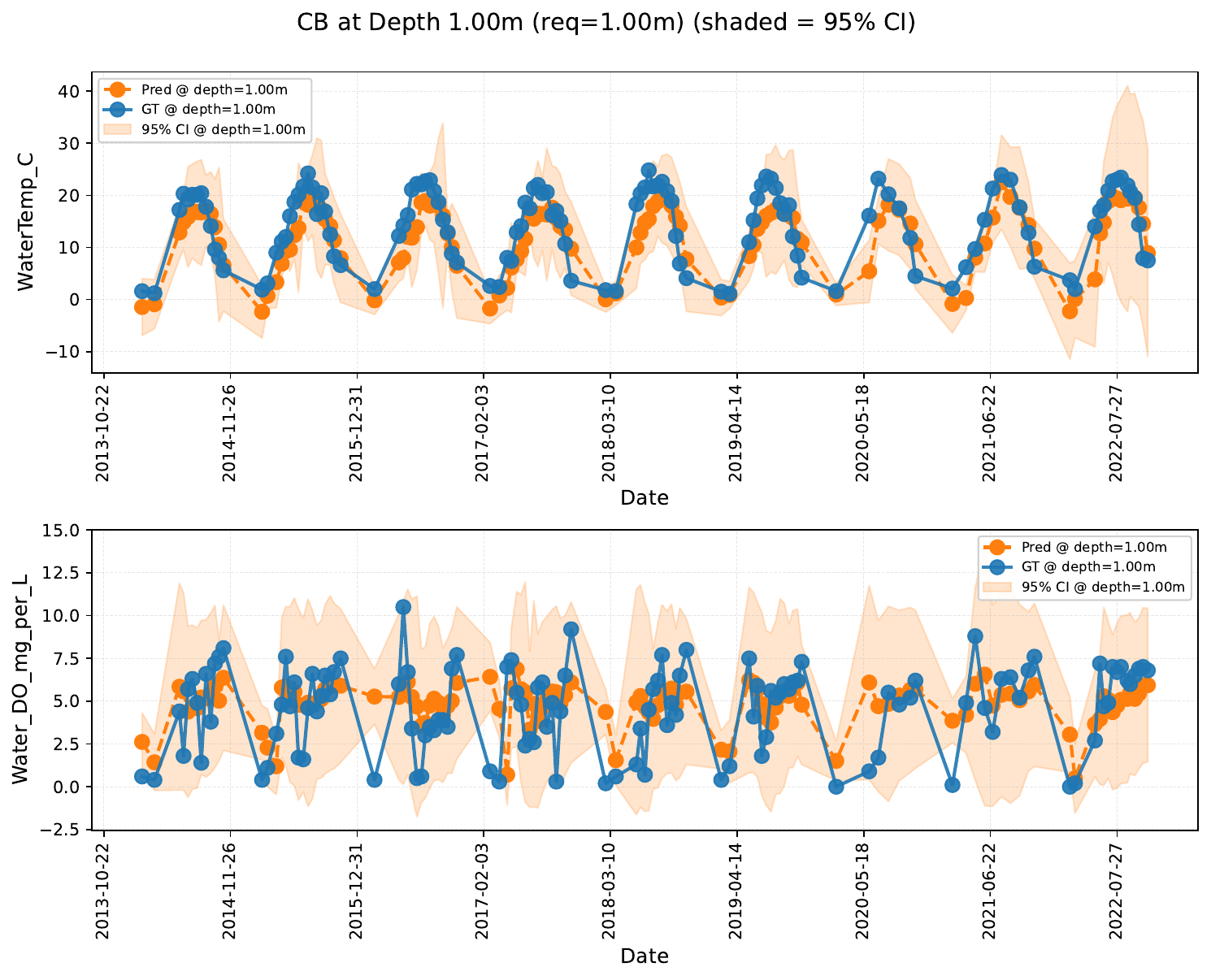}
        \caption{No Masking. CB @ 1.0m}
    \end{subfigure}
    \hfill
    \begin{subfigure}[t]{0.32\textwidth}
        \centering
        \includegraphics[width=\linewidth]{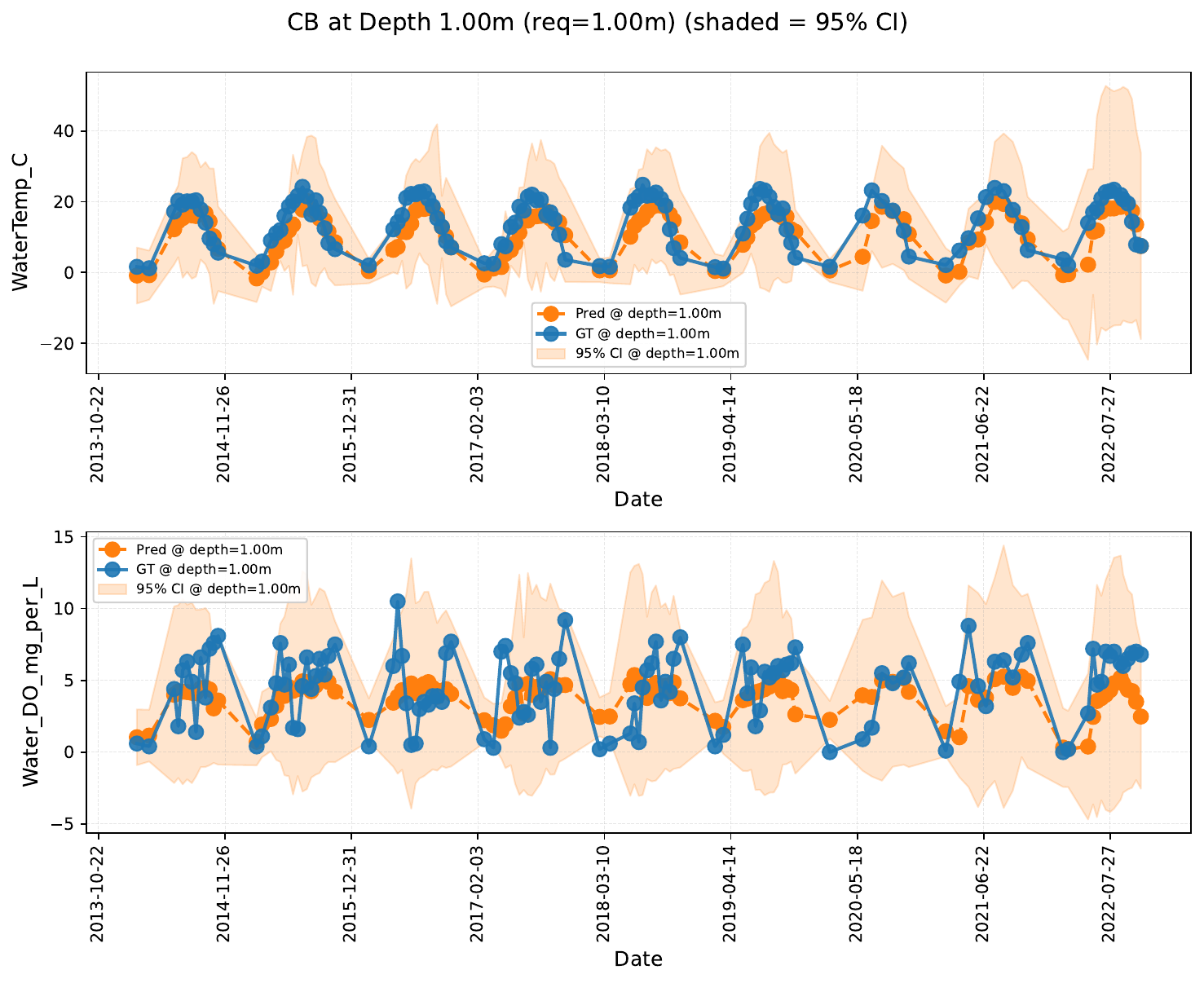}
        \caption{DO Masked. CB @ 1.0m}
    \end{subfigure}
    \hfill
    \begin{subfigure}[t]{0.32\textwidth}
        \centering
        \includegraphics[width=\linewidth]{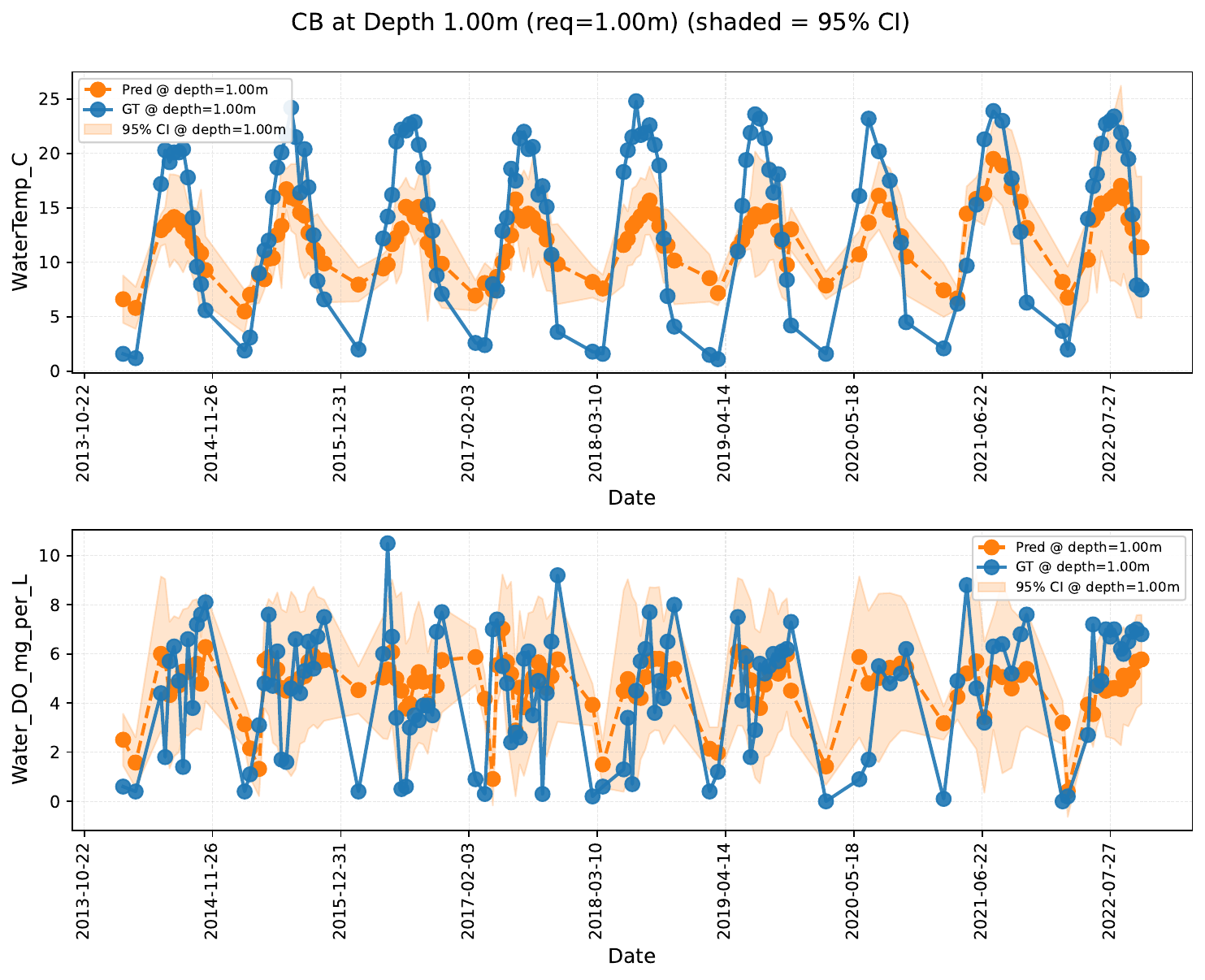}
        \caption{Temp Masked. CB @ 1.0m}
    \end{subfigure}

    \begin{subfigure}[t]{0.32\textwidth}
        \centering
        \includegraphics[width=\linewidth]{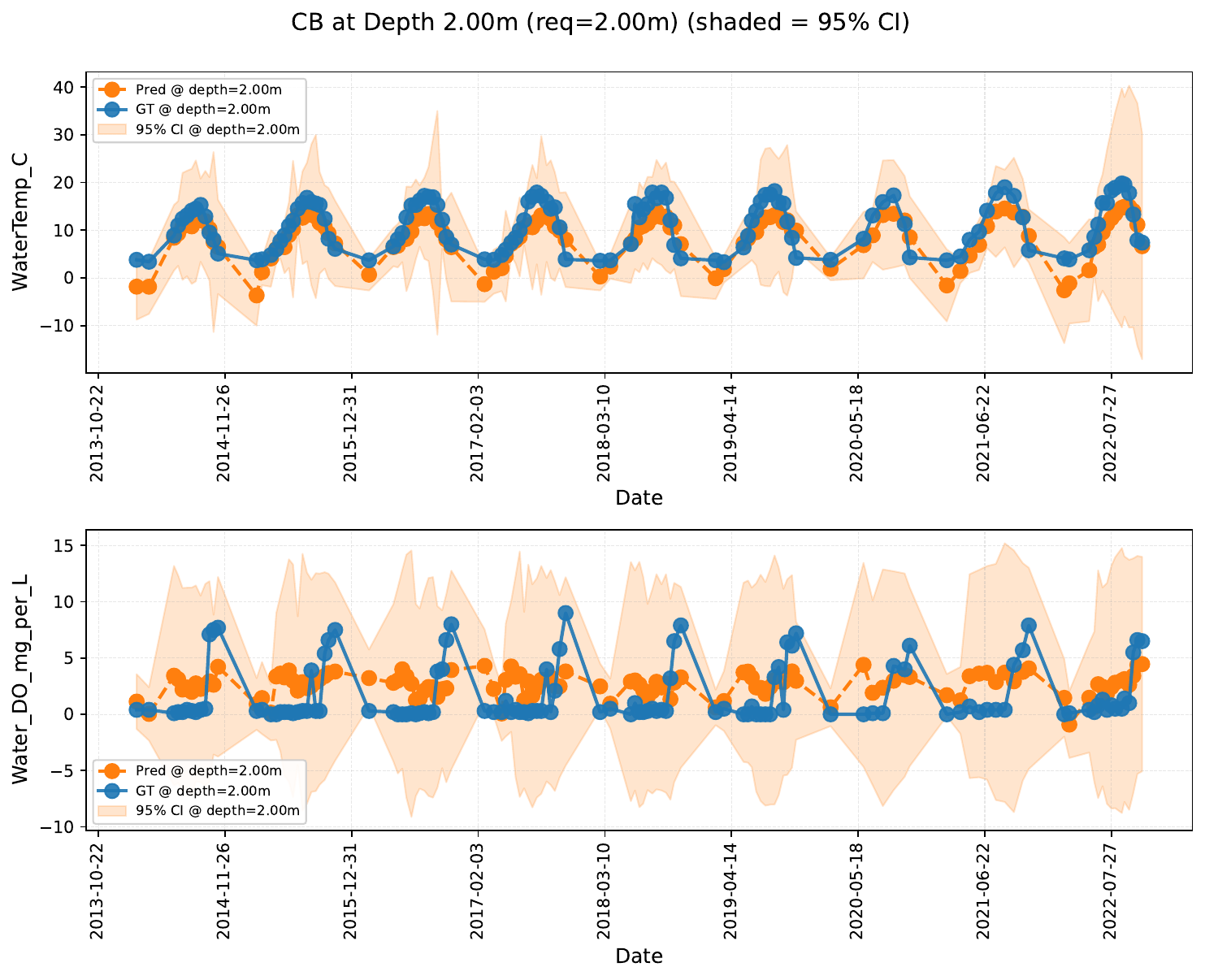}
        \caption{No Masking. CB @ 2.0m}
    \end{subfigure}
    \hfill
    \begin{subfigure}[t]{0.32\textwidth}
        \centering
        \includegraphics[width=\linewidth]{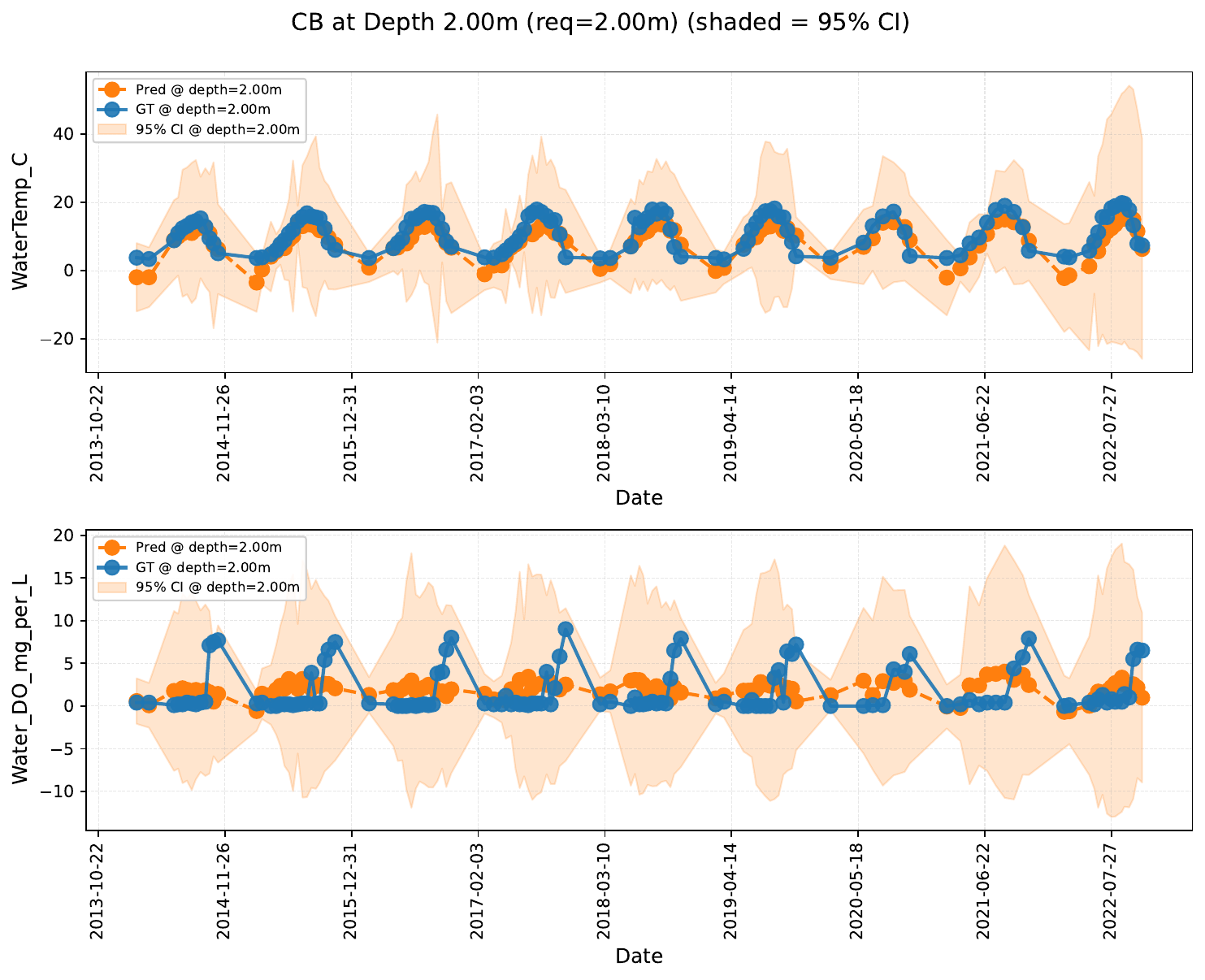}
        \caption{DO Masked. CB @ 2.0m}
    \end{subfigure}
    \hfill
    \begin{subfigure}[t]{0.32\textwidth}
        \centering
        \includegraphics[width=\linewidth]{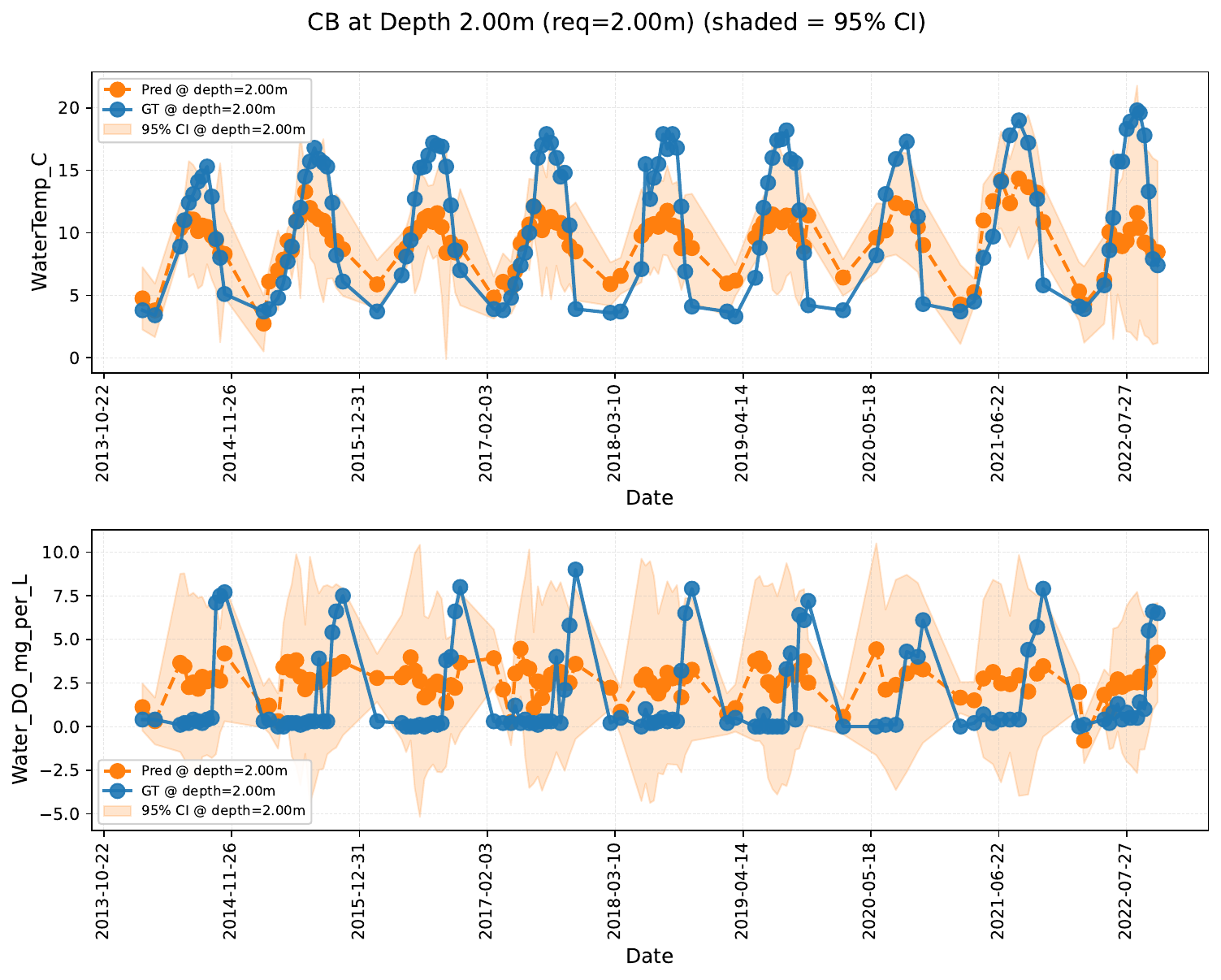}
        \caption{Temp Masked. CB @ 2.0m}
    \end{subfigure}
    \caption{Variate Masking - No masking vs Masked Prediction Plots for Lake CB}
    \label{fig:cb_masking}
\end{figure*}

\subsection{Variate Importance - Single Variate Masking}
\label{sec:appendix-single-variate-masking}
We mask out one variable in the model's input/historical window, and measure the change in the predictive performance of each of the variables (including the variate masked). This experiment enables us to discover the influencing variables, as well as least sensitive variables (w.r.t. the other variates) within each lake system. Figure \ref{fig:per_lake_delta_heatmaps} shows the heatmaps corresponding to the change in the error metrics based on masking out a single variable, for each lake. 

While we generally see an increase in the error metrics on masking an input variable, this is however, not always the case. For e.g. in SUGG, we see that removing Water DO improves the prediction performance of Water Temp. Here are some ecological reasons based on these observations - 
DO dynamics reflect the interaction of biological processes (e.g., primary production and respiration) and physical processes (e.g., mixing and air–water exchange) \cite{staehr2010lake}, with their relative importance shifting across depths and temporal scales \cite{langman2010control}. Under stable physical conditions, warm temperatures and high irradiance drive photosynthesis and produce a predictable diel DO signal, while deeper waters below the photic zone are dominated by respiration and declining DO. This structure can be disrupted by mixing events, which obscure the biological signal. At shorter time scales, internal waves, organismal advection past sensors, and local biotic interactions introduce additional variability that may be sensitive to weather or effectively stochastic \cite{langman2010control, mcafee2025lakebed}, making DO an instructive example of a predictor whose information content can vary dramatically across contexts.


\begin{figure*}[t]
\centering
\caption{Per-lake performance deltas visualized as heatmaps.
For each lake, we show $\Delta$MSE relative to the baseline.}
\label{fig:per_lake_delta_heatmaps}

\includegraphics[width=0.45\linewidth]{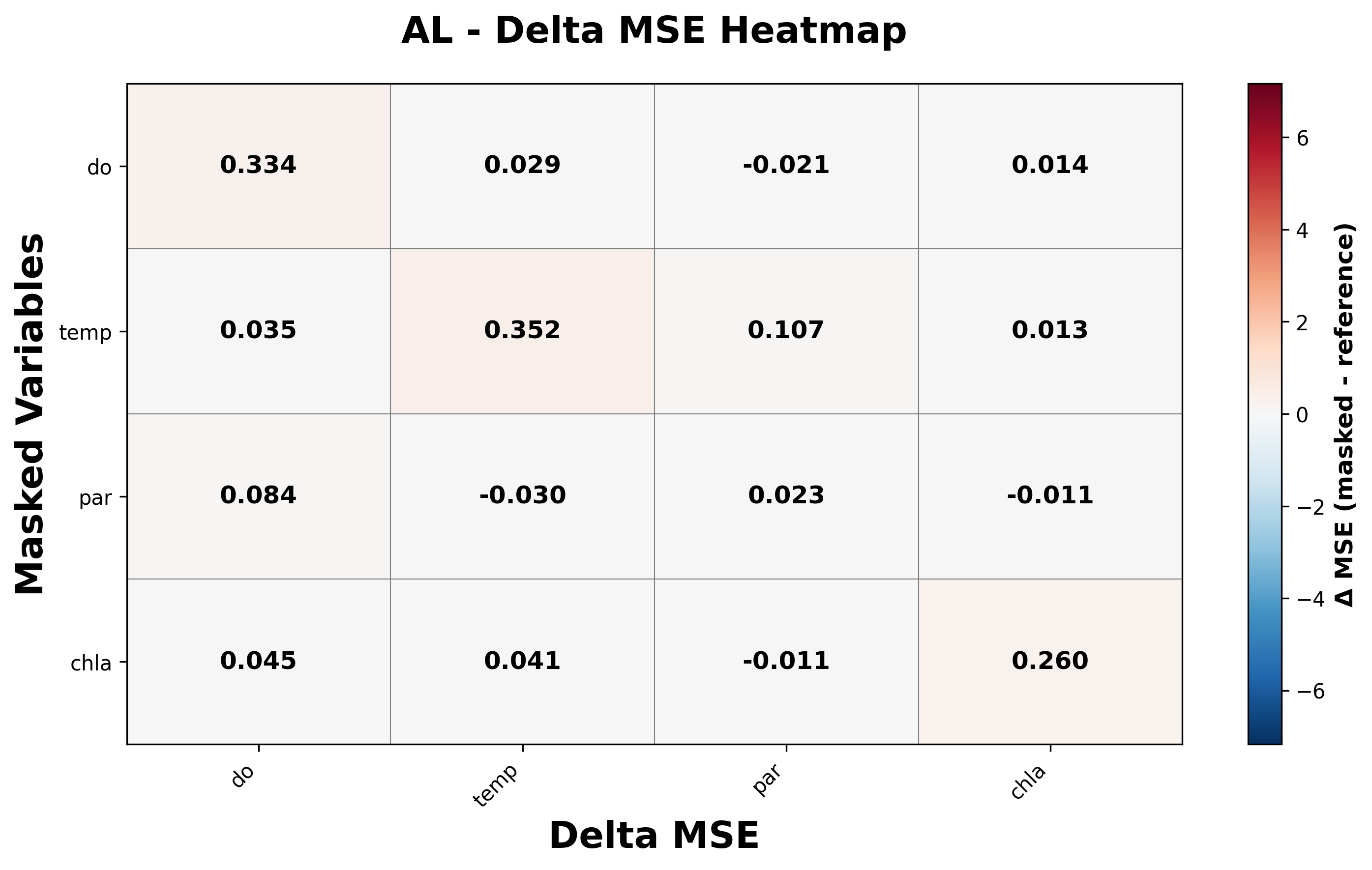}
\hspace{0.03\linewidth}
\includegraphics[width=0.45\linewidth]{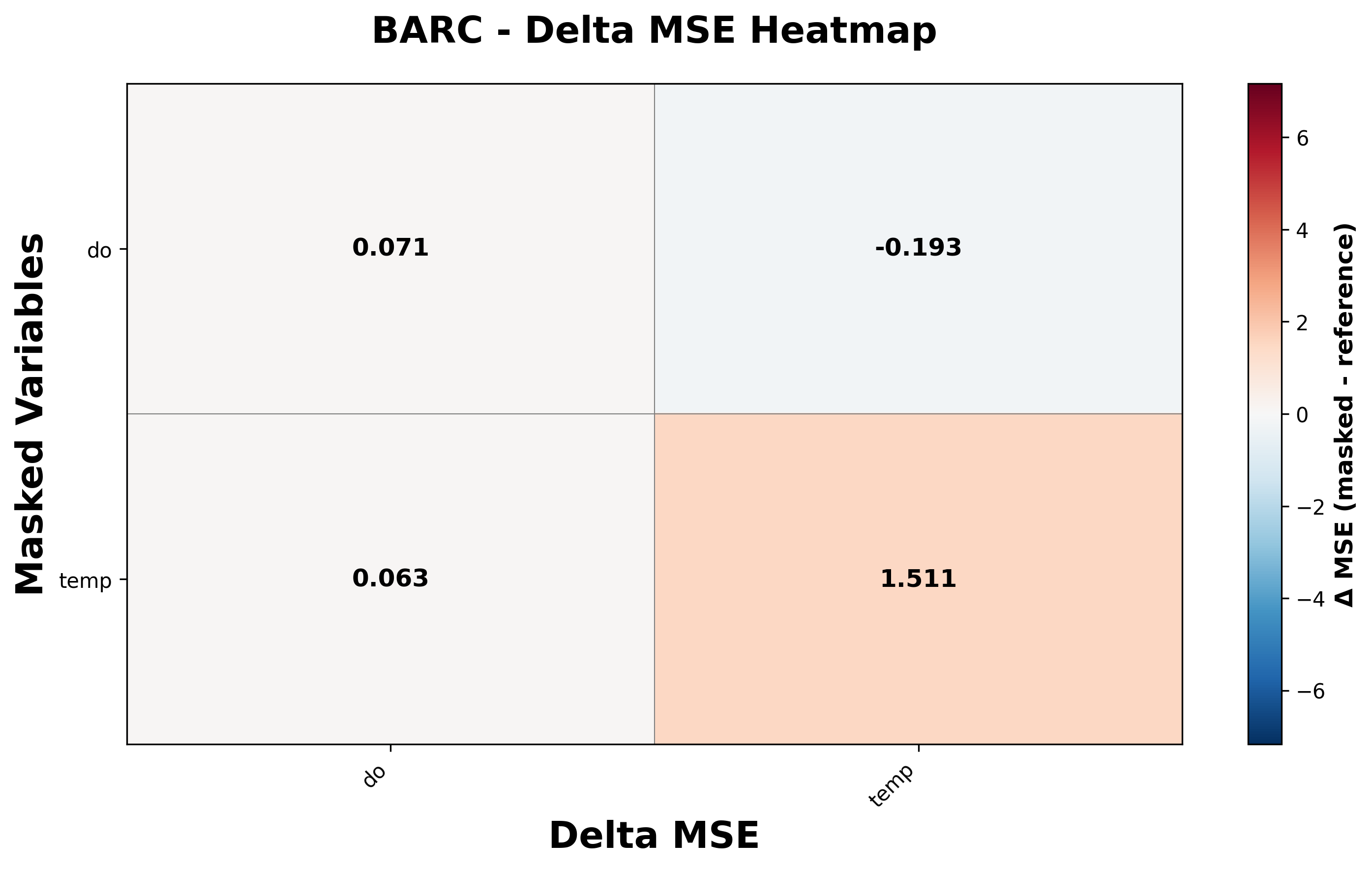}

\includegraphics[width=0.45\linewidth]{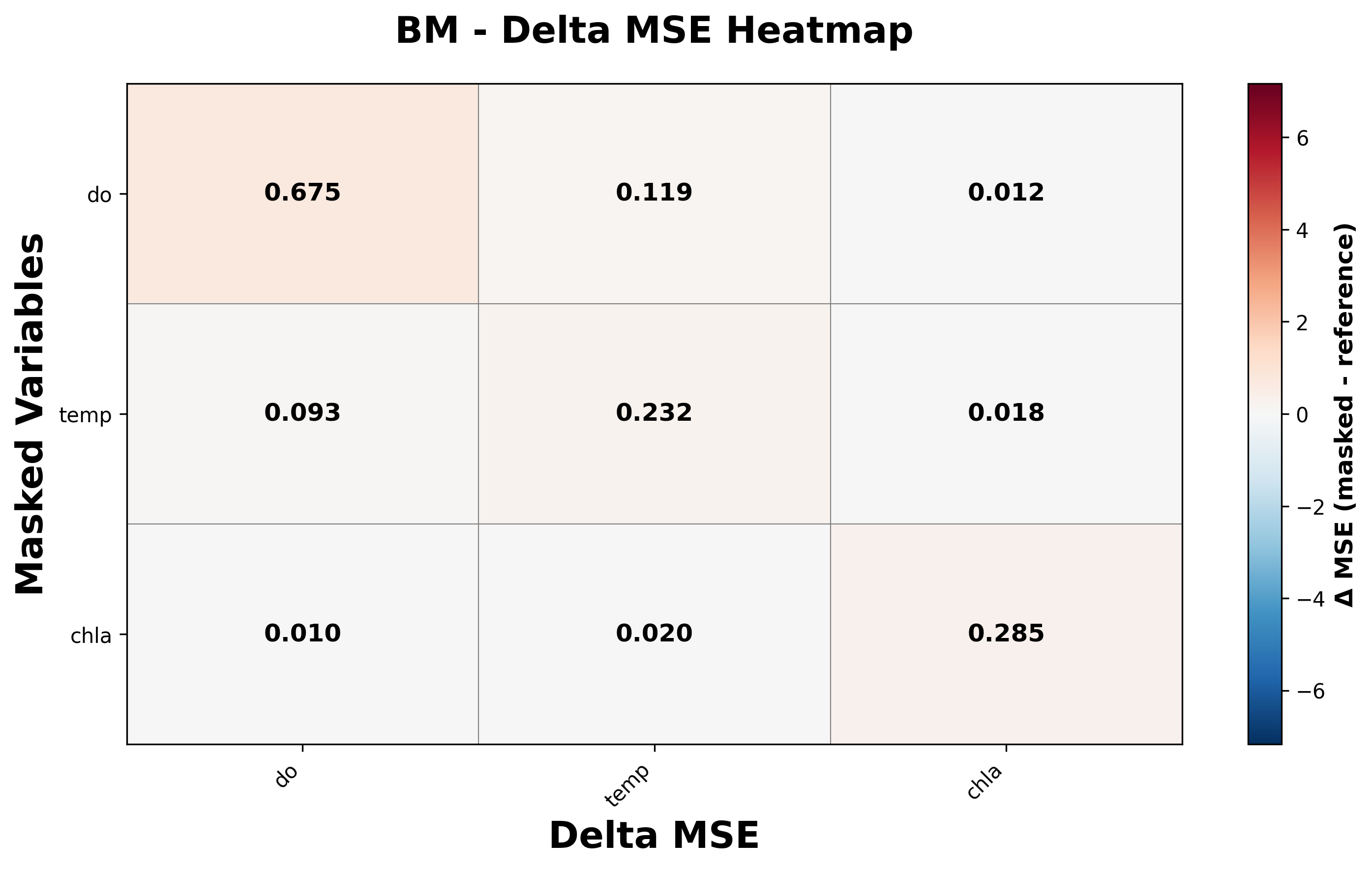}
\includegraphics[width=0.45\linewidth]{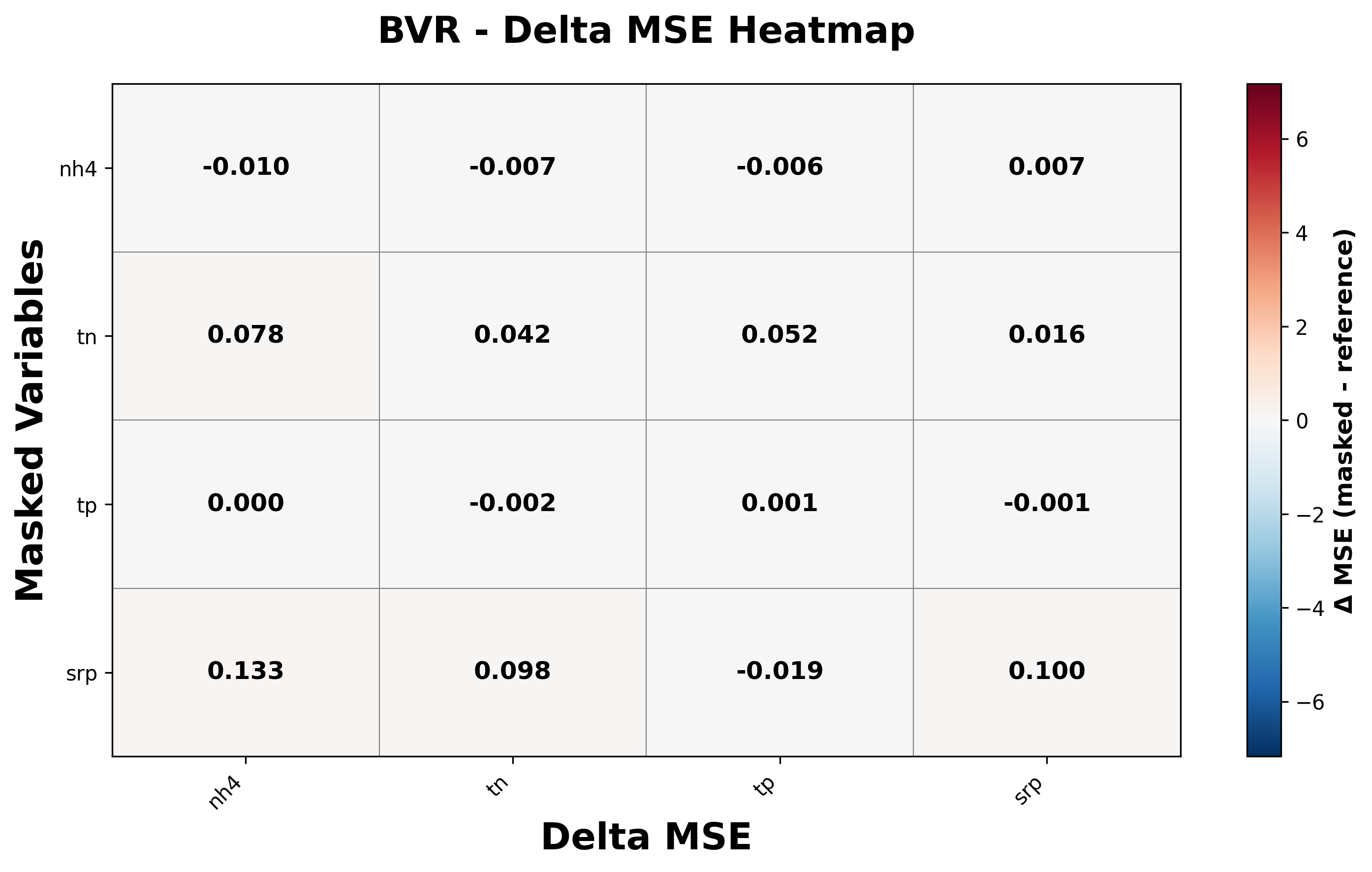}

\includegraphics[width=0.45\linewidth]{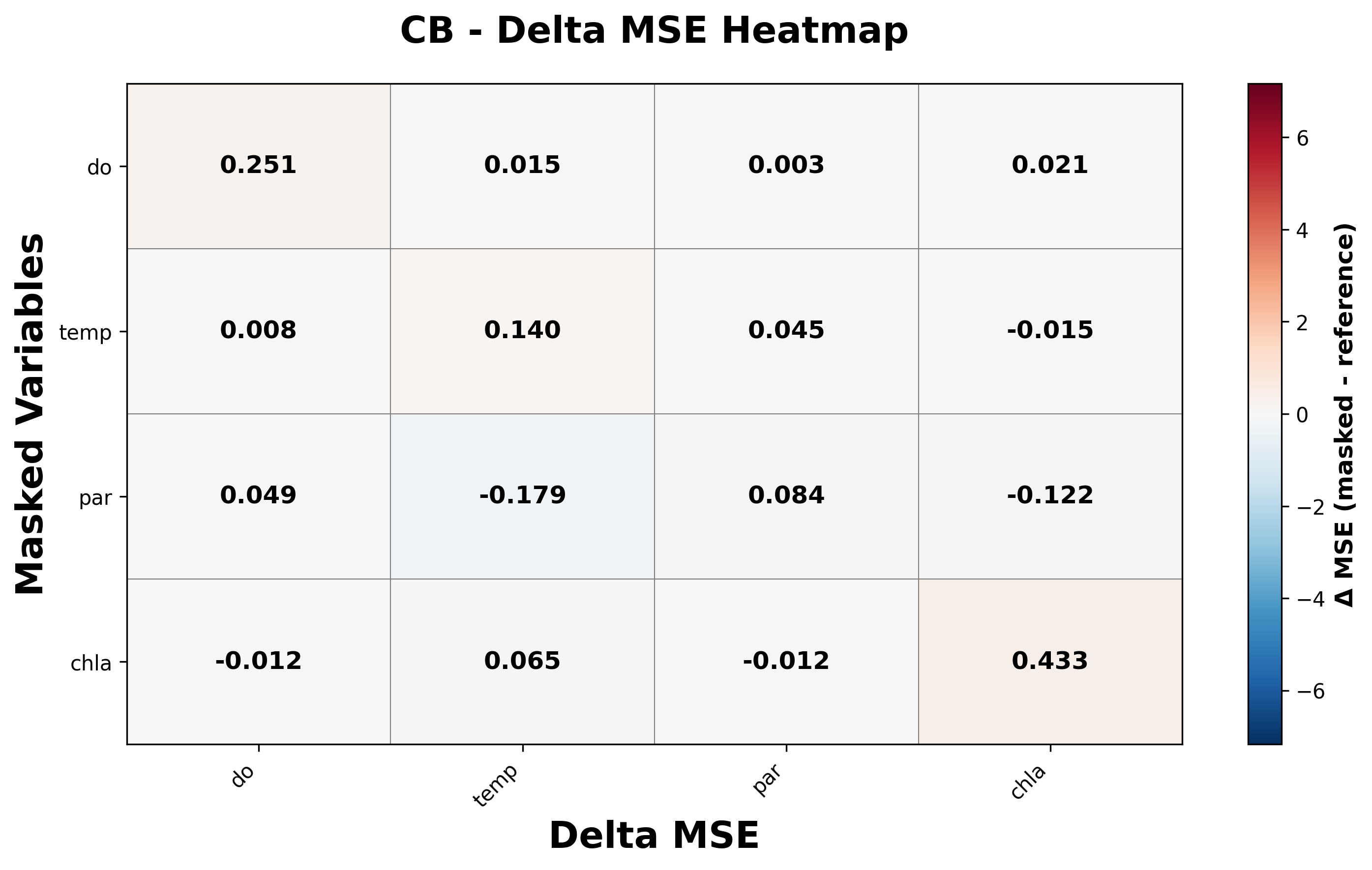}
\includegraphics[width=0.45\linewidth]{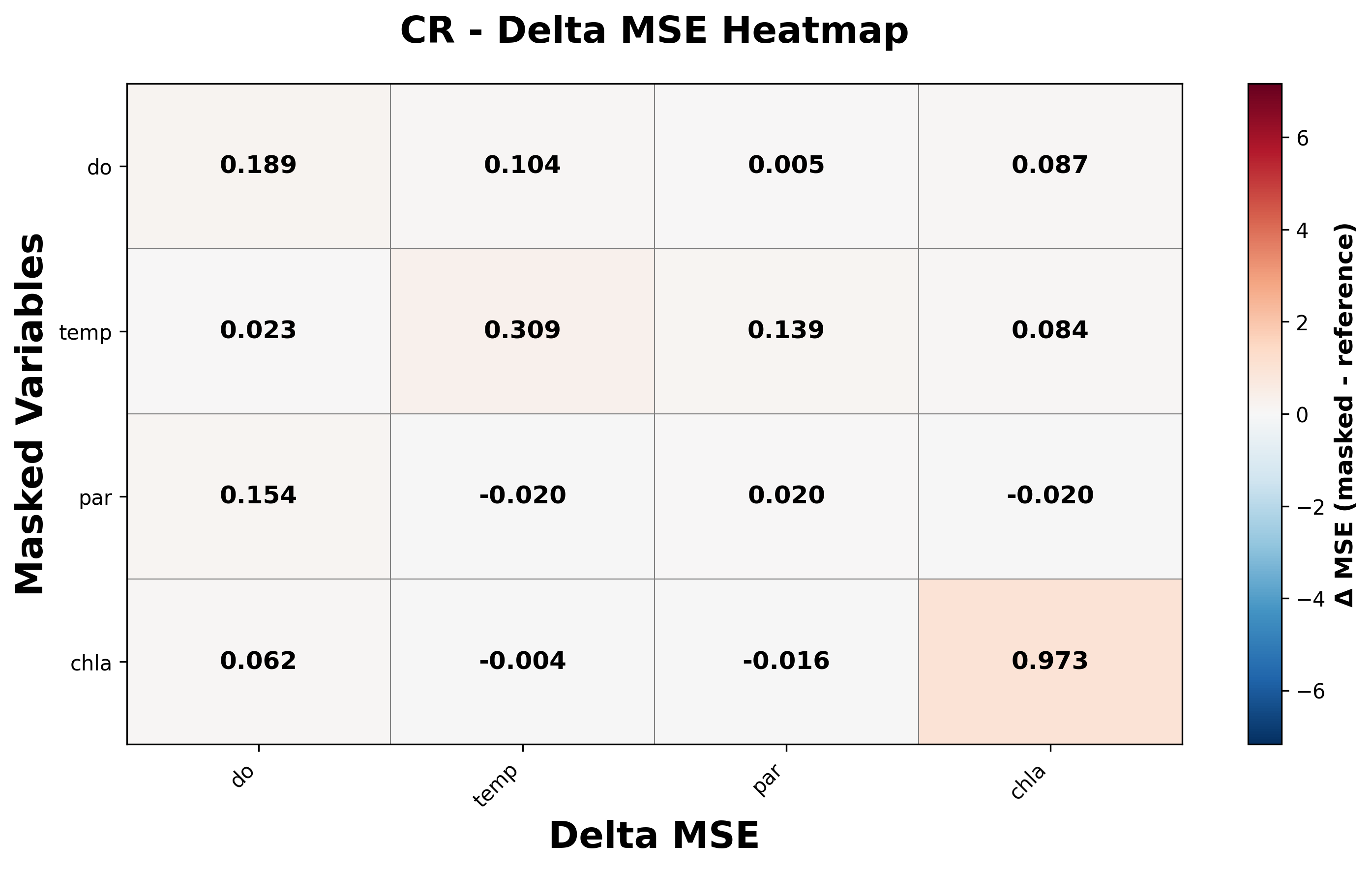}

\includegraphics[width=0.45\linewidth]{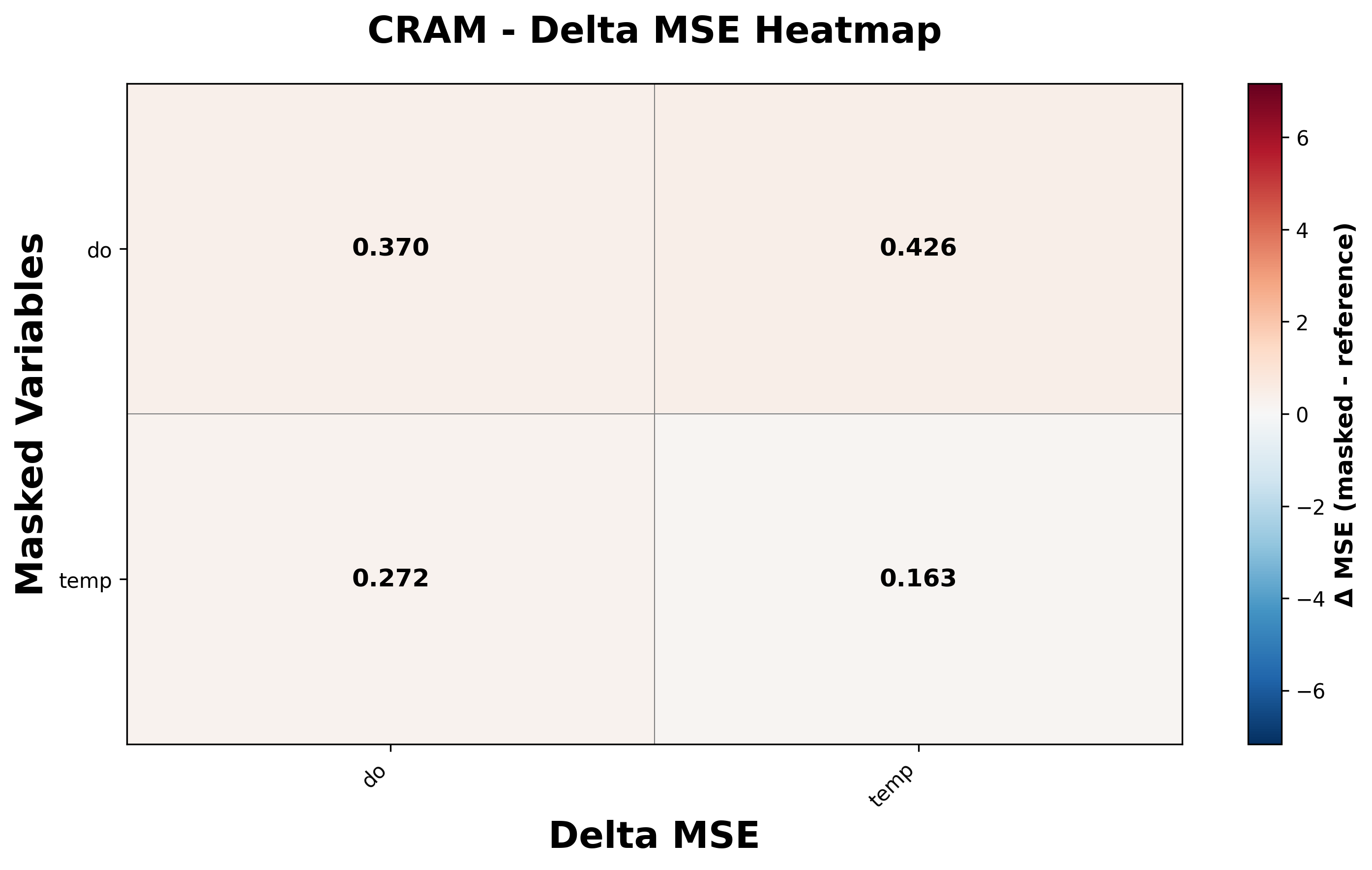}
\includegraphics[width=0.45\linewidth]{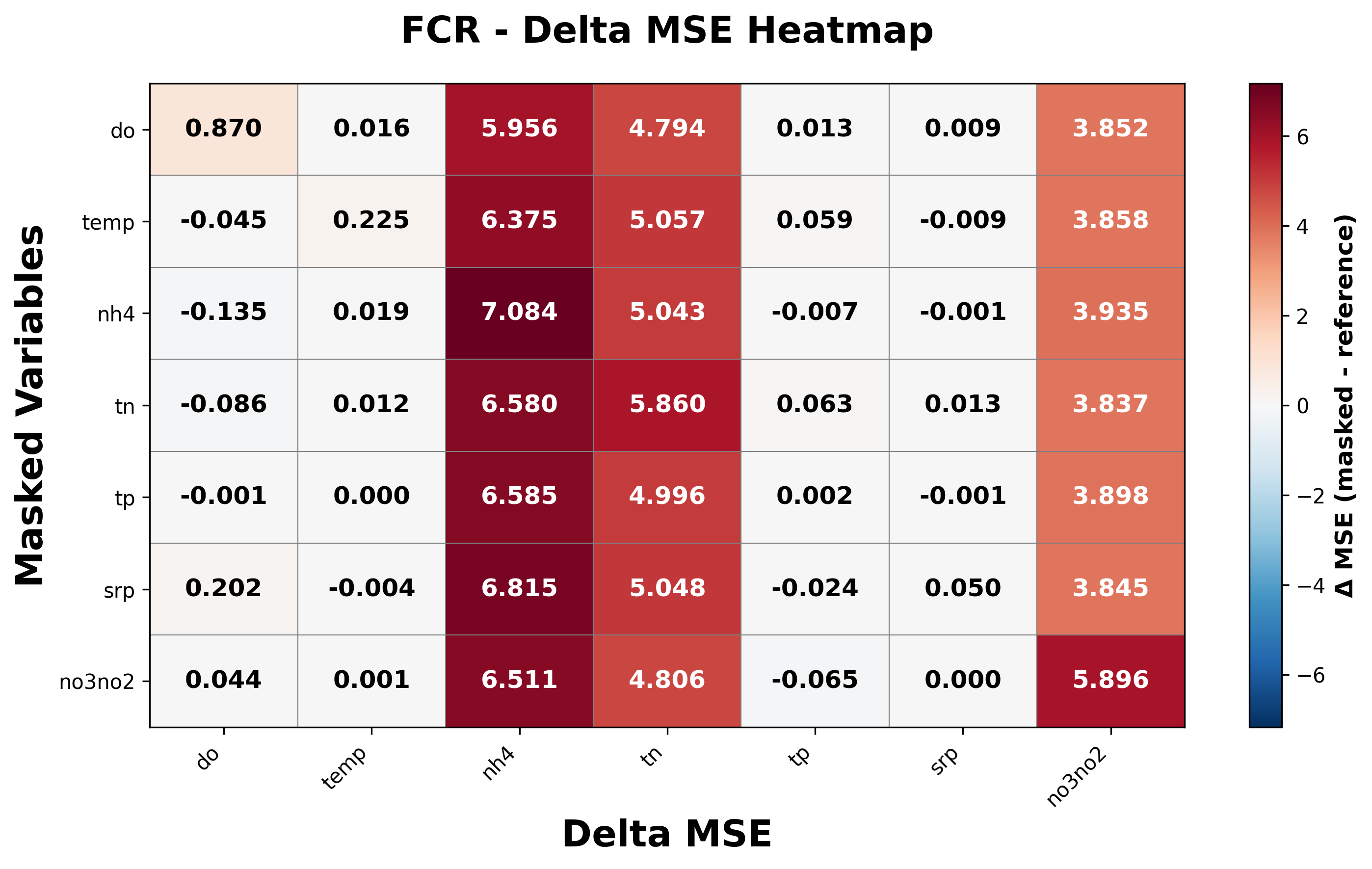}

\end{figure*}

\clearpage
\begin{figure*}[t]
\ContinuedFloat
\centering

\includegraphics[width=0.45\linewidth]{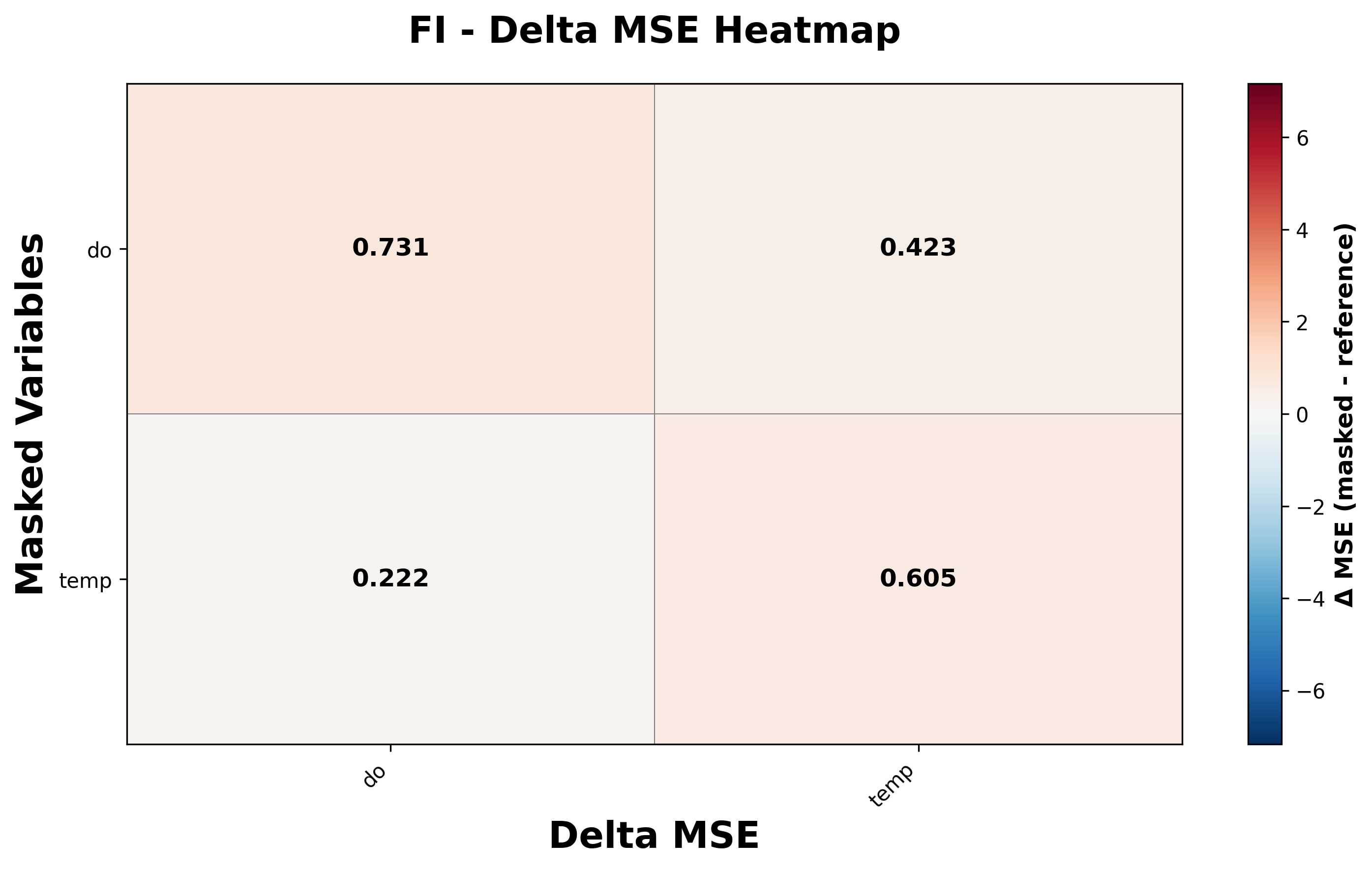}
\includegraphics[width=0.45\linewidth]{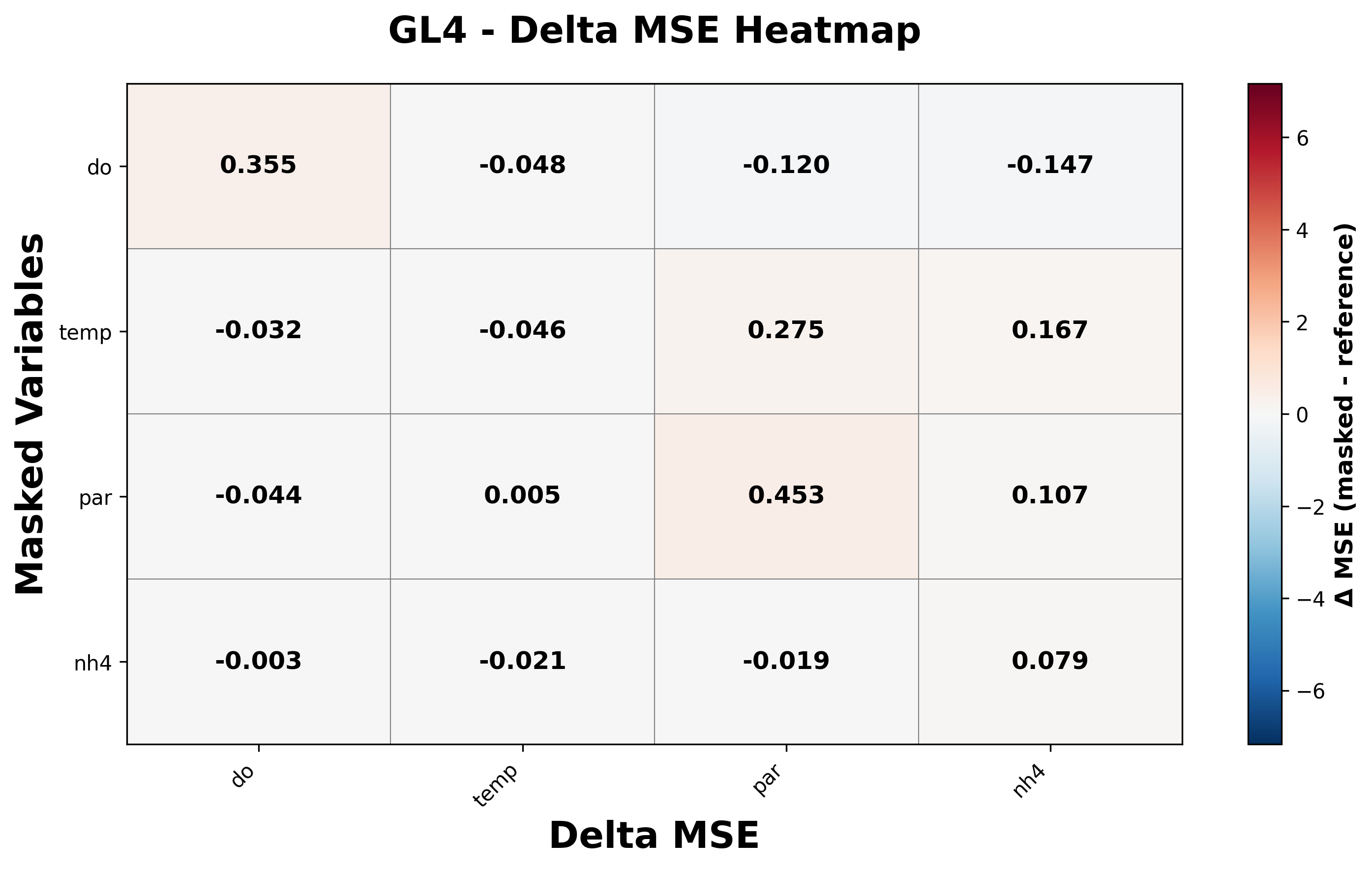}

\includegraphics[width=0.45\linewidth]{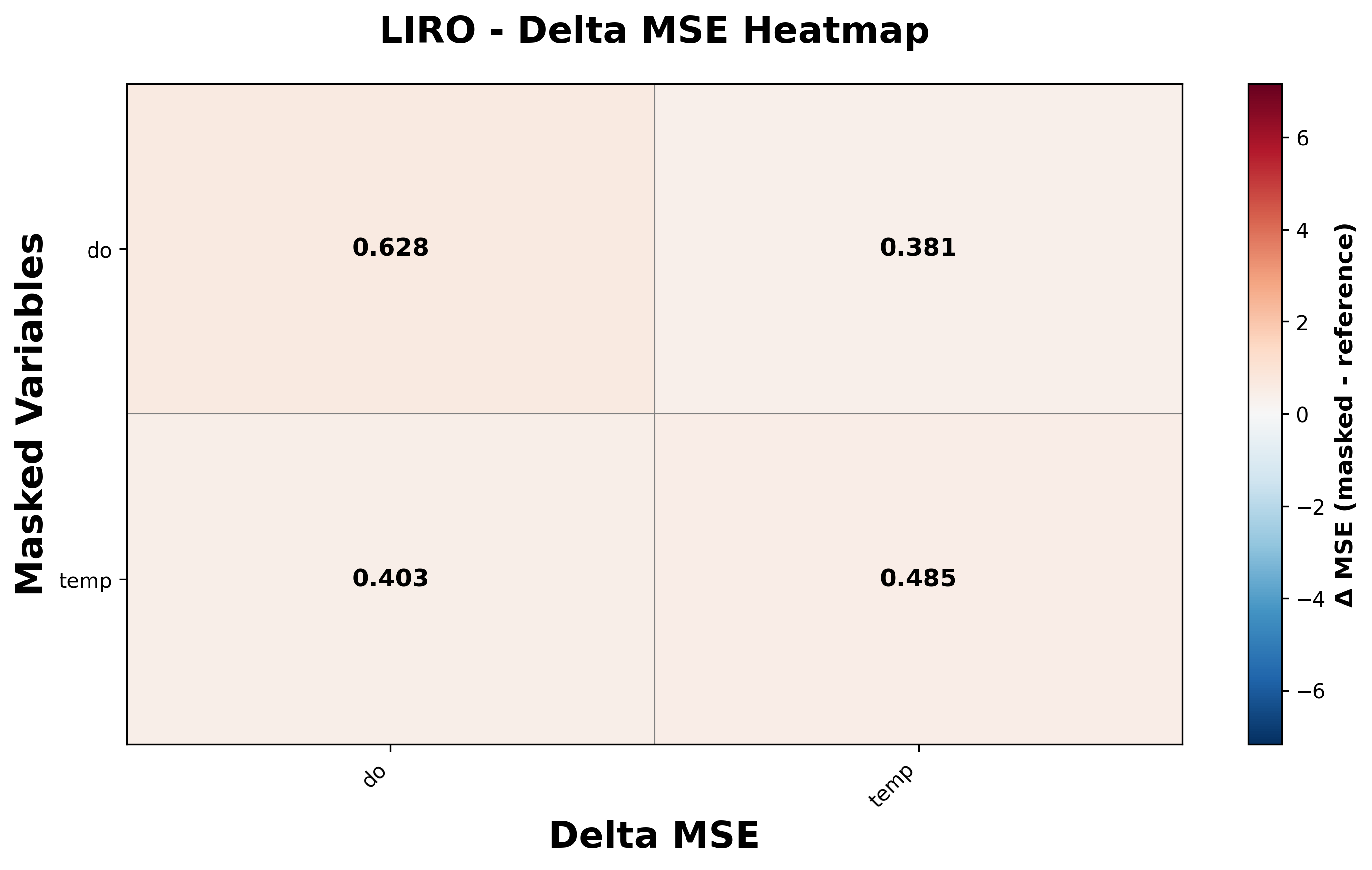}
\includegraphics[width=0.45\linewidth]{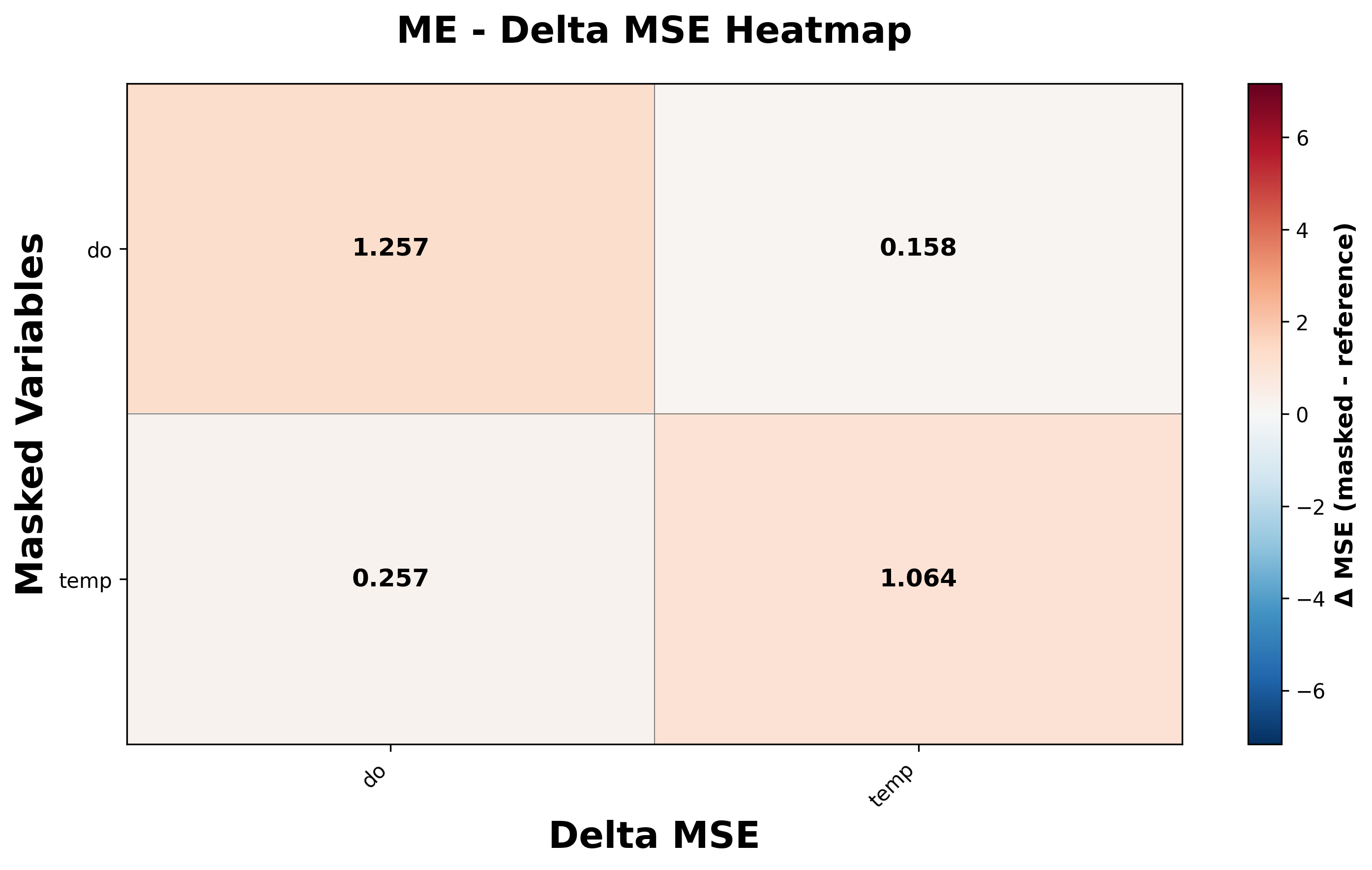}

\includegraphics[width=0.45\linewidth]{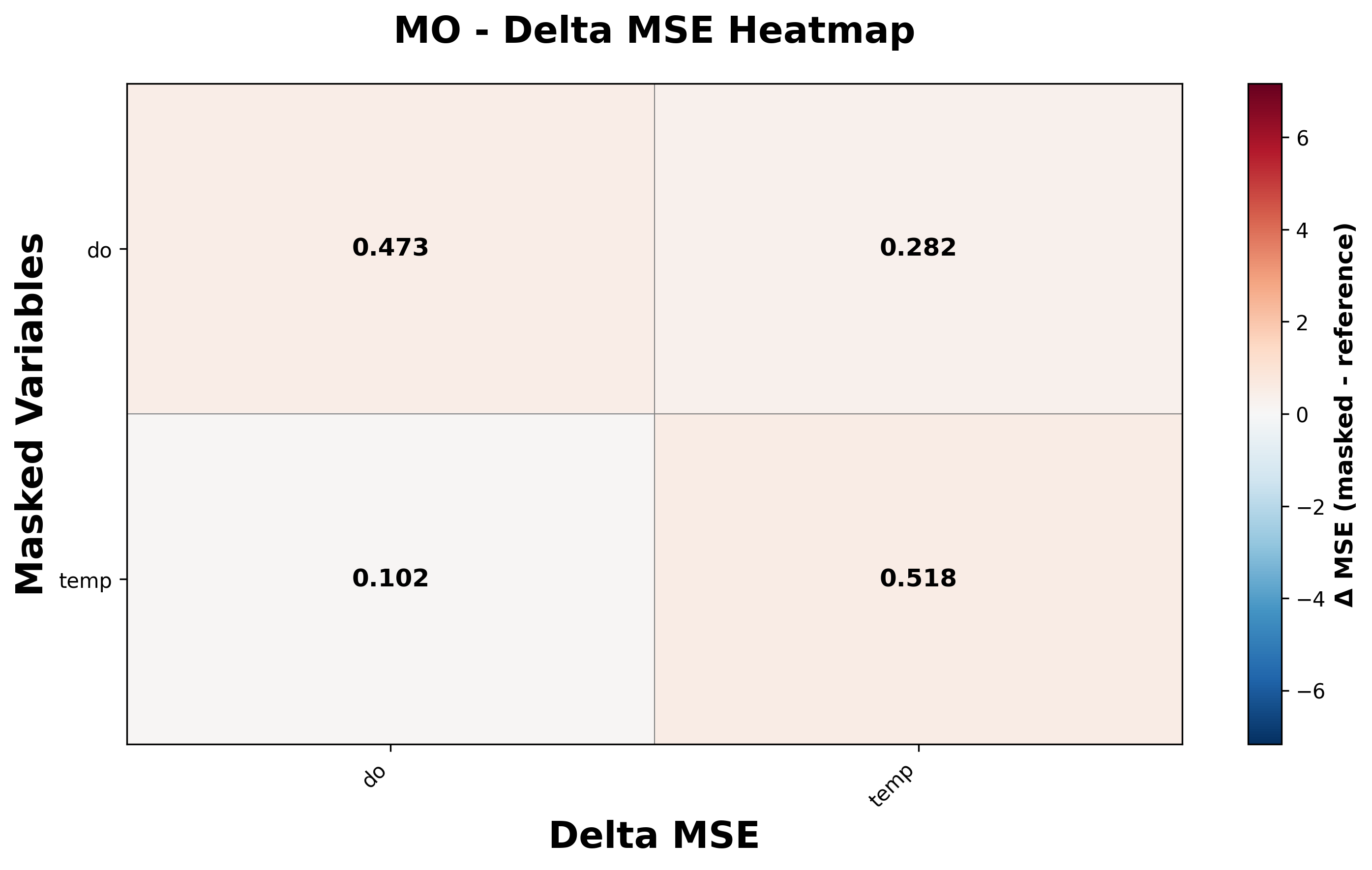}
\includegraphics[width=0.45\linewidth]{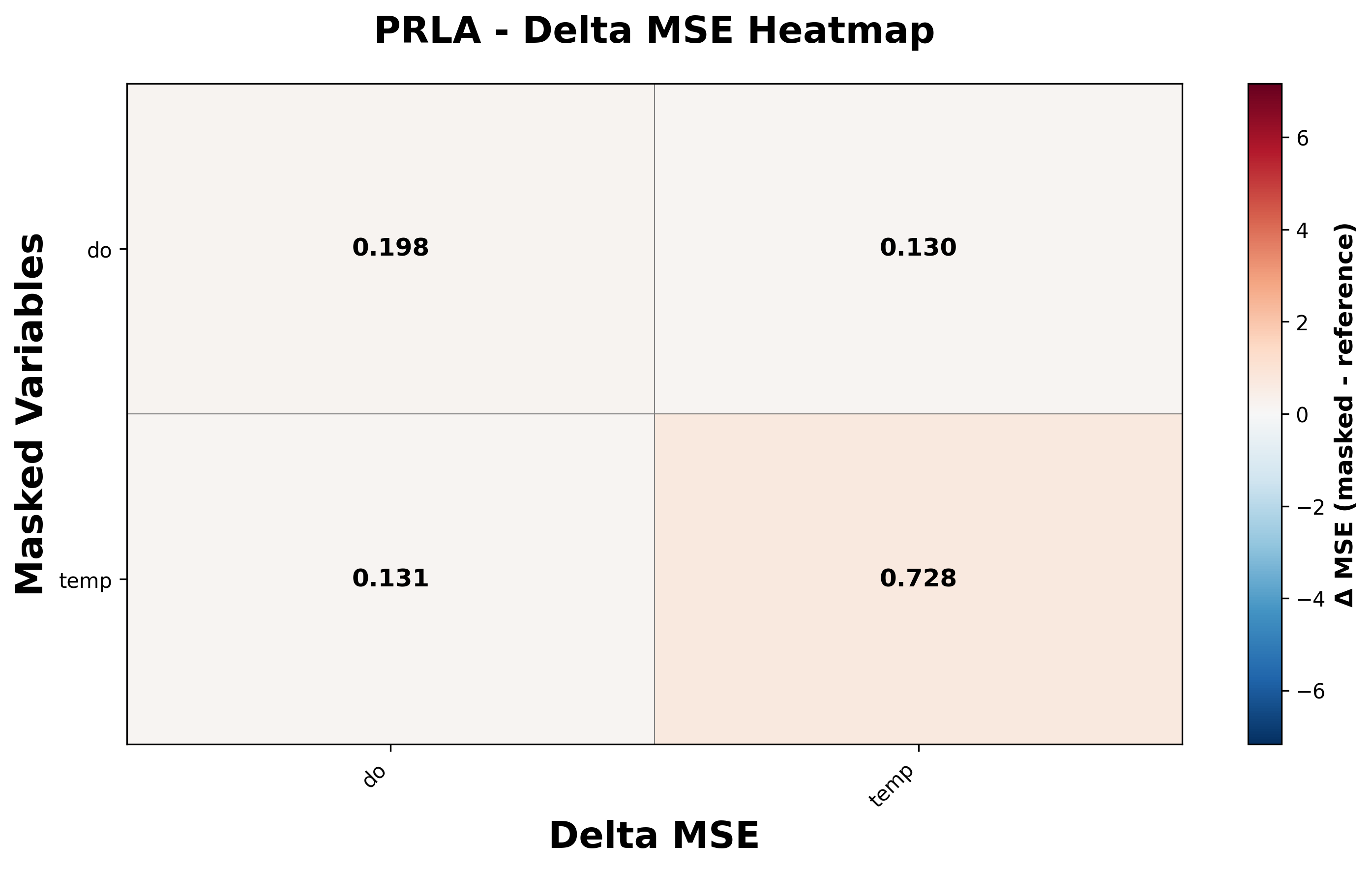}

\includegraphics[width=0.45\linewidth]{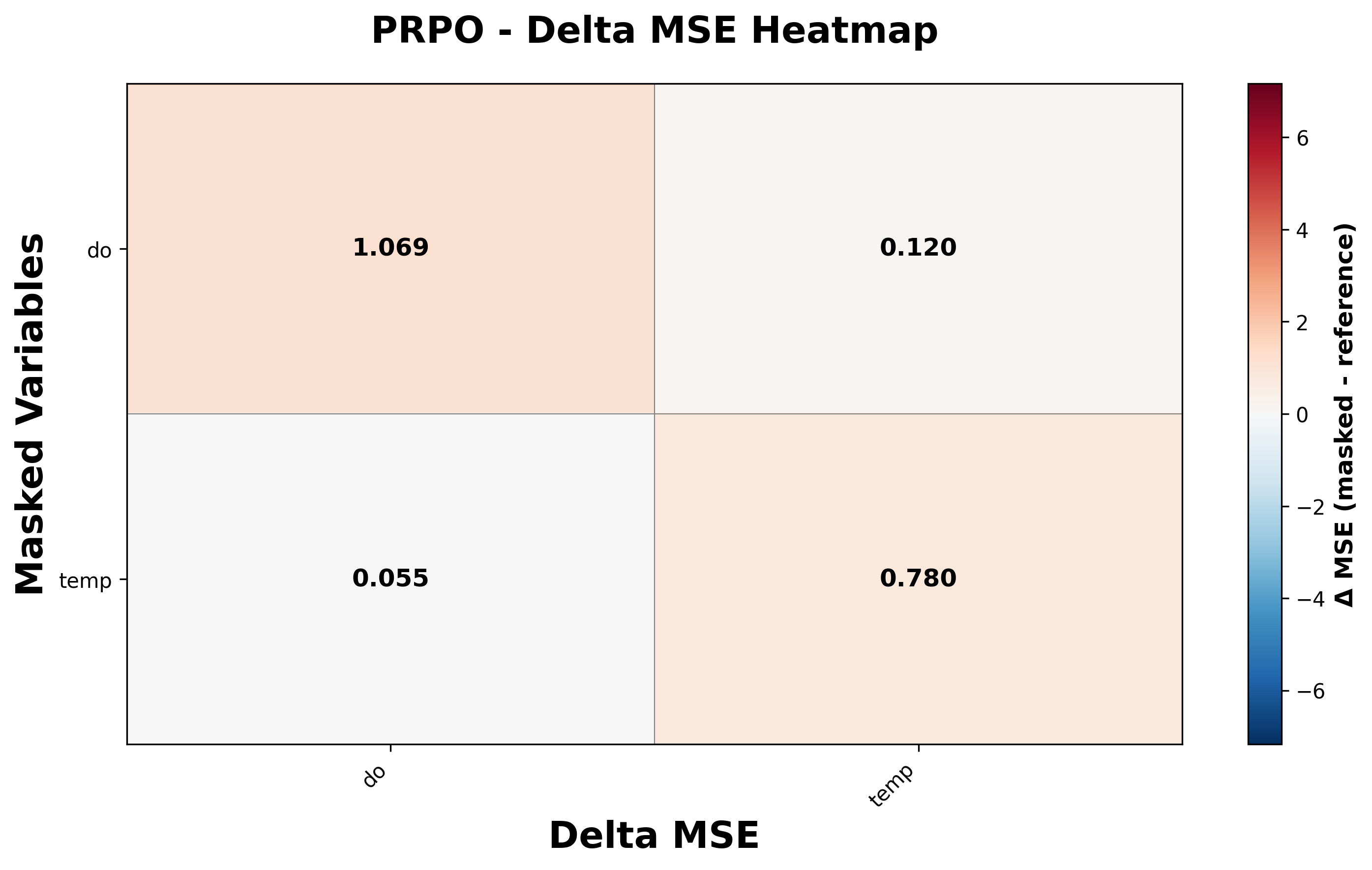}
\includegraphics[width=0.45\linewidth]{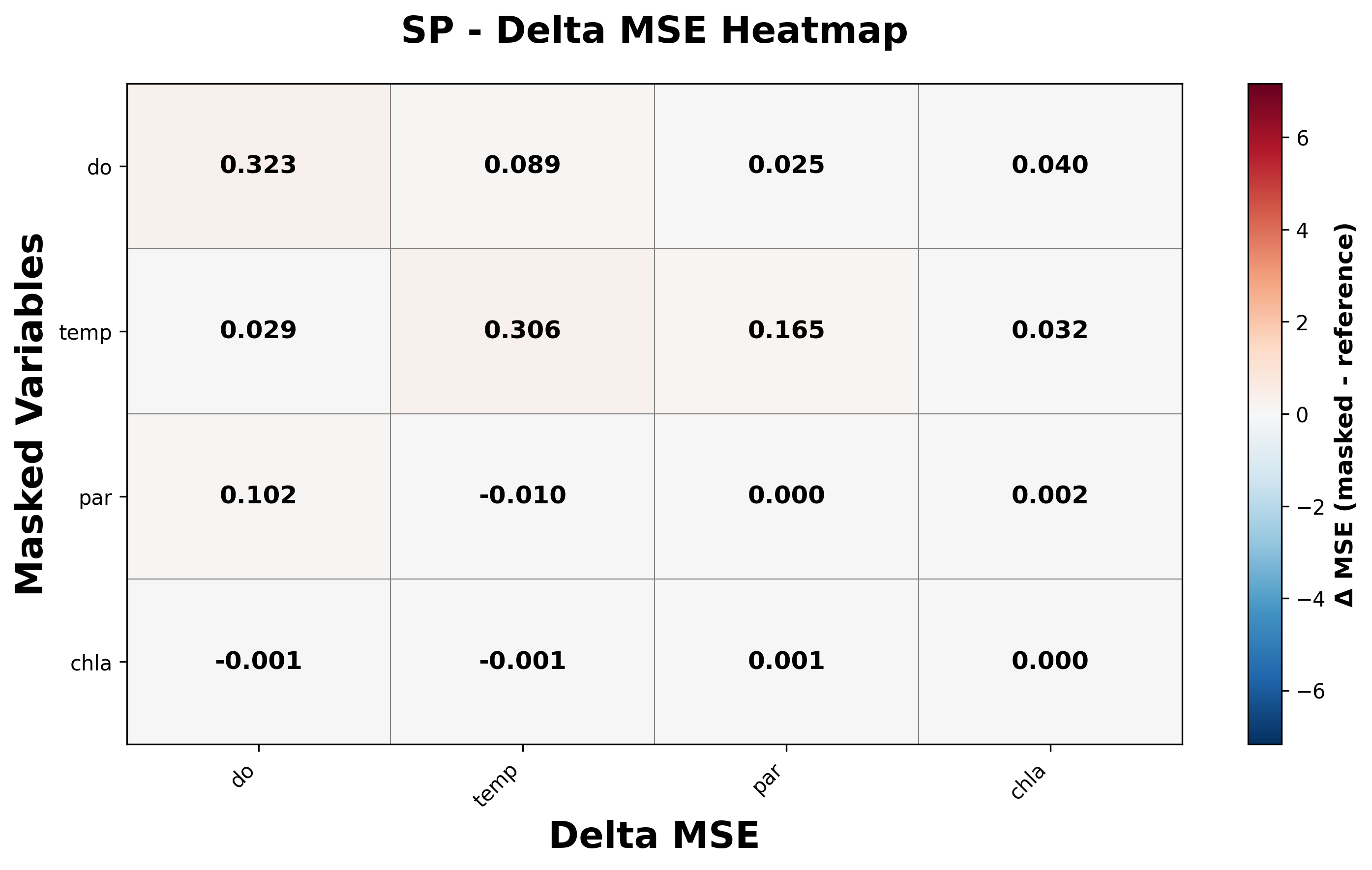}
\end{figure*}

\clearpage
\begin{figure*}[t]
\ContinuedFloat
\centering
\includegraphics[width=0.4\linewidth]{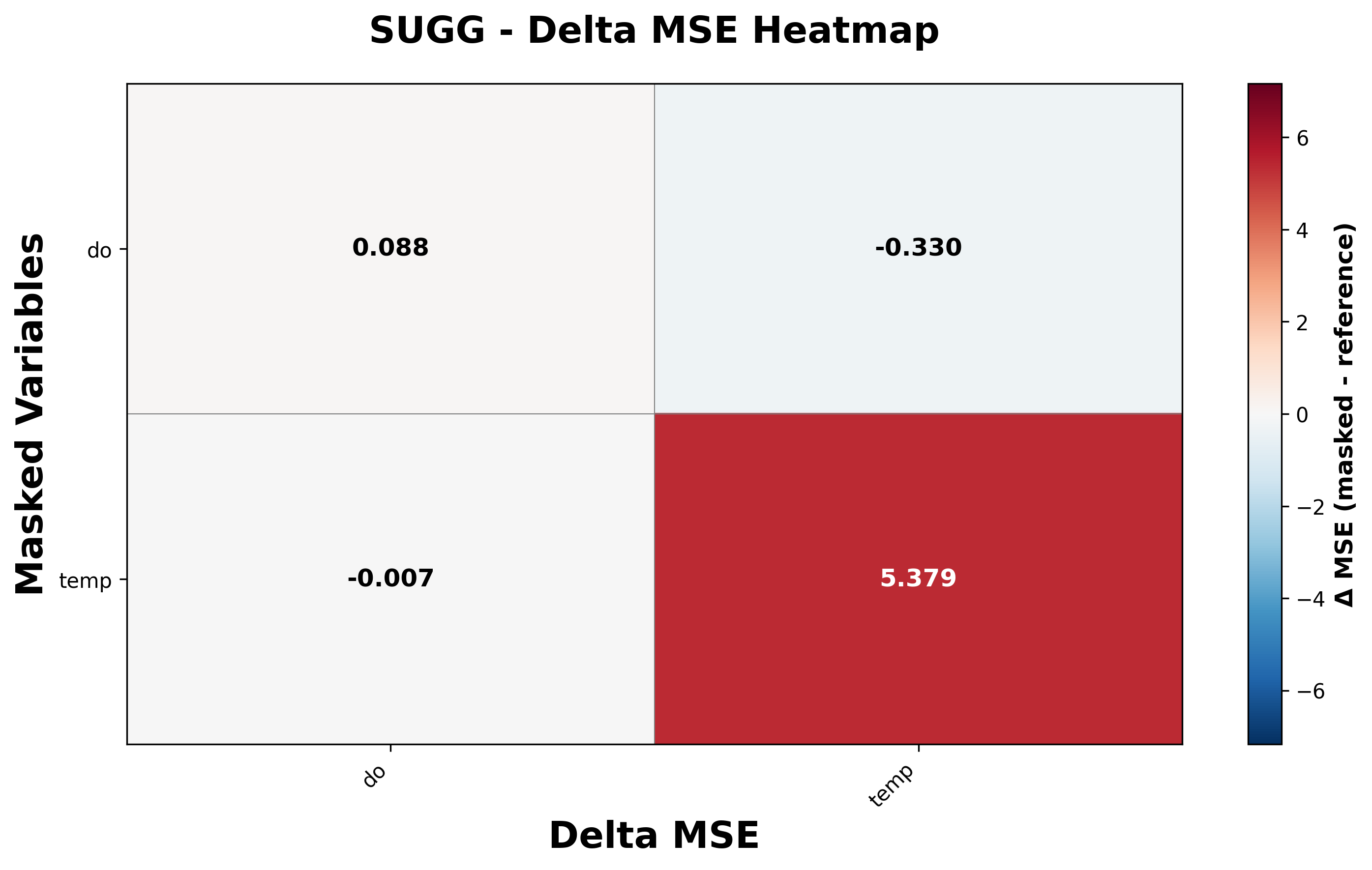}
\includegraphics[width=0.4\linewidth]{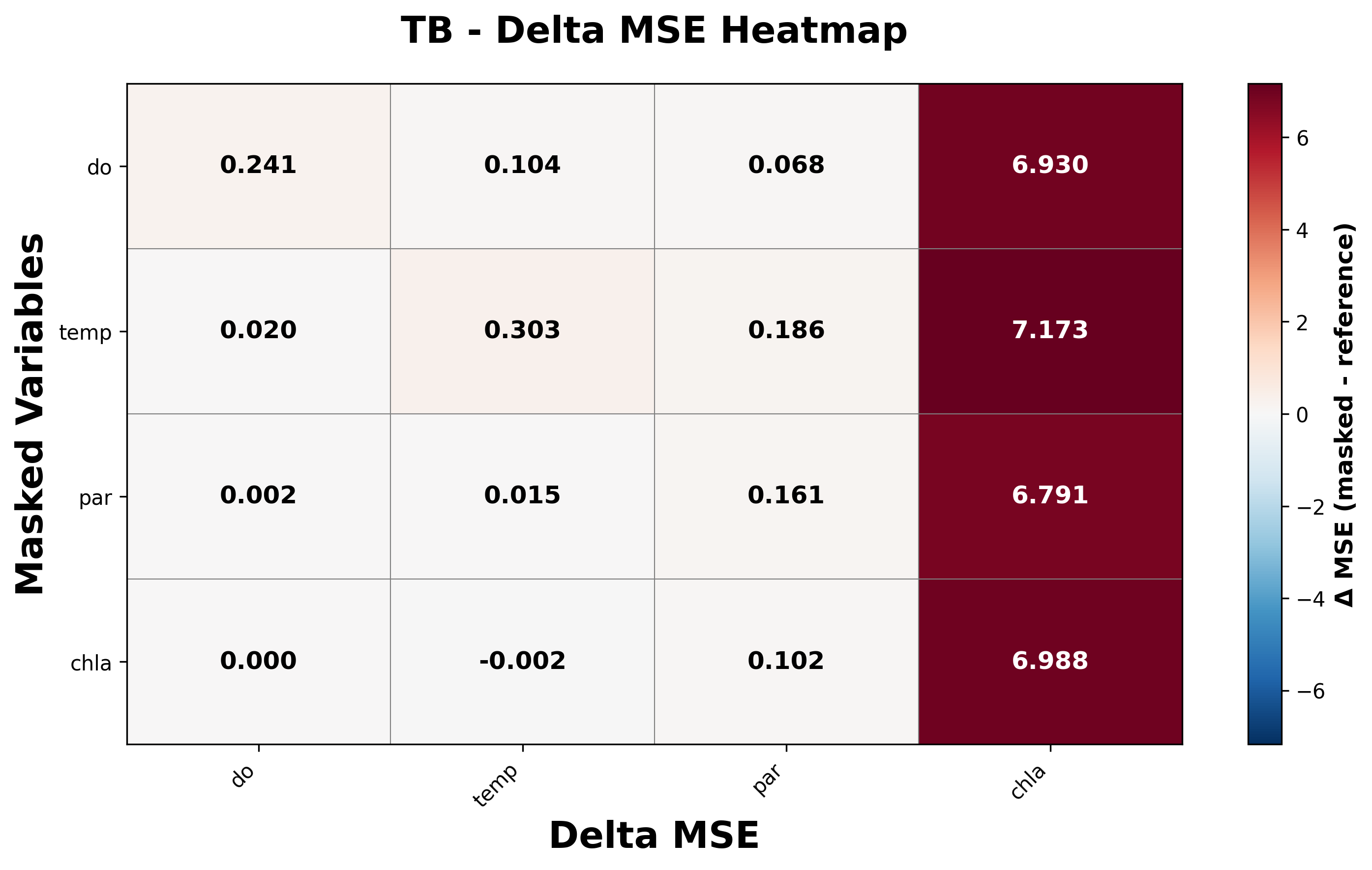}

\includegraphics[width=0.4\linewidth]{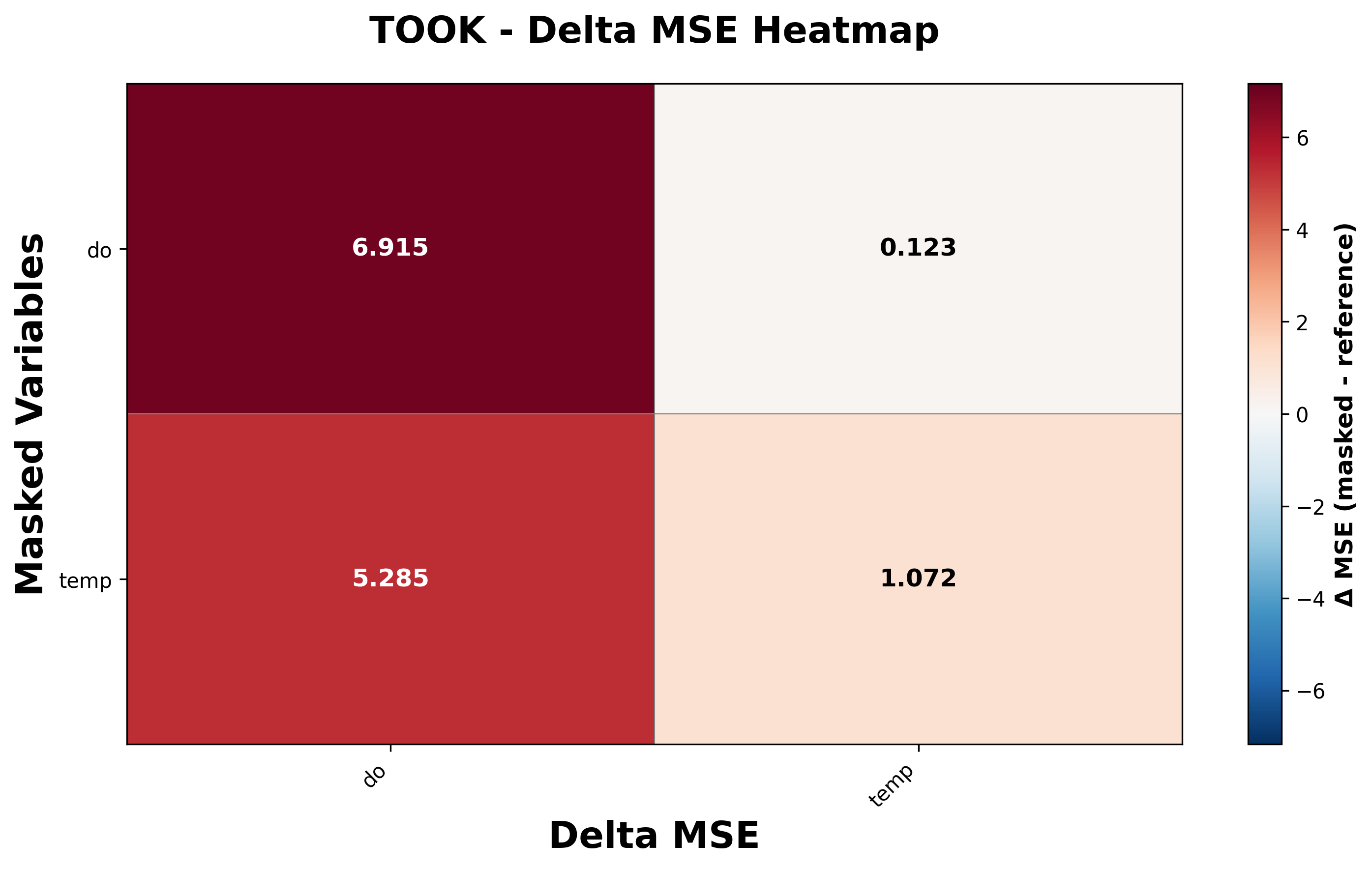}
\includegraphics[width=0.4\linewidth]{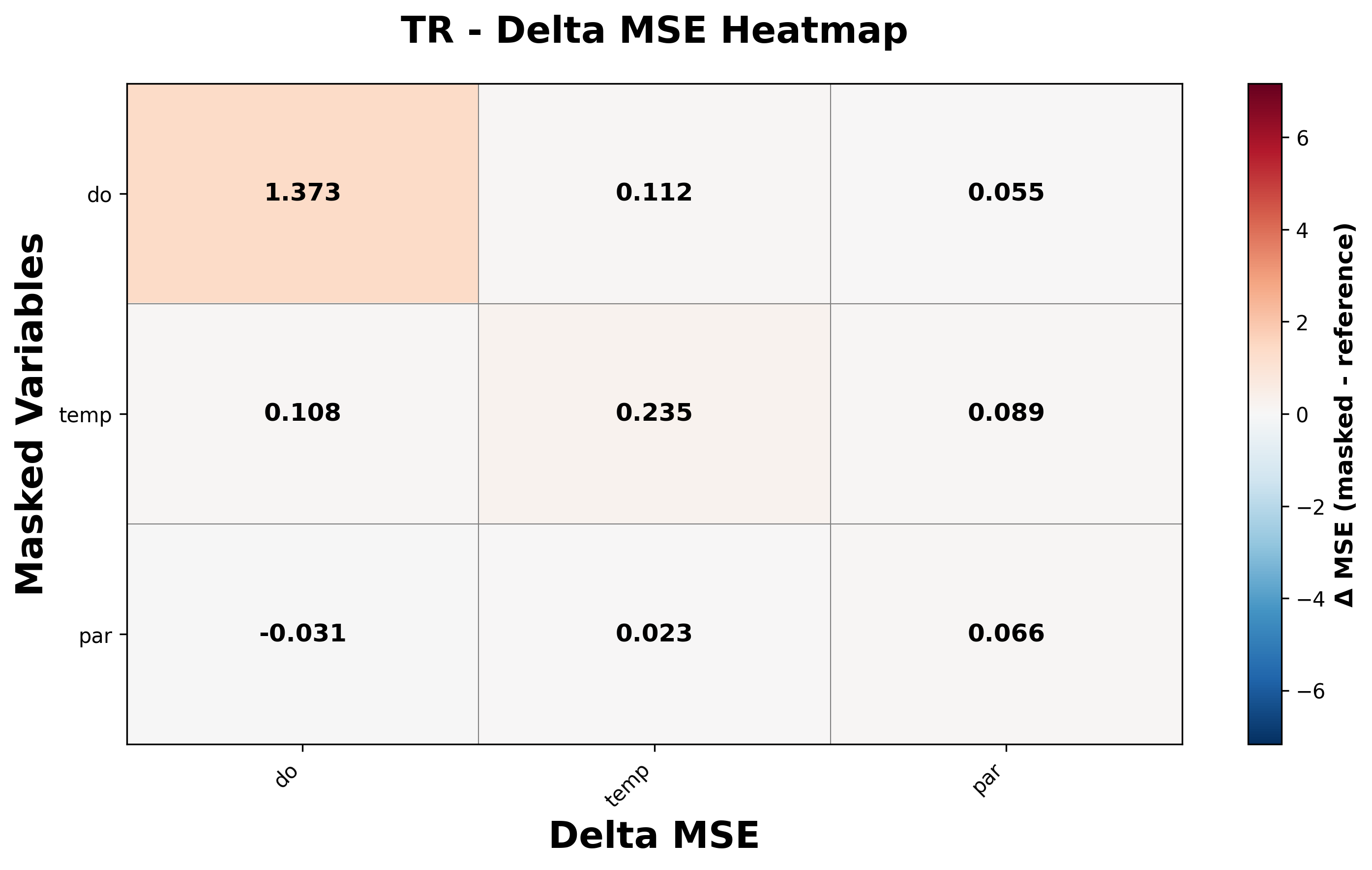}

\includegraphics[width=0.4\linewidth]{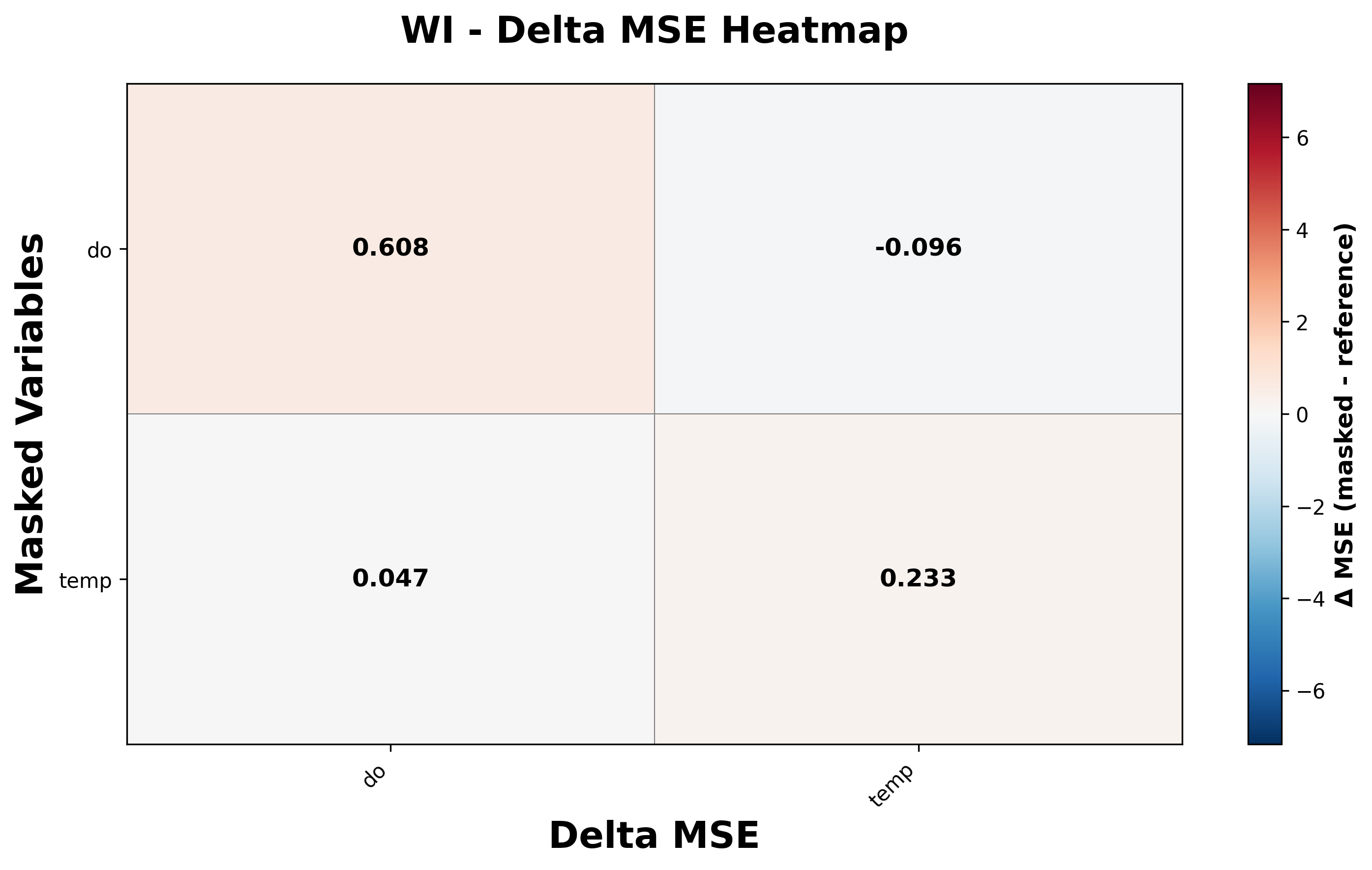}
\end{figure*}

\newpage
\subsection{Variate Importance - Single Variate Input}
\label{sec:appendix-single-variate-input}
In this experiment, we mask out $V-1$ variates (out of $V$ variates). That is, only one variable is present in the model's input/historical window. We measure the change in the predictive performance of each of the variables (including the variate masked). Figure \ref{fig:per_lake_delta_heatmaps_v1} shows the heatmaps corresponding to each lake. 
\begin{figure*}[t]
\centering
\caption{Per-lake performance deltas, on passing a single variate as input, visualized as heatmaps. For each lake, we show $\Delta$MSE relative to the baseline (no masking). The left axis represents the single input variable passed.}
\label{fig:per_lake_delta_heatmaps_v1}

\includegraphics[width=0.45\linewidth]{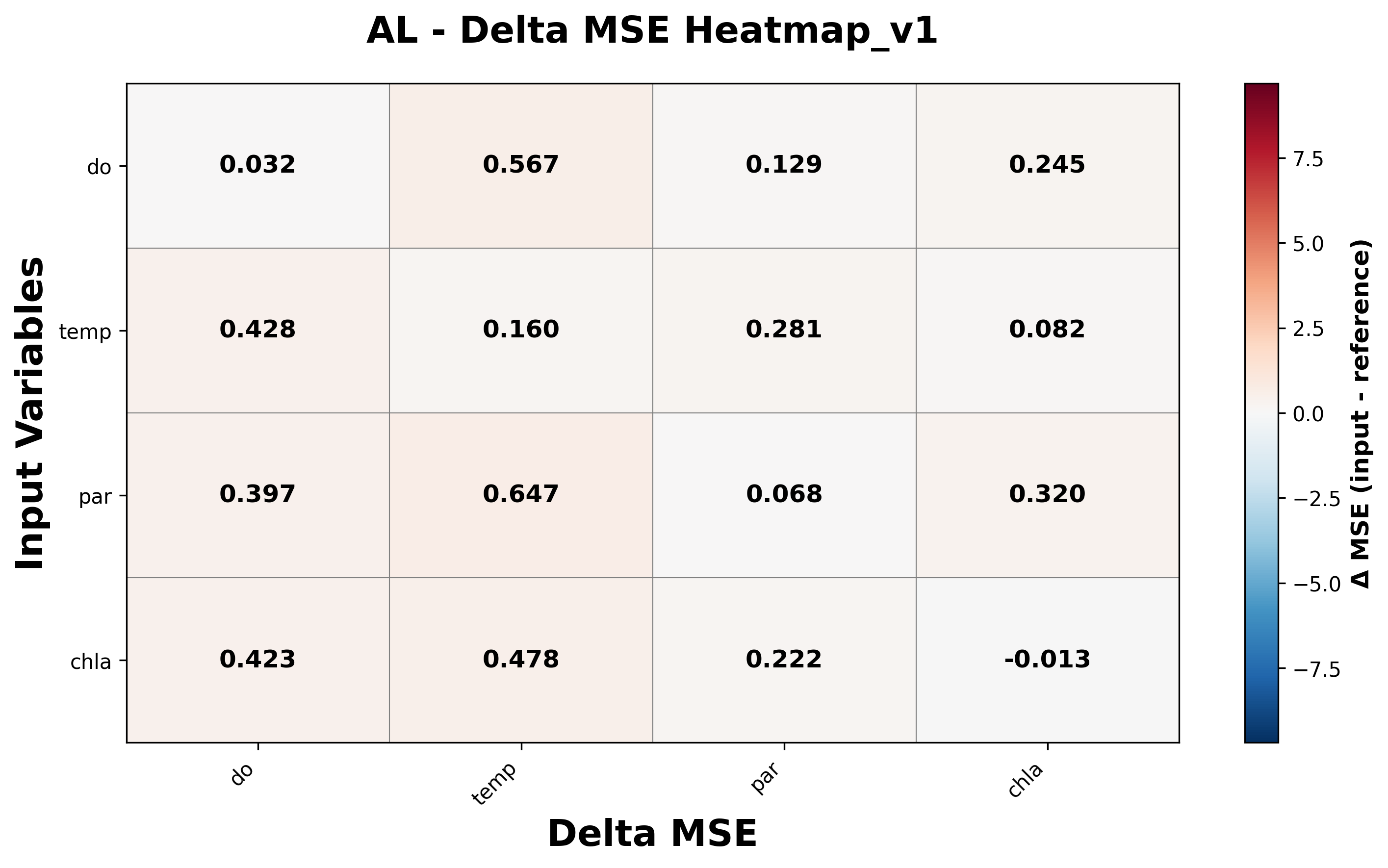}
\includegraphics[width=0.45\linewidth]{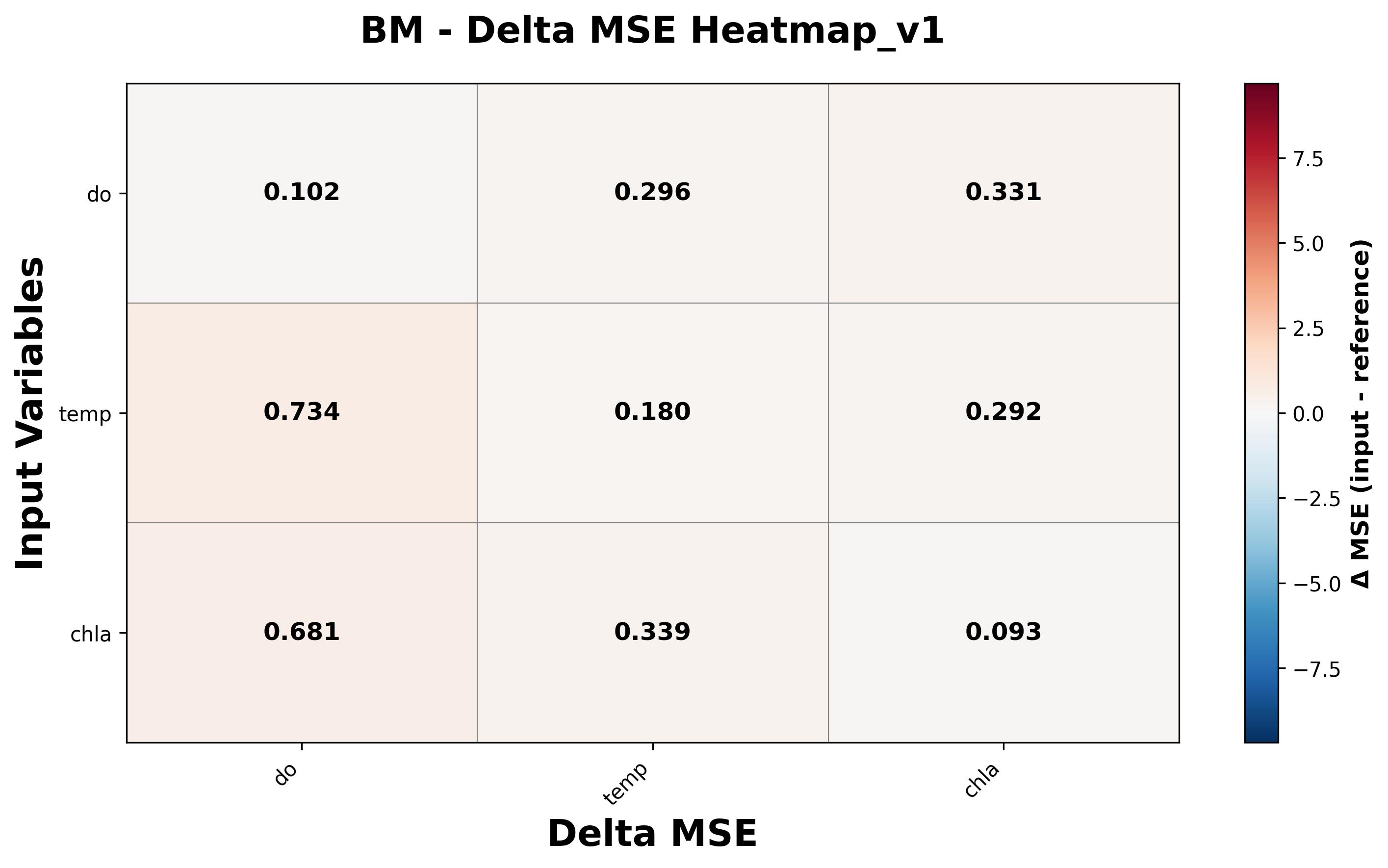}

\includegraphics[width=0.45\linewidth]{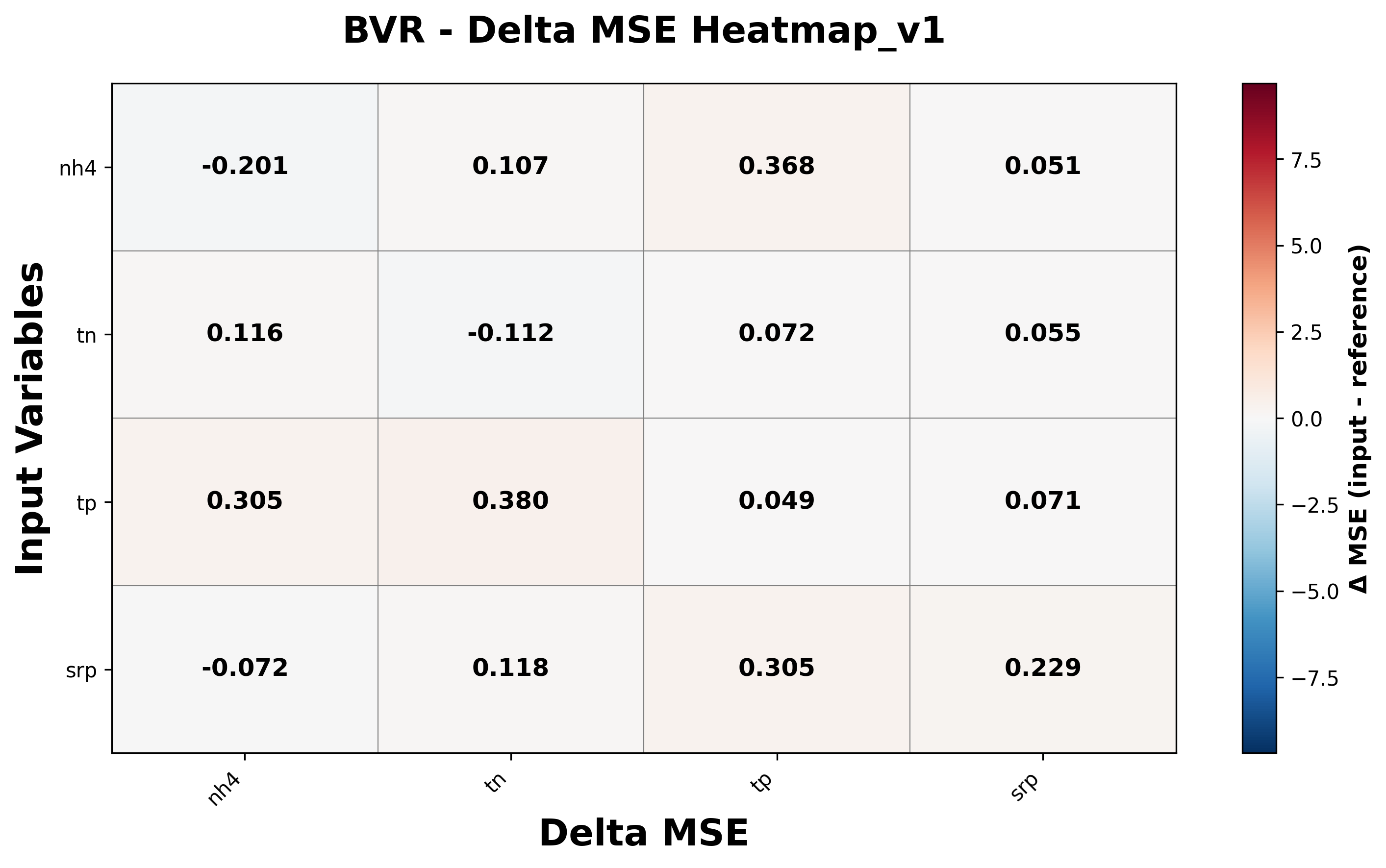}
\includegraphics[width=0.45\linewidth]{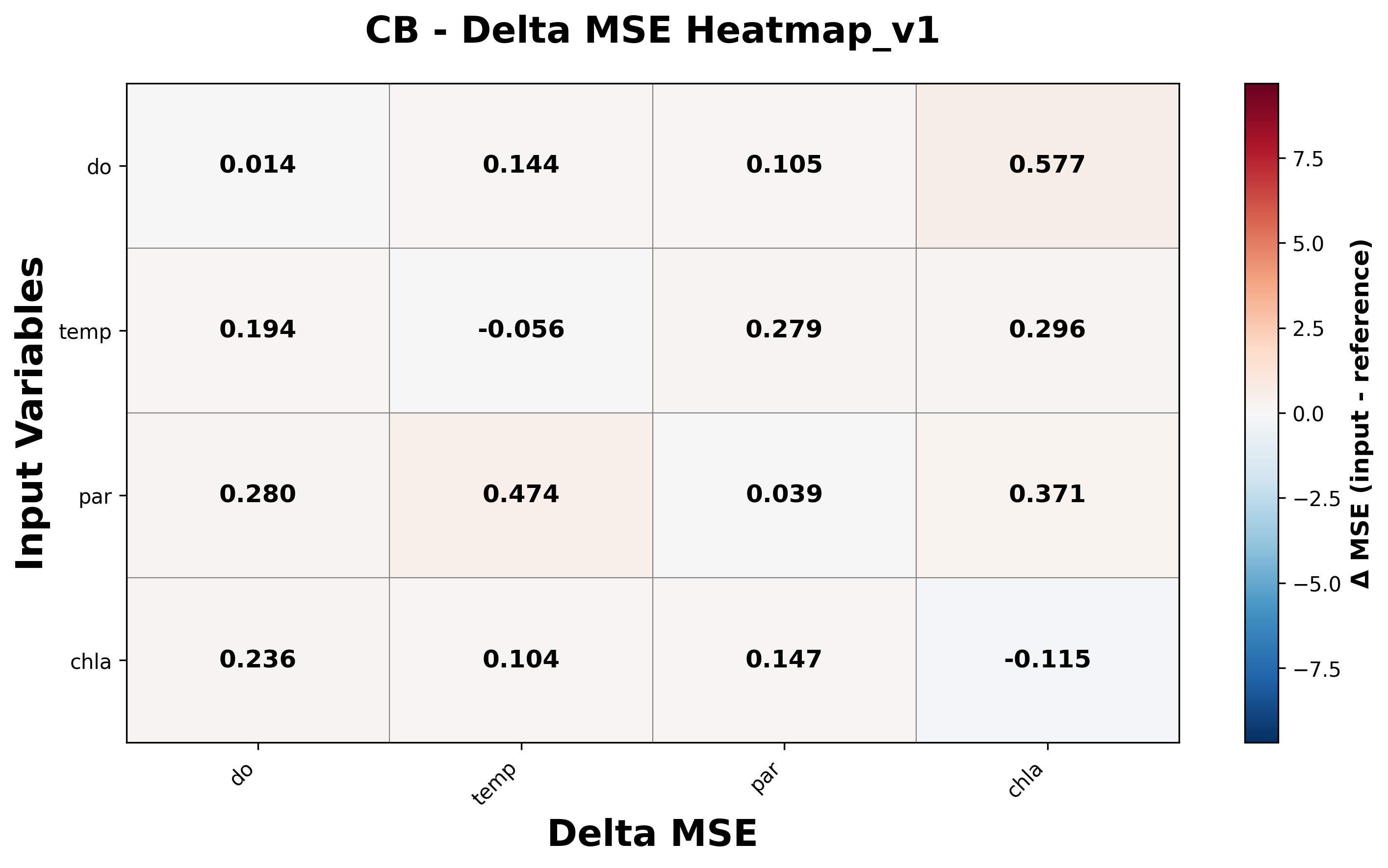}

\includegraphics[width=0.45\linewidth]{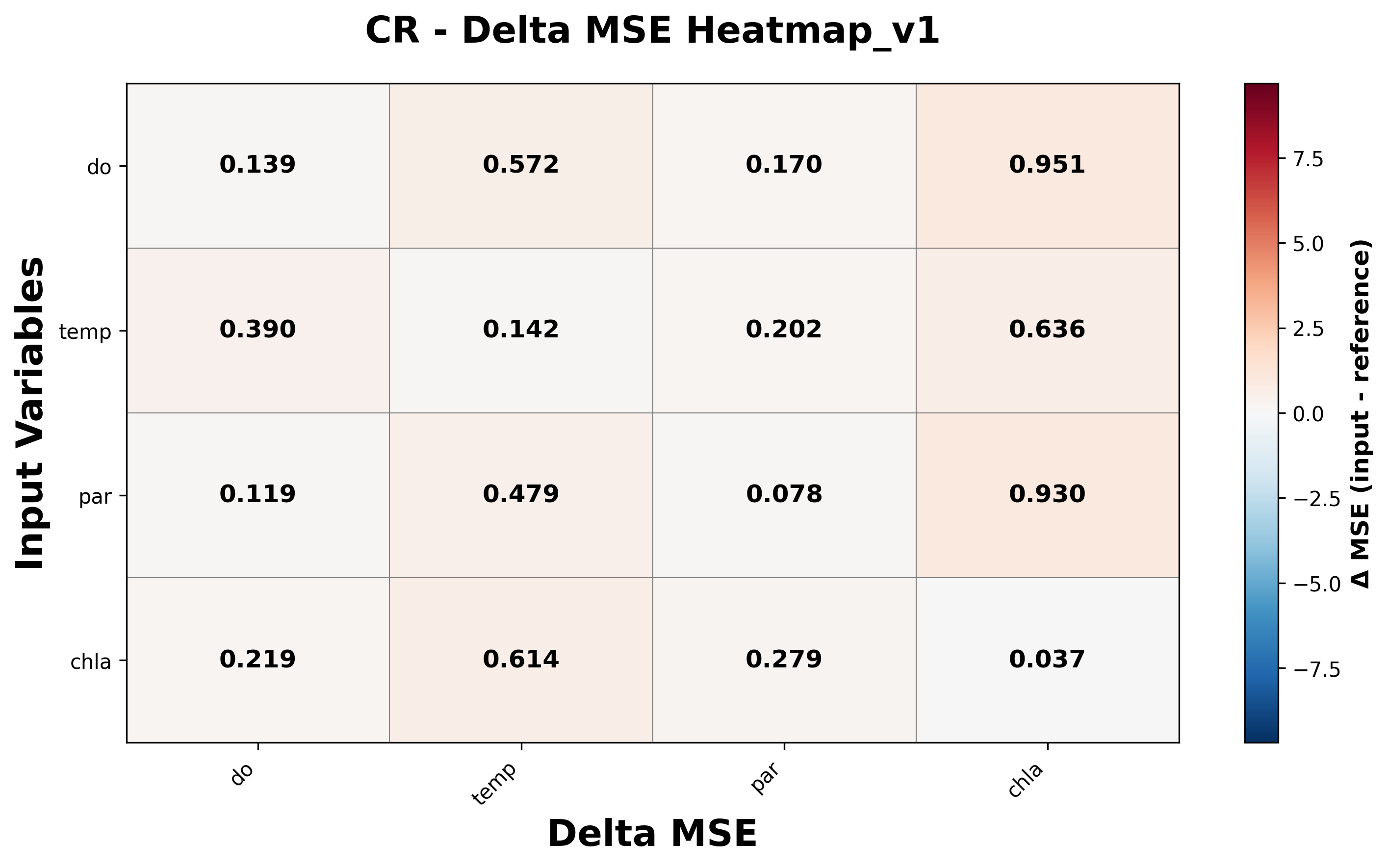}
\includegraphics[width=0.45\linewidth]{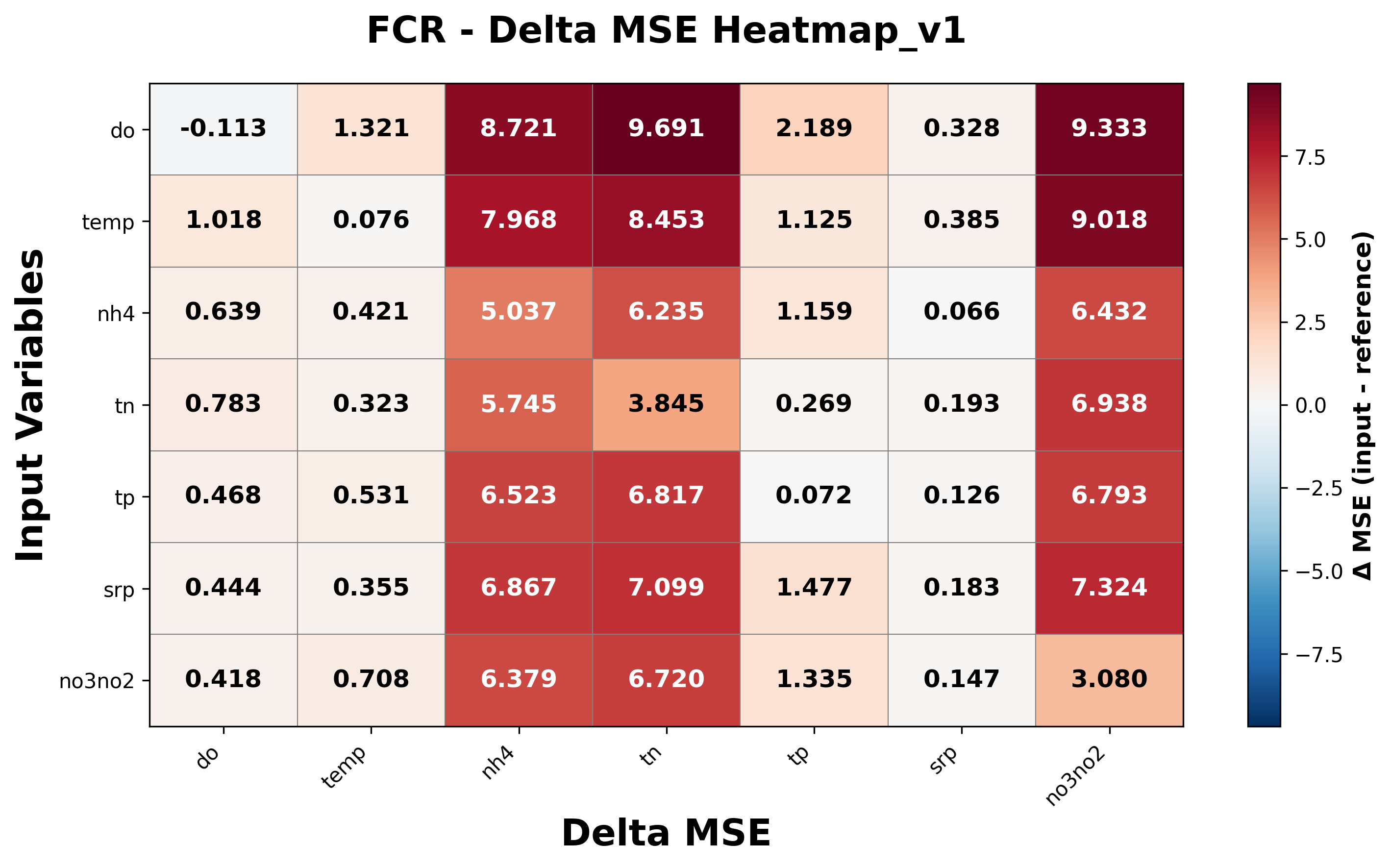}

\includegraphics[width=0.45\linewidth]{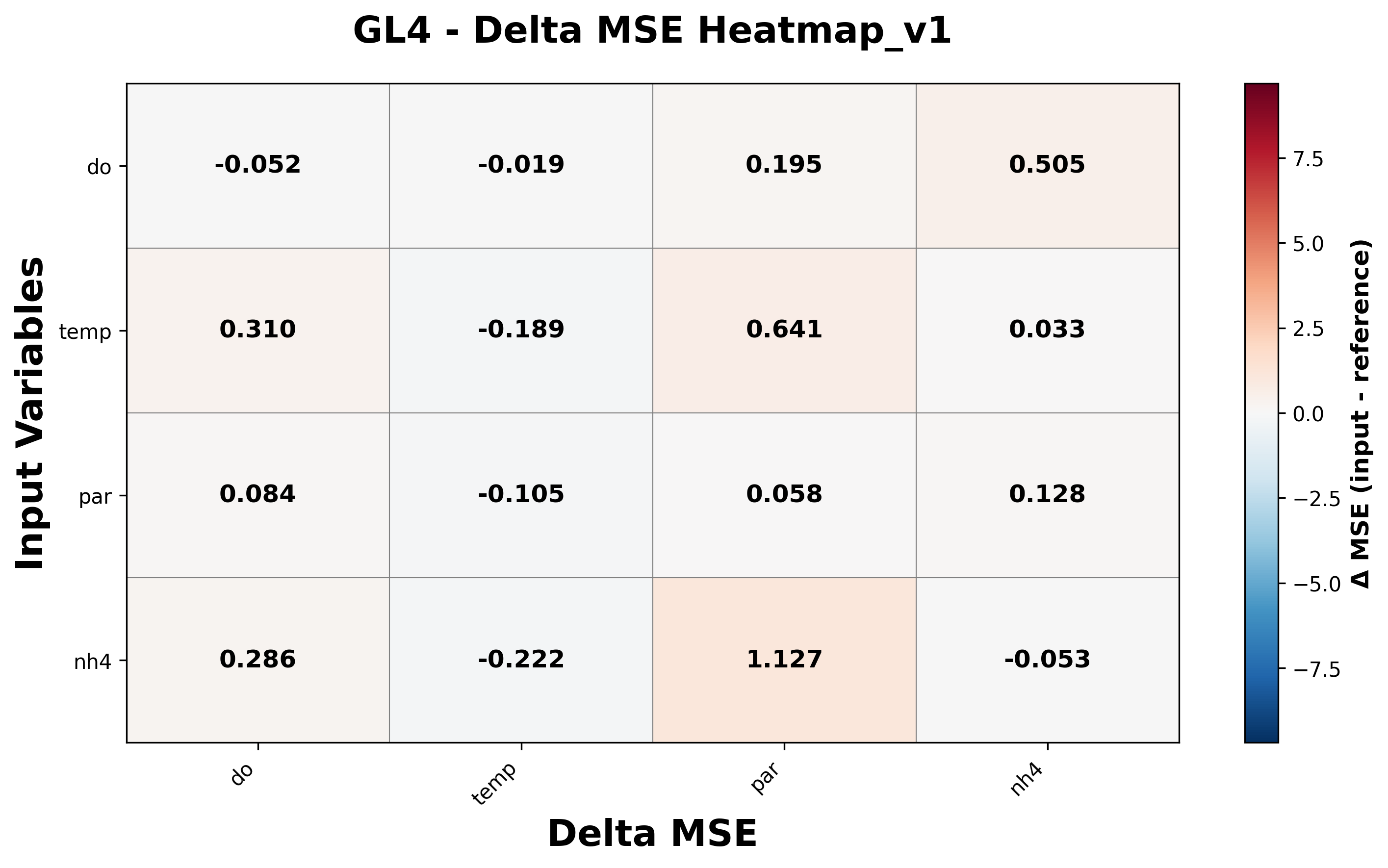}
\includegraphics[width=0.45\linewidth]{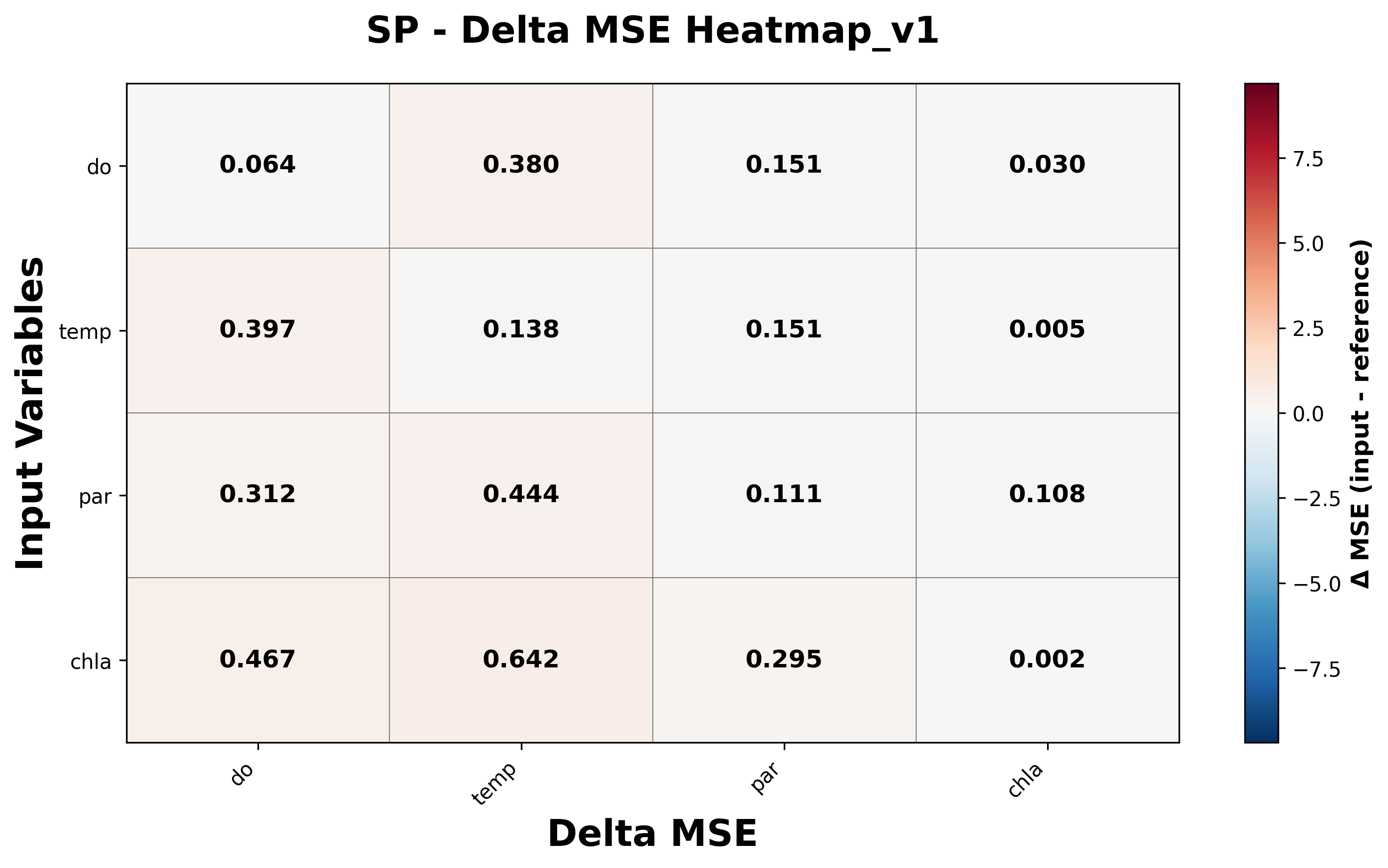}

\end{figure*}

\clearpage
\begin{figure*}[hbtp]
\ContinuedFloat
\centering

\includegraphics[width=0.45\linewidth]{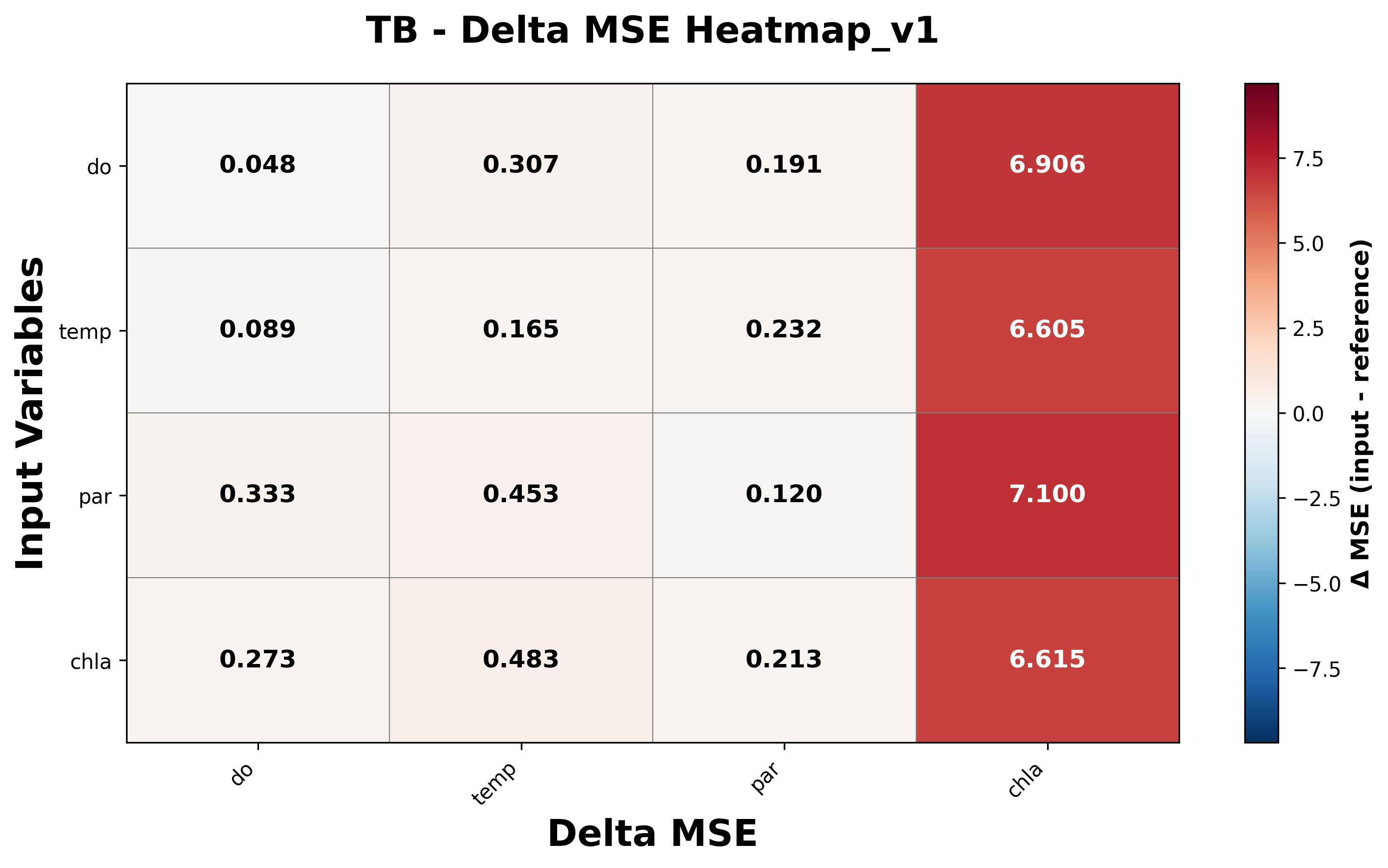}
\includegraphics[width=0.45\linewidth]{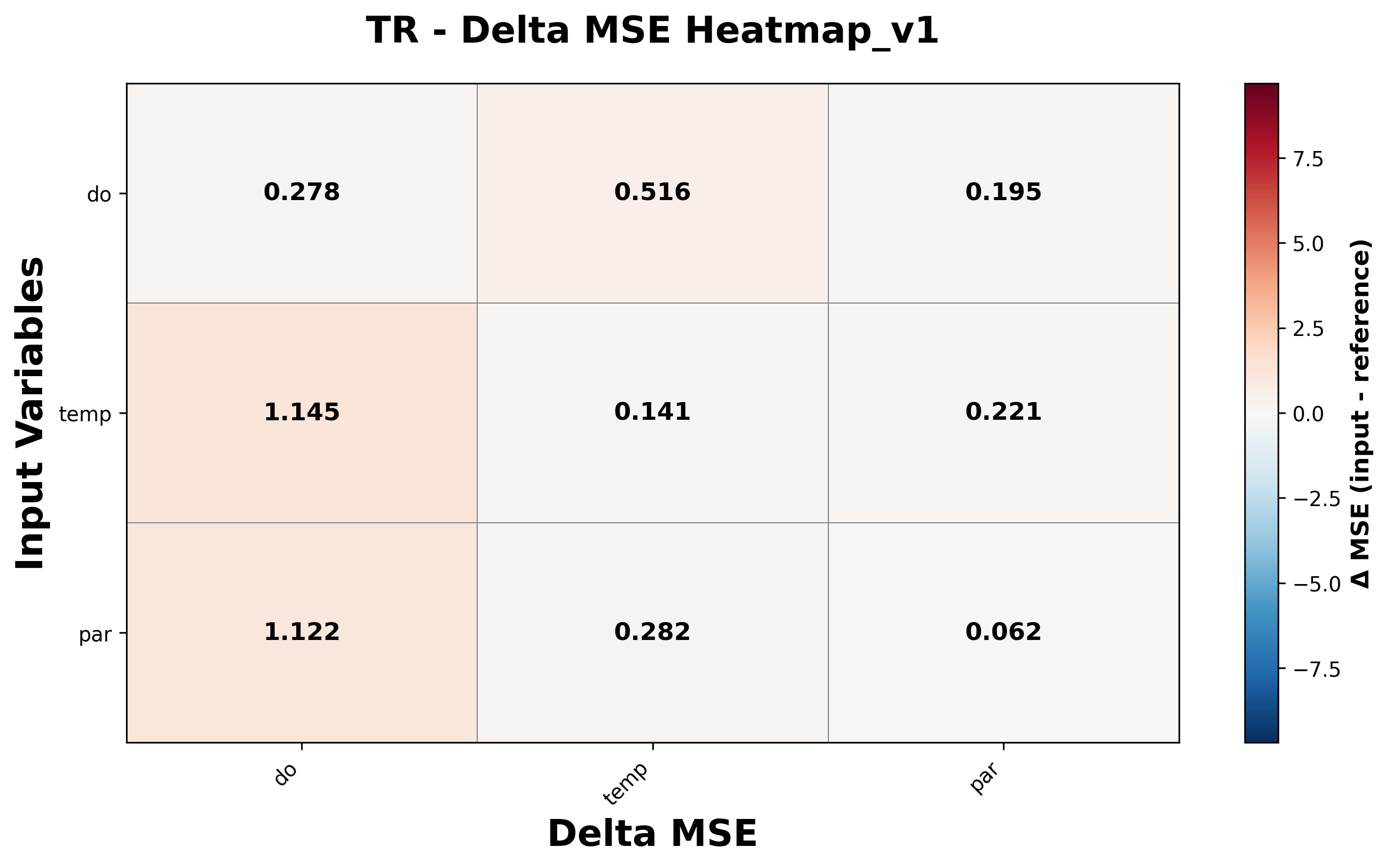}

\end{figure*}

\subsection{Qualitative Analysis - Depth Masking}
\label{sec:appendix-depth-masking}
Figures \ref{fig:cram_masking_shallow_viz},  \ref{fig:barc_depth_masking_viz} show the effects of masking shallow and deep layers of a lake on lake forecasts.
For CRAM, shallow depths considered are 1.0, 2.0, 3.0 m and deeper depths are 14.0, 14.5, 15.0 m. Similarly, for BARC, we consider 0.5, 1.0, 1.5 m as shallow depths, and 5.5 m as the deeper depth.

\begin{figure*}[!hbpt]
    \centering
    \begin{subfigure}[t]{0.32\textwidth}
        \centering
        \includegraphics[width=\linewidth]{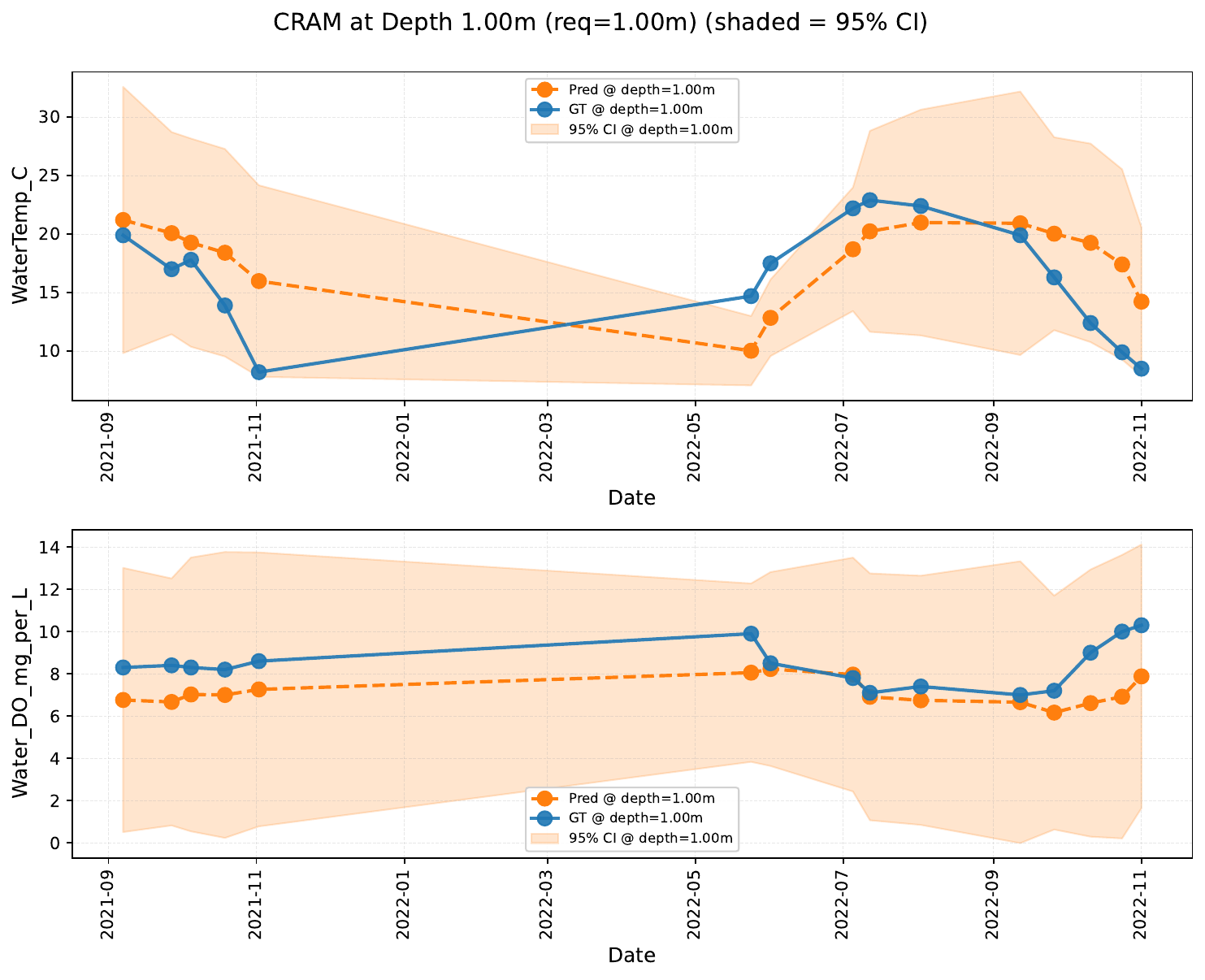}
        \caption{No Masking. CRAM @ 1.0m}
    \end{subfigure}
    \hfill
    \begin{subfigure}[t]{0.32\textwidth}
        \centering
        \includegraphics[width=\linewidth]{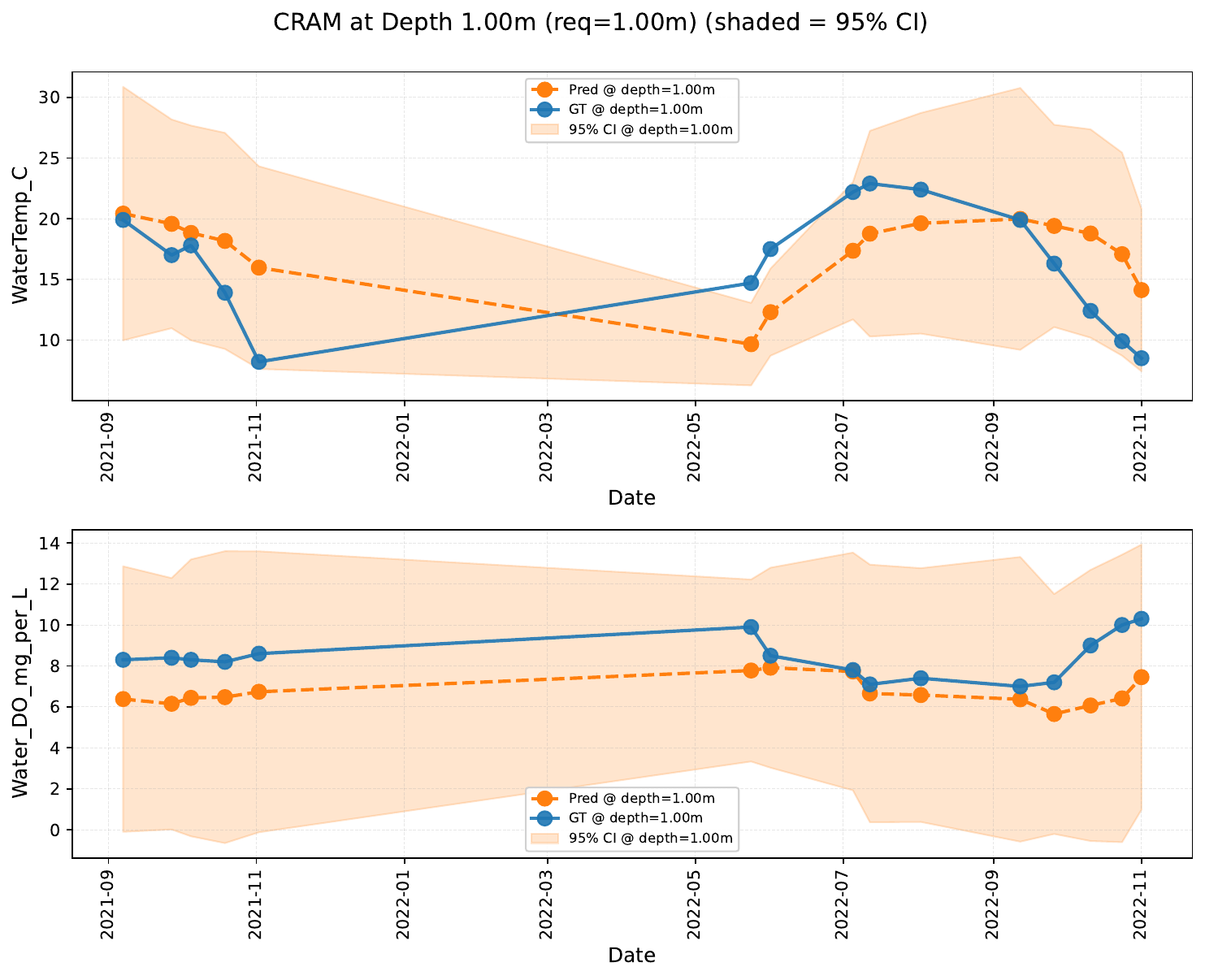}
        \caption{Shallow Layers Masked. CRAM @ 1.0m}
    \end{subfigure}
    \hfill
    \begin{subfigure}[t]{0.32\textwidth}
        \centering
        \includegraphics[width=\linewidth]{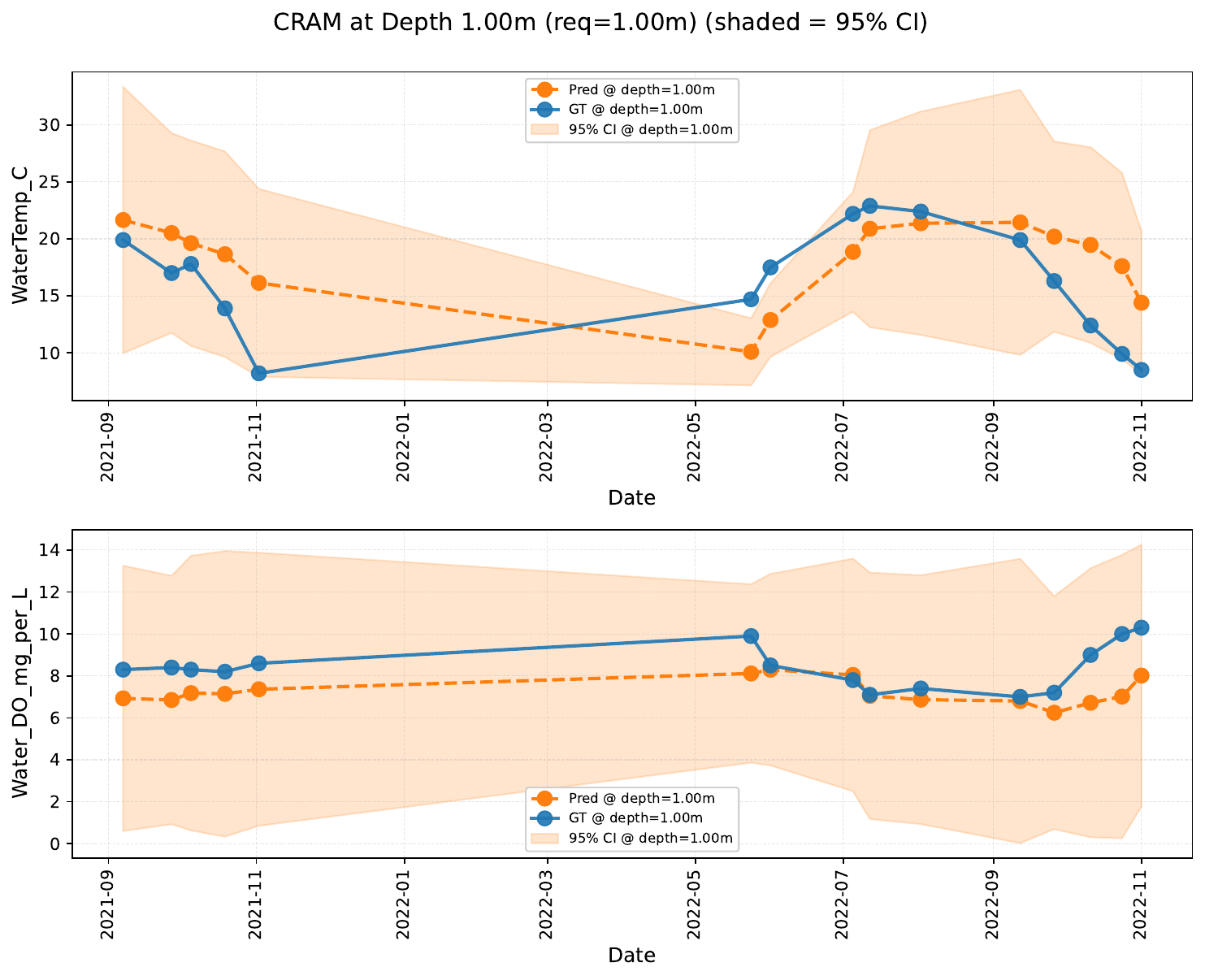}
        \caption{Deeper Layers Masked. CRAM @ 1.0m}
    \end{subfigure}

    \begin{subfigure}[t]{0.32\textwidth}
        \centering
        \includegraphics[width=\linewidth]{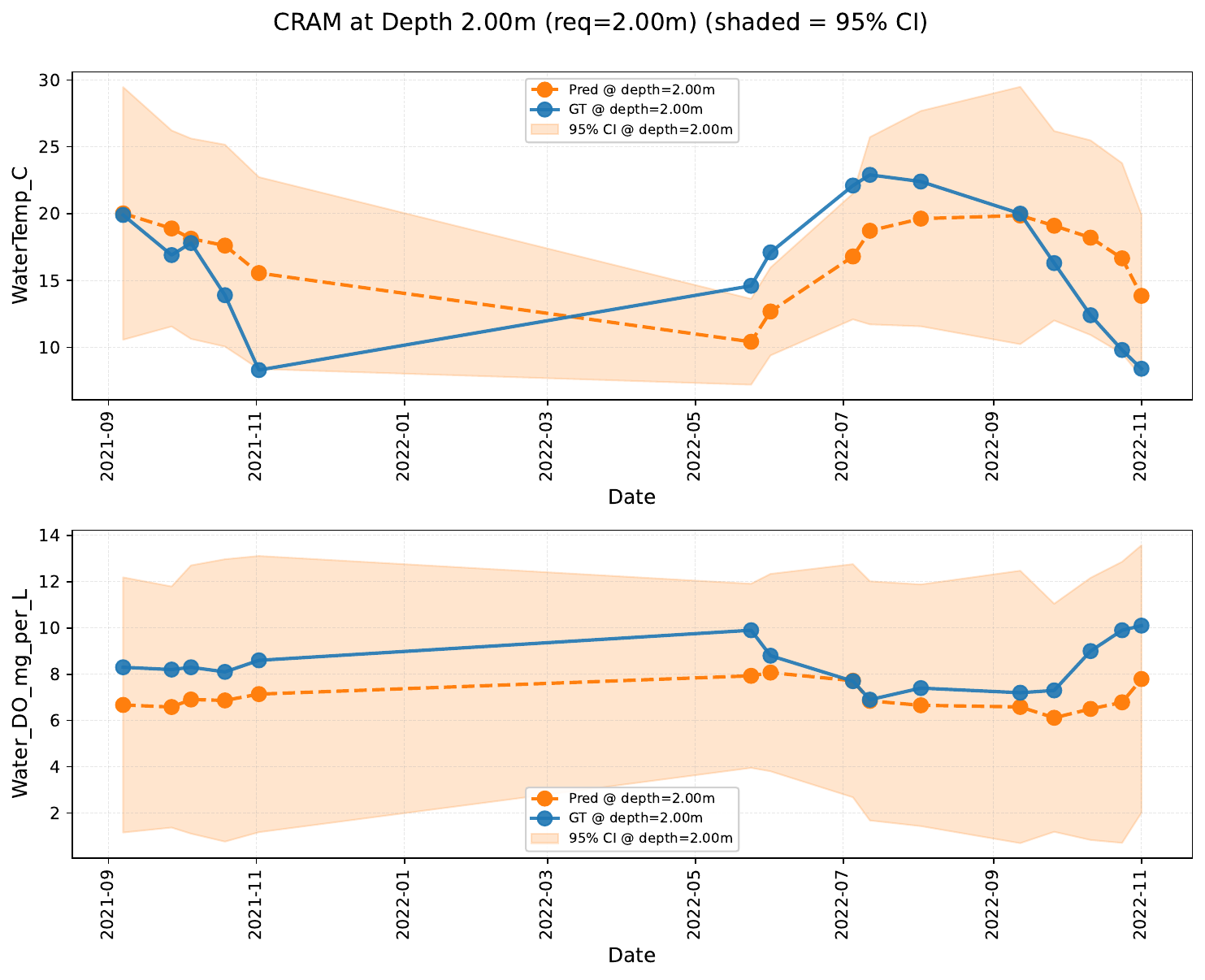}
        \caption{No Masking. CRAM @ 2.0m}
    \end{subfigure}
    \hfill
    \begin{subfigure}[t]{0.32\textwidth}
        \centering
        \includegraphics[width=\linewidth]{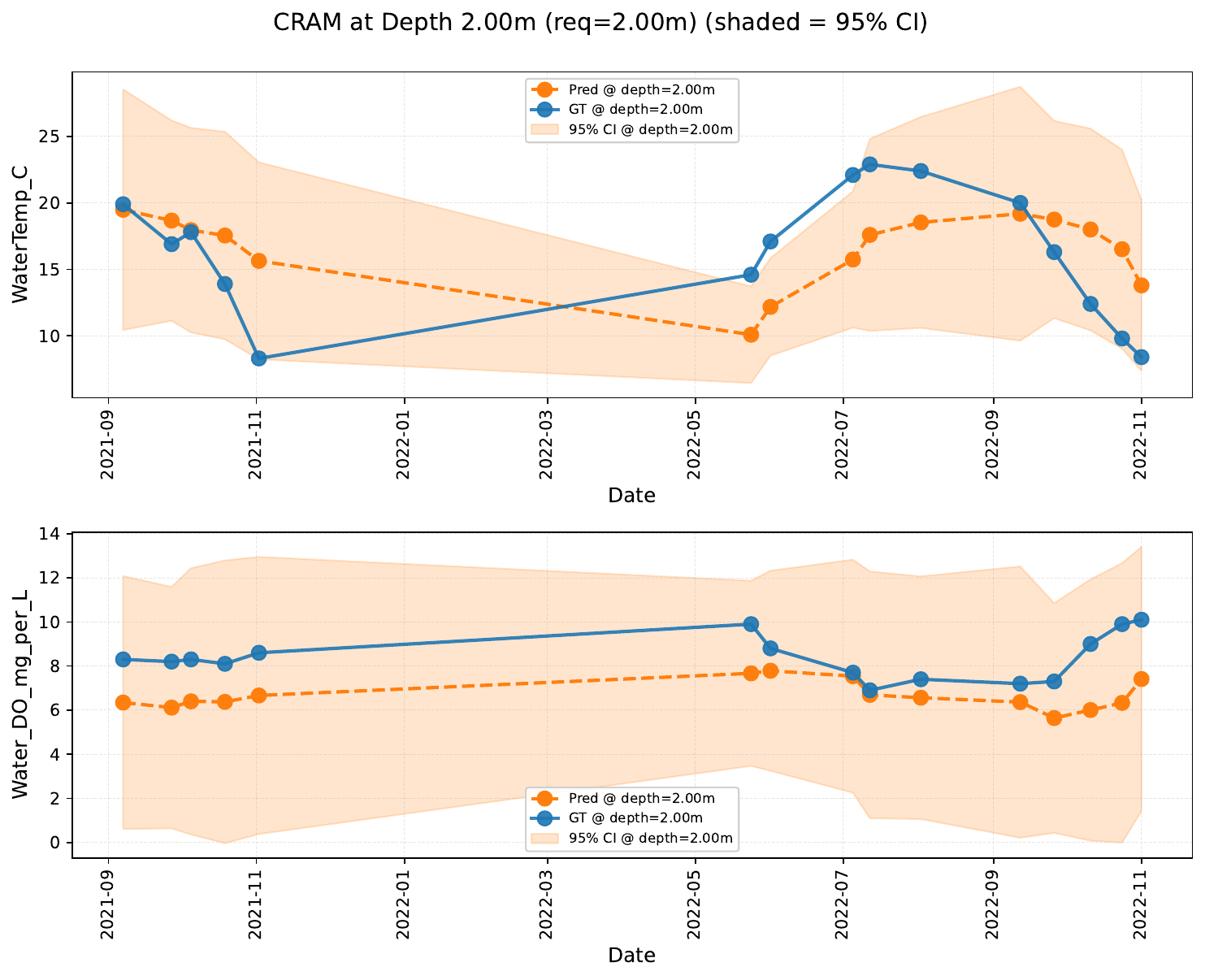}
        \caption{Shallow Layers Masked. CRAM @ 2.0m}
    \end{subfigure}
    \hfill
    \begin{subfigure}[t]{0.32\textwidth}
        \centering
        \includegraphics[width=\linewidth]{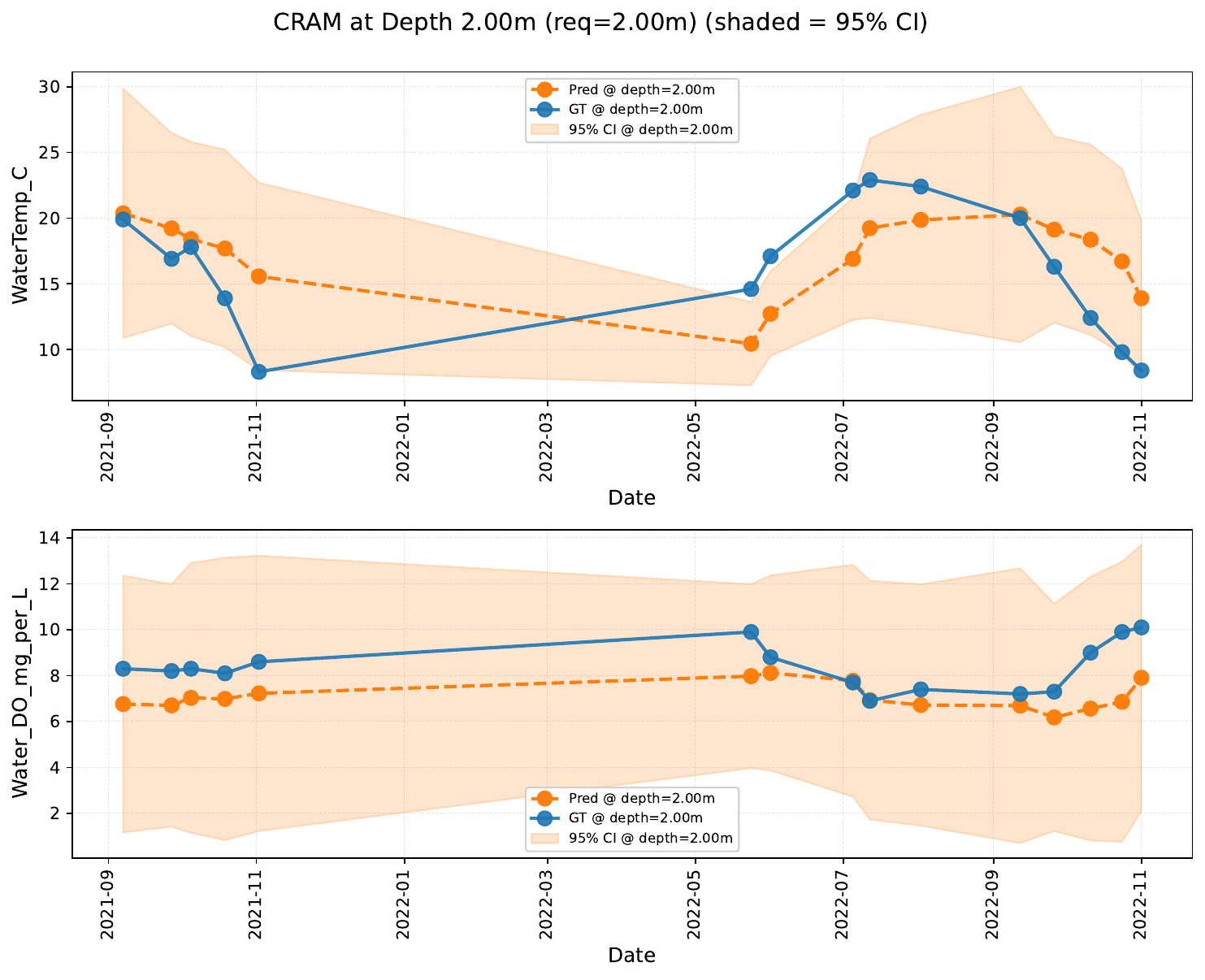}
        \caption{Deeper Layers Masked. CRAM @ 2.0m}
    \end{subfigure}

    \caption{Depth Masking - Prediction performance visualization in the shallow region under no masking, masking the shallow layers and masking the deeper layers in the input, for Lake CRAM}
    \label{fig:cram_masking_shallow_viz}
\end{figure*}

\begin{figure*}[t]
    \centering
    \begin{subfigure}[t]{0.32\textwidth}
        \centering
        \includegraphics[width=\linewidth]{figures/v_d_masking/BARC_Masking/NO_MASK_upd_BARC_plot0.5_evaluation.pdf}
        \caption{No Masking. BARC @ 0.5m}
    \end{subfigure}
    \hfill
    \begin{subfigure}[t]{0.32\textwidth}
        \centering
        \includegraphics[width=\linewidth]{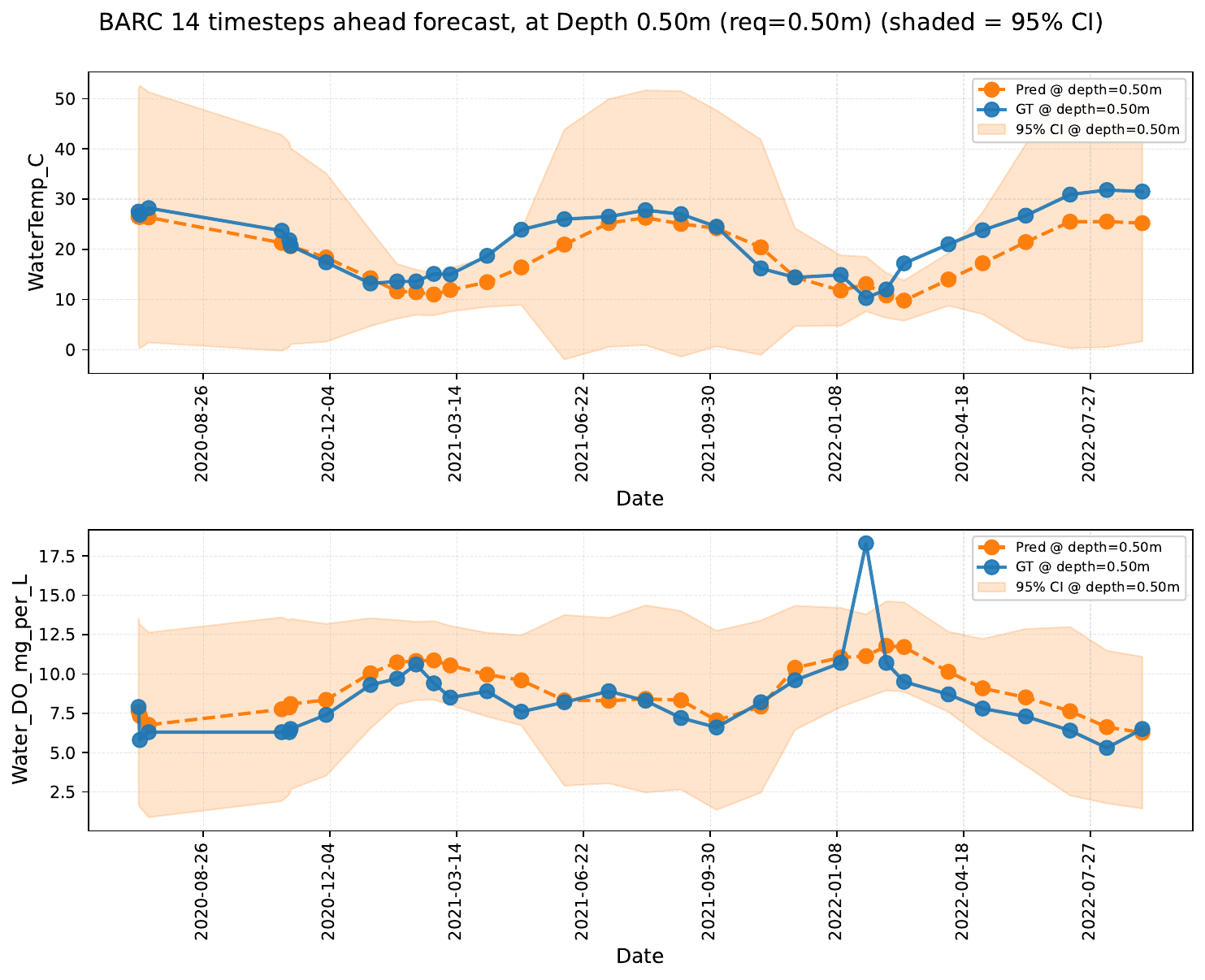}
        \caption{Shallow Layers Masked. BARC @ 0.5m}
    \end{subfigure}
    \hfill
    \begin{subfigure}[t]{0.32\textwidth}
        \centering
        \includegraphics[width=\linewidth]{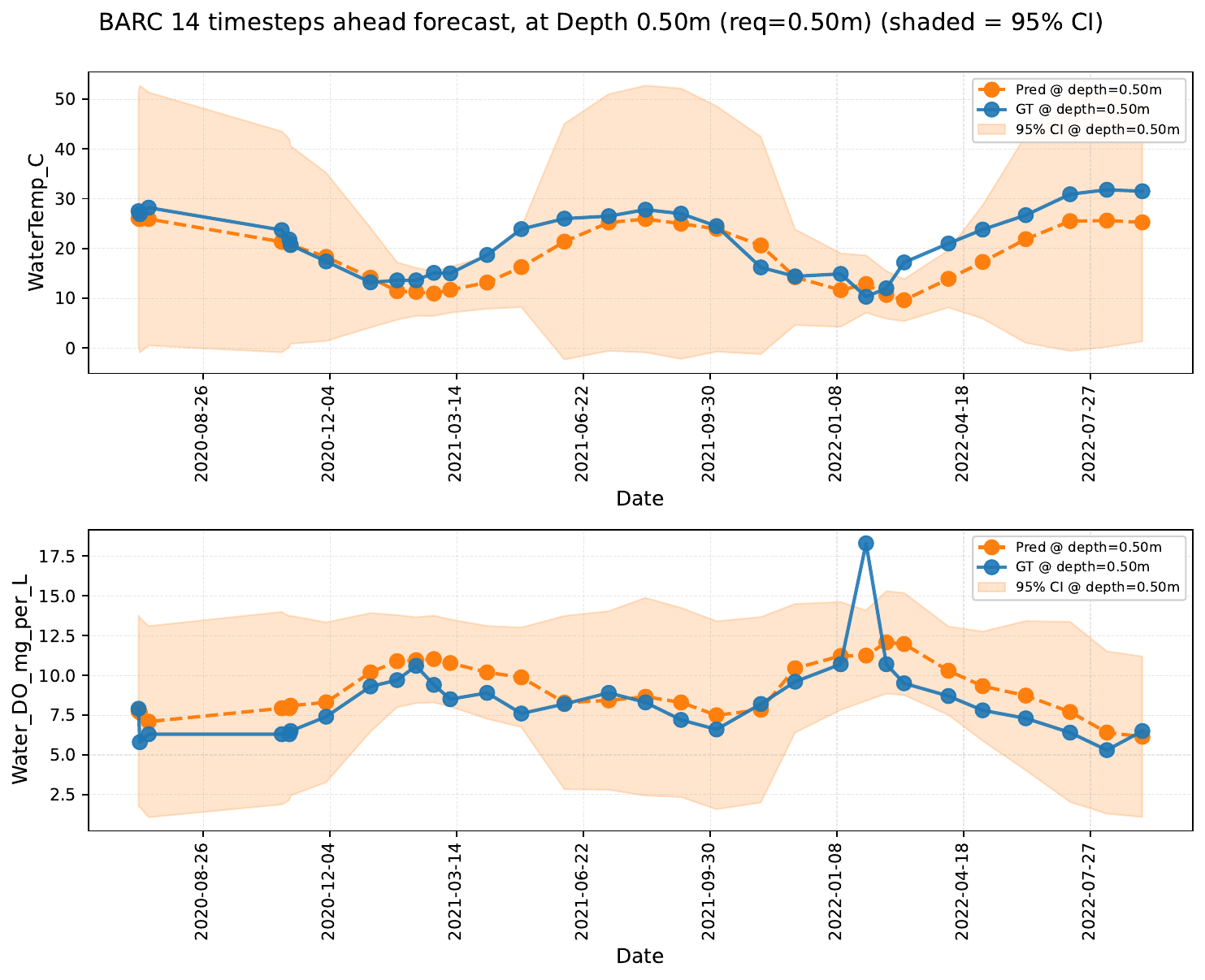}
        \caption{Deeper Layers Masked. BARC @ 0.5m}
    \end{subfigure}

    \begin{subfigure}[t]{0.32\textwidth}
        \centering
        \includegraphics[width=\linewidth]{figures/v_d_masking/BARC_Masking/NO_MASK_upd_BARC_plot1.5_evaluation.pdf}
        \caption{No Masking. BARC @ 1.5m}
    \end{subfigure}
    \hfill
    \begin{subfigure}[t]{0.32\textwidth}
        \centering
        \includegraphics[width=\linewidth]{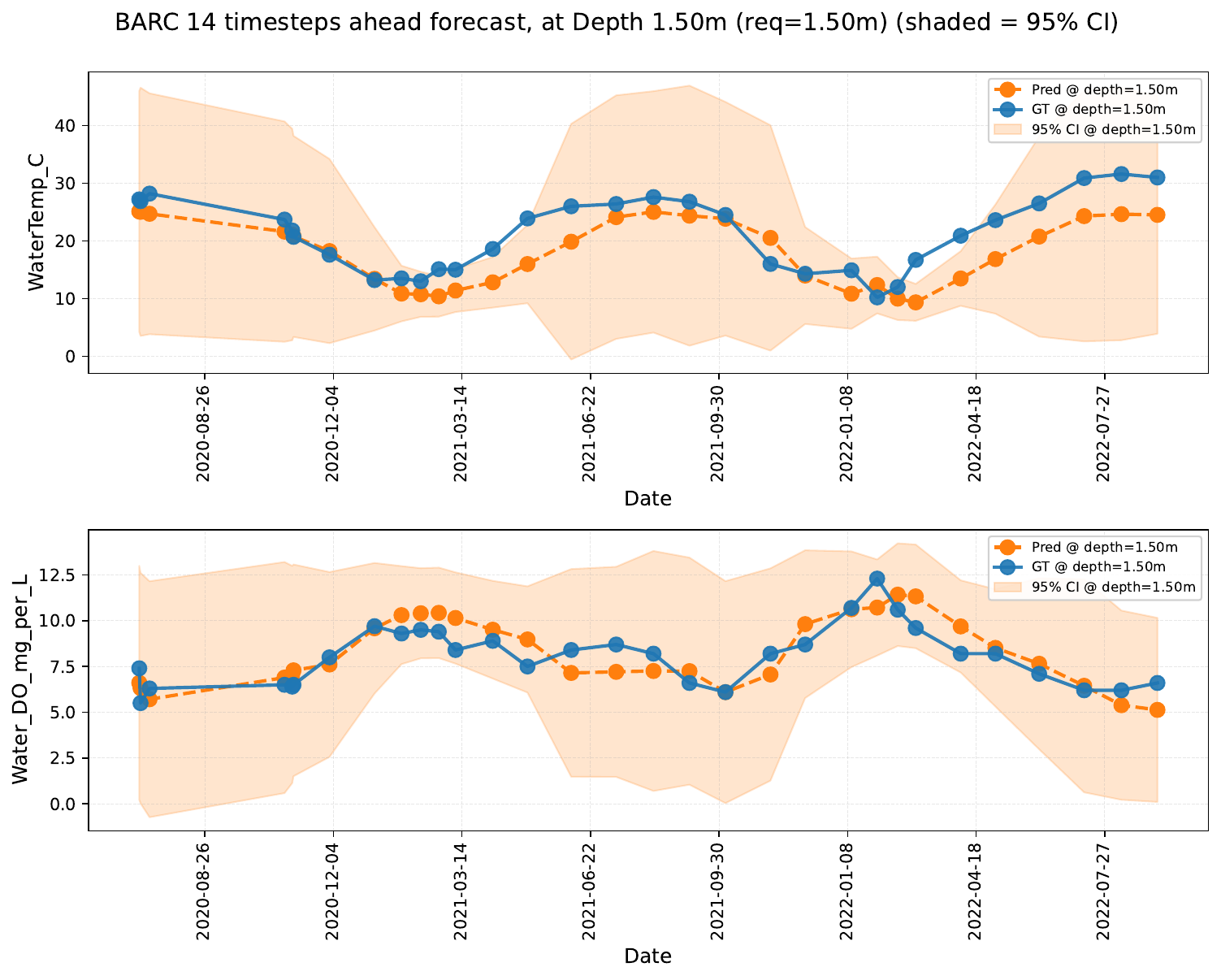}
        \caption{Shallow Layers Masked. BARC @ 1.5m}
    \end{subfigure}
    \hfill
    \begin{subfigure}[t]{0.32\textwidth}
        \centering
        \includegraphics[width=\linewidth]{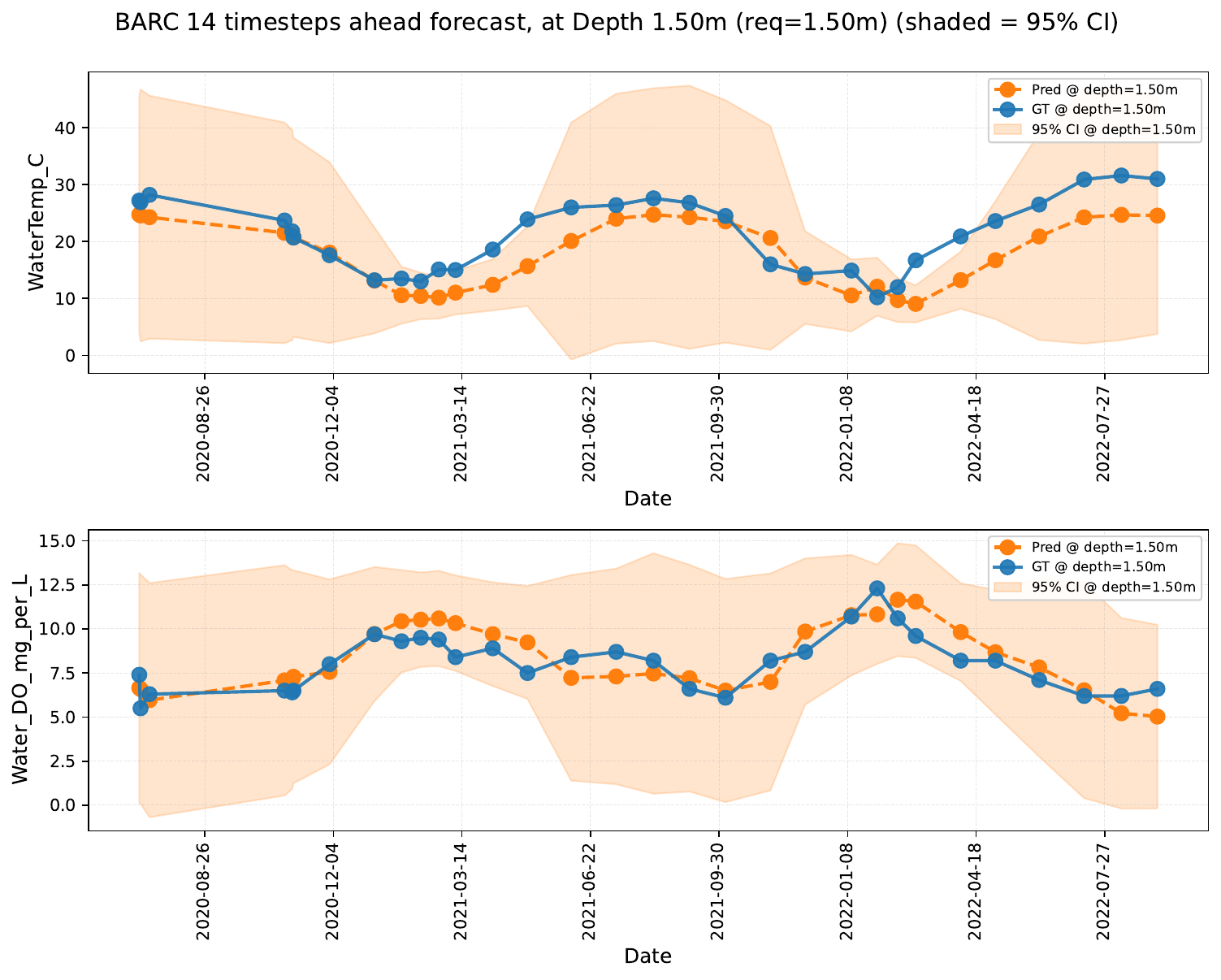}
        \caption{Deeper Layers Masked. BARC @ 1.5m}
    \end{subfigure}

    \caption{Depth Masking - Prediction performance visualization under no masking, masking the shallow layers and masking the deeper layers in the input, for Lake BARC}
    \label{fig:barc_depth_masking_viz}
\end{figure*}
\newpage
\newpage
\section{Full Results}\label{sec:full_results}
\camready{Tables~\ref{tab:lakebed_full_mse_mae_OOD} and~\ref{tab:lakebed_full_mse_mae_ID} report full results for \lakefm{} and non-IMTS baselines; IMTS comparisons are provided in Tables \ref{tab:imts_ood} and \ref{tab:imts_id}}
\begin{table*}[!htbp]
\centering
\caption{Performance comparison on OOD LakeBeD-US data. 
}
\label{tab:lakebed_full_mse_mae_OOD}
\resizebox{0.95\textwidth}{!}{%
\begin{tabular}{cc*{10}{c}}
\toprule
\textbf{Lake} & \textbf{Baseline} & \multicolumn{2}{c}{\textbf{Lake}} & \multicolumn{2}{c}{\textbf{WaterTemp\_C}} & \multicolumn{2}{c}{\textbf{Water\_DO\_mg\_per\_L}} & \multicolumn{2}{c}{\textbf{par}} & \multicolumn{2}{c}{\textbf{nh4}} \\
\cmidrule(lr){3-4}
\cmidrule(lr){5-6}
\cmidrule(lr){7-8}
\cmidrule(lr){9-10}
\cmidrule(lr){11-12}
 &  & \textbf{MSE} & \textbf{MAE} & \textbf{MSE} & \textbf{MAE} & \textbf{MSE} & \textbf{MAE} & \textbf{MSE} & \textbf{MAE} & \textbf{MSE} & \textbf{MAE} \\
\midrule
\multirow{6}{*}{\textbf{BARC}}
& LPTM           & 1.38 & \underline{0.56} & \underline{0.58} & \underline{0.59} & 2.18 & 0.53 & -- & -- & -- & -- \\
& MOMENT         & 1.38 & \underline{0.56} & \underline{0.58} & \underline{0.59} & 2.18 & 0.53 & -- & -- & -- & -- \\
& Chronos2       & 1.38 & \underline{0.56} & 0.59 & 0.60 & 2.17 & \underline{0.52} & -- & -- & -- & -- \\
& iTransformer   & \textbf{0.94} & \textbf{0.53} & \textbf{0.50} & \underline{0.59} & \textbf{1.38} & \textbf{0.48} & -- & -- & -- & -- \\
& MOIRAI         & 1.41 & \underline{0.56} & 0.60 & 0.60 & 2.20 & \underline{0.52} & -- & -- & -- & -- \\
& LakeFM         & \underline{1.05} & 0.65 & \textbf{0.50} & \textbf{0.56} & \underline{1.61} & 0.74 & -- & -- & -- & -- \\
\midrule
\multirow{6}{*}{\textbf{GL4}}
& LPTM           & 1.80 & 0.91 & 1.84 & 1.13 & 2.32 & 1.06 & \textbf{0.70} & \underline{0.54} & \underline{1.61} & 1.14 \\
& MOMENT         & 1.80 & 0.91 & 1.81 & 1.12 & 2.31 & 1.06 & \textbf{0.70} & \underline{0.54} & \underline{1.61} & 1.14 \\
& Chronos2       & \underline{1.59} & 0.85 & 1.77 & 1.10 & \underline{2.15} & \underline{0.97} & 0.78 & 0.57 & \textbf{1.59} & \textbf{1.08} \\
& iTransformer   & 5.90 & 1.56 & 1.68 & 1.14 & 16.45 & 2.88 & 0.93 & 0.60 & 3.70 & 1.76 \\
& MOIRAI         & 1.80 & \underline{0.84} & \underline{1.41} & \underline{0.90} & 2.70 & 1.00 & 0.89 & 0.59 & 1.67 & \underline{1.11} \\
& LakeFM         & \textbf{0.76} & \textbf{0.61} & \textbf{0.63} & \textbf{0.64} & \textbf{0.63} & \textbf{0.55} & \underline{0.71} & \textbf{0.52} & 4.74 & 1.96 \\
\midrule
\multirow{6}{*}{\textbf{ME}}
& LPTM           & 0.62 & 0.59 & 0.65 & 0.58 & 0.58 & 0.60 & -- & -- & -- & -- \\
& MOMENT         & 0.62 & 0.59 & 0.65 & 0.58 & 0.58 & 0.60 & -- & -- & -- & -- \\
& Chronos2       & 0.61 & 0.59 & 0.67 & 0.61 & 0.55 & 0.56 & -- & -- & -- & -- \\
& iTransformer   & 0.76 & 0.76 & 0.61 & 0.64 & 0.91 & 0.89 & -- & -- & -- & -- \\
& MOIRAI         & \underline{0.58} & \underline{0.54} & \underline{0.59} & \underline{0.53} & \underline{0.53} & \underline{0.55} & -- & -- & -- & -- \\
& LakeFM         & \textbf{0.33} & \textbf{0.44} & \textbf{0.21} & \textbf{0.37} & \textbf{0.45} & \textbf{0.51} & -- & -- & -- & -- \\
\midrule
\multirow{6}{*}{\textbf{SUGG}}
& LPTM           & 0.87 & 0.55 & 1.34 & 0.76 & 0.41 & \textbf{0.34} & -- & -- & -- & -- \\
& MOMENT         & 0.87 & 0.55 & 1.34 & 0.76 & 0.41 & \textbf{0.34} & -- & -- & -- & -- \\
& Chronos2       & 0.88 & 0.55 & 1.35 & 0.76 & 0.41 & \textbf{0.34} & -- & -- & -- & -- \\
& iTransformer   & \textbf{0.43} & \underline{0.54} & \underline{0.77} & 0.70 & 0.43 & 0.38 & -- & -- & -- & -- \\
& MOIRAI         & 0.87 & \underline{0.54} & 1.30 & \textbf{0.54} & \underline{0.40} & \underline{0.36} & -- & -- & -- & -- \\
& LakeFM         & \underline{0.45} & \textbf{0.48} & \textbf{0.57} & \underline{0.60} & \textbf{0.34} & 0.36 & -- & -- & -- & -- \\
\midrule
\multirow{6}{*}{\textbf{TOOK}}
& LPTM           & 2.12 & 0.98 & 1.88 & \underline{1.11} & 2.36 & 0.84 & -- & -- & -- & -- \\
& MOMENT         & 2.11 & 0.98 & 1.88 & \underline{1.11} & 2.35 & 0.84 & -- & -- & -- & -- \\
& Chronos2       & 2.06 & 0.98 & 1.89 & 1.13 & 2.23 & 0.82 & -- & -- & -- & -- \\
& iTransformer   & \underline{1.54} & \underline{0.90} & \textbf{1.43} & \textbf{1.04} & \underline{1.64} & \underline{0.76} & -- & -- & -- & -- \\
& MOIRAI         & 2.23 & 1.00 & 1.84 & 1.12 & 2.61 & 0.87 & -- & -- & -- & -- \\
& LakeFM         & \textbf{1.41} & \textbf{0.71} & \underline{1.82} & \underline{1.11} & \textbf{1.00} & \textbf{0.30} & -- & -- & -- & -- \\
\midrule
\multirow{6}{*}{\textbf{TR}}
& LPTM           & \underline{0.60} & 0.49 & \underline{0.38} & \underline{0.43} & 0.56 & 0.54 & \textbf{0.87} & \underline{0.51} & -- & -- \\
& MOMENT         & \underline{0.60} & 0.49 & \underline{0.38} & \underline{0.43} & 0.56 & 0.54 & \textbf{0.87} & \underline{0.51} & -- & -- \\
& Chronos2       & \underline{0.60} & \underline{0.48} & 0.38 & 0.44 & 0.55 & \underline{0.51} & \underline{0.88} & \textbf{0.50} & -- & -- \\
& iTransformer   & 2.74 & 1.23 & 2.20 & 1.02 & 4.02 & 1.63 & 2.00 & 0.82 & -- & -- \\
& MOIRAI         & \textbf{0.55} & \textbf{0.45} & \textbf{0.30} & \textbf{0.40} & \underline{0.52} & \textbf{0.49} & 1.06 & 0.55 & -- & -- \\
& LakeFM         & 0.71 & 0.66 & 0.85 & 0.75 & \textbf{0.45} & 0.53 & 0.97 & 0.72 & -- & -- \\
\midrule
\textbf{Average Rank} & \textbf{LPTM} & 4.17 & 4.08 & 4.25 & 3.67 & 4.42 & 4.08 & \textbf{1.50} & \textbf{2.50} & \underline{2.50} & 3.50 \\
\textbf{Average Rank} & \textbf{MOMENT} & 4.00 & 4.08 & 4.08 & 3.50 & 4.08 & 4.08 & \textbf{1.50} & \textbf{2.50} & \underline{2.50} & 3.50 \\
\textbf{Average Rank} & \textbf{Chronos2} & \underline{3.50} & 3.58 & 5.17 & 4.75 & \underline{3.00} & \textbf{2.42} & \underline{3.50} & \textbf{2.50} & \textbf{1.00} & \textbf{1.00} \\
\textbf{Average Rank} & \textbf{iTransformer} & 3.67 & 3.92 & \underline{2.75} & 4.17 & 4.50 & 4.50 & 6.00 & 6.00 & 5.00 & 5.00 \\
\textbf{Average Rank} & \textbf{MOIRAI} & 3.67 & \underline{2.83} & 2.83 & \underline{2.75} & 3.83 & 3.08 & 5.00 & 4.50 & 4.00 & \underline{2.00} \\
\textbf{Average Rank} & \textbf{LakeFM} & \textbf{2.00} & \textbf{2.50} & \textbf{1.92} & \textbf{2.17} & \textbf{1.17} & \underline{2.83} & \underline{3.50} & \underline{3.00} & 6.00 & 6.00 \\
\bottomrule
\end{tabular}%
}
\end{table*}

\begin{table*}[!htbp]
\centering
\caption{Performance comparison on ID LakeBeD-US data.}
\label{tab:lakebed_full_mse_mae_ID}
\renewcommand{\arraystretch}{0.85}
\resizebox{1.0\textwidth}{!}{%
\begin{tabular}{cc*{20}{c}}
\toprule
\textbf{Lake} & \textbf{Baseline} & \multicolumn{2}{c}{\textbf{Lake}} & \multicolumn{2}{c}{\textbf{WaterTemp\_C}} & \multicolumn{2}{c}{\textbf{Chla\_ugL}} & \multicolumn{2}{c}{\textbf{Water\_DO\_mg\_per\_L}} & \multicolumn{2}{c}{\textbf{par}} & \multicolumn{2}{c}{\textbf{nh4}} & \multicolumn{2}{c}{\textbf{SRP\_ugL}} & \multicolumn{2}{c}{\textbf{Water\_TP\_mg\_per\_L}} & \multicolumn{2}{c}{\textbf{tn}} & \multicolumn{2}{c}{\textbf{no3no2}} \\
\cmidrule(lr){3-4}
\cmidrule(lr){5-6}
\cmidrule(lr){7-8}
\cmidrule(lr){9-10}
\cmidrule(lr){11-12}
\cmidrule(lr){13-14}
\cmidrule(lr){15-16}
\cmidrule(lr){17-18}
\cmidrule(lr){19-20}
\cmidrule(lr){21-22}
 &  & \textbf{MSE} & \textbf{MAE} & \textbf{MSE} & \textbf{MAE} & \textbf{MSE} & \textbf{MAE} & \textbf{MSE} & \textbf{MAE} & \textbf{MSE} & \textbf{MAE} & \textbf{MSE} & \textbf{MAE} & \textbf{MSE} & \textbf{MAE} & \textbf{MSE} & \textbf{MAE} & \textbf{MSE} & \textbf{MAE} & \textbf{MSE} & \textbf{MAE} \\
\midrule
\multirow{6}{*}{\textbf{AL}}
& LPTM           & 0.95 & 0.58 & 0.49 & 0.60 & 1.99 & 0.67 & 0.58 & \textbf{0.56} & 0.95 & \underline{0.56} & -- & -- & -- & -- & -- & -- & -- & -- & -- & -- \\
& MOMENT         & 0.95 & 0.58 & 0.49 & 0.59 & 1.98 & 0.67 & 0.58 & \textbf{0.56} & 0.94 & \textbf{0.55} & -- & -- & -- & -- & -- & -- & -- & -- & -- & -- \\
& Chronos2       & 0.94 & 0.59 & 0.49 & 0.59 & 1.97 & \underline{0.66} & 0.63 & \underline{0.57} & 0.96 & \textbf{0.55} & -- & -- & -- & -- & -- & -- & -- & -- & -- & -- \\
& iTransformer   & 0.94 & 0.63 & 0.49 & 0.62 & \underline{1.90} & 0.76 & \underline{0.56} & 0.61 & \underline{0.93} & 0.58 & -- & -- & -- & -- & -- & -- & -- & -- & -- & -- \\
& MOIRAI         & \underline{0.85} & \underline{0.58} & \underline{0.44} & \underline{0.53} & 2.00 & 0.67 & \textbf{0.53} & \textbf{0.56} & 1.03 & 0.58 & -- & -- & -- & -- & -- & -- & -- & -- & -- & -- \\
& LakeFM         & \textbf{0.71} & \textbf{0.55} & \textbf{0.36} & \textbf{0.47} & \textbf{1.14} & \textbf{0.46} & 0.63 & \textbf{0.56} & \textbf{0.87} & 0.64 & -- & -- & -- & -- & -- & -- & -- & -- & -- & -- \\
\midrule
\multirow{6}{*}{\textbf{BM}}
& LPTM           & 0.75 & 0.60 & \underline{0.44} & \underline{0.53} & 1.23 & 0.78 & \underline{0.56} & 0.53 & \underline{0.83} & \underline{0.53} & -- & -- & -- & -- & -- & -- & -- & -- & -- & -- \\
& MOMENT         & 0.76 & 0.60 & \underline{0.44} & \underline{0.53} & 1.24 & 0.78 & \underline{0.56} & 0.53 & 0.84 & \underline{0.53} & -- & -- & -- & -- & -- & -- & -- & -- & -- & -- \\
& Chronos2       & \underline{0.74} & \underline{0.58} & \underline{0.44} & \underline{0.53} & \underline{1.11} & \underline{0.75} & 0.60 & \underline{0.52} & \textbf{0.82} & \textbf{0.52} & -- & -- & -- & -- & -- & -- & -- & -- & -- & -- \\
& iTransformer   & 14.91 & 3.23 & 13.85 & 3.32 & 29.91 & 4.98 & 14.33 & 3.50 & 11.11 & 2.15 & -- & -- & -- & -- & -- & -- & -- & -- & -- & -- \\
& MOIRAI         & \textbf{0.71} & \textbf{0.54} & \textbf{0.38} & \textbf{0.50} & 1.41 & 0.80 & \textbf{0.51} & \textbf{0.50} & 1.00 & 0.58 & -- & -- & -- & -- & -- & -- & -- & -- & -- & -- \\
& LakeFM         & \underline{0.73} & 0.62 & 0.48 & 0.58 & \textbf{1.07} & \textbf{0.71} & 0.69 & 0.55 & 0.95 & 0.71 & -- & -- & -- & -- & -- & -- & -- & -- & -- & -- \\
\midrule
\multirow{6}{*}{\textbf{BVR}}
& LPTM           & 1.59 & 0.87 & 5.13 & 1.81 & -- & -- & \underline{1.41} & \underline{0.91} & -- & -- & \underline{0.64} & \underline{0.50} & 1.04 & 0.77 & 0.64 & 0.61 & 0.70 & \underline{0.55} & -- & -- \\
& MOMENT         & 1.60 & 0.87 & 5.13 & 1.81 & -- & -- & \underline{1.41} & \underline{0.91} & -- & -- & 0.65 & \underline{0.50} & 1.04 & 0.77 & 0.65 & 0.62 & 0.70 & \underline{0.55} & -- & -- \\
& Chronos2       & 1.58 & 0.85 & 5.12 & 1.81 & -- & -- & 1.43 & 0.92 & -- & -- & \textbf{0.59} & \textbf{0.49} & 1.15 & 0.76 & \underline{0.58} & 0.60 & \textbf{0.63} & \textbf{0.53} & -- & -- \\
& iTransformer   & 2.66 & 1.16 & \underline{3.68} & \underline{1.48} & -- & -- & \textbf{0.84} & \textbf{0.87} & -- & -- & 1.37 & 0.75 & 2.99 & 1.23 & 5.38 & 1.66 & 1.75 & 1.00 & -- & -- \\
& MOIRAI         & \underline{1.00} & \underline{0.71} & 5.20 & 1.82 & -- & -- & 1.50 & 0.92 & -- & -- & 1.21 & 0.78 & \textbf{0.61} & \underline{0.61} & 0.66 & \underline{0.50} & 0.68 & \underline{0.55} & -- & -- \\
& LakeFM         & \textbf{0.66} & \textbf{0.53} & \textbf{1.76} & \textbf{1.10} & -- & -- & 1.65 & \underline{0.91} & -- & -- & 0.65 & 0.51 & \underline{0.81} & \textbf{0.52} & \textbf{0.24} & \textbf{0.39} & \underline{0.65} & \underline{0.55} & -- & -- \\
\midrule
\multirow{6}{*}{\textbf{CB}}
& LPTM           & \underline{0.80} & \underline{0.54} & 0.52 & 0.62 & 1.31 & 0.45 & \textbf{0.59} & \underline{0.60} & \underline{0.80} & 0.51 & -- & -- & -- & -- & -- & -- & -- & -- & -- & -- \\
& MOMENT         & \underline{0.80} & \underline{0.54} & 0.52 & 0.62 & 1.31 & 0.45 & \textbf{0.59} & \underline{0.60} & \underline{0.80} & 0.51 & -- & -- & -- & -- & -- & -- & -- & -- & -- & -- \\
& Chronos2       & 0.83 & \underline{0.54} & 0.51 & 0.61 & 1.31 & \underline{0.44} & \underline{0.67} & 0.63 & 0.84 & \underline{0.50} & -- & -- & -- & -- & -- & -- & -- & -- & -- & -- \\
& iTransformer   & 1.07 & 0.74 & 0.82 & 0.82 & \underline{1.18} & 0.52 & 0.96 & 0.88 & 1.20 & 0.68 & -- & -- & -- & -- & -- & -- & -- & -- & -- & -- \\
& MOIRAI         & 0.84 & 0.55 & \underline{0.45} & \underline{0.55} & 1.41 & 0.45 & \textbf{0.59} & \textbf{0.59} & 1.00 & 0.56 & -- & -- & -- & -- & -- & -- & -- & -- & -- & -- \\
& LakeFM         & \textbf{0.67} & \textbf{0.50} & \textbf{0.34} & \textbf{0.47} & \textbf{1.16} & \textbf{0.29} & 0.73 & 0.65 & \textbf{0.67} & \textbf{0.48} & -- & -- & -- & -- & -- & -- & -- & -- & -- & -- \\
\midrule
\multirow{6}{*}{\textbf{CR}}
& LPTM           & \textbf{0.56} & \textbf{0.50} & 0.38 & 0.48 & 1.10 & 0.64 & \underline{0.46} & \underline{0.47} & \underline{0.70} & \textbf{0.51} & -- & -- & -- & -- & -- & -- & -- & -- & -- & -- \\
& MOMENT         & \textbf{0.56} & \textbf{0.50} & 0.38 & 0.48 & 1.10 & 0.64 & \underline{0.46} & \underline{0.47} & \underline{0.70} & \textbf{0.51} & -- & -- & -- & -- & -- & -- & -- & -- & -- & -- \\
& Chronos2       & \underline{0.57} & \textbf{0.50} & 0.38 & 0.48 & \underline{1.08} & \underline{0.63} & \textbf{0.45} & \textbf{0.45} & 0.75 & \textbf{0.51} & -- & -- & -- & -- & -- & -- & -- & -- & -- & -- \\
& iTransformer   & 0.59 & \textbf{0.50} & \textbf{0.35} & \underline{0.45} & 1.09 & 0.64 & 0.53 & 0.50 & 0.77 & \underline{0.54} & -- & -- & -- & -- & -- & -- & -- & -- & -- & -- \\
& MOIRAI         & 0.75 & \underline{0.53} & 0.42 & \textbf{0.43} & 1.30 & 0.66 & 0.48 & \textbf{0.45} & 0.83 & 0.55 & -- & -- & -- & -- & -- & -- & -- & -- & -- & -- \\
& LakeFM         & 0.72 & 0.58 & \underline{0.37} & 0.51 & \textbf{0.98} & \textbf{0.56} & 1.04 & 0.66 & \textbf{0.67} & 0.59 & -- & -- & -- & -- & -- & -- & -- & -- & -- & -- \\
\midrule
\multirow{6}{*}{\textbf{CRAM}}
& LPTM           & \underline{0.53} & 0.52 & 0.46 & 0.54 & -- & -- & 0.60 & \underline{0.50} & -- & -- & -- & -- & -- & -- & -- & -- & -- & -- & -- & -- \\
& MOMENT         & \underline{0.53} & 0.52 & 0.46 & 0.54 & -- & -- & 0.59 & \underline{0.50} & -- & -- & -- & -- & -- & -- & -- & -- & -- & -- & -- & -- \\
& Chronos2       & \underline{0.53} & \underline{0.51} & 0.40 & \underline{0.50} & -- & -- & 0.66 & 0.52 & -- & -- & -- & -- & -- & -- & -- & -- & -- & -- & -- & -- \\
& iTransformer   & 0.59 & 0.58 & \textbf{0.38} & \textbf{0.46} & -- & -- & 0.80 & 0.70 & -- & -- & -- & -- & -- & -- & -- & -- & -- & -- & -- & -- \\
& MOIRAI         & \textbf{0.48} & \textbf{0.49} & \underline{0.39} & \underline{0.50} & -- & -- & \underline{0.56} & \textbf{0.48} & -- & -- & -- & -- & -- & -- & -- & -- & -- & -- & -- & -- \\
& LakeFM         & 0.57 & 0.63 & 0.62 & 0.68 & -- & -- & \textbf{0.52} & 0.58 & -- & -- & -- & -- & -- & -- & -- & -- & -- & -- & -- & -- \\
\midrule
\multirow{6}{*}{\textbf{FCR}}
& LPTM           & 1.01 & 0.75 & 1.00 & 0.70 & -- & -- & 0.71 & 0.61 & -- & -- & 0.64 & 0.61 & 1.98 & 1.15 & \underline{1.01} & \underline{0.81} & 0.75 & 0.69 & 0.48 & 0.59 \\
& MOMENT         & 1.02 & 0.75 & 1.00 & 0.70 & -- & -- & 0.71 & 0.61 & -- & -- & 0.64 & 0.61 & 1.99 & 1.15 & \underline{1.01} & \underline{0.81} & 0.75 & 0.69 & 0.48 & 0.59 \\
& Chronos2       & \underline{0.96} & \underline{0.71} & \underline{0.96} & 0.68 & -- & -- & 0.66 & 0.58 & -- & -- & \textbf{0.55} & \underline{0.56} & 2.03 & 1.14 & 1.05 & \underline{0.81} & \underline{0.67} & 0.64 & \textbf{0.41} & \textbf{0.54} \\
& iTransformer   & 1.44 & 0.98 & 1.52 & 1.10 & -- & -- & 0.79 & 0.83 & -- & -- & 1.40 & 0.91 & \underline{1.37} & \underline{0.94} & 2.08 & 1.21 & 1.73 & 1.07 & 1.34 & 0.88 \\
& MOIRAI         & 1.01 & \underline{0.71} & 1.00 & \underline{0.67} & -- & -- & \underline{0.60} & \underline{0.54} & -- & -- & 2.13 & 1.21 & \textbf{1.10} & \textbf{0.83} & \textbf{0.58} & \textbf{0.55} & \underline{0.67} & \underline{0.62} & \underline{0.48} & 0.59 \\
& LakeFM         & \textbf{0.94} & \textbf{0.65} & \textbf{0.20} & \textbf{0.37} & -- & -- & \textbf{0.37} & \textbf{0.45} & -- & -- & \underline{0.59} & \textbf{0.55} & 1.67 & 0.95 & 1.04 & 1.07 & \textbf{0.43} & \textbf{0.47} & 0.88 & \underline{0.58} \\
\midrule
\multirow{6}{*}{\textbf{FI}}
& LPTM           & 0.64 & 0.57 & 0.65 & 0.52 & -- & -- & 0.63 & 0.61 & -- & -- & -- & -- & -- & -- & -- & -- & -- & -- & -- & -- \\
& MOMENT         & 0.64 & 0.57 & 0.65 & 0.52 & -- & -- & 0.63 & 0.61 & -- & -- & -- & -- & -- & -- & -- & -- & -- & -- & -- & -- \\
& Chronos2       & 0.63 & 0.55 & 0.68 & 0.55 & -- & -- & \underline{0.58} & \underline{0.56} & -- & -- & -- & -- & -- & -- & -- & -- & -- & -- & -- & -- \\
& iTransformer   & 1.07 & 0.78 & 1.07 & 0.76 & -- & -- & 1.08 & 0.80 & -- & -- & -- & -- & -- & -- & -- & -- & -- & -- & -- & -- \\
& MOIRAI         & \underline{0.61} & \underline{0.53} & \underline{0.62} & \underline{0.50} & -- & -- & \underline{0.58} & 0.58 & -- & -- & -- & -- & -- & -- & -- & -- & -- & -- & -- & -- \\
& LakeFM         & \textbf{0.38} & \textbf{0.47} & \textbf{0.23} & \textbf{0.40} & -- & -- & \textbf{0.54} & \textbf{0.54} & -- & -- & -- & -- & -- & -- & -- & -- & -- & -- & -- & -- \\
\midrule
\multirow{6}{*}{\textbf{LIRO}}
& LPTM           & 0.64 & 0.65 & 0.56 & 0.65 & -- & -- & 0.72 & 0.66 & -- & -- & -- & -- & -- & -- & -- & -- & -- & -- & -- & -- \\
& MOMENT         & 0.64 & 0.65 & 0.56 & 0.65 & -- & -- & 0.72 & 0.66 & -- & -- & -- & -- & -- & -- & -- & -- & -- & -- & -- & -- \\
& Chronos2       & 0.63 & 0.65 & \underline{0.53} & 0.63 & -- & -- & 0.74 & 0.66 & -- & -- & -- & -- & -- & -- & -- & -- & -- & -- & -- & -- \\
& iTransformer   & 0.69 & 0.66 & 0.59 & 0.64 & -- & -- & 0.79 & 0.68 & -- & -- & -- & -- & -- & -- & -- & -- & -- & -- & -- & -- \\
& MOIRAI         & \underline{0.59} & \underline{0.62} & \textbf{0.50} & \underline{0.62} & -- & -- & \underline{0.70} & \underline{0.63} & -- & -- & -- & -- & -- & -- & -- & -- & -- & -- & -- & -- \\
& LakeFM         & \textbf{0.57} & \textbf{0.59} & 0.58 & \textbf{0.61} & -- & -- & \textbf{0.56} & \textbf{0.56} & -- & -- & -- & -- & -- & -- & -- & -- & -- & -- & -- & -- \\
\midrule
\multirow{6}{*}{\textbf{MO}}
& LPTM           & 0.66 & 0.58 & 0.74 & 0.59 & -- & -- & 0.60 & 0.56 & -- & -- & -- & -- & -- & -- & -- & -- & -- & -- & -- & -- \\
& MOMENT         & 0.66 & 0.58 & 0.74 & 0.59 & -- & -- & 0.59 & 0.56 & -- & -- & -- & -- & -- & -- & -- & -- & -- & -- & -- & -- \\
& Chronos2       & 0.66 & 0.58 & 0.77 & 0.63 & -- & -- & 0.56 & \underline{0.53} & -- & -- & -- & -- & -- & -- & -- & -- & -- & -- & -- & -- \\
& iTransformer   & 0.71 & 0.66 & \underline{0.65} & 0.63 & -- & -- & 0.78 & 0.69 & -- & -- & -- & -- & -- & -- & -- & -- & -- & -- & -- & -- \\
& MOIRAI         & \underline{0.63} & \underline{0.54} & 0.70 & \underline{0.55} & -- & -- & \underline{0.54} & \underline{0.53} & -- & -- & -- & -- & -- & -- & -- & -- & -- & -- & -- & -- \\
& LakeFM         & \textbf{0.32} & \textbf{0.41} & \textbf{0.18} & \textbf{0.35} & -- & -- & \textbf{0.47} & \textbf{0.50} & -- & -- & -- & -- & -- & -- & -- & -- & -- & -- & -- & -- \\
\midrule
\multirow{6}{*}{\textbf{PRLA}}
& LPTM           & \underline{0.91} & \underline{0.68} & 0.41 & 0.51 & -- & -- & 1.42 & \textbf{0.84} & -- & -- & -- & -- & -- & -- & -- & -- & -- & -- & -- & -- \\
& MOMENT         & \underline{0.91} & \underline{0.68} & 0.41 & 0.51 & -- & -- & \underline{1.41} & \textbf{0.84} & -- & -- & -- & -- & -- & -- & -- & -- & -- & -- & -- & -- \\
& Chronos2       & 0.93 & 0.69 & 0.41 & 0.51 & -- & -- & 1.46 & 0.88 & -- & -- & -- & -- & -- & -- & -- & -- & -- & -- & -- & -- \\
& iTransformer   & 1.29 & 0.89 & 0.58 & 0.62 & -- & -- & 2.00 & 1.14 & -- & -- & -- & -- & -- & -- & -- & -- & -- & -- & -- & -- \\
& MOIRAI         & 0.92 & \underline{0.68} & \underline{0.40} & \underline{0.50} & -- & -- & 1.44 & 0.89 & -- & -- & -- & -- & -- & -- & -- & -- & -- & -- & -- & -- \\
& LakeFM         & \textbf{0.76} & \textbf{0.61} & \textbf{0.23} & \textbf{0.37} & -- & -- & \textbf{1.28} & \underline{0.84} & -- & -- & -- & -- & -- & -- & -- & -- & -- & -- & -- & -- \\
\midrule
\multirow{6}{*}{\textbf{PRPO}}
& LPTM           & 1.06 & \underline{0.80} & 0.68 & 0.71 & -- & -- & 1.43 & 0.90 & -- & -- & -- & -- & -- & -- & -- & -- & -- & -- & -- & -- \\
& MOMENT         & \underline{1.05} & \underline{0.80} & 0.68 & 0.71 & -- & -- & 1.42 & 0.90 & -- & -- & -- & -- & -- & -- & -- & -- & -- & -- & -- & -- \\
& Chronos2       & 1.10 & \underline{0.80} & 0.71 & 0.72 & -- & -- & 1.50 & \underline{0.88} & -- & -- & -- & -- & -- & -- & -- & -- & -- & -- & -- & -- \\
& iTransformer   & 1.06 & 0.84 & 0.74 & 0.74 & -- & -- & \underline{1.39} & 0.95 & -- & -- & -- & -- & -- & -- & -- & -- & -- & -- & -- & -- \\
& MOIRAI         & 1.30 & 0.84 & \underline{0.65} & \underline{0.70} & -- & -- & 1.80 & 1.00 & -- & -- & -- & -- & -- & -- & -- & -- & -- & -- & -- & -- \\
& LakeFM         & \textbf{0.73} & \textbf{0.62} & \textbf{0.33} & \textbf{0.47} & -- & -- & \textbf{1.13} & \textbf{0.75} & -- & -- & -- & -- & -- & -- & -- & -- & -- & -- & -- & -- \\
\midrule
\multirow{6}{*}{\textbf{SP}}
& LPTM           & \textbf{0.62} & \textbf{0.49} & \underline{0.38} & \underline{0.44} & 1.07 & \underline{0.64} & 0.42 & \underline{0.45} & \underline{0.79} & \textbf{0.51} & -- & -- & -- & -- & -- & -- & -- & -- & -- & -- \\
& MOMENT         & \textbf{0.62} & \textbf{0.49} & \underline{0.38} & \underline{0.43} & 1.07 & \underline{0.64} & 0.42 & 0.45 & \underline{0.79} & \textbf{0.51} & -- & -- & -- & -- & -- & -- & -- & -- & -- & -- \\
& Chronos2       & \underline{0.64} & \textbf{0.49} & 0.38 & 0.45 & \underline{1.02} & \textbf{0.62} & \textbf{0.41} & \textbf{0.44} & 0.88 & \underline{0.52} & -- & -- & -- & -- & -- & -- & -- & -- & -- & -- \\
& iTransformer   & 1.74 & 1.08 & 1.73 & 1.09 & 3.03 & 1.52 & 1.52 & 1.13 & 1.53 & 0.83 & -- & -- & -- & -- & -- & -- & -- & -- & -- & -- \\
& MOIRAI         & 0.75 & \underline{0.51} & \textbf{0.32} & \textbf{0.40} & 1.21 & \underline{0.64} & \underline{0.42} & \textbf{0.44} & 1.05 & 0.57 & -- & -- & -- & -- & -- & -- & -- & -- & -- & -- \\
& LakeFM         & 0.71 & 0.60 & 0.45 & 0.56 & \textbf{1.01} & 0.65 & 0.84 & 0.62 & \textbf{0.78} & 0.61 & -- & -- & -- & -- & -- & -- & -- & -- & -- & -- \\
\midrule
\multirow{6}{*}{\textbf{TB}}
& LPTM           & \textbf{0.56} & \textbf{0.42} & \underline{0.36} & \textbf{0.40} & 1.58 & 0.49 & \underline{0.48} & \textbf{0.40} & \textbf{0.75} & \textbf{0.45} & -- & -- & -- & -- & -- & -- & -- & -- & -- & -- \\
& MOMENT         & \textbf{0.56} & \textbf{0.42} & \underline{0.36} & \textbf{0.40} & 1.57 & 0.49 & \textbf{0.47} & \textbf{0.40} & \textbf{0.75} & \textbf{0.45} & -- & -- & -- & -- & -- & -- & -- & -- & -- & -- \\
& Chronos2       & \textbf{0.56} & \textbf{0.42} & \underline{0.36} & \underline{0.41} & \underline{1.45} & \underline{0.46} & 0.49 & \underline{0.41} & \underline{0.75} & \textbf{0.45} & -- & -- & -- & -- & -- & -- & -- & -- & -- & -- \\
& iTransformer   & 3.59 & 1.14 & 4.67 & 1.45 & \underline{1.45} & 0.68 & 2.55 & 1.02 & 3.87 & 1.00 & -- & -- & -- & -- & -- & -- & -- & -- & -- & -- \\
& MOIRAI         & 1.10 & \underline{0.48} & \textbf{0.34} & \textbf{0.40} & 2.80 & 0.56 & \underline{0.48} & \textbf{0.40} & 0.83 & \underline{0.47} & -- & -- & -- & -- & -- & -- & -- & -- & -- & -- \\
& LakeFM         & \underline{0.71} & 0.59 & 0.58 & 0.63 & \textbf{1.19} & \textbf{0.37} & 0.65 & 0.58 & 0.88 & 0.58 & -- & -- & -- & -- & -- & -- & -- & -- & -- & -- \\
\midrule
\multirow{6}{*}{\textbf{WI}}
& LPTM           & \underline{1.07} & 0.81 & 1.15 & 0.82 & -- & -- & \underline{0.99} & \underline{0.79} & -- & -- & -- & -- & -- & -- & -- & -- & -- & -- & -- & -- \\
& MOMENT         & \underline{1.07} & 0.81 & 1.15 & 0.82 & -- & -- & \underline{0.99} & \underline{0.79} & -- & -- & -- & -- & -- & -- & -- & -- & -- & -- & -- & -- \\
& Chronos2       & 1.11 & 0.82 & 1.21 & 0.87 & -- & -- & 1.01 & \underline{0.79} & -- & -- & -- & -- & -- & -- & -- & -- & -- & -- & -- & -- \\
& iTransformer   & 1.90 & 1.23 & 2.68 & 1.58 & -- & -- & 1.12 & 0.89 & -- & -- & -- & -- & -- & -- & -- & -- & -- & -- & -- & -- \\
& MOIRAI         & 1.10 & \underline{0.80} & \underline{1.14} & \underline{0.81} & -- & -- & 1.00 & 0.80 & -- & -- & -- & -- & -- & -- & -- & -- & -- & -- & -- & -- \\
& LakeFM         & \textbf{0.46} & \textbf{0.51} & \textbf{0.29} & \textbf{0.41} & -- & -- & \textbf{0.62} & \textbf{0.60} & -- & -- & -- & -- & -- & -- & -- & -- & -- & -- & -- & -- \\
\midrule
\textbf{Average Rank} & \textbf{LPTM} & \underline{3.27} & 3.40 & 3.73 & 3.83 & 4.17 & 3.58 & 3.33 & 3.20 & \underline{2.50} & 2.42 & \underline{2.75} & 3.00 & 3.75 & 5.00 & \underline{2.75} & 3.50 & 4.50 & 4.00 & 3.50 & 4.00 \\
\textbf{Average Rank} & \textbf{MOMENT} & 3.40 & 3.40 & 3.83 & 3.73 & 4.00 & 3.58 & \underline{2.97} & 3.27 & \underline{2.50} & \underline{2.17} & 3.50 & 3.00 & 4.25 & 5.00 & 3.25 & 4.00 & 4.50 & 4.00 & 3.50 & 4.00 \\
\textbf{Average Rank} & \textbf{Chronos2} & 3.40 & 3.20 & 3.77 & 3.90 & \underline{2.58} & \underline{1.83} & 3.80 & 3.20 & 3.50 & \textbf{1.92} & \textbf{1.00} & \textbf{1.50} & 5.50 & 3.50 & 3.50 & \underline{3.00} & \underline{1.75} & \textbf{2.00} & \textbf{1.00} & \textbf{1.00} \\
\textbf{Average Rank} & \textbf{iTransformer} & 5.53 & 5.67 & 4.70 & 4.97 & 3.58 & 5.67 & 5.07 & 5.53 & 5.17 & 5.42 & 5.50 & 5.00 & 4.00 & 4.00 & 6.00 & 6.00 & 6.00 & 6.00 & 6.00 & 6.00 \\
\textbf{Average Rank} & \textbf{MOIRAI} & 3.37 & \underline{2.87} & \underline{2.50} & \textbf{2.10} & 5.67 & 4.67 & \textbf{2.80} & \textbf{2.77} & 5.17 & 4.42 & 5.50 & 6.00 & \textbf{1.00} & \textbf{1.50} & 3.00 & \textbf{1.50} & 2.75 & 2.75 & \underline{2.00} & 4.00 \\
\textbf{Average Rank} & \textbf{LakeFM} & \textbf{2.03} & \textbf{2.47} & \textbf{2.47} & \underline{2.47} & \textbf{1.00} & \textbf{1.67} & 3.03 & \underline{3.03} & \textbf{2.17} & 4.67 & \underline{2.75} & \underline{2.50} & \underline{2.50} & \underline{2.00} & \textbf{2.50} & \underline{3.00} & \textbf{1.50} & \underline{2.25} & 5.00 & \underline{2.00} \\
\bottomrule
\end{tabular}%
}
\end{table*}


\begin{table}[!htbp]
\centering
\caption{Performance comparison against IMTS baselines on OOD LakeBeD-US data.}
\label{tab:imts_ood}
\resizebox{0.85\textwidth}{!}{%
\begin{tabular}{cc*{10}{c}}
\toprule
\textbf{Lake} & \textbf{Baseline} & \multicolumn{2}{c}{\textbf{Lake}} & \multicolumn{2}{c}{\textbf{WaterTemp\_C}} & \multicolumn{2}{c}{\textbf{Water\_DO\_mg\_per\_L}} & \multicolumn{2}{c}{\textbf{par}} & \multicolumn{2}{c}{\textbf{nh4}} \\
\cmidrule(lr){3-4}
\cmidrule(lr){5-6}
\cmidrule(lr){7-8}
\cmidrule(lr){9-10}
\cmidrule(lr){11-12}
 &  & \textbf{MSE} & \textbf{MAE} & \textbf{MSE} & \textbf{MAE} & \textbf{MSE} & \textbf{MAE} & \textbf{MSE} & \textbf{MAE} & \textbf{MSE} & \textbf{MAE} \\
\midrule
\multirow{3}{*}{\textbf{BARC}}
& HyperIMTS            & \underline{4.11} & \textbf{0.64} & \textbf{0.47} & \textbf{0.56} & 7.75 & \underline{0.72} & -- & -- & -- & -- \\
& ReIMTS               & 4.19 & \underline{0.65} & 0.69 & 0.58 & \underline{7.69} & \textbf{0.72} & -- & -- & -- & -- \\
& LakeFM               & \textbf{1.05} & 0.65 & \underline{0.50} & \underline{0.56} & \textbf{1.61} & 0.74 & -- & -- & -- & -- \\
\midrule
\multirow{3}{*}{\textbf{GL4}}
& HyperIMTS            & 2.08 & 2.29 & 2.54 & 2.16 & 2.52 & 2.73 & 2.73 & 12.64 & 4e4 & 2e2 \\
& ReIMTS               & \underline{2.29} & \underline{22.64} & \underline{2.56} & \underline{23.27} & \underline{2.00} & \underline{37.35} & \underline{15.46} & \underline{1.24} & \underline{3e3} & \underline{47.64} \\
& LakeFM               & \textbf{0.76} & \textbf{0.61} & \textbf{0.63} & \textbf{0.64} & \textbf{0.63} & \textbf{0.55} & \textbf{0.71} & \textbf{0.52} & \textbf{4.74} & \textbf{1.96} \\
\midrule
\multirow{3}{*}{\textbf{ME}}
& HyperIMTS            & 1.48 & 0.71 & 1.38 & 0.60 & 1.58 & 0.82 & -- & -- & -- & -- \\
& ReIMTS               & \underline{0.89} & \underline{0.50} & \underline{0.95} & \underline{0.40} & \underline{0.84} & \underline{0.60} & -- & -- & -- & -- \\
& LakeFM               & \textbf{0.33} & \textbf{0.44} & \textbf{0.21} & \textbf{0.37} & \textbf{0.45} & \textbf{0.51} & -- & -- & -- & -- \\
\midrule
\multirow{3}{*}{\textbf{SUGG}}
& HyperIMTS            & \underline{2.78} & \underline{0.62} & 5.04 & 0.93 & \underline{0.52} & \textbf{0.31} & -- & -- & -- & -- \\
& ReIMTS               & 3.47 & 0.78 & \underline{3.61} & \underline{0.84} & 3.34 & 0.72 & -- & -- & -- & -- \\
& LakeFM               & \textbf{0.45} & \textbf{0.48} & \textbf{0.57} & \textbf{0.60} & \textbf{0.34} & \underline{0.36} & -- & -- & -- & -- \\
\midrule
\multirow{3}{*}{\textbf{TOOK}}
& HyperIMTS            & \underline{3.92} & \underline{0.81} & \underline{1.22} & \underline{0.84} & \underline{6.62} & \underline{0.78} & -- & -- & -- & -- \\
& ReIMTS               & 8.43 & 0.81 & \textbf{1.01} & \textbf{0.78} & 15.85 & 0.85 & -- & -- & -- & -- \\
& LakeFM               & \textbf{1.41} & \textbf{0.71} & 1.82 & 1.11 & \textbf{1.00} & \textbf{0.30} & -- & -- & -- & -- \\
\midrule
\multirow{3}{*}{\textbf{TR}}
& HyperIMTS            & 1.13 & 0.69 & \underline{0.55} & \underline{0.61} & 0.60 & 0.56 & 3.40 & 1.13 & -- & -- \\
& ReIMTS               & \underline{0.78} & \underline{0.68} & \textbf{0.52} & \textbf{0.44} & \underline{0.47} & \textbf{0.38} & \underline{2.47} & \underline{1.00} & -- & -- \\
& LakeFM               & \textbf{0.71} & \textbf{0.66} & 0.85 & 0.75 & \textbf{0.45} & \underline{0.53} & \textbf{0.97} & \textbf{0.72} & -- & -- \\
\midrule
\textbf{Average Rank} & \textbf{HyperIMTS} & \underline{2.50} & \underline{2.33} & \underline{2.33} & \underline{2.33} & 2.67 & 2.33 & 3.00 & 3.00 & 3.00 & 3.00 \\
\textbf{Average Rank} & \textbf{ReIMTS} & \underline{2.50} & \underline{2.33} & \textbf{1.83} & \textbf{1.83} & \underline{2.33} & \underline{2.00} & \underline{2.00} & \underline{2.00} & \underline{2.00} & \underline{2.00} \\
\textbf{Average Rank} & \textbf{LakeFM} & \textbf{1.00} & \textbf{1.33} & \textbf{1.83} & \textbf{1.83} & \textbf{1.00} & \textbf{1.67} & \textbf{1.00} & \textbf{1.00} & \textbf{1.00} & \textbf{1.00} \\
\bottomrule
\end{tabular}%
}
\end{table}

\begin{table}[!htbp]
\centering
\caption{Performance comparison against IMTS baselines on ID LakeBeD-US data.}
\label{tab:imts_id}
\resizebox{1.0\textwidth}{!}{%
\begin{tabular}{cc*{20}{c}}
\toprule
\textbf{Lake} & \textbf{Baseline} & \multicolumn{2}{c}{\textbf{Lake}} & \multicolumn{2}{c}{\textbf{WaterTemp\_C}} & \multicolumn{2}{c}{\textbf{Chla\_ugL}} & \multicolumn{2}{c}{\textbf{Water\_DO\_mg\_per\_L}} & \multicolumn{2}{c}{\textbf{par}} & \multicolumn{2}{c}{\textbf{SRP\_ugL}} & \multicolumn{2}{c}{\textbf{Water\_TP\_mg\_per\_L}} & \multicolumn{2}{c}{\textbf{nh4}} & \multicolumn{2}{c}{\textbf{tn}} & \multicolumn{2}{c}{\textbf{no3no2}} \\
\cmidrule(lr){3-4}
\cmidrule(lr){5-6}
\cmidrule(lr){7-8}
\cmidrule(lr){9-10}
\cmidrule(lr){11-12}
\cmidrule(lr){13-14}
\cmidrule(lr){15-16}
\cmidrule(lr){17-18}
\cmidrule(lr){19-20}
\cmidrule(lr){21-22}
 &  & \textbf{MSE} & \textbf{MAE} & \textbf{MSE} & \textbf{MAE} & \textbf{MSE} & \textbf{MAE} & \textbf{MSE} & \textbf{MAE} & \textbf{MSE} & \textbf{MAE} & \textbf{MSE} & \textbf{MAE} & \textbf{MSE} & \textbf{MAE} & \textbf{MSE} & \textbf{MAE} & \textbf{MSE} & \textbf{MAE} & \textbf{MSE} & \textbf{MAE} \\
\midrule
\multirow{3}{*}{\textbf{AL}}
& HyperIMTS            & 1.08 & 0.81 & 0.74 & 0.70 & \textbf{1.25} & \underline{0.64} & 0.71 & \underline{0.66} & 1.52 & 1.03 & -- & -- & -- & -- & -- & -- & -- & -- & -- & -- \\
& ReIMTS               & \textbf{0.81} & \underline{0.67} & \textbf{0.31} & \textbf{0.42} & \underline{1.41} & 0.71 & \underline{0.71} & 0.68 & \textbf{0.75} & \textbf{0.67} & -- & -- & -- & -- & -- & -- & -- & -- & -- & -- \\
& LakeFM               & \underline{0.87} & \textbf{0.65} & \underline{0.59} & \underline{0.62} & 1.42 & \textbf{0.62} & \textbf{0.68} & \textbf{0.60} & \underline{1.02} & \underline{0.74} & -- & -- & -- & -- & -- & -- & -- & -- & -- & -- \\
\midrule
\multirow{3}{*}{\textbf{BM}}
& HyperIMTS            & 1.24 & 0.83 & 1.34 & 0.87 & 1.42 & 0.83 & 1.05 & 0.77 & 1.29 & 0.84 & -- & -- & -- & -- & -- & -- & -- & -- & -- & -- \\
& ReIMTS               & \textbf{0.47} & \textbf{0.50} & \textbf{0.25} & \textbf{0.36} & \textbf{0.74} & \textbf{0.61} & \textbf{0.40} & \textbf{0.47} & \textbf{0.70} & \textbf{0.66} & -- & -- & -- & -- & -- & -- & -- & -- & -- & -- \\
& LakeFM               & \underline{0.73} & \underline{0.62} & \underline{0.48} & \underline{0.58} & \underline{1.07} & \underline{0.71} & \underline{0.69} & \underline{0.55} & \underline{0.95} & \underline{0.71} & -- & -- & -- & -- & -- & -- & -- & -- & -- & -- \\
\midrule
\multirow{3}{*}{\textbf{BVR}}
& HyperIMTS            & \underline{2.23} & \underline{29.02} & 2.69 & 2.46 & -- & -- & 2.56 & 2.02 & -- & -- & 2.83 & 12.33 & 2.57 & 18.13 & 22.50 & 1.80 & 2.02 & 6.45 & -- & -- \\
& ReIMTS               & 2.55 & 32.17 & \underline{2.23} & \underline{10.23} & -- & -- & \underline{2.50} & \underline{12.78} & -- & -- & \underline{0.90} & \underline{0.55} & \underline{0.37} & \underline{0.45} & \underline{1.01} & \textbf{0.51} & \underline{0.92} & \textbf{0.53} & -- & -- \\
& LakeFM               & \textbf{0.66} & \textbf{0.53} & \textbf{1.76} & \textbf{1.10} & -- & -- & \textbf{1.65} & \textbf{0.91} & -- & -- & \textbf{0.81} & \textbf{0.52} & \textbf{0.24} & \textbf{0.39} & \textbf{0.65} & \underline{0.51} & \textbf{0.65} & \underline{0.56} & -- & -- \\
\midrule
\multirow{3}{*}{\textbf{CB}}
& HyperIMTS            & 0.91 & 0.72 & \underline{0.41} & \underline{0.54} & 2.49 & \underline{0.73} & \textbf{0.64} & 0.61 & 0.89 & 0.84 & -- & -- & -- & -- & -- & -- & -- & -- & -- & -- \\
& ReIMTS               & \underline{0.80} & \underline{0.68} & \textbf{0.26} & \textbf{0.49} & \underline{2.37} & 0.79 & 0.65 & \textbf{0.57} & \textbf{0.78} & \underline{0.62} & -- & -- & -- & -- & -- & -- & -- & -- & -- & -- \\
& LakeFM               & \textbf{0.76} & \textbf{0.58} & 0.57 & 0.62 & \textbf{1.18} & \textbf{0.35} & \underline{0.63} & \underline{0.59} & \underline{0.80} & \textbf{0.60} & -- & -- & -- & -- & -- & -- & -- & -- & -- & -- \\
\midrule
\multirow{3}{*}{\textbf{CR}}
& HyperIMTS            & 1.14 & 0.76 & \underline{0.56} & 0.56 & 2.15 & 0.90 & 1.60 & 0.92 & 0.93 & 0.75 & -- & -- & -- & -- & -- & -- & -- & -- & -- & -- \\
& ReIMTS               & \underline{0.85} & \underline{0.60} & 0.67 & \textbf{0.45} & \underline{2.00} & \underline{0.77} & \textbf{0.86} & \underline{0.68} & \textbf{0.63} & \underline{0.61} & -- & -- & -- & -- & -- & -- & -- & -- & -- & -- \\
& LakeFM               & \textbf{0.72} & \textbf{0.58} & \textbf{0.37} & \underline{0.51} & \textbf{0.98} & \textbf{0.56} & \underline{1.04} & \textbf{0.66} & \underline{0.67} & \textbf{0.59} & -- & -- & -- & -- & -- & -- & -- & -- & -- & -- \\
\midrule
\multirow{3}{*}{\textbf{CRAM}}
& HyperIMTS            & 0.83 & 0.72 & 0.83 & 0.77 & -- & -- & 0.83 & \underline{0.68} & -- & -- & -- & -- & -- & -- & -- & -- & -- & -- & -- & -- \\
& ReIMTS               & \underline{0.69} & \underline{0.68} & \textbf{0.54} & \textbf{0.69} & -- & -- & \underline{0.75} & 0.68 & -- & -- & -- & -- & -- & -- & -- & -- & -- & -- & -- & -- \\
& LakeFM               & \textbf{0.63} & \textbf{0.66} & \underline{0.68} & \underline{0.69} & -- & -- & \textbf{0.57} & \textbf{0.63} & -- & -- & -- & -- & -- & -- & -- & -- & -- & -- & -- & -- \\
\midrule
\multirow{3}{*}{\textbf{FCR}}
& HyperIMTS            & \underline{0.74} & \underline{0.61} & \underline{0.38} & \textbf{0.46} & -- & -- & \underline{0.46} & \underline{0.50} & -- & -- & \underline{1.20} & \underline{0.80} & 1.07 & 0.77 & \textbf{0.71} & \textbf{0.62} & \underline{0.77} & \underline{0.67} & \textbf{0.50} & \textbf{0.47} \\
& ReIMTS               & 1.94 & 0.92 & 0.66 & 0.50 & -- & -- & 2.67 & 1.17 & -- & -- & \textbf{0.98} & \textbf{0.74} & \textbf{0.63} & \textbf{0.57} & 4.20 & 1.44 & 2.70 & 1.16 & 3.05 & 1.19 \\
& LakeFM               & \textbf{0.71} & \textbf{0.57} & \textbf{0.33} & \underline{0.45} & -- & -- & \textbf{0.42} & \textbf{0.48} & -- & -- & 1.36 & 0.85 & \underline{0.69} & \underline{0.59} & \underline{0.92} & \underline{0.63} & \textbf{0.75} & \textbf{0.55} & \underline{0.88} & \underline{0.58} \\
\midrule
\multirow{3}{*}{\textbf{FI}}
& HyperIMTS            & 1.37 & 0.75 & 1.49 & 0.66 & -- & -- & 1.25 & 0.85 & -- & -- & -- & -- & -- & -- & -- & -- & -- & -- & -- & -- \\
& ReIMTS               & \underline{0.88} & \underline{0.52} & \underline{1.20} & \textbf{0.42} & -- & -- & \underline{0.56} & \underline{0.57} & -- & -- & -- & -- & -- & -- & -- & -- & -- & -- & -- & -- \\
& LakeFM               & \textbf{0.39} & \textbf{0.49} & \textbf{0.28} & \underline{0.43} & -- & -- & \textbf{0.51} & \textbf{0.55} & -- & -- & -- & -- & -- & -- & -- & -- & -- & -- & -- & -- \\
\midrule
\multirow{3}{*}{\textbf{LIRO}}
& HyperIMTS            & 0.97 & 0.79 & 0.85 & 0.79 & -- & -- & 1.08 & 0.79 & -- & -- & -- & -- & -- & -- & -- & -- & -- & -- & -- & -- \\
& ReIMTS               & \underline{0.69} & \textbf{0.63} & \textbf{0.59} & \textbf{0.55} & -- & -- & \underline{0.79} & \underline{0.72} & -- & -- & -- & -- & -- & -- & -- & -- & -- & -- & -- & -- \\
& LakeFM               & \textbf{0.67} & \underline{0.66} & \underline{0.70} & \underline{0.69} & -- & -- & \textbf{0.64} & \textbf{0.64} & -- & -- & -- & -- & -- & -- & -- & -- & -- & -- & -- & -- \\
\midrule
\multirow{3}{*}{\textbf{MO}}
& HyperIMTS            & 0.89 & 0.68 & 0.75 & 0.62 & -- & -- & 1.03 & 0.75 & -- & -- & -- & -- & -- & -- & -- & -- & -- & -- & -- & -- \\
& ReIMTS               & \underline{0.45} & \underline{0.47} & \underline{0.27} & \textbf{0.36} & -- & -- & \underline{0.63} & \underline{0.55} & -- & -- & -- & -- & -- & -- & -- & -- & -- & -- & -- & -- \\
& LakeFM               & \textbf{0.33} & \textbf{0.46} & \textbf{0.24} & \underline{0.41} & -- & -- & \textbf{0.41} & \textbf{0.51} & -- & -- & -- & -- & -- & -- & -- & -- & -- & -- & -- & -- \\
\midrule
\multirow{3}{*}{\textbf{PRLA}}
& HyperIMTS            & \underline{0.65} & \underline{0.67} & 0.82 & 0.81 & -- & -- & \textbf{0.48} & \textbf{0.53} & -- & -- & -- & -- & -- & -- & -- & -- & -- & -- & -- & -- \\
& ReIMTS               & \textbf{0.47} & \textbf{0.50} & \underline{0.39} & \textbf{0.45} & -- & -- & \underline{0.56} & \underline{0.54} & -- & -- & -- & -- & -- & -- & -- & -- & -- & -- & -- & -- \\
& LakeFM               & 1.01 & 0.75 & \textbf{0.38} & \underline{0.51} & -- & -- & 1.65 & 1.00 & -- & -- & -- & -- & -- & -- & -- & -- & -- & -- & -- & -- \\
\midrule
\multirow{3}{*}{\textbf{PRPO}}
& HyperIMTS            & 0.90 & \underline{0.70} & \underline{0.56} & \underline{0.64} & -- & -- & 1.23 & \underline{0.77} & -- & -- & -- & -- & -- & -- & -- & -- & -- & -- & -- & -- \\
& ReIMTS               & \underline{0.84} & 0.71 & 0.58 & 0.67 & -- & -- & \textbf{1.09} & \textbf{0.75} & -- & -- & -- & -- & -- & -- & -- & -- & -- & -- & -- & -- \\
& LakeFM               & \textbf{0.79} & \textbf{0.67} & \textbf{0.40} & \textbf{0.55} & -- & -- & \underline{1.18} & 0.80 & -- & -- & -- & -- & -- & -- & -- & -- & -- & -- & -- & -- \\
\midrule
\multirow{3}{*}{\textbf{SP}}
& HyperIMTS            & 0.79 & 0.64 & 0.49 & 0.61 & 1.24 & \textbf{0.66} & \underline{0.79} & \underline{0.67} & \textbf{0.65} & \underline{0.63} & -- & -- & -- & -- & -- & -- & -- & -- & -- & -- \\
& ReIMTS               & \underline{0.75} & \underline{0.62} & \textbf{0.40} & \textbf{0.32} & \underline{1.24} & 0.69 & \textbf{0.65} & 0.67 & \underline{0.70} & 0.70 & -- & -- & -- & -- & -- & -- & -- & -- & -- & -- \\
& LakeFM               & \textbf{0.71} & \textbf{0.60} & \underline{0.45} & \underline{0.56} & \textbf{1.01} & \underline{0.65} & 0.84 & \textbf{0.62} & 0.78 & \textbf{0.61} & -- & -- & -- & -- & -- & -- & -- & -- & -- & -- \\
\midrule
\multirow{3}{*}{\textbf{TB}}
& HyperIMTS            & 1.50 & \underline{0.60} & 0.90 & \underline{0.47} & \underline{12.85} & 1.03 & 1.17 & 0.66 & \underline{0.86} & \underline{0.60} & -- & -- & -- & -- & -- & -- & -- & -- & -- & -- \\
& ReIMTS               & \underline{0.91} & \textbf{0.46} & \textbf{0.24} & \textbf{0.34} & 13.37 & \underline{0.98} & \textbf{0.44} & \textbf{0.52} & \textbf{0.36} & \textbf{0.44} & -- & -- & -- & -- & -- & -- & -- & -- & -- & -- \\
& LakeFM               & \textbf{0.76} & 0.62 & \underline{0.74} & 0.68 & \textbf{1.21} & \textbf{0.48} & \underline{0.58} & \underline{0.55} & 0.94 & 0.64 & -- & -- & -- & -- & -- & -- & -- & -- & -- & -- \\
\midrule
\multirow{3}{*}{\textbf{WI}}
& HyperIMTS            & 1.37 & 0.96 & 1.17 & 0.92 & -- & -- & 1.58 & 1.01 & -- & -- & -- & -- & -- & -- & -- & -- & -- & -- & -- & -- \\
& ReIMTS               & \underline{0.78} & \underline{0.64} & \textbf{0.34} & \textbf{0.42} & -- & -- & \underline{1.22} & \underline{0.86} & -- & -- & -- & -- & -- & -- & -- & -- & -- & -- & -- & -- \\
& LakeFM               & \textbf{0.60} & \textbf{0.62} & \underline{0.49} & \underline{0.59} & -- & -- & \textbf{0.72} & \textbf{0.65} & -- & -- & -- & -- & -- & -- & -- & -- & -- & -- & -- & -- \\
\midrule
\textbf{Average Rank} & \textbf{HyperIMTS} & 2.80 & 2.67 & 2.73 & 2.67 & 2.50 & \underline{2.33} & 2.60 & 2.53 & 2.50 & \underline{2.67} & 2.50 & 2.50 & \underline{3.00} & \underline{3.00} & \underline{2.00} & \textbf{2.00} & \underline{2.50} & 2.50 & \textbf{1.00} & \textbf{1.00} \\
\textbf{Average Rank} & \textbf{ReIMTS} & \underline{1.93} & \underline{1.93} & \underline{1.67} & \textbf{1.33} & \underline{2.00} & \underline{2.33} & \underline{1.80} & \underline{2.00} & \textbf{1.17} & \textbf{1.67} & \textbf{1.50} & \textbf{1.50} & \textbf{1.50} & \textbf{1.50} & 2.50 & \textbf{2.00} & \underline{2.50} & \underline{2.00} & 3.00 & 3.00 \\
\textbf{Average Rank} & \textbf{LakeFM} & \textbf{1.27} & \textbf{1.40} & \textbf{1.60} & \underline{2.00} & \textbf{1.50} & \textbf{1.33} & \textbf{1.60} & \textbf{1.47} & \underline{2.33} & \textbf{1.67} & \underline{2.00} & \underline{2.00} & \textbf{1.50} & \textbf{1.50} & \textbf{1.50} & \textbf{2.00} & \textbf{1.00} & \textbf{1.50} & \underline{2.00} & \underline{2.00} \\
\bottomrule
\end{tabular}%
}
\end{table}
\section{\camready{Visualization}}
\label{sec:visualization}
Figure \ref{fig:prediction_plots} shows forecasting plots across four lakes for the Water temperature variable observed at Depth 0m
\begin{figure*}[t]
    \centering
    \captionsetup[subfigure]{font=scriptsize,skip=1pt}
    \captionsetup[figure]{font=small,skip=2pt}

    \begin{subfigure}[b]{0.47\textwidth}
        \centering
        \includegraphics[width=\linewidth,height=0.28\textheight,keepaspectratio]{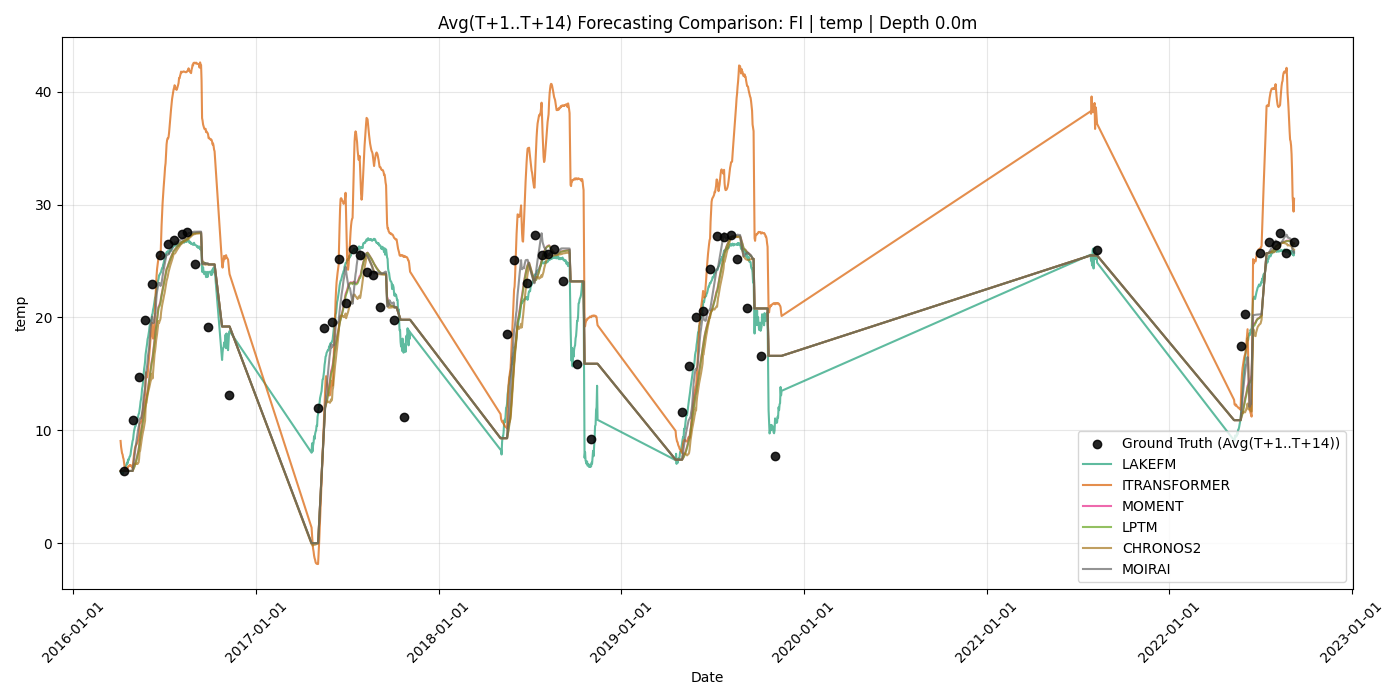}
        \caption{Lake FI}
        \label{fig:pred_fi}
    \end{subfigure}
    \hfill
    \begin{subfigure}[b]{0.47\textwidth}
        \centering
        \includegraphics[width=\linewidth,height=0.28\textheight,keepaspectratio]{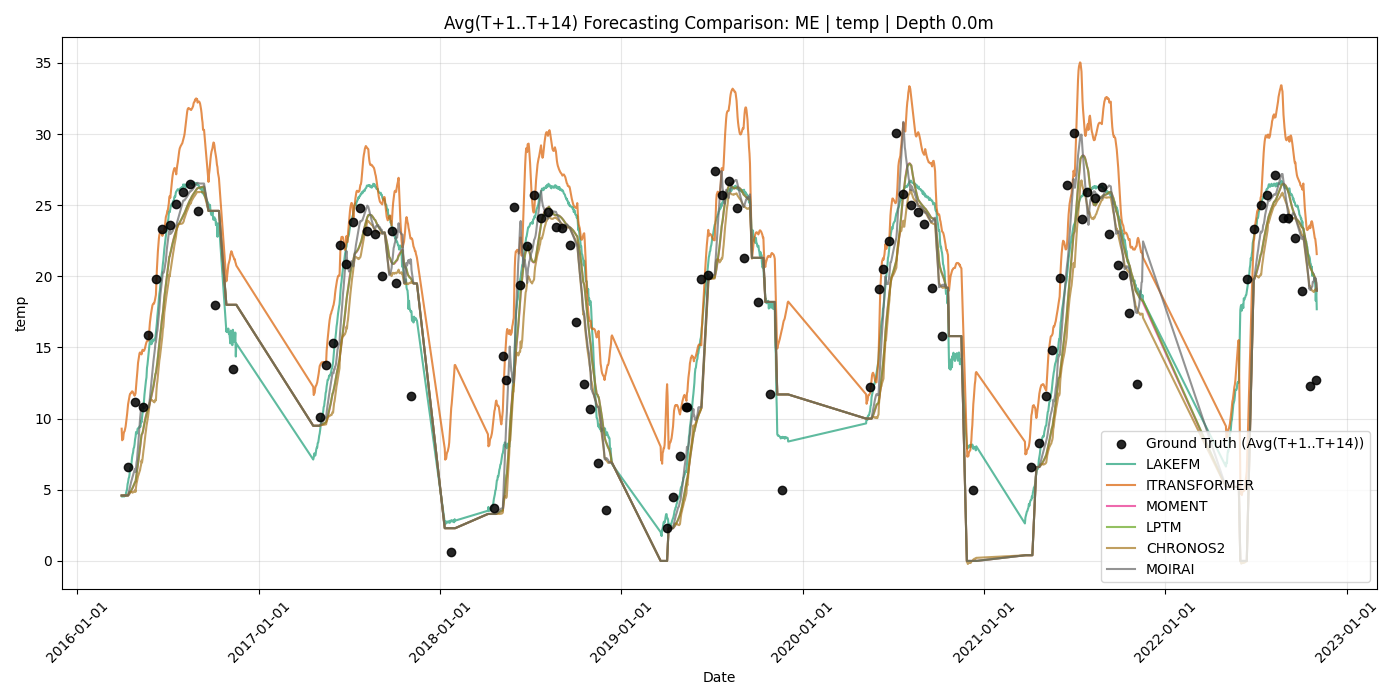}
        \caption{Lake ME}
        \label{fig:pred_me}
    \end{subfigure}

    \vspace{-0.4em}

    \begin{subfigure}[b]{0.47\textwidth}
        \centering
        \includegraphics[width=\linewidth,height=0.28\textheight,keepaspectratio]{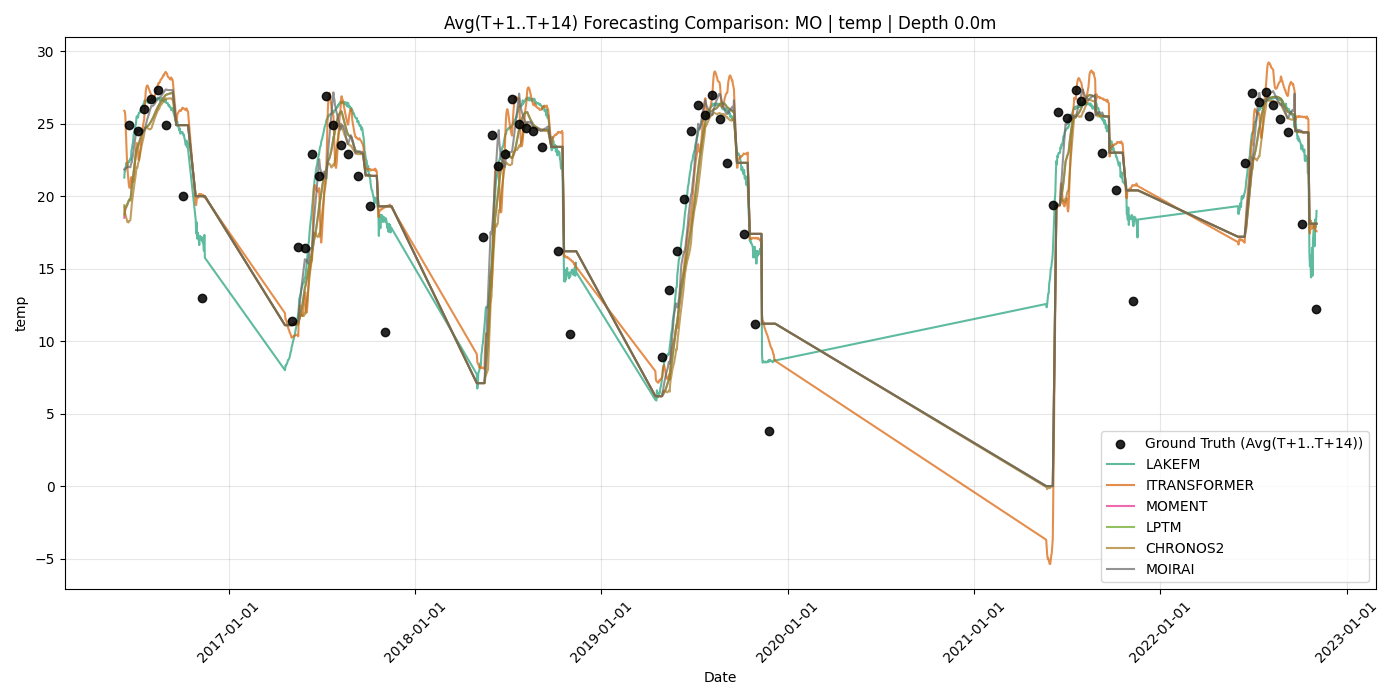}
        \caption{Lake MO}
        \label{fig:pred_mo}
    \end{subfigure}
    \hfill
    \begin{subfigure}[b]{0.47\textwidth}
        \centering
        \includegraphics[width=\linewidth,height=0.28\textheight,keepaspectratio]{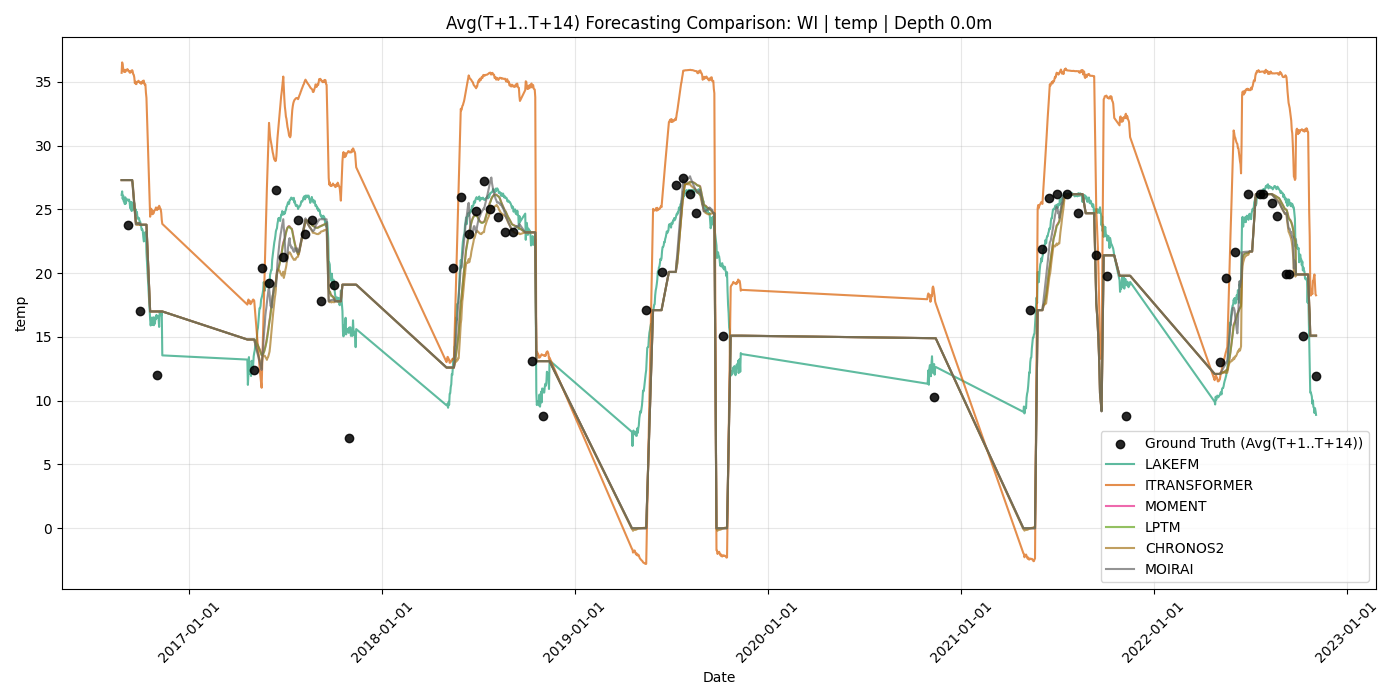}
        \caption{Lake WI}
        \label{fig:pred_wi}
    \end{subfigure}

    \vspace{-0.3em}

    \caption{Prediction plots for four lakes (FI, ME, MO, WI) for water temperature at depth 0.0m.}
    \label{fig:prediction_plots}
\end{figure*}




\end{document}